\documentclass[12pt]{report}
\usepackage{times}
\usepackage{url}
\usepackage{latexsym}
\usepackage[T1]{fontenc}
\usepackage{setspace}

\newlength\titlebox
\newenvironment{references}%
{\begin{list}{}{\setlength{\leftmargin}{\parindent}
               \setlength{\labelsep}{0in} 
               \setlength{\labelwidth}{0in}
               \setlength{\itemindent}{-\leftmargin}}}%
{\end{list}}

\title{Indirectly Supervised English Sentence Break Prediction Using
  Paragraph Break Probability Estimates}

\author{Robert C. Moore \\
  Google Research \\
  {\tt bobmoore@google.com}\thanks{Alternate email address: {\tt robert.carter.moore@gmail.com}}}

\begin{document}
\maketitle
\begin{abstract}

This report explores the use of paragraph break probability estimates
to help predict the location of sentence breaks in English natural
language text. We show that a sentence break predictor based almost
solely on paragraph break probability estimates can achieve high
accuracy on this task. This sentence break predictor is trained almost
entirely on a large amount of naturally occurring text without
sentence break annotations, with only a small amount of annotated data
needed to tune two hyperparameters. We also show that even better
results can be achieved across in-domain and out-of-domain test data,
if paragraph break probability signals are combined with a support
vector machine classifier trained on a somewhat larger amount of
sentence-break-annotated data. Numerous related issues are addressed
along the way.

\end{abstract}

\tableofcontents

\chapter{Introduction}

Segmenting extended texts into individual sentences (or sentence-like
sequences) has always been, and still remains, an important early step
in most NLP pipelines. Even in the age of deep neural networks, many
state-of-the-art systems are designed to work one sentence (or one
sentence pair) at a time; e.g., Wu et al.\ (2016), Vaswani et
al.\ (2017), and Devlin et al.\ (2018).

At the same time, relatively little research has gone into accurately
segmenting extended texts into sentences, compared to other core NLP
problems such as parsing or part-of-speech tagging. For English text,
this task is nontrivial, because the three punctuation marks
canonically used to signal the end of a sentence -- period (\verb|.|),
question mark (\verb|?|), and exclamation point (\verb|!|) -- can all
be used in contexts other than the end of a sentence. The use of
periods to mark initials (\verb|J.K. Rowling|) or abbreviations
(\verb|Alphabet Inc.|) is most common, but even question marks or
exclamation points may not signal the end of a sentence if they are
used in quotations, titles, or names (\verb|Yahoo!|) embedded in
longer sentences.

As far as we are aware, Gillick's (2009) relatively simple directly
supervised linear support vector machine (SVM) binary classifier
remains the state of the art, at least for in-domain test data. As we
will see, the type of sentence break predictor developed by Gillick is
very accurate on formal, well-edited text from the same domain it was
trained on (Wall Street Journal), but performs much less well on more
informal, less-well-edited, out-of-domain text.

In this report we investigate a signal not generally used in sentence
break prediction: a paragraph break probability estimate. Our interest
in this approach was sparked by noticing that in the newswire text
constituting the LDC English Gigaword Fifth Edition corpus (Parker et
al., 2011), it seemed that a very high proportion of the paragraphs
were just one sentence long. Since paragraph breaks are naturally
annotated in found text (by indented first sentences, blank lines, or
explicit tags), we can train a directly supervised model to
probabilistically predict the location of paragraph breaks in a text
from which the paragraph breaks have been removed, without extensive
training data annotation. If a high proportion of the sentence breaks
in the training text are also paragraph breaks, a model that estimates
the probability of a paragraph break should be a good proxy for a
sentence break prediction model, if we simply adjust the probability
threshold at which we predict a break.

Formally, if paragraph breaks in text are simply a random sample of
the sentence breaks, then
\[p(\langle {\rm S} \rangle|t) = 1/\alpha \times p(\langle {\rm P} \rangle|t),\]
where $p(\langle {\rm S} \rangle|t)$ is the probability of a sentence
break, given a text context $t$, $p(\langle {\rm P} \rangle|t)$ is the
probability of a paragraph break given $t$, and $\alpha$ is the
proportion of sentence breaks that are paragraph breaks. While
paragraph breaks are not simply a random sample of sentence breaks, we
nevertheless find it effective to predict a sentence break whenever
\[p(\langle {\rm P} \rangle|t) > \beta,\]
for an estimate of $p(\langle {\rm P} \rangle|t)$ based on a large
corpus of naturally occurring text, and a value of the threshold
$\beta$ optimized on a much smaller text corpus annotated with
sentence breaks.

In this report, we explore this idea in depth. In Chapter~2, we review
what we consider to be the most relevant previous work on sentence
break prediction, by Riley (1989), Palmer and Hearst (1997), Kiss and
Strunk (2006), and Gillick (2009). In Chapter~3, we describe the
various data sets we use: the LDC English Gigaword Fifth Edition
corpus (Parker et al., 2011); the Wall Street Journal portion of the
Penn Treebank corpus (Marcus et al., 1993); the Satz WSJ corpus, which
consists of more than 420,000 words of text extracted from the ACL/DCI
corpus (ACL/DCI, 1993) and sentence-break-annotated by Palmer and
Hearst (1997); and the LDC English Web Treebank corpus (Bies et al.,
2012). In Chapter~4, we examine three problematic issues in sentence
break prediction: whether the problem is one of classifying
punctuation marks or segmenting text; what should count as a sentence
break; and how sentence break prediction accuracy should be measured.

In Chapter~5, we present the details of how we predict sentence breaks
using paragraph break probability estimates. We develop a simple
initial approach based on an N-gram model, and then explore a number
of refinements. With these refinements, we find that we can achieve an
error rate of less than one percent on the in-domain test set used by
Gillick, which is very low compared to many NLP tasks, but is still
more than twice the error rate of Gillick's SVM classifier, when
tested under comparable conditions. In Chapter~6, we explore
incorporating a paragraph break probability signal into an SVM similar
to Gillick's, trying several different ways of incorporating a
paragraph break probability signal into the SVM. The one that performs
the best on our in-domain development set also slightly out-performs a
baseline SVM without such a signal on the in-domain test set, but not
significantly so.

In Chapter~7, we turn our attention to poorly-edited, out-of-domain
text, and observe that the results change dramatically. All error
rates increase from a fraction of one percent to several percent, but
overall the sentence break predictor that uses only paragraph break
probability estimates performs the best, and the SVM without any
paragraph break probability signal performs the worst. The SVM
incorporating a paragraph break probability signal performs almost as
well on the out-of-domain test set as the predictor based on paragraph
break probability estimates alone, hence that SVM classifier performs
well in both test conditions.

Historically, actual paragraph breaks in test data have not been used
to help predict sentence breaks between paragraphs in evaluations
of sentence break prediction. In effect, the entire test set has been
treated as a single paragraph. If, however, we take into account the
fact that a sentence break candidate occurs at a paragraph break, then
the probability of a sentence break dramatically increases. In Chapter~8,
We look at two ways of incorporating paragraph breaks in the test
data into our statistical models, plus the simple heuristic of
declaring a sentence break whenever a paragraph ends in standard
sentence-final punctuation. All three methods substantially
out-perform ignoring test data paragraph breaks, but none
consistently out-performed the others on in-domain data. The heuristic
method, however, clearly out-performs either of the model-based
methods on our out-of-domain test set.

In Chapter~9, we look at the question of how much annotated data is
needed to tune a sentence break predictor based only on paragraph
break probability estimates. In most of our experiments we take
advantage of having a training set of more than 35,000 sentences
annotated with sentence breaks, derived from the Penn Treebank (Marcus
et al., 1993). Our paragraph-break-probability-based sentence break
predictor, however, requires data annotated with sentence breaks only
to optimize two hyperparameters: a count-cutoff used in estimating
paragraph break probabilities, and the paragraph break probability
threshold used to predict a sentence break.  With only those two
parameters, it seems likely that something much less than 35,000
annotated sentences should be sufficient to optimize the predictor,
which could be important in applying the method to low-resource
languages.

To test this conjecture, we reoptimize our two hyperparameters using
only 1000 annotated sentences. We find that tuning the
paragraph-break-probability-based sentence break predictor on the
small annotated data set, instead of the full 35,000 annotated
sentences, actually performs slightly better on the in-domain test
set, and only slightly worse on the out-of-domain test set. For
comparison, we also train an SVM-based predictor on this small
annotated data set, and find that it performs substantially worse than
the paragraph-break-probability-based predictor even on the in-domain
test set.

In Chapter~10, we look at the issue of sentence breaks without white
space. In conventionally formatted English text, with the exception of
some ellipses (\verb|...|), every sentence break occurs at white
space. Most sentence break predictors, including ours, do not even
consider the possibility of a sentence break except at white space or
following an ellipsis. In preprocessing and formatting our
out-of-domain test set, however, we noticed many sentence breaks
without white space and not following ellipses, due to nonstandard
formatting by internet users.

Predicting sentence breaks without white space in out-of-domain data,
when there are almost none in our in-domain training data, presents us
with a problem. If we just retrain a model on clean in-domain data,
but allow for the possibility of sentence breaks without white space,
the model will simply learn that this never happens, except following
ellipses.  The approach we have adopted to get around this problem is
to try to predict where there should be additional white space, insert
a space character, and then apply a sentence break predictor trained
on clean, in-domain data. We use the English Gigaword Fifth Edition
corpus to train a missing-space model, treating the presence or absence
of white space following possible sentence-final punctuation as a
hidden token. In addition, we use a set of heuristics implemented as
regular expressions to identify classes of embedded punctuation marks
that should not be considered candidates for missing spaces (in
hyphenated abbreviations, decimal numbers, strings of multiple
initials, inflected forms of abbreviations, email addresses, and
computer filenames, domain names, URLs, and related
expressions). Using this model and these heuristics to predict missing
spaces, we obtain additional improvements in predicting sentence
breaks in the out-of-domain test set, with no losses in the
well-edited in-domain test set.

In Chapter~11, we undertake a manual analysis of the remaining errors
produced by our best overall sentence break predictor, and of the
differences in the results produced between some alternative
predictors.  In the course of this analysis, however, we observe what
seem to be numerous errors in the reference annotations of sentence
breaks, particularly sentence breaks signaled by ellipses.

In Chapter~12, we propose some principles for deciding when an
ellipsis signals a sentence break, apply these principles to our test
data, and re-evaluate a number of sentence break predictors on the
corrected test data. We carry out some additional analysis of the
errors made on this corrected test data, and find that the advantage
of the predictor based only on estimated paragraph break probabilities
over the SVM incorporating a paragraph break probability signal
increases substantially on the corrected test data, but this advantage
occurs only on sentence breaks signaled by question marks or
exclamation points. We then note that the best accuracy on both
in-domain and out-of-domain test sets would be obtained by a combined
predictor that applies the predictor based only on estimated paragraph
break probabilities on sentence break candidates signaled by question
marks and exclamation points, and applies the SVM incorporating a
paragraph break probability signal everywhere else.

In Chapter~13, we state our conclusions, reviewing our achievements,
noting the limitations of our work, proposing some approaches for
improving sentence break prediction accuracy, and suggesting how the
sentence break prediction problem might be reconceived.

Following the main body of the report, we include a number of
appendices.  In Appendix~A, we give further details of the
preprocessing we carry out on each of the corpora used in our research
that we describe in Chapter~3.  In Appendix~B, we give more details of
how we normalize both training texts and test texts with respect to
the use of spaces and the representation of single quotes, double
quotes, and ellipses, as described in Section~5.1.

Appendices~C and D give details relating to our work on predicting
sentence breaks in the absence of white space described in
Chapter~10. Appendix~C gives the details of the heuristics,
implemented as regular expressions, that we use to identify classes of
embedded punctuation marks that should not be considered candidates
for missing spaces. Appendix~D gives the criteria we use for
classifying examples not ruled out by these heuristics as likely a
sentence break, likely not a sentence break, or ``don't care.''

Appendix~E gives more detailed analyses than those presented in
Chapter~11 of the behavior of our sentence break predictors on our
test data. Finally, in Appendix~F, we provide resources for extracting
from standard LDC releases, and sentence-break annotating, the data we
use from the Penn Treebank WSJ corpus, the Satz-3 WSJ corpus, and the
English Web Treebank corpus.

\chapter{Previous work}

A variety of approaches have been taken to sentence break prediction,
including hand-constructed rule-based systems, supervised
statistical/machine-learning-based systems, and at least one largely
unsupervised statistical system. We will not review all this preceding
work, but we note that Read et al.\ (2012) contains a fairly
comprehensive survey. Four previous studies, however, deserve
particular mention.

\section{Riley (1989)}

In one of the earliest published empirical studies of sentence break
prediction, Riley (1989) trained a classification-tree-based predictor
using the million-word tagged Brown corpus as annotated data to
predict which periods in a text signal the end of a sentence. Among
the features used by Riley are:
\begin{itemize}
\item
  the probability of a paragraph break after a particular period,
  given the period and the word occurring before the period
\item
  the probability of a paragraph break after a particular period,
  given the period and the word occurring after the period
\end{itemize}
These probabilities were estimated from 25 million words of
Associated Press news.

Our approach effectively generalizes Riley's two paragraph break
probability features by estimating the probability of a paragraph
break at a certain position in a text, given several words of context
both before and after the break candidate position. Our probability
estimate is based on N-gram statistics for multiple text sequences
overlapping the candidate position. The particular statistics used by
Riley to compute his estimates are sometimes combined into our overall
probability estimate, in cases where reliable statistics are not
available for longer N-grams.

The clearest advantage our approach has over simply using Riley's
paragraph break probability features is that we can include statistics
for N-grams spanning both sides of a break candidate position. For
instance, consider trying to predict whether there is a sentence break
following \verb|U.S.| in the sequences \verb|U.S. The| (very likely a
sentence break) and \verb|U.S. Government| (very likely not a sentence
break). In the LDC English Gigaword Fifth Edition corpus (Parker et
al., 2011), the probabilities of a paragraph break, estimated by
relative frequencies, for some relevant sequences are
\begin{center}
  \begin{tabular}{ll}
    0.005858035 & \verb|U.S.| $\|$ \\
    0.701347585 & \verb|.| $\|$ \verb|The| \\
    0.716562857 & \verb|.| $\|$ \verb|Government| \\
    0.626943005 & \verb|U.S.| $\|$ \verb|The| \\
    0.003553976 & \verb|U.S.| $\|$ \verb|Government| \\
    \end{tabular}
\end{center}
with $\|$ indicating the possible paragraph break position.

With Riley's paragraph break probability features, only the first
three of these probability estimates would be used to predict whether
there is a sentence break in these two examples. They share the same
token before the sentence break candidate, and the two tokens after
the sentence break candidate have almost the same probability of
signaling a paragraph break. In fact, \verb|Government| is slightly
more likely than \verb|The| to signal a paragraph break; hence, these
signals are not going to help predict a sentence break for
\verb|U.S. The| while predicting no sentence break for
\verb|U.S. Government|.

If we compare the two probability estimates that incorporate the
tokens on both sides of the sentence break candidate, however, the
picture changes dramatically. Using these estimates, \verb|U.S. The| is
predicted to be 176.4 times as likely as \verb|U.S. Government| to signal
a paragraph break, and hence much more likely to signal a sentence
break as well. In Chapter~5 we show how to combine statistics for
N-grams before, after, and spanning a sentence break candidate into a
single estimate for the probability of a paragraph break at a certain
position in a text.

\section{Palmer and Hearst (1997)}

Palmer and Hearst (1997) describe a system to predict sentence breaks
signaled by periods, question marks, or exclamation
points.\footnote{However, their principal English test set contained
  no exclamation points, because they had been systematically replaced
  by right square brackets. See Appendix~A.3 for details.}
Their system was based on a neural network classifier that was highly
accurate for its time, but is perhaps of less interest in light of
more recent work. Its main relevance to us is that it introduced what
has become a standard test set, which we refer to as the Satz WSJ
corpus. This test set has subsequently been used by Reynar and
Ratnaparkhi (1997), Kiss and Strunk (2006), and Gillick (2009), among
others, and we also use it as our in-domain test set.

\section{Kiss and Strunk (2006)}

As we will see in Chapters~5 and 6, with the right choice of
threshold, a well-trained estimate of paragraph break probability by
itself can be an accurate predictor of sentence breaks, but not as
accurate on well-edited in-domain text as a classifier trained
directly on a sufficient amount of hand-annotated text. The pure
paragraph break probability model, however, has the advantage of being
almost unsupervised in terms of data hand-annotated with sentence
breaks, requiring only a small amount of such data to optimize two
hyperparameters. This makes it a plausible approach to use in
low-resource situations where obtaining a large amount of
hand-annotated data may not be practical.

There is one other previous approach that is also almost unsupervised
in this sense, that of Kiss and Strunk (2006). In their paper, the
authors describe a two-stage process for determining the location of
sentence breaks in a test corpus. Like Riley (1989), Kiss and Strunk
attempt only to identify sentence breaks signaled by periods.

In the first stage of Kiss and Strunk's method, predictions are made
about which word types function as abbreviations followed by a period,
using a heuristic formula based partly on a log-likelihood ratio
measuring the strength of association between the word type and the
following period, as measured on the test corpus itself. The rationale
for this is the observation that the more strongly a word-type
associates with a following period, the more likely it is to be an
abbreviation. All single periods not following words predicted to be
abbreviations are tentatively labeled as sentence breaks.

In the second stage, statistics are collected from the corpus based on
the tentatively labeled sentence breaks, other than those following
single letters or numeral sequences, which the first stage tends to
make mistakes on. These statistics are used by a number of additional
heuristics to relabel certain single letters followed by periods
(initials) and numeral sequences followed by periods (ordinal numbers,
in some languages) as non-sentence-breaks. Other heuristics are used
to decide which ellipses and abbreviations with periods are also
sentence breaks, based on the following token and the statistics
collected from the sentence breaks labeled in the first stage.

The exact forms of the heuristics and three numerical thresholds were
tuned on a 10 MB development corpus of English text from The Wall
Street Journal. While Kiss and Strunk do not say that they hand
annotated any of this data, they do say that they determined the exact
form of the heuristics and the values of the thresholds by ``manual
inspection'' of the results on this development corpus.

For our sentence break predictor based only on paragraph break
probability estimates, our use of annotated training data to tune the
exact form of the estimate and two numerical hyperparameters seems
quite comparable to Kiss and Strunk's manual inspection of results on
their development corpus. Our use of annotated data is perhaps just
more systematic. Our paragraph-break-probability-based predictor,
however, is much more accurate when measured on Kiss and Strunk's WSJ
test set, making only about 32\% as many errors, when tested under
comparable conditions.

\section{Gillick (2009)}

In previous work on sentence break prediction, our main point of
comparison will be Gillick (2009), whose approach is simple to
understand, and which has the highest reported accuracy on the widely
shared Satz WSJ corpus, a result that was independently verified
by Read et al.\ (2012). Gillick carried out supervised training of
binary classifiers of punctuation marks that might signal the end of a
sentence. He used eight feature templates that look at the token to
the left and the token to the right of the punctuation mark to be
classified. Gillick trained his classifier on Wall Street Jounal (WSJ)
text from the Penn Treebank (Marcus et al., 1993), to produce a
weighted linear model, using either a Naive Bayes approach, or a
linear support-vector-machine (SVM) classifier, trained with the SVM
Light package (Joachims, 1999). Gillick performed in-domain testing on
the Satz WSJ corpus, as well as out-of-domain testing with text from
the works of Edgar Allen Poe and the Brown Corpus.

Gillick's feature templates seem well-specified, but they are defined
in terms of a tokenization that is, unfortunately, not completely
specified. Thus we were not able to exactly replicate Gillick's
classifiers for our experiments. We did, however, obtain Gillick's
in-domain test set, and created our own linear SVM classifier, using
different feature templates, that produced similar accuracy to
Gillick's SVM on that test set.

Like Riley (1989) and Kiss and Strunk (2006), Gillick trained and
evaluated his classifiers only on recognizing sentences breaks
signaled by periods. In our work, we consider sentence breaks signaled
by periods, question marks, and exclamation points, but in addition,
we also measure errors only on periods to compare directly with
Gillick and Kiss and Strunk.

\chapter{Data}

In the research reported here, we make use of four corpora. In order
to maximize the applicability of our work, we focus on applying
sentence break prediction to texts in raw, untokenized form. All four
of our corpora were available to us in a relatively raw form, but each
corpus had somewhat different formatting and markup, so
corpus-dependent preprocessing was needed to reduce all the corpora to
a consistent plain text representation.

For training paragraph break probability estimates, we use the LDC
English Gigaword Fifth Edition corpus (Parker et al., 2011). This
corpus consists of almost ten million documents containing over four
billion words, from seven news wire services, in the form of printable
ASCII and white space with SGML markup. The sources for this data are
Agence France-Presse, English Service; Associated Press Worldstream,
English Service; Central News Agency of Taiwan, English Service; Los
Angeles Times/Washington Post Newswire Service; Washington
Post/Bloomberg Newswire Service; New York Times Newswire Service; and
Xinhua News Agency, English Service.  This corpus was prepared by
extracting all text between ``begin paragraph'' and ``end paragraph''
tags and replacing SGML tags representing special characters with
plain text ASCII equivalents or stand-ins.

For directly supervised training, tuning, and development testing of
sentence break predictors, we use the Wall Street Journal portion of
the Penn Treebank (Marcus et al., 1993). We omit sections 03--06 of
this corpus because they overlap with the Satz WSJ corpus that we
use as our in-domain test set.\footnote{Kiss and Strunk (2006) and
Gillick (2009) mistakenly assert that the Satz WSJ corpus
consists only of sections 03--06 of the Penn Treebank WSJ corpus, but
it in fact contains much additional Wall Street Journal text not found
in the Penn Treebank.}  Of the remaining sections, we use 00--02 and
07--21 for training (approximately 733,000 words) and 22--24 for
hyperparameter tuning and development testing (approximately 110,000
words). We use the untokenized, raw form of the WSJ corpus included in
the Treebank-2 release (Marcus et al., 1995). Sentence breaks were
annotated by aligning this version of the corpus to a tokenized,
sentence-broken version of the corpus extracted from the Treebank-2
parse trees.

For in-domain testing of sentence break prediction, we use the Wall
Street Journal text constituting the Satz WSJ corpus, which
consists of more than 420,000 words of text extracted from the ACL/DCI
corpus (ACL/DCI, 1993) and sentence-break-annotated by Palmer and
Hearst (1997). We also obtained a second version of this corpus in
tokenized and sentence-broken form that was used as an in-domain test
set by Gillick (2009), and which we believe is also the version used
by Kiss and Strunk (2006).\footnote{The Satz WSJ corpus was also
used by Reynar and Ratnaparkhi (1997), but we do not know whether
their version corresponds exactly to either of the versions we
obtained.} When we compared these two versions of the corpus, we found
many differences, over and above differences in markup and
tokenization, including clear sentence break annotation errors in the
original version of the corpus, differences in sentence break
annotation that seemed to be due to how the task is interpreted (see
Section~4.1), sentence break differences in cases where it is unclear
whether to say there is a sentence break or not, and minor differences
in the text itself.

In light of the differences between the two versions of the the Satz
WSJ corpus and a desire both to report results that would be
comparable to previous work and also to report on what we consider to
be the most accurate representation and annotation of the original
text, we decided to report results on three verions:
\begin{itemize}
\item
  Satz-1, corresponding to Palmer and Hearst's version of the corpus.
\item
  Satz-2, corresponding to Gillick's version of the corpus.
\item
  Satz-3, reconciling the differences between Satz-1 and Satz-2.
\end{itemize}

For out-of-domain testing of sentence break prediction, we use the LDC
English Web Treebank corpus (Bies et al., 2012), consisting of over
250,000 words of weblogs, newsgroups, email, reviews, and
question-answers. This corpus is provided in three forms: raw,
sentence-broken and tokenized, and parsed. We again based our test
corpus on the raw form, aligning it with the sentence-broken tokenized
form to determine the position of sentence breaks in the raw text.

In Appendix~A, we give details of the preprocessing we carry out on
each of these corpora. In Appendix~F, we provide resources for
extracting from standard LDC releases, and sentence-break annotating,
the data we use from the Penn Treebank WSJ corpus, the Satz-3 WSJ
corpus, and the English Web Treebank corpus.

\chapter{Problematic issues}

In the simplest case, an English text consists of a sequence of
paragraphs, and each paragraph is a sequence of sentences, with each
sentence begining with a capitalized word and ending with a period,
question mark, or exclamation point, followed by a space. In this
case, segmenting a text into sentences requires only identifying which
periods, question marks, and exclamation points signal the end of a
sentence, and inserting a sentence break after all that do so.

Real text, however, is much more complex and messy than this. In this
chapter we discuss a number of issues that arise when text does not
fit the simple structure presented above, and we explain how we deal
with those issues.

\section{Classifying punctuation marks, or segmenting text?}

Examination of corpora annotated for sentence break prediction
suggests two different ideas of what the goal of sentence break
prediction should be:
\begin{itemize}
\item
  Determining which punctuation marks signal the end of a sentence
\item
  Segmenting text into individual sentences and other sentence-like
  units
\end{itemize}

In our work, we take the primary goal of sentence break prediction to
be segmenting text into individual sentences and other sentence-like
units. We believe this is the most pragmatic approach to sentence
break prediction, because it addresses its primary use: dividing an
extended text into smaller segments that can be independently
processed for certain purposes, such as parsing. Under either
interpretation, sentence break prediction is a classification problem;
in one case, classifying punctuation marks as signaling sentence
breaks, and in the other case, classifying text positions as being
sentence breaks.

In simple cases, indentifying punctuation marks that signal the end of
a sentence and breaking text into sentences are tightly coupled; the
sentence break immediately follows the punctuation mark. Not all text
is this simple, however, and in some cases the relation between the
positions of sentence breaks and the punctuation marks that signal the
end of a sentence is more complex.

\subsection{Sentence-closing characters}

One case in which determining that a punctuation mark signals the end
of a sentence does not by itself solve the problem of where to insert
the corresponding sentence break occurs when a sentence-end-signaling
period, question mark, or exclamation point is followed by closing
quotation marks, closing brackets\footnote{We use the term ``bracket''
  to include parentheses, curly braces, and square brackets.}, or
both. If the task is viewed as being text segmentation, closing
quotation marks or brackets clearly belong with the sentence that is
ending, rather than with the following sentence, so the sentence break
should be placed after closing quotation marks and brackets, before
the following sentence begins.

In the version of the Satz corpus used by Gillick (2009) and Kiss and
Strunk (2006), however, sentence break labels are always attached
directly to periods, question marks, and exclamation points, and no
attempt is made to identify the exact location of the sentence break.
Furthermore, both opening and closing double quotes are indicated by
the ASCII double quote character (\verb|"|), and tokenization results
in space characters on both sides of all double quotes. Thus in this
version of the Satz corpus, it is often impossible to tell using only
local information whether a particular double quote opens or closes a
quotation, and therefore, exactly where to place the sentence break
when a sentence-ending punctuation mark is followed by a double
quote. Using this version of the Satz corpus implicitly makes the task
that of classifying punctuation marks rather than segmenting text.

Under standard orthographic conventions (in untokenized text), the
actual sentence break following a sentence-end-signaling period,
question mark, or exclamation point is normally at the next white
space, which may follow other sentence-ending punctuation characters,
especially closing quotation marks or closing brackets. Since our
sentence break predictors are trained to place the sentence break in
this spot, in our Satz-2 test corpus, which in all other respects is
an untokenized version of the test corpus used by Gillick, we move the
sentence break tags past any such sentence-closing punctuation.

\subsection{Sentences embedded in longer sentences}

A second situation in which viewing sentence break prediction as
identifying sentence-ending punctuation marks or as segmenting text
into sentences can lead to different results occurs when one sentence
appears within a longer sentence. This often happens with direct
quotations, titles, and parentheticals, as illustrated by the
following examples from our Penn Treebank
training data:
\begin{itemize}
\item
  \verb|"Remember Pinocchio?" says a female voice.|
\item
  \verb|His recent speech, provocatively titled "Blacks? Animals?|
  \verb|Homosexuals? What is a Minority?" caused an uproar when its|
  \verb|title leaked out.|
\item
  \verb|I was in the avenues, on the third floor of an old building,|
  \verb|and except for my heart (Beat, BEAT!) I'm OK.|
\end{itemize}

The embedded question marks and exclamation point in these examples
might be considered to signal the end of a sentence, as similar cases
typically are annotated in Gillick's version of the Satz corpus. In
the Penn Treebank, however, none of them are treated as ending
sentences, presumably because segmenting the text at these points
would result in sentence fragments on one side or both of the dividing
point. In some cases, introducing such dividing points can result in
``sentences'' that consist of only punctuation characters, of which
there are three examples in the annotation of the Satz corpus used by
Gillick.

Except when directly comparing results with Gillick, we follow the
implicit Penn Treebank standard of only annotating a sentence break
for the outermost sentence, when embedded sentences occur. The one
situation that might be considered an exception to this generalization
occurs when a direct quotation consists of multiple sentences,
embedded in a clause that attributes the quotation to a speaker, such
as
\begin{quote}
\verb|A Lorillard spokewoman said, "This is an old story.| $\|$ \\
\verb|We're talking about years ago before anyone heard of| \\
\verb|asbestos having any questionable properties.| $\|$ \verb|There is| \\
\verb|no asbestos in our products now."| $\|$
\end{quote}

In this example of a multi-sentence direct quotation, $\|$
indicates the sentence breaks annotated in the Penn Treebank, even
though the entire quotation can be considered embedded in the clause
that begins \verb|A Lorillard spokewoman said,|. We follow this same
convention of breaking apart multi-sentence quotations in any
adjustments we make to the annotations we report on.

\section{What should count as a sentence break?}

So far, we have focused on sentences ending in periods, question
marks, or exclamation points. To the best of our knowledge virtually
all previous work on predicting sentence breaks addresses only those
signaled by these punctuation marks, and in some cases, only periods
are addressed. The survey carried out by Read et al.\ (2012), is a
partial exception to this generalization, as they evaluate nine
publicly-available sentence break prediction systems on data sets
annotated with a broader set of sentence breaks, but there is no
indication that any of the systems evaluated consider the possibility
of sentence breaks not signaled by a period, question mark, or
exclamation point.\footnote{Ellipses also sometimes signal a sentence
  break, but in standard test sets ellipses are represented as a
  series of periods, so they become a subcase of sentence breaks
  signaled by periods.}

Real texts often contain syntactically independent segments that are
not complete, grammatical sentences, and are not punctuated with
periods, question marks or, exclamation points, but which we might
still wish to separate from surrounding sentences. The annotators of
the LDC English Web Treebank corpus coined the term ``sentence unit''
to include such segments along with complete, grammatical
sentences. Laury et al.\ (2011) give a partial list of types of
examples to be treated as separate sentence units in the annotation of
the English Web Treebank. These include titles/headings/headlines,
datelines and similar lines for author/place/source information,
epistular salutations, epistular valedictions and signatures, as well
as postal addresses and other blocks of contact information.

In the rest of this section, we consider in what situations data
annotators have inserted sentence breaks in the absence of periods,
question marks, and exclamation points, and we assess the feasibility
of training statistical models to predict these breaks.

\subsection{Penn Treebank WSJ corpus}

The annotators of the Penn Treebank definitely thought that not all
sentence breaks are signaled by periods, question marks, or
exclamation points. If we put aside paragraph-final sentences, where
the paragraph break itself might be the most important sentence break
signal, we find that in our corrected Penn Treebank training set, 202
out of 20,054 paragraph-internal sentence breaks are not signaled by a
period, question mark, or exclamation point. All but five of these are
signaled either by a colon (\verb|:|) followed by a space or by a dash
(\verb|--|)\footnote{In the Penn Treebank, dashes are usually
  represented by a pair of hyphens, although occasionally one hyphen
  or three hyphens are used.} surrounded by spaces.

Colons and dashes are principally treated as signaling sentence breaks
in the Penn Treebank WSJ corpus when they follow a word or phrase that
introduces, but is syntactically independent from the text that
follows it. Consider the following examples, where $\|$ indicates
positions treated as sentence breaks in the Penn Treebank:
\begin{itemize}
\item
  \verb|SWITCHING TO THE DEFENSE:| $\|$ \verb|A former member of the| \\
  \verb|prosecution team in the Iran/Contra affair joined the| \\
  \verb|Chicago firm of Mayer, Brown & Platt.| $\|$
\item
  \verb|ENERGY:| $\|$ \verb|Petroleum futures were generally higher with| \\
  \verb|heating oil leading the way.| $\|$
\item
  \verb|BRIEFS:| $\|$ \verb|Guarana Antarctica, a Brazilian soft drink, is| \\
  \verb|brought to the U.S. by Amcap, Chevy Chase, Md.| $\|$
\item
  \verb|International Business Machines Corp. --| $\|$ \verb|$750 million| \\
  \verb|of 8 3/8% debentures due Nov. 1, 2019, priced at 99 to| \\
  \verb|yield 8.467%.| $\|$
\item
  \verb|WASHINGTON --| $\|$ \verb|United Technologies Corp. won an $18| \\
  \verb|million Army contract for helicopter modifications and| \\
  \verb|spare parts.| $\|$
\end{itemize}

\subsection{Satz WSJ corpus}

When we analyzed our test sets, we found that they were annotated very
differently from the Penn Treebank regarding colons and
dashes. Although the Satz WSJ corpus is drawn from the same WSJ data
source as the Penn Treebank, it has no paragraph-internal sentence
breaks annotated following colons or dashes, even though many cases
were of exactly the same form as examples in the Penn Treebank WSJ
corpus that were annotated as sentence breaks. Examples from from the
Satz WSJ corpus include:
\begin{itemize}
\item
  \verb|Other issues: Rep. Stephen L. Neal has submitted a joint|
  \verb|resolution that would instruct the Fed to stabilize prices|
  \verb|-- and keep them stable.| $\|$
\item
  \verb|BRIEFS: Airline pilot Charles J. Collins of Boulder, Colo.,|
  \verb|had a credibility problem in Tax Court that cost him all his|
  \verb|claims.| $\|$
\item
  \verb|PRECIOUS METALS: Futures prices showed modest changes in light|
  \verb|trading volume.| $\|$
\item
  \verb|Beneficial Corp. -- $248.3 million of securities backed by|
  \verb|home-equity loans through Merrill Lynch Capital Markets.| $\|$
\item
  \verb|AMERICAN FEDERAL SAVINGS BANK OF DUVAL COUNTY (Jacksonville,|
  \verb|Fla.) -- Charles Bates, president, chief executive and chief|
  \verb|operating officer will resign from these positions and the|
  \verb|board effective Oct. 31.| $\|$
\end{itemize}
In all, only three out of 11,176 paragraph-internal sentence breaks in
the Satz-3 corpus are not signaled by a period, question mark, or
exclamation point.

\subsection{English Web Treebank corpus}

Our observations about the differences between the Penn Treebank WSJ
corpus and the Satz WSJ corpus are purely empirical; there are no
documented annotation standards for sentence breaking for either
corpus.  When we turn to our out-of-domain test set based on the LDC
English Web Treebank corpus, there is such a document by Laury et
al.\ (2011), distributed by LDC along with the data. This document is
not comprehensive, and dashes are not addressed, but colons are. It
turns out that whether a colon can be taken to signal a sentence break
depends on whether it immediately precedes a hard\footnote{By a
``hard'' line break, we mean one inserted by human author or editor of
the text, for reasons other than keeping text from exceeding a desired
line length.}  line break. Within a line ``colons do not count as
final punctuation.'' By default, however, line breaks are treated as
sentence breaks, unless ``there is syntactic structure to preserve''
that spans the line break. Hence, if a colon appears at the end of a
line in the original text, and there are no syntactic dependencies
between the text before and the text after the colon, the colon will
be treated as the final punctuation of a sentence.

The results of applying these standards can be seen in the following
examples from LDC English Web Treebank:
\begin{itemize}
\item  
  \verb|For more info on lo-fi photography, check put my website:| $\|$ \\
  \verb|http://dianacamera.com| $\|$
\item
  \verb|I don't know about birding in that part of the world but you| \\
  \verb|might possibly find help at this site:| \\
  \verb|http://www.wildlifeofpakistan.com/PakistanBirdClub/index.html|~$\|$
\end{itemize}
These two examples have a very similar structure, but in the first
there is a sentence break following the colon that precedes the URL,
and in the second, there is no sentence break following the colon that
precedes the URL. The first example is line-broken as in the original
text; there are no syntactic dependencies between the two lines, hence
the colon ending the first line is followed by a sentence break. In
the second example, the original text was all one line, hence there is
no sentence break between the colon and the URL.

In our Penn Treebank WSJ training set and Satz WSJ test set, virtually
all paragraph-internal sentence breaks are signaled by periods,
question marks, or exclamation points (sometimes followed by closing
quotes or closing brackets), and in the case of the Penn Treebank,
dashes or colons. There are only five exceptions to this in the Penn
Treebank training set, out of 20,054 paragraph-internal sentence
breaks, and only three in the Satz test set, out of 11,176
paragraph-internal sentence breaks. Most of these exceptions are
problematical in some way, representing either possible sentence break
annotation errors, or paragraph break annotation errors making them
not truly paragraph internal.

The LDC English Web Treebank is entirely different. In this corpus,
1,057 out of 10,127 total paragraph-internal sentence breaks are not
signaled by standard sentence-final punctuation. Of these, 779 have no
sentence-final punctuation at all. Examples suggest that in this
user-generated text, line breaks in the original text sometimes signal
sentence breaks without any sentence-final punctuation, for instance:
\begin{quote}
\verb|an island called jejudo/jeju island| $\|$ \\
\verb|its really amazing there, itll blow your mind| $\|$ \\
\verb|it's almost like paradise there| $\|$
\end{quote}
Many of the examples of ``sentence units'' delimited only by line
breaks are nonsentential items in a list, with one item per line:
\begin{quote}
\verb|- Big Mac| $\|$ \\
\verb|- 10 piece Chicken McNuggets| $\|$ \\
\verb|- Medium Fountain Drink| $\|$ \\
\verb|- McCAFE Drink| $\|$ \\
\verb|- Filet-O-Fish| $\|$ \\
\verb|- Hash Browns| $\|$ \\
\verb|- Egg McMuffin, Sausage| $\|$ \\
\verb|- Fruit & Maple Oatmeal| $\|$
\end{quote}
According to the annotation guidelines by Laury et al.\ (2011),
``postal addresses and other blocks of contact information are joined
into sentence units,'' but many of the annotations violate this, such as:
\begin{quote}
\verb|The Breakfast Club & Grill| $\|$ \\
\verb|1381 W. Hubbard| $\|$ \\
\verb|Chicago, IL 60622| $\|$ \\
\verb|312-666-2372 Tel| $\|$ \\
\verb|Email Us| $\|$
\end{quote}
Finally, in a few cases there are clear annotation errors created by
marking individual lines as separate sentences even though sentences
obviously span multiple lines:
\begin{quote}
\verb|You are busy, you don't have to sit face to face to try|~$\|$ \\
\verb|to make convo, it's short, so if things go great, you can|~$\|$ \\
\verb|extend, and if you want to end it, you just say bye,|~$\|$ \\
\verb|gotta go.|~$\|$
\end{quote}

Of the remaining paragraph-internal sentence breaks with nonstandard
sentence-final punctuation, many occur after ``emoticons'':
\begin{itemize}
\item
  \verb|And no, I do not work for them. :)|~$\|$
\item
  \verb|I didn't get to read the answers lol :(|~$\|$
\item
  \verb|The motel is very well maintained, and the managers| \\
  \verb|are so accomodating, it's kind of like visiting family| \\
  \verb|each year! ;-)|~$\|$
\end{itemize}

Also, when lines of nonalphanumeric characters are used for
formatting, these are often annotated to separate them from the text
that comes before and after, resulting in ``sentence units'' that
contain nothing other than nonalphanumeric characters, as in this
example:
\begin{quote}
\verb|====================================================|~$\|$ \\
\verb|(Harrison) Kyle Jones| \\
\verb|Technical Solutions Engineer|~$\|$ \\
\verb|====================================================|~$\|$ \\
\verb|Radianz| \\
\verb|1251 Avenue of the Americas| \\
\verb|7th Floor| \\
\verb|New York, NY 10016| \\
\verb|USA| \\
\verb|Phone:  +1 (212) 899-4425| \\
\verb|Fax:      +1 (212) 899-4310| \\
\verb|Cell:      +1 (917) 859-7187| \\
\verb|Email:   kyle.jones@radianz.com| $\|$ \\
\verb|====================================================|~$\|$ \\
\verb|See our web page at"     http://www.radianz.com|~$\|$ \\
\verb|====================================================|~$\|$
\end{quote}

\subsection{Which sentence breaks we test on}

To summarize, according to the annotations in our training set from
the Penn Treebank WSJ corpus, the only punctuation marks that signal
sentence breaks in significant numbers are periods, question marks,
exclamation points, colons, and dashes. In our in-domain test set from
the Satz WSJ corpus and our out-of-domain test set from the LDC
English Web Treebank corpus, the annotation of segments ending in
colons or dashes is very different from the Penn Treebank. Also, in
the English Web Treebank, there are many segments for which the only
sentence break signal seems to be a hard line break. In our training
data, however, we do not have access to original line breaks, since in
the ``raw'' version of the Penn Treebank, line breaks are used to
annotate an errorful guess at sentence breaks. Finally, the English
Web Treebank also includes other sentence breaks that follow sequences
of characters not found in our Penn Treebank training data.

We would like to train models to try to predict sentence breaks beyond
those signaled by periods, question marks, and exclamation
points. Given the inconsistencies in annotations between our training
set and our test sets, the inconsistencies in annotation between our
two test sets, and the lack of information in the training data needed
to predict many types of examples in the out-of-domain test set,
predicting additional sentence breaks just does not seem possible.
Thus we end up following previous work in the field, trying to predict
only sentence breaks signaled by periods, question marks, and
exclamation points. We should keep in mind, however, that accurately
predicting sentence breaks in the presence of periods, question marks,
and exclamation points is not a complete solution to the problem of
sentence break prediction.

\section{How should sentence break prediction accuracy be measured?}

\subsection{The difficulty of counting sentence break candidates}

Comparing results across different studies of sentence break
prediction turns out to be surprisingly difficult, even when the same
test set is used. Typically, an error rate or an accuracy rate (1 -
error rate) is given as the primary metric, but these are often
incomparable. Error rate is simply defined as the number of errors
divided by the number of sentence break candidates, but even when
counting only periods, question marks, and exclamation points,
researchers disagree about what constitutes a sentence break
candidate. In the annotated version of the Satz WSJ test set used by
Gillick (2009), 27,207 positions in the text are marked as sentence
break candidates. Palmer and Hearst's (1997) count of sentence break
candidates for nominally the same corpus is 27,294, and the count
given by Reynar and Ratnaparkhi (1997) is 32,173.

It appears that Reynar and Ratnaparkhi's count of sentence break
candidates for the Satz WSJ test set includes literally every period,
question mark, and exclamation point, even periods used as decimal
points preceded and followed by digits, and all periods in
representations of ellipses, including nonfinal ones (e.g., the first
two periods in ``\verb|...|'').\footnote{Our actual counts of all periods,
  question marks, and exclamation points were 32,170 for Palmer and
  Hearst's version of the corpus, and 32,179 for Gillick's version, so
  we assume Reynar and Ratnaparkhi's version differs slightly from
  either of those.}

Palmer and Hearst mention that they exclude periods in ``numbers
containing periods acting as decimal points,'' but applying only this
exclusion to their version of the Satz WSJ test set leaves 28,805
sentence break candidates. If we also exclude nonfinal periods in
representations of ellipses, we still have 28,586 sentence break
candidates, 292 more than the 27,294 Palmer and Hearst report in their
paper. Although sentence breaks are annotated in their version of the
corpus, other candidates are not marked in any special way, so we have
no way to know exactly what additional criteria they used to get to
their count of sentence break candidates.

In his paper, Gillick says that only punctuation marks ``followed by
white space or punctuation'' are considered as sentence break
candidates. Applying that filter to the version of the Satz WSJ test set
used by Gillick yields 27,455 sentence break candidates. Also
excluding nonfinal periods in representations of ellipses leaves
27,241 sentence break candidates, which is 34 more than the number of
positions marked as sentence break candidates in Gillick's version of
the corpus. Since all sentence break candidates are marked in this
version, we performed a detailed examimation and discovered that the 33
instances of a period followed by a hyphen (e.g., \verb|U.S.-owned|) are
not counted as sentence break candidates. Three other idiosyncratic
examples account for the exact set of sentence break candidates
annotated in Gillick's version of the corpus:
\begin{itemize}
\item
  One exclamation point surrounded by parentheses is not marked as a
  sentence break candidate:
  \begin{quote}
    \verb|Alix's pursuit of the serial killer's mother (!),|
  \end{quote}
\item
  In one ellipsis represented by four periods, the position after the
  third period is marked as a sentence break candidate:
  \begin{quote}
    \verb|we didn't feel it would be worth it...|$\|$\verb|.trying to get|
    \verb|misdemeanor convictions|\footnote{This is the only example
      in the corpus in which an ellipsis is not followed by white space,
      suggesting the placement of the sentence-break-candidate mark
      may be due to a bug in code for recognizing the end of an
      ellipsis.} 
  \end{quote}
\item
  One question mark internal to a sentence is not marked as a sentence
  break candidate:
  \begin{quote}
    \verb|In the ongoing turf battle over who has the right to|
    \verb|regulate advertising -- the federal government or the|
    \verb|states? -- score one for the federal government.|
  \end{quote}
\end{itemize}

Our own preferred definition of a sentence break candidate for
well-edited, conventionally formatted text includes:
\begin{itemize}
\item
  All positions that
  \begin{itemize}
  \item
    immediately precede white space that is not immediately followed
    by a period,\footnote{This rules out nonfinal periods in an
    ellipsis that includes spaces between the periods.} and
  \item
    immediately follow
    \begin{itemize}
    \item
      a period, question mark, or exclamation point and
    \item
      any optional following combination of closing brackets and
      closing quotes
    \end{itemize}
  \end{itemize}
\item
  All positions that
  \begin{itemize}
  \item
    immediately preceed an alphanumeric (i.e.,
    non-punctuation) character and
  \item
    immediately follow the final period of an
    ellipsis
  \end{itemize}
\end{itemize}
This definition rules out many easy cases such as the period ending an
abbreviation that is followed by a comma, which in well-edited,
conventionally formatted text is never a sentence break. By this
standard, there are 25,781 sentence break candidates in Palmer and
Hearst's version of the Satz WSJ test set and 25,796 in
Gillick's.\footnote{Most of the difference between these counts is due
  to the systematic replacement of exclamation points with right
  square brackets (]) in Palmer and Hearst's version of the
corpus. See Appendix~A.3}

In most publications describing previous work, there is enough
information given to restate the error rates given in terms of a
single consistent definition of sentence break candidate, such as the
one we prefer. However, it becomes exceedingly difficult to give a
reasonable but simple definition of sentence break candidate once we
consider poorly-edited or non-conventionally-formatted text, in which
sentence-ending punctuation may be followed by the start of the next
sentence without any white space (See Chapter~10). In the English Web
Treebank corpus, we count 106 cases where a sentence break occurs
without any white space. Of these, 32 follow ellipses, which are
easy to identify as sentence break candidates, but 74 times the author
of the text has simply failed to insert the space that would normally
follow the end of a sentence.

When we start considering periods, question marks, and exclamation
points not immediately followed by other punctuation or white space as
sentence break candidates, we find a host of special cases we might
like to exclude. We have already mentioned some of these: decimal
points followed by by digits, nonfinal periods in ellipses, and
hyphenated expressions incorporating abbreviations ending in
periods. Other situations involving periods not followed by
punctuation or white space that we might wish to rule out as sentence
break candidates include internal periods in inflected forms of
abbreviations, strings of multiple initials, email addresses, and
computer filenames, domain names, URLs, and related expressions; but
sometimes these are not easy to recognize.

Taking all this into account, it seems that even if we consider only
sentence breaks signaled by periods, question marks, and exclamation
points, there is no simple criterion we can use to define which ones
should be seriously considered as sentence break candidates; and
therefore we cannot easily define an error rate as simply the number
of errors divided by the number of candidates.

\subsection{Evaluating sentence break prediction using $\rm{F_{1}}$ score}

The difficulty of giving a simple, reasonable definition of a sentence
break candidate suggests looking for an alternative measure that does
not depend on counting sentence break candidates. One such measure is
$\rm{F_{1}}$ score, originally defined as way of combining precision and recall
scores for information retrieval:
\[\rm{F_{1}} = \frac{2 \, \rm{P} \, \rm{R}}{\rm{P} + \rm{R}}\]
\[\rm{P} = \frac{\rm{TPP}}{\rm{TPP} + \rm{FPP}}\]
\[\rm{R} = \frac{\rm{TPP}}{\rm{TPP} + \rm{FNP}}\]
In these formulas, $\rm{F_{1}}$ is defined in terms of precision P and recall R,
which can be in turn defined in terms of the number of true positive
predictions TPP, false positive predictions FPP, and false negative
predictions FNP.

Since the number of true negative predictions is not required to
compute either precision or recall, $\rm{F_{1}}$ score is not
dependent on a definition of sentence break candidate. Read et
al.\ (2012) have previously noted the difficulty of giving a
reasonable definition of sentence break candidate other than every
position in the text, which would cause simple error rate or accuracy
percentages to be dominated by uninteresting, obvious negative
cases. Hence they choose precision, recall, and $\rm{F_{1}}$ score as
their prefered evaluation metrics.

We considered using $\rm{F_{1}}$ score as our primary evaluation
metric and training objective function, but decided against it when we
noticed that sometimes one sentence break predictor produces a better
$\rm{F_{1}}$ score but makes more errors than another sentence break
predictor. This happens because $\rm{F_{1}}$ score inherrently
penalizes false negative errors more than false positive errors. While
one might argue that false positives and false negatives are not
necessarily equally harmful, it seems that if there is a difference
for sentence break predictions, false positives would be more harmful
than false negatives. A false positive may cause irrecoverable errors
in downstream processing, while a false negative leaves open the
possiblity of discovering the missed sentence break later. For example,
if sentence break prediction is followed by dependency parsing and a
sentence break has been missed (false negative), the parser may be
able to discover that a text segment consists of two sentences rather
than one by labeling two words as sentence heads. On the other hand, a
mistakenly inserted sentence break (false positive) will prevent a
dependency parser from considering dependencies that span the mistaken
sentence break, which will necessarily cause parsing errors.

To see that $\rm{F_{1}}$ score penalizes false negatives more than
false positives, we can substitute the definitions of P and R into the
definition of $\rm{F_{1}}$ and simplify to give us another well-known
formula for $\rm{F_{1}}$:
\[\rm{F_{1} = \frac{2 \, TPP}{2 \, TPP + FPP + FNP}}\]
If we let PE represent the number of positive examples, we can
re-express TPP as $\rm{TPP = PE - FNP}$. Applying this substitution to
the previous formula for $\rm{F_{1}}$ and simplifying gives us
\[\rm{F_{1} = \frac{2 \, PE - 2 \, FNP}{2 \, PE + FPP - FNP}}\]

At this point, we can see that if we hold the total number of errors
constant, but replace one false negative with a false positive, the
numerator and denominator in this formula will both increase by 2. It
follows that this replacement will increase the $\rm{F_{1}}$ score
towards its maximum value of 1.0, meaning that false positive
predictions are not penalized as much as false negative
predictions. For example, if PE = 100, FNP = 25, and FPP = 75, then
$\rm{F_{1}} = 0.6000$. However, if if PE = 100, FNP = 75, and FPP = 25
(the same number of positive examples and the same number of total
errors, but more false negatives and fewer false positives), then
$\rm{F_{1}} = 0.3333$.

\subsection{Evaluating sentence break prediction using $\rm{F_{\beta}}$ score}

We conclude that $\rm{F_{1}}$ score is unsuitable for evaluating
sentence break prediction, because it inherrently penalizes false
negatives more than false positives, and we would like to retain the
possibilty of penalizing all errors equally. $\rm{F_{1}}$ score,
however, is a special case of $\rm{F_{\beta}}$ score, where $\beta$
serves as a weighting of the relative importance of recall and
precision. $\rm{F_{\beta}}$ score can be defined by the equation
\[\rm{F_{\beta} = \frac{(1+\beta^{2}) \, TPP}{(1+\beta^{2}) \, TPP +
    \beta^{2} \, FNP + FPP}}\]

By setting the value of $\beta$, we can make the $\rm{F_{\beta}}$
score change the relative penalty for false positives and false
negatives, but not in a linear way. More specifically, there is no
value of $\beta$ that always penalizes a false negative and a false
positive exactly the same amount. It turns out that for a given number
of positive examples and total errors, there is a value of $\beta$
that makes $\rm{F_{\beta}}$ insensitive how the total errors are
distributed between false positives and false negatives, but this
value of $\beta$ depends on the ratio of total errors to positive
examples. Hence there is no value of $\beta$ that always gives the
same $\rm{F_{\beta}}$ score to all sets of predictions that have the
same total number of errors on a given test set.

To see that this claim is true, we derive the following equation from
the definition of $\rm{F_{\beta}}$ score:
\[(\rm{F_{\beta}} - 1) \, (1+\beta^{2}) \, \rm{PE} = (\rm{F_{\beta}} -
(1+\beta^{2})) \, \rm{FNP} + (-\rm{F_{\beta}}) \, \rm{FPP}\] The
property we would like to maintain is that $\rm{F_{\beta}}$ should
remain constant as long as the total number of errors remains
constant. If this property holds and the total number of errors is
constant, then the left side of this equation will be constant, hence
the right side must also be constant. But the only way the right side
can remain constant as we shift the same total number of errors
between false positives and false negatives is for the coefficients of
FNP and FPP to be equal; hence, the following equation must hold:
\[\rm{F_{\beta}} -(1+\beta^{2}) = -\rm{F_{\beta}}\]
which gives us
\[\beta = \sqrt{2\rm{F_{\beta}}-1}\]
If we substitute this value for $\beta$ into the equation defining
$\rm{F_{\beta}}$ and then solve for $\rm{F_{\beta}}$, we get
\[\rm{F_{\beta}} = 1 - \frac{\rm{FPP} + \rm{FNP}}{2 \, \rm{PE}}\]
If we substitute this value of $\rm{F_{\beta}}$ into the formula above
for $\beta$, we in turn get
\[\beta = \sqrt{1 - \frac{\rm{FPP} + \rm{FNP}}{\rm{PE}}}\]
At this point, we can see that with this value of $\beta$, the value
of $\rm{F_{\beta}}$ for a given total number of errors is insensitive
to how they are distributed between false positives and false
negatives, but the value of $\beta$ needed to make this work depends
on the ratio of total errors to positive examples. Hence, there is no
single value of $\beta$ that makes the value of $\rm{F_{\beta}}$
insensitive to how the total number of errors are distributed, for
different ratios of total errors to positive examples.

\subsection{Evaluating sentence break prediction using errors per sentence break}

At this point, we have shown how difficult it is to give any simple
definition of a straightforward error rate for sentence break
prediction, unless we count every position in the text as a sentence
break candidate, which would give misleadingly low error rates due to
the preponderance of obvious negative cases. We have further shown
that using either $\rm{F_{1}}$ score or its generalization
$\rm{F_{\beta}}$ score as the primary metric for evaluating sentence
break prediction would make it impossible to give equal weight under
all conditions to false positive errors and false negative errors.

Finally, we turn to a metric inspired by the definition of word error
rate used in speech recognition: the total number of word errors
(substitutions, insertions, and deletions) divided by the number of
actual words. Analogously, the metric we will use for evaluating
sentence break prediction is total sentence break prediction errors
divided by the number of actual sentence breaks:
\[\frac{\rm{FNP} + \rm{FPP}}{\rm{PE}}\]
Since error rates are typically reported as percentages, we will scale
this measure by a factor of 100, to report total errors per 100 actual
sentence breaks:
\[100 \times \frac{\rm{FNP} + \rm{FPP}}{\rm{PE}}\]
We will indicate this scaling by using a percent sign (\%), as is done
in reports of speech recognition error rates.

One final note is that our counts of actual sentence breaks will be
counts of the actual sentence breaks we are trying to
predict. Usually, this will be sentence breaks signaled by periods,
question marks, and exclamation points; so our metric will be the
number of errors in predicting sentence breaks signaled by periods,
question marks, and exclamation points, per 100 actual sentence breaks
signaled by periods, question marks, and exclamation points. In some
cases, for comparison to previous work, we will consider only sentence
breaks signaled by periods, and report the number of errors in
predicting sentence breaks signaled by periods, per 100 actual
sentence breaks signaled by periods. When making comparisons to
previous work, whenever possible we will restate previously reported
results in terms of errors per 100 actual sentence breaks.

\chapter{Predicting sentence breaks from estimated paragraph break probabilities}

As we explained in Chapter~1, our main idea is to use an estimate of
the probability of a paragraph break as a signal for predicting a
sentence break. In our simplest sentence break predictors, if there is
a significant estimated probability of a paragraph break at a
particular text position, given the surrounding text context, then we
predict a sentence break at the position; if the probability of a
paragraph break is very low, we predict that there is no sentence
break. Formally, if $p(\verb|<P>| \, | \, t)$ is the estimated
probability of a paragraph break at a particular text position, given
the surrounding text context $t$, we predict a sentence break whenever
\[p(\verb|<P>| \, | \, t) > \beta,\]
for an estimate of $p(\verb|<P>| \, | \, t)$ based on a large
corpus of naturally occurring text, and a value of the threshold
$\beta$ optimized on a much smaller text corpus annotated with
sentence breaks.

Since paragraph breaks are easily recovered from most naturally
occurring text, we can obtain virtually unlimited amounts of data to
train a paragraph break probability estimate in a directly supervised
way. For this reason, we opt for a very simple, fast-to-estimate
model of paragraph break probability, expecting that simply using
more data to train the model will produce better results than training
a more sophisticated and expensive model on less data.

Our approach therefore uses a simple N-gram model to estimate the
probability of a hidden token, \verb|<P>| indicating a paragraph
break, given a tokenization of the surrounding text designed to
capture information relevant to the prediction. To build and apply
such an N-gram model, we proceed in several stages. First we normalize
our training text, so that the use of spaces and the representation of
single quotes, double quotes, and ellipses is as uniform as
possible. Next we tokenize the normalized training text in a way that
preserves spacing information and incorporates an abstract
representation of the use of capitalization, alphabetic characters,
and numerical characters, so that we can condition on features such as
``a five-letter capitalized word'' or ``a four digit number'' even
when we have not seen enough instances of the particular word or
particular number in question to collect reliable statistics at that
level of specificity. We then collect N-gram counts from the
normalized, tokenized training text. Finally, when presented with a
similarly normalized and tokenized text to divide into sentences, we
use the N-gram counts to estimate the probability of a paragraph break
at each sentence break candidate position in that text, and we predict
a sentence break whenever the paragraph break probability exceeds an
empirically optimized threshold. Details of these stages are presented
below.

\section{Text normalization}

We normalize both our training text and texts to be sentence-broken,
with respect to the use of spaces and the representation of single
quotes, double quotes, and ellipses. For texts to be sentence broken,
these normalizations are applied only internal to the sentence break
predictor, and the sentence breaks are projected back onto the
original text.

The normalizations we perform are motivated by differences in the
representation of ellipses, single quotes, and double quotes that we
have observed in various parts of the English Gigaword corpus, which
we use as training data for our paragraph break probability model. We
have also noticed occasional extra spaces before (or after) certain
punctuation marks that are normally not preceded (or followed) by a
space. Since our models are based on N-grams, and these N-grams
preserve information about the location of spaces, the exact character
strings, including spaces, are important both for collecting N-gram
statistics in building models and for assuring a good match to the
texts to which the models are applied. If our training data
predominantly uses one representation of ellipses or quotation marks
near sentence boundaries, and the model we learn is applied to a text
that uses a different representation, errors in predicting sentence
breaks near ellipses or quotation marks will likely result.

Some of the differences we have observed in the representation of
ellipses include whether there are spaces (\verb|. . .|) or no spaces
(\verb|...|) within the sequence of periods representing an ellipsis,
whether or not a space precedes an ellipsis, and whether or not a
space follows an ellipsis. Single quotes are at times represented by
an ordinary single quote character (\verb|'|) or a back quote
character (\verb|`|), and double quotes are at times represented by
one double quote character (\verb|"|), two single quote characters
(\verb|''|), or two back quote characters (\verb|``|).

We perform normalization using an extensive set of heuristic
regular-expression-matching substitution commands intended to achieve
the following:
\begin{itemize}
\item
  All ellipses should be represented by a sequence of periods without
  internal spaces, should not be preceded by a space, and should be
  followed by a space (e.g., \verb|abcd... efgh|) unless at the very
  end of a paragraph.
\item
  All double quotes opening quoted text should be represented by one
  double quote character, should not be followed by space, and should
  be preceded by a space or an opening bracket unless at the very
  beginning of a paragraph.
\item
  All double quotes closing quoted text should be represented by one
  double quote character, should not be preceded by space, and should
  be followed by a space, a closing bracket, or a punctuation
  character unless at the very end of a paragraph.
\item
  All single quotes opening quoted text should be represented by one
  single quote character, should not be followed by space, and should
  be preceded by a space, an opening bracket, or a double quote unless
  at the very beginning of a paragraph.
\item
  All single quotes closing quoted text should be represented by one
  single quote character, should not be preceded by space, and
  should be followed by a space, a closing bracket, a double quote, or
  another punctuation character unless at the very end of a paragraph.
\item
  No opening bracket should be followed by a space.
\item
  No period followed by a space should be preceded by a space.
\item
  No comma, semicolon, colon, question mark, exclamation point, or
  closing bracket should be preceded by a space.
\item
  No apostrophe indicating a possessive construction should be
  preceded by a space.
\end{itemize}

Our text normalization procedure does not achieve these goals
perfectly, but seems to perform well overall. Most of the errors that
it makes seem to be the result of confusing opening and closing quotation
marks, or confusing apostrophes and single quotes. More detailed
information on the text normalization process is given in Appendix~B.

\section{Text tokenization}

As mentioned above, for simplicity, our model for estimating paragraph
break probabilities is based on N-gram statistics. In order to define
an N-gram model, we need to specify what the tokens are. Our
tokenization is defined over normalized training data text as follows:
\begin{itemize}
\item
  A \verb|<P>| token is inserted at every paragraph break that is a
  sentence break candidate.
\item
  A \verb|<NO-P>| token is inserted at every other position that we
  consider to be a sentence break candidate.
\end{itemize}
Sentence break candidates are considered to be spaces or paragraph
breaks,\footnote{As noted previously, well-edited text may also have
  sentence breaks without any white space following an ellipsis, but
  our text normalization procedure ensures that every ellipsis is
  followed by white space. Our normalization procedure also
  substitutes spaces for paragraph-internal line breaks, so we do not
  have to consider line breaks as possible sentence breaks.} not at
the end of a document,\footnote{We also treat the end of any paragraph
  not immediately followed by another text paragraph as if it were the
  end of a document, and do not treat it as an ordinary paragraph
  break.} preceded by sentence-final punctuation (periods, question
marks, or exclamation points), possibly with one or more closing
quotes and/or closing brackets between the sentence-final punctuation
and the space or paragraph break.
\begin{itemize}
\item
  Every white space is treated as a token break.
\item
  Strings between white spaces containing both ``word'' characters
  (\verb|_|, \verb|0|--\verb|9|, \verb|A|--\verb|Z|,
  \verb|a|--\verb|z|) and ``non-word'' characters are broken into
  tokens that are maximal substrings containing either only word
  characters, or only non-word characters.
\item
  \verb|_| is added as a prefix to a non-word-character token not
  preceded by white space.
\item
  \verb|_| is added as a suffix to a non-word-character token not
  followed by white space.
\item
  Each word-character token (base token) is accompanied by a ``shape'' token,
  prefaced with \verb|<SH>|, in which:
  \begin{itemize}
  \item
    Each digit in the base token is replaced by \verb|9|.
  \item
    Each lower-case letter in the base token is replaced by \verb|a|.
  \item
    Each upper-case letter in the base token is replaced by \verb|A|.
  \end{itemize}
\item
  In the N-grams extracted from a tokenized text:
  \begin{itemize}
  \item
    Base tokens preceding a sentence break candidate are followed by
    their corresponding shape tokens.
  \item
    Base tokens following a sentence break candidate are preceded by
    their corresponding shape tokens.
  \end{itemize}
\end{itemize}
For example, \verb|U.S. Army|, with no paragraph break between
\verb|U.S.| and \verb|Army|, is tokenized as
\begin{quote}
  \verb|U <SH>A _._ S <SH>A _. <NO-P> <SH>Aaaa Army|.
\end{quote}

The reason for introducing shape tokens is so that if an N-gram
beginning or ending in a particular base token has not been seen in
training data often enough to be included in our model, we can back
off to a shorter N-gram while still retaining some information about
the base token, including whether it is capitalized, whether it
contains digits, and how long it is. For example, if in a text we are
trying to find sentence breaks in, the abbreviation \verb|U.S.| is
followed by a word that we never saw \verb|U.S.| followed by in
training data, we could still use N-grams ending in the shape token
for the word, e.g., the N-gram
\verb|U <SH>A _._ S <SH>A _. <NO-P> <SH>aaaaaaa|, and training data
statistics might inform us that \verb|U.S.| is not likely to mark a
paragraph break when followed by a 7-letter uncapitalized word.

This illustrates why we want a shape token placed closer to the
sentence break candidate position than its base token; following the
base token when it precedes the sentence break candidate, and
preceding the base token when it follows the sentence break
candidate. That way, when we shorten N-grams from either end, we will
always discard a base token before we discard its shape token.

Note that in practice, when we tokenize an entire text, we insert all
the shape tokens preceding their corresponding base tokens. Then as we
scan the text from left to right to extract N-grams, when we pass a
base token, we reverse the positions of the base token and its shape
token. When we encounter the next sentence break candidate position,
all the the shape tokens preceding the sentence break candidate will
follow their base tokens, and all shape tokens following the sentence
break candidates will still precede their base tokens.

This defines the tokenization of normalized training texst.
Tokenization of normalized text in which sentence breaks are to be
predicted is identical, except that all sentence break candidate
positions are marked with the token \verb|<S?>|, instead of \verb|<P>|
or \verb|<NO-P>|.

\section{Collecting N-gram counts}

We collected N-gram counts from our normalized and tokenized version of
the LDC English Gigaword Fifth Edition corpus, but only for N-grams
that contain either a \verb|<P>| token or a \verb|<NO-P>|. The way we
build our models, no other N-grams can have any effect in estimating
the probability of a paragraph break.

We decided not to count N-grams occurring within direct quotations,
because paragraph breaks occur within quotations much less frequently
than in non-quoted text. In the English Gigaword corpus, we observed
59\% of positions we considered to be sentence break candidates were
paragraph breaks when not within quoted text, but less than 10\% of
such positions within quoted text were paragraph breaks. We concluded
that the relationship between paragraph breaks and sentence breaks in
quoted text was atypical of general text and it might bias our models
if we included N-gram counts from within direct quotations.

After some initial experimentation with our training set, we decided
to limit the N-grams we collect to length 10 or less. This might seem
rather long, but recall that in our tokenization, even the short phrase
\verb|U.S. Army| has a token length of 9:
\begin{quote}
  \verb|U <SH>A _._ S <SH>A _. <NO-P> <SH>Aaaa Army|
\end{quote}

In our models, the selection of N-grams to use in predicting sentence
breaks is also subject to an empirically optimized count cutoff. The
way we use count cutoffs in our initial approach is a bit
nonstandard. This approach requires estimating the probabilities of
two versions of a string of tokens surrounding a sentence break
candidate position, one with the token \verb|<P>| at the candidate
position, and one with the token \verb|<NO-P>| at that position. We
want to be sure that these estimates are comparable, so whenever we
use an N-gram $t_{i-m} \ldots t_{i-1} \, \verb|<P>| \, t_{i+1} \ldots
t_{i+n}$ in estimating the probability of the \verb|<P>| version
of the string, we use $t_{i-m} \ldots t_{i-1} \, \verb|<NO-P>| \,
t_{i+1} \ldots t_{i+n}$ in estimating the probability of the \verb|<NO-P>|
version, and vice versa. We therefore apply the count cutoff to the
sum of the counts of the \verb|<P>| version of the N-gram and the
\verb|<NO-P>| version of the N-gram, requiring, for the N-grams to be
used in our model, that
\[C(t_{i-m} \ldots t_{i-1} \, \verb|<P>| \, t_{i+1} \ldots t_{i+n}) + C(t_{i-m} \ldots t_{i-1} \,
\verb|<NO-P>| \, t_{i+1} \ldots t_{i+n}) \ge CC\] where $CC$ is the
selected count cutoff. We use this sum, because if one of the versions
of the N-gram has a large number of occurrences in our training data
and the other has very few, that is very informative, and we would not
want to disregard the pattern simply because the less likely version
of the N-gram has a low count.

There are two other important details of how we collect N-gram
counts. First, if an N-gram pattern has a count of 0 for either the
\verb|<P>| version or the \verb|<NO-P>| version of the N-gram, we
don't save counts for any longer N-grams that extend that pattern by
adding tokens on either the left or right end. Any such extension must
also have a count of 0 for either the \verb|<P>| or the
\verb|<NO-P>| version, and will not add any additional information
beyond that supplied by the shorter N-grams.

Second, for our initial experiments, we include at most one \verb|<P>|
or \verb|<NO-P>| token per N-gram, at the candidate sentence break
position that the N-gram is used to make a prediction for. Any other
\verb|<P>| or \verb|<NO-P>| tokens are dropped. When there are
multiple \verb|<P>| or \verb|<NO-P>| tokens within what would be a
single N-gram, we produce multiple versions, with only one of the
\verb|<P>| or \verb|<NO-P>| tokens in each version of the N-gram.
For example, the tokenized version of the string \verb|P. & L.|
occurring in the middle of a paragraph, would produce two separate
N-grams spanning the entire string:
\begin{quote}
  \verb|P <SH>A _. <NO-P> & <SH>A L _.| \\
  \verb|P <SH>A _. & L <SH>A _. <NO-P>|
\end{quote}

Thus, in predicting whether there is a sentence break at a particular
position, we do not have access to information about other nearby
paragraph breaks. We do this for comparability with previous work,
almost none of which makes any mention of using information about
paragraph breaks in the test data to predict sentence breaks. In
particular, we know from examining Gillick's (2009) version of the
Satz WSJ test set that he could not have used such information,
because paragraph breaks are not represented in his test data. We
revisit this issue in Chapter~8, when we examine the use of paragraph
break information at prediction time.

\section{N-gram paragraph break probability estimates}

We now explain how we estimate the probability of a paragraph break at
a particular place in a text, using the N-gram statistics collected
from our normalized, tokenized training set as described above.  We
begin by describing our initial approach, and then introduce a number
of refinements that improve accuracy in predicting sentence breaks,
using part of the Penn Treebank WSJ corpus as a tuning/development data
set.

\subsection{Our initial approach}

Let a hidden paragraph break candidate token $t_i$, which might be
either \verb|<P>| (paragraph break) or \verb|<NO-P>| (no paragraph
break), be the $i$-th token in a normalized, tokenized text. We estimate
the probability that $t_i = \verb|<P>|$, given that $t_i = \verb|<P>|$
or $t_i = \verb|<NO-P>|$, and given the sequence of $j$ tokens
preceding $t_i$ and the sequence of $k$ tokens following $t_i$, as
follows:
\[p(t_{i-j} \ldots t_{i-1} \, \verb|<P>| \, t_{i+1} \ldots t_{i+k} \,
| \, t_{i-j} \ldots t_{i-1} \, \verb|<P?>| \, t_{i+1} \ldots t_{i+k}) =
\frac{P_{+}}{P_{?}}\]
where \verb|<P?>| represents the condition that the token at the
indicated position is either \verb|<P>| or \verb|<NO-P>|, and
\begin{eqnarray*}
P_{+} & = & p(t_{i-j} \ldots t_{i-1} \, \verb|<P>| \, t_{i+1} \ldots
t_{i+k}) \\
P_{?} & = & p(t_{i-j} \ldots t_{i-1} \, \verb|<P?>| \, t_{i+1} \ldots
t_{i+k})
\end{eqnarray*}
$P_{+}$ is the probability of the token string from $i-j$ to $i+k$
with $t_i = \verb|<P>|$, and $P_{?}$ is the probability of the token
string from $i-j$ to $i+k$ with either $t_i = \verb|<P>|$ or $t_i =
\verb|<NO-P>|$. Hence $P_{+}/P_{?}$ is the probability of a paragraph
break at position $i$, given the surrounding token string from $i-j$
to $i+k$ and the fact that either $t_i = \verb|<P>|$ or $t_i =
\verb|<NO-P>|$. The values of $j$ and $k$, delimiting the extent of
the surrounding token string used to estimate the probability of a
paragraph break at position $i$, are determined separately for each
example, using count cutoffs applied to the N-gram statistics of the
training data.

Note that the equations for $P_{+}$ and $P_{?}$ can be rewritten as
follows:
\begin{eqnarray*}
P_{+} & = & p(t_{i-j} \ldots t_{i-1} \, \verb|<P?>|) \times p(t_{i-j} \ldots t_{i-1} \, \verb|<P>| \, t_{i+1} \ldots t_{i+k} \, | \,
t_{i-j} \ldots t_{i-1} \verb|<P?>|) \\
P_{?} & = & p(t_{i-j} \ldots t_{i-1} \, \verb|<P?>|) \times p(t_{i-j} \ldots t_{i-1} \, \verb|<P?>| \, t_{i+1} \ldots t_{i+k} \, | \,
t_{i-j} \ldots t_{i-1} \verb|<P?>|)
\end{eqnarray*}
Thus both numerator and denomenator in $P_{+}/P_{?}$ include a factor
of $p(t_{i-j} \ldots t_{i-1} \, \verb|<P?>|)$, which can cancel,
letting us reformulate the estimation of the paragraph break
probability as:
\[p(t_{i-j} \ldots t_{i-1} \, \verb|<P>| \, t_{i+1} \ldots t_{i+k} \,
| \, t_{i-j} \ldots t_{i-1} \, \verb|<P?>| \, t_{i+1} \ldots t_{i+k}) =
\frac{P'_{+}}{P'_{?}}\]
where
\begin{eqnarray*}
P'_{+} & = & p(t_{i-j} \ldots t_{i-1} \, \verb|<P>| \, t_{i+1} \ldots t_{i+k} \, | \,
t_{i-j} \ldots t_{i-1} \verb|<P?>|) \\
P'_{?} & = & p(t_{i-j} \ldots t_{i-1} \, \verb|<P?>| \, t_{i+1} \ldots t_{i+k} \, | \,
t_{i-j} \ldots t_{i-1} \verb|<P?>|)
\end{eqnarray*}

In our initial approach, we further break down $P'_{?}$ into two
subcases, $P'_{?} = P'_{+}+P'_{-}$, where $P'_{+}$ is as defined above
and
\[P'_{-} = p(t_{i-j} \ldots t_{i-1} \, \verb|<NO-P>| \, t_{i+1} \ldots t_{i+k} \, | \,
t_{i-j} \ldots t_{i-1} \verb|<P?>|)\]
so that
\[p(t_{i-j} \ldots t_{i-1} \, \verb|<P>| \, t_{i+1} \ldots t_{i+k} \,
| \, t_{i-j} \ldots t_{i-1} \, \verb|<P?>| \, t_{i+1} \ldots t_{i+k}) =
\frac{P'_{+}}{P'_{+}+P'_{-}}\]
$P'_{+}$ and $P'_{-}$ are then decomposed into products of conditional
probabilities as follows:
\begin{eqnarray*}
P'_{+} & = & p(t_{i-j} \ldots t_{i-1} \, \verb|<P>| \, | \, t_{i-j} \ldots t_{i-1} \, \verb|<P?>|) \times \\
      &   & \prod_{n=1}^{k} p(t_{i-j} \ldots t_{i-1} \, \verb|<P>|
\, t_{i+1} \ldots t_{i+n} \, | \, t_{i-j} \ldots t_{i-1} \, \verb|<P>|
\, t_{i+1} \ldots t_{(i+n)-1}) \\
\\
P'_{-} & = & p(t_{i-j} \ldots t_{i-1} \, \verb|<NO-P>| \, | \, t_{i-j} \ldots t_{i-1} \, \verb|<P?>|) \times \\
      &   & \prod_{n=1}^{k} p(t_{i-j} \ldots  t_{i-1} \, \verb|<NO-P>| \, t_{i+1} \ldots t_{i+n} \, | \, t_{i-j} \ldots t_{i-1} \,
\verb|<NO-P>| \, t_{i+1} \ldots t_{(i+n)-1})
\end{eqnarray*}
In interpreting these equations, note that for any $m$ and $n$, $t_m
\ldots t_m$ is simply $t_m$, and $t_m \ldots t_n$ is the empty
sequence if $n<m$.

The conditional probabilities on the right hand side of the above
equations are estimated by ratios expressing relative frequencies in our
training data. The conditional probabilities for the paragraph break
candidate at position $i$, given the preceding text, are
estimated by
\begin{eqnarray*}
p(t_{i-j} \ldots t_{i-1} \, \verb|<P>| \, | \, t_{i-j} \ldots t_{i-1}
\, \verb|<P?>|) & = & \frac{C_{+}}{C_{+}+C_{-}} \\
p(t_{i-j} \ldots t_{i-1} \, \verb|<NO-P>| \, | \, t_{i-j} \ldots
t_{i-1} \, \verb|<P?>|) & = & \frac{C_{-}}{C_{+}+C_{-}}
\end{eqnarray*}
where
\begin{eqnarray*}
C_{+} & = & C(t_{i-j} \ldots t_{i-1} \, \verb|<P>|) \\
C_{-} & = & C(t_{i-j} \ldots t_{i-1} \, \verb|<NO-P>|)
\end{eqnarray*}
and $C(t_{i-j} \ldots t_{i})$ is the number of occurrences of
$t_{i-j} \ldots t_{i}$ in the training data.

To stay within the length and count cutoff limits described in
Section~5.3, for each example, we set $j$ to the maximum value such
that $j \le 9$ and
\[C(t_{i-j} \ldots t_{i-1} \, \verb|<P>|)+C(t_{i-j} \ldots t_{i-1} \,
\verb|<NO-P>|) \ge CC\] where $CC$ is the selected count cutoff.

The remaining conditional probabilities used in our equations for
$P'_{+}$ and $P'_{-}$ are all of the form $p(t_{i-j} \ldots t_{i+n} \, | \, t_{i-j}
\ldots t_{(i+n)-1})$, which we estimate by
\[p(t_{i-j} \ldots t_{i+n} \, | \, t_{i-j} \ldots t_{(i+n)-1}) = \frac{C(t_{(i+n)-m} \ldots t_{i+n})}{C(t_{(i+n)-m} \ldots t_{(i+n)-1})}\]
where, for each conditional probability needed, $m$ is set to the
maximum value such that $m \le 9$ and
\[C(t_{(i+n)-m} \ldots t_{i-1} \, \verb|<P>| \, t_{i+1} \ldots t_{i+n})+
C(t_{(i+n)-m} \ldots t_{i-1} \, \verb|<NO-P>| \, t_{i+1} \ldots t_{i+n}) \ge
CC\]
Given the way we selected $j$, it must be the case that $(i+n)-m \ge
i-j$. If $(i+n)-m > i-j$, we are simply backing off the estimate of the
conditional probability $p(t_{i-j} \ldots t_{i+n} \, | \, t_{i-j}
\ldots t_{(i+n)-1})$, using shorter left contexts, so that the length and
count cutoff limits on the required N-grams will be satisfied.

We set $k$, the number of tokens we take into account following
the paragraph break candidate position, to be the maximum value such
that $k \le 9$ and
\[C(\verb|<P>| \, t_{i+1} \ldots t_{i+k})+C(\verb|<NO-P>| \,
t_{i+1} \ldots t_{i+k}) \ge CC\] 
Note that if we let $k$ take on a larger value, then 
\[C(\verb|<P>| \, t_{i+1} \ldots t_{i+k})+C(\verb|<NO-P>| \,
t_{i+1} \ldots t_{i+k}) < CC\]
and our estimates of
\[p(\verb|<P>| \, t_{i+1} \ldots t_{i+k} \, | \, \verb|<P>| \, t_{i+1} \ldots t_{(i+k)-1})\]
and
\[p(\verb|<NO-P>| \, t_{i+1} \ldots t_{i+k} \, | \, \verb|<NO-P>| \, t_{i+1} \ldots t_{(i+k)-1})\]
would both back off to the estimate for
\[p(t_{i+1} \ldots t_{i+k} \, | \, t_{i+1} \ldots t_{(i+k)-1})\]
and cancel out in the computation of $P_{+}/(P_{+}+P_{-})$, giving us the
same result as with the value of $k$ that we have proposed.

One final detail of our paragraph break probability estimation
deserves mention at this point. Given the way we have defined our
count cutoff test and the way we compute our probability estimates, it
is possible for one of the frequency ratios to have a denomenator of
0. However, if we compute each of the factors in our product of
conditional probabilities from left to right, we will always encounter
a numerator of 0 before we encounter a denomenator of 0. Whenever we
encounter a numerator of 0, we treat the final product as 0, even if
some of the following factors are undefined because they incorporate a
denomenator of 0.

\subsection{Paragraph break probability estimate refinements}

We now consider combinations of three modifications to the initial
approach presented above:
\begin{enumerate}
\item
  Additional count cutoff constraints
\item
  Directly estimating
  \begin{eqnarray*}
    P'{?} & = & p(t_{i-j} \ldots t_{i-1} \, \verb|<P?>| \, t_{i+1} \ldots t_{i+k} \, | \,
    t_{i-j} \ldots t_{i-1} \verb|<P?>|)
  \end{eqnarray*}
  instead of using the sum of estimates for the two subcases $P'_{?} = P'_{+}+P'_{-}$:
  \begin{eqnarray*}
    P'_{+} & = & p(t_{i-j} \ldots t_{i-1} \, \verb|<P>| \, t_{i+1} \ldots t_{i+k} \, | \,
    t_{i-j} \ldots t_{i-1} \verb|<P?>|) \\
    P'_{-} & = & p(t_{i-j} \ldots t_{i-1} \, \verb|<NO-P>| \, t_{i+1} \ldots t_{i+k} \, | \,
    t_{i-j} \ldots t_{i-1} \verb|<P?>|)
  \end{eqnarray*}
\item
  Averaging left-to-right and right-to-left estimates of
  \[p(t_{i-j} \ldots t_{i-1} \, \verb|<P>| \, t_{i+1} \ldots
  t_{i+k}) \, | \, t_{i-j} \ldots t_{i-1} \, \verb|<P?>| \, t_{i+1}
  \ldots t_{i+k})\]
\end{enumerate}

\subsubsection{Additional count cutoff constraints}

In debugging the implementation of our initial approach to estimating
paragraph break probablities, we observed a weakness in the way we
were applying count cutoffs. The problem is illustrated in estimating
$P'_{+}$ and $P'_{-}$ for the abstract example $X \, \verb|_.| \,
\verb|<P?>| \, Y \, Z$ using the following N-gram counts:
\begin{center}
  \begin{tabular}{|l|r|r|}
    \hline
    N-gram pattern & \verb|<P>| count & \verb|<NO-P>| count \\
    \hline
    $X \, \verb|_.| \, \verb|<P?>|$             & 100 & 9900 \\
    $X \, \verb|_.| \, \verb|<P?>| \, Y$        & 1   & 99 \\
    $X \, \verb|_.| \, \verb|<P?>| \, Y \, Z$   & 1   & 90 \\
    $\, \verb|_.| \verb|<P?>| \, Y$             & 1   & 10000 \\
    $\, \verb|_.| \verb|<P?>| \, Y \, Z$        & 1   & 100 \\
    \hline
  \end{tabular}
\end{center}

Since all the N-gram counts for \verb|<NO-P>| are much larger than the
corresponding N-gram counts for \verb|<P>|, it looks like, in this
context, \verb|<NO-P>| should have a much higher estimated probability
than \verb|<P>|. But depending on what count cutoff we use, that might
not be the case. Suppose $CC=95$. In that case, we compute the
estimated probability of \verb|<P>| given the surrounding text as
\[p(X \, \verb|_.| \, \verb|<P>| \, Y \, Z \, | \, X \, \verb|_.| \, \verb|<P?>| \, Y \, Z) =
\frac{P'_{+}}{P'_{+}+P'_{-}}\]
where
\begin{eqnarray*}
  P'_{+} & = & p(X \, \verb|_.| \, \verb|<P>| \, | \, X \, \verb|_.| \, \verb|<P?>|) \times \\
        &   & p(X \, \verb|_.| \, \verb|<P>| \, Y \, | \, X \, \verb|_.| \, \verb|<P>|) \times \\
        &   & p(X \, \verb|_.| \, \verb|<P>| \, Y \, Z \, | \, X \, \verb|_.| \, \verb|<P>| \, Y) \\
  \\
  P'_{-} & = & p(X \, \verb|_.| \, \verb|<NO-P>| \, | \, X \, \verb|_.| \, \verb|<P?>|) \times \\
        &   & p(X \, \verb|_.| \, \verb|<NO-P>| \, Y \, | \, X \, \verb|_.| \, \verb|<NO-P>|) \times \\
        &   & p(X \, \verb|_.| \, \verb|<NO-P>| \, Y \, Z \, | \, X \, \verb|_.| \, \verb|<NO-P>| \, Y) \\
\end{eqnarray*}

Given the count cutoff $CC=95$, the three factors of $P'_{+}$ are
estimated as
\begin{eqnarray*}
p(X \, \verb|_.| \, \verb|<P>| \, | \, X \, \verb|_.| \, \verb|<P?>|) & = & C(X \, \verb|_.| \, \verb|<P>|)/(C(X \,
\verb|_.| \, \verb|<P>|) + C(X \, \verb|_.| \, \verb|<NO-P>|)) \\
& = & 100/(100+9900) \\
\\
p(X \, \verb|_.| \, \verb|<P>| \, Y \, | \, X \, \verb|_.| \, \verb|<P>|) & = & C(X \, \verb|_.| \, \verb|<P>| \, Y)/C(X \,
\verb|_.| \, \verb|<P>|) \\
& = & 1/100 \\
\\
p(X \, \verb|_.| \, \verb|<P>| \, Y \, Z \, | \, X \, \verb|_.| \, \verb|<P>| \, Y) & = & C(\verb|_.| \, \verb|<P>| \, Y \,
Z)/C(\verb|_.| \, \verb|<P>| \, Y) \\
& = & 1/1 
\end{eqnarray*}
Multiplying these three factors gives us
\[P'_{+} = [100/(100+9900)] \times [1/100] \times [1/1] = 0.0001\]

Repeating the calculation for $P'_{-}$, we get
\begin{eqnarray*}
p(X \, \verb|_.| \, \verb|<NO-P>| \, | \, X \, \verb|_.| \, \verb|<P?>|) & = & C(X \, \verb|_.| \, \verb|<NO-P>|)/(C(X \,
\verb|_.| \, \verb|<P>|) + C(X \, \verb|_.| \, \verb|<NO-P>|)) \\
& = & 9900/(100+9900) \\
\\
p(X \, \verb|_.| \, \verb|<NO-P>| \, Y \, | \, X \, \verb|_.| \, \verb|<NO-P>|) & = & C(X \, \verb|_.| \, \verb|<NO-P>| \, Y)/C(X \,
\verb|_.| \, \verb|<NO-P>|) \\
& = & 99/9900 \\
\\
p(X \, \verb|_.| \, \verb|<NO-P>| \, Y \, Z \, | \, X \, \verb|_.| \, \verb|<NO-P>| \, Y) & = & C(\verb|_.| \, \verb|<NO-P>| \, Y \,
Z)/C(\verb|_.| \, \verb|<NO-P>| \, Y) \\
& = & 100/10000
\end{eqnarray*}

\[P'_{-} = [9900/(100+9900)] \times [99/9900] \times [100/10000] = 0.000099\]

Perhaps suprisingly, our estimate for $P'_{+}$ turns out to be
slightly larger than that for $P'_{-}$, which means that, according to our
model, a paragraph break is slightly more likely than not. The key
turns out to be our backoff estimate for the conditional probabilities
\begin{eqnarray*}
& & p(X \, \verb|_.| \, \verb|<P>| \, Y \, Z \, | \, X \, \verb|_.| \,
\verb|<P>| \, Y) \\
& & p(X \, \verb|_.| \, \verb|<NO-P>| \, Y \, Z \, | \, X \, \verb|_.|
\, \verb|<NO-P>| \, Y)
\end{eqnarray*}
Our count cutoffs block us from using the counts for $X \, \verb|_.|
\, \verb|<P>| \, Y \, Z$ and $X \, \verb|_.| \, \verb|<NO-P>| \, Y \, Z$,
which would have suggested that a paragraph
break is very unlikely, but allow us to estimate
\[p(X \, \verb|_.| \, \verb|<P>| \, Y \, Z \, | \, X \, \verb|_.| \, \verb|<P>| \, Y) = C(\verb|_.| \, \verb|<P>| \, Y \,
Z)/C(\verb|_.| \, \verb|<P>| \, Y) = 1/1 \]
based on a single example. To prevent conditional probabilities from
being estimated based on such low counts, we incorporate additional
count cutoffs into our model.

With our additional count cutoff constraints, in computing $P'_{+}$
and $P'_{-}$, for each conditional probability estimated by an
equation of the form
\[p(t_{i-j} \ldots t_{i+n} \, | \, t_{i-j} \ldots t_{(i+n)-1}) = \frac{C(t_{(i+n)-m} \ldots t_{i+n})}{C(t_{(i+n)-m} \ldots t_{(i+n)-1})}\]
$m$ is set to the maximum value such that $m \le 9$ and
\begin{eqnarray*}
C(t_{(i+n)-m} \ldots t_{i-1} \, \verb|<P>| \, t_{i+1} \ldots t_{i+n}) +
C(t_{(i+n)-m} \ldots t_{i-1} \, \verb|<NO-P>| \, t_{i+1} \ldots t_{i+n})&\ge&CC \\
C(t_{(i+n)-m} \ldots t_{i-1} \, \verb|<P>| \, t_{i+1} \ldots t_{(i+n)-1})&\ge&CC \\
C(t_{(i+n)-m} \ldots t_{i-1} \, \verb|<NO-P>| \, t_{i+1} \ldots t_{(i+n)-1})&\ge&CC
\end{eqnarray*}
The additional count cutoff constraints ensure that, for every
conditional probability we estimate as a frequency ratio, the
denomenator of the ratio meets our count cutoff, so that no
conditional probability is estimated based on a very small sample.

These additional count cutoffs also imply that the largest useful
value of $k$ in the equation
\[p(t_{i-j} \ldots t_{i-1} \, \verb|<P>| \, t_{i+1} \ldots t_{i+k} \,
| \, t_{i-j} \ldots t_{i-1} \, \verb|<P?>| \, t_{i+1} \ldots t_{i+k}) =
\frac{P'_{+}}{P'_{+}+P'_{-}}\]
is the maximum value such that $k \le 9$ and
\begin{eqnarray*}
C(\verb|<P>| \, t_{i+1} \ldots t_{i+k})+C(\verb|<NO-P>| \,
t_{i+1} \ldots t_{i+k})&\ge&CC \\
C(\verb|<P>| \, t_{i+1} \ldots t_{(i+k)-1})&\ge&CC \\
C(\verb|<NO-P>| \, t_{i+1} \ldots t_{(i+k)-1})&\ge&CC
\end{eqnarray*}

With these additional count cutoff constraints and $CC=95$, we are forced to approximate $p(X \, \verb|_.| \, \verb|<P>| \, Y \, Z \,
| \, X \, \verb|_.| \, \verb|<P?>| \, Y \, Z)$ in our example by
\[p(X \, \verb|_.| \, \verb|<P>| \, Y \, | \, X \, \verb|_.| \, \verb|<P?>| \, Y) =
\frac{P'_{+}}{P'_{+}+P'_{-}}\]
where
\begin{eqnarray*}
  P'_{+} & = & p(X \, \verb|_.| \, \verb|<P>| \, | \, X \, \verb|_.| \, \verb|<P?>|) \times \\
        &   & p(X \, \verb|_.| \, \verb|<P>| \, Y \, | \, X \, \verb|_.| \, \verb|<P>|) \\
  \\
  P'_{-} & = & p(X \, \verb|_.| \, \verb|<NO-P>| \, | \, X \, \verb|_.| \, \verb|<P?>|) \times \\
        &   & p(X \, \verb|_.| \, \verb|<NO-P>| \, Y \, | \, X \, \verb|_.| \, \verb|<NO-P>|)
\end{eqnarray*}
We now estimate $P'_{+}$ by
\begin{eqnarray*}
p(X \, \verb|_.| \, \verb|<P>| \, | \, X \, \verb|_.| \, \verb|<P?>|) & = & C(X \, \verb|_.| \, \verb|<P>|)/(C(X \,
\verb|_.| \, \verb|<P>|) + C(X \, \verb|_.| \, \verb|<NO-P>|)) \\
& = & 100/(100+9900) \\
\\
p(X \, \verb|_.| \, \verb|<P>| \, Y \, | \, X \, \verb|_.| \, \verb|<P>|) & = & C(X \, \verb|_.| \, \verb|<P>| \, Y)/C(X \,
\verb|_.| \, \verb|<P>|) \\
& = & 1/100 
\end{eqnarray*}
\[P'_{+} = [100/(100+9900)] \times [1/100] = 0.0001\]
Repeating the calculation for $P'_{-}$, we get
\begin{eqnarray*}
p(X \, \verb|_.| \, \verb|<NO-P>| \, | \, X \, \verb|_.| \, \verb|<P?>|) & = & C(X \, \verb|_.| \, \verb|<NO-P>|)/(C(X \,
\verb|_.| \, \verb|<P>|) + C(X \, \verb|_.| \, \verb|<NO-P>|)) \\
& = & 9900/(100+9900) \\
\\
p(X \, \verb|_.| \, \verb|<NO-P>| \, Y \, | \, X \, \verb|_.| \, \verb|<NO-P>|) & = & C(X \, \verb|_.| \, \verb|<NO-P>| \, Y)/C(X \,
\verb|_.| \, \verb|<NO-P>|) \\
& = & 99/9900
\end{eqnarray*}
\[P'_{-} = [9900/(100+9900)] \times [99/9900] = 0.0099\]
Thus, by applying count cutoff constraints to the denomenator of every
conditional probability estimate, we get the more reasonable estimated
result that a paragraph break is 99 times less likely than no
paragraph break.

\subsubsection{Directly estimating $P'_{?}$}

In our initial approach, we estimated
\[p(t_{i-j} \ldots t_{i-1} \, \verb|<P>| \, t_{i+1} \ldots t_{i+k} \,
| \, t_{i-j} \ldots t_{i-1} \, \verb|<P?>| \, t_{i+1} \ldots t_{i+k}) =
\frac{P'_{+}}{P'_{?}}\]
using the equations
\begin{eqnarray*}
P'_{?} & = & P'_{+}+P'_{-} \\
P'_{+} & = & p(t_{i-j} \ldots t_{i-1} \, \verb|<P>| \, t_{i+1} \ldots t_{i+k} \, | \,
t_{i-j} \ldots t_{i-1} \verb|<P?>|) \\
P'_{-} & = & p(t_{i-j} \ldots t_{i-1} \, \verb|<NO-P>| \, t_{i+1} \ldots t_{i+k} \, | \,
t_{i-j} \ldots t_{i-1} \verb|<P?>|)
\end{eqnarray*}
to compute $P'_{?}$ by separately estimating $P'_{+}$ and
$P'_{-}$ and summing the results.

Alternatively, we can directly estimate 
\begin{eqnarray*}
P'_{?} & = & p(t_{i-j} \ldots t_{i-1} \, \verb|<P?>| \, t_{i+1} \ldots t_{i+k} \, | \,
t_{i-j} \ldots t_{i-1} \verb|<P?>|)
\end{eqnarray*}
by decomposing the right hand side of this equation into a product of
conditional probabilities as follows:
\[P'_{?} = \prod_{n=1}^{k} p(t_{i-j} \ldots t_{i-1} \, \verb|<P?>|
\, t_{i+1} \ldots t_{i+n} \, | \, t_{i-j} \ldots t_{i-1} \, \verb|<P?>| \,
t_{i+1} \ldots t_{(i+n)-1})\]
Note that the factor $p(t_{i-j} \ldots t_{i-1} \, \verb|<P>| \, | \,
t_{i-j} \ldots t_{i-1} \, \verb|<P?>|)$ that occurs in the
decomposition of $P'_{+}$ into a product of conditional probabilities
would become $p(t_{i-j} \ldots t_{i-1} \, \verb|<P?>| \, | \, t_{i-j}
\ldots t_{i-1} \, \verb|<P?>|)$ in the definition of $P'_{?}$, but
that conditional probability is identically equal to 1 and is
therefore omitted.

The individual factors in the decomposition of $P'_{?}$ are estimated by
frequency ratios, as in our initial approach:
\[p(t_{i-j} \ldots t_{i+n} \, | \, t_{i-j} \ldots t_{(i+n)-1}) =
\frac{C(t_{(i+n)-m} \ldots t_{i+n})}{C(t_{(i+n)-m} \ldots t_{(i+n)-1})}\]
where $t_i = \verb|<P?>|$ and
\begin{eqnarray*}
C(t_{(i+n)-m} \ldots t_{i-1} \, \verb|<P?>| \, t_{i+1} \ldots t_{i+n}) & = &
C(t_{(i+n)-m} \ldots t_{i-1} \, \verb|<P>| \, t_{i+1} \ldots t_{i+n}) + \\
& & C(t_{(i+n)-m} \ldots t_{i-1} \, \verb|<NO-P>| \, t_{i+1} \ldots
t_{i+n}) \\
\\
C(t_{(i+n)-m} \ldots t_{i-1} \, \verb|<P?>| \, t_{i+1} \ldots t_{(i+n)-1}) & = &
C(t_{(i+n)-m} \ldots t_{i-1} \, \verb|<P>| \, t_{i+1} \ldots t_{(i+n)-1}) + \\
& & C(t_{(i+n)-m} \ldots t_{i-1} \, \verb|<NO-P>| \, t_{i+1} \ldots
t_{(i+n)-1})
\end{eqnarray*}

We now consider how to set $j$ and $k$ in
\[p(t_{i-j} \ldots t_{i-1} \, \verb|<P>| \, t_{i+1} \ldots t_{i+k} \,
| \, t_{i-j} \ldots t_{i-1} \, \verb|<P?>| \, t_{i+1} \ldots t_{i+k}) =
\frac{P'_{+}}{P'_{?}}\]
and how to set $m$ for each conditional probability estimated as
\[p(t_{i-j} \ldots t_{i+n} \, | \, t_{i-j} \ldots t_{(i+n)-1}) =
\frac{C(t_{(i+n)-m} \ldots t_{i+n})}{C(t_{(i+n)-m} \ldots t_{(i+n)-1})}\]
when directly estimating $P'_{?}$.

If we apply count cutoffs as in our initial approach, $j$, $k$, and
$m$ are all set exactly as before. If we apply the additional count
cutoff constraints of Section~5.4.2.1, the conditions on $j$ still
do not change, but something interesting happens in setting $k$ and
$m$. The conditional probability estimates
\[p(t_{i-j} \ldots t_{i+n} \, | \, t_{i-j} \ldots t_{(i+n)-1}) =
\frac{C(t_{(i+n)-m} \ldots t_{i+n})}{C(t_{(i+n)-m} \ldots t_{(i+n)-1})}\]
always come in pairs. In our initial approach, if one of the pair is
\[p(t_{(i+n)-m} \ldots t_{i-1} \, \verb|<P>| \, t_{i+1} \ldots t_{i+n} \, |
\, t_{(i+n)-m} \ldots t_{i-1} \, \verb|<P>| \, t_{i+1} \ldots t_{(i+n)-1})\]
and the other would be
\[p(t_{(i+n)-m} \ldots t_{i-1} \, \verb|<NO-P>| \, t_{i+1} \ldots t_{i+n} \, |
\, t_{(i+n)-m} \ldots t_{i-1} \, \verb|<NO-P>| \, t_{i+1} \ldots t_{(i+n)-1})\]
We therefore added the count cutoff constraints
\begin{eqnarray*}
C(t_{(i+n)-m} \ldots t_{i-1} \, \verb|<P>| \, t_{i+1} \ldots t_{(i+n)-1})&\ge&CC \\
C(t_{(i+n)-m} \ldots t_{i-1} \, \verb|<NO-P>| \, t_{i+1} \ldots t_{(i+n)-1})&\ge&CC
\end{eqnarray*}
to ensure that all the frequency ratios used to estimate conditional
probabilities had reasonably large denomentators.

In the approach we are currently considering, the paired conditional
probabilities are of the form 
\[p(t_{(i+n)-m} \ldots t_{i-1} \, \verb|<P>| \, t_{i+1} \ldots t_{i+n} \, |
\, t_{(i+n)-m} \ldots t_{i-1} \, \verb|<P>| \, t_{i+1} \ldots t_{(i+n)-1})\]
and 
\[p(t_{(i+n)-m} \ldots t_{i-1} \, \verb|<P?>| \, t_{i+1} \ldots t_{i+n} \, |
\, t_{(i+n)-m} \ldots t_{i-1} \, \verb|<P?>| \, t_{i+1} \ldots t_{(i+n)-1})\]
suggesting that the additional count cutoff constraints should be
\begin{eqnarray*}
C(t_{(i+n)-m} \ldots t_{i-1} \, \verb|<P>| \, t_{i+1} \ldots t_{(i+n)-1})&\ge&CC \\
C(t_{(i+n)-m} \ldots t_{i-1} \, \verb|<P?>| \, t_{i+1} \ldots t_{(i+n)-1})&\ge&CC
\end{eqnarray*}
But necessarily
\[C(t_{(i+n)-m} \ldots t_{i-1} \, \verb|<P?>| \, t_{i+1} \ldots
t_{(i+n)-1}) \ge C(t_{(i+n)-m} \ldots t_{i-1} \, \verb|<P>| \, t_{i+1} \ldots
t_{(i+n)-1})\]
so the second added constraint becomes redundant. Similar reasoning
leads to dropping the constraint that $C(\verb|<NO-P>| \, t_{i+1}
\ldots t_{(i+k)-1}) \ge CC$, without replacement, in setting the value
of $k$, the number of tokens we take into account following the
paragraph break candidate position.

Putting it all together, combining the additional count cutoff
constraints of Section~5.4.2.1 with direct estimation of $P_{?}$:
\[p(t_{i-j} \ldots t_{i-1} \, \verb|<P>| \, t_{i+1} \ldots t_{i+k} \,
| \, t_{i-j} \ldots t_{i-1} \, \verb|<P?>| \, t_{i+1} \ldots t_{i+k}) =
\frac{P'_{+}}{P'_{?}}\]
where $j$ is the maximum value such that $j \le 9$ and 
\[C(t_{i-j} \ldots t_{i-1} \, \verb|<P>|)+C(t_{i-j} \ldots t_{i-1} \,
\verb|<NO-P>|) \ge CC\]
and $k$ is the maximum value such that $k \le 9$ and
\begin{eqnarray*}
C(\verb|<P>| \, t_{i+1} \ldots t_{i+k})+C(\verb|<NO-P>| \,
t_{i+1} \ldots t_{i+k})&\ge&CC \\
C(\verb|<P>| \, t_{i+1} \ldots t_{(i+k)-1})&\ge&CC
\end{eqnarray*}
Then
\begin{eqnarray*}
P'_{+} & = & p(t_{i-j} \ldots t_{i-1} \, \verb|<P>| \, | \, t_{i-j} \ldots t_{i-1} \, \verb|<P?>|) \times \\
      &   & \prod_{n=1}^{k} p(t_{i-j} \ldots t_{i-1} \, \verb|<P>|
\, t_{i+1} \ldots t_{i+n} \, | \, t_{i-j} \ldots t_{i-1} \, \verb|<P>|
\, t_{i+1} \ldots t_{(i+n)-1}) \\
\\
P'_{?} & = & \prod_{n=1}^{k} p(t_{i-j} \ldots t_{i-1} \, \verb|<P?>|
\, t_{i+1} \ldots t_{i+n} \, | \, t_{i-j} \ldots t_{i-1} \, \verb|<P?>| \,
t_{i+1} \ldots t_{(i+n)-1})
\end{eqnarray*}
where
\[p(t_{i-j} \ldots t_{i-1} \, \verb|<P>| \, | \, t_{i-j} \ldots t_{i-1}
\, \verb|<P?>|) = \frac{C_{+}}{C_{+}+C_{-}}\]
\begin{eqnarray*}
C_{+} & = & C(t_{i-j} \ldots t_{i-1} \, \verb|<P>|) \\
C_{-} & = & C(t_{i-j} \ldots t_{i-1} \, \verb|<NO-P>|)
\end{eqnarray*}
and for all remaining conditional probabilities, 
\[p(t_{i-j} \ldots t_{i+n} \, | \, t_{i-j} \ldots t_{(i+n)-1}) =
\frac{C(t_{(i+n)-m} \ldots t_{i+n})}{C(t_{(i+n)-m} \ldots t_{(i+n)-1})}\]
where, if $t_i = \verb|<P?>|$,
\begin{eqnarray*}
C(t_{(i+n)-m} \ldots t_{i-1} \, \verb|<P?>| \, t_{i+1} \ldots t_{i+n}) & = &
C(t_{(i+n)-m} \ldots t_{i-1} \, \verb|<P>| \, t_{i+1} \ldots t_{i+n}) + \\
& & C(t_{(i+n)-m} \ldots t_{i-1} \, \verb|<NO-P>| \, t_{i+1} \ldots
t_{i+n}) \\
\\
C(t_{(i+n)-m} \ldots t_{i-1} \, \verb|<P?>| \, t_{i+1} \ldots t_{(i+n)-1}) & = &
C(t_{(i+n)-m} \ldots t_{i-1} \, \verb|<P>| \, t_{i+1} \ldots t_{(i+n)-1}) + \\
& & C(t_{(i+n)-m} \ldots t_{i-1} \, \verb|<NO-P>| \, t_{i+1} \ldots
t_{(i+n)-1})
\end{eqnarray*}
Finally, $m$ is set to the maximum value such that $m \le 9$ and
\begin{eqnarray*}
C(t_{(i+n)-m} \ldots t_{i-1} \, \verb|<P>| \, t_{i+1} \ldots t_{i+n}) +
C(t_{(i+n)-m} \ldots t_{i-1} \, \verb|<NO-P>| \, t_{i+1} \ldots t_{i+n})&\ge&CC \\
C(t_{(i+n)-m} \ldots t_{i-1} \, \verb|<P>| \, t_{i+1} \ldots t_{(i+n)-1})&\ge&CC
\end{eqnarray*}

An advantage of direct estimation of $P'_{?}$ when using our
additional count cutoff constraints is that low counts for N-grams
including \verb|<NO-P>| will not necessarily block us from using
longer N-grams. Consider the abstract example $X \, \verb|_.| \,
\verb|<P?>| \, Y \, Z$ that we looked at before, but
with different N-gram counts and a different count cutoff:
\begin{center}
  \begin{tabular}{|l|r|r|}
    \hline
    N-gram pattern & \verb|<P>| count & \verb|<NO-P>| count \\
    \hline
    $X \, \verb|_.| \, \verb|<P?>|$             & 9000 & 1000 \\
    $X \, \verb|_.| \, \verb|<P?>| \, Y$        & 900  & 100 \\
    $X \, \verb|_.| \, \verb|<P?>| \, Y \, Z$   & 10   & 100 \\
    $\, \verb|_.| \verb|<P?>| \, Y$             & 9000 & 1000 \\
    $\, \verb|_.| \verb|<P?>| \, Y \, Z$        & 900   & 100 \\
    \hline
  \end{tabular}
\end{center}
Let our count cutoff $CC = 105$.

If we apply our additional count cutoffs, and we estimate $P'_{?} =
P'_{+} + P'_{-}$ by separately estimating $P'_{+}$ and $P'_{-}$ and
summing the results, we cannot use the counts
\begin{eqnarray*}
& & C(X \, \verb|_.| \, \verb|<P>| \, Y) = 900 \\
& & C(X \, \verb|_.| \, \verb|<P>| \, Y \, Z) = 10 \\
& & C(X \, \verb|_.| \, \verb|<NO-P>| \, Y) = 100 \\
& & C(X \, \verb|_.| \, \verb|<NO-P>| \, Y \, Z) = 100
\end{eqnarray*}
to estimate
\begin{eqnarray*}
  & & p(X \, \verb|_.| \, \verb|<P>| \, Y \, Z \, | \, X \, \verb|_.|
\, \verb|<P>| \, Y) \\
& & p(X \, \verb|_.| \, \verb|<NO-P>| \, Y \, Z \, | \, X \, \verb|_.|
\, \verb|<NO-P>| \, Y)
\end{eqnarray*}
because the denomenator count for the \verb|<NO-P>| case,
$C(X \, \verb|_.| \, \verb|<NO-P>| \, Y) = 100$, is less than
the count cutoff $CC = 105$.

Without going through all the detailed calculations, using $P'_{?} =
P'_{+} + P'_{-}$, with the back-offs required by our count cutoff,
gives us the estimate
\[p(X \, \verb|_.| \, \verb|<P>| \, Y \, Z \, | \, X \, \verb|_.| \,
\verb|<P?>| \, Y \, Z) = 9/10\]
On the other hand, using direct estimation of $P'_{?}$, even with the additional count
cutoffs, we arrive at the estimate
\[p(X \, \verb|_.| \, \verb|<P>| \, Y \, Z \, | \, X \, \verb|_.| \,
\verb|<P?>| \, Y \, Z) = 1/11\]
because we can use the counts
\begin{eqnarray*}
& & C(X \, \verb|_.| \, \verb|<P>| \, Y) = 900 \\
& & C(X \, \verb|_.| \, \verb|<P>| \, Y \, Z) = 10 \\
& & C(X \, \verb|_.| \, \verb|<NO-P>| \, Y) = 100 \\
& & C(X \, \verb|_.| \, \verb|<NO-P>| \, Y \, Z) = 100
\end{eqnarray*}
to estimate
\begin{eqnarray*}
  & & p(X \, \verb|_.| \, \verb|<P>| \, Y \, Z \, | \, X \, \verb|_.|
\, \verb|<P>| \, Y) \\
& & p(X \, \verb|_.| \, \verb|<P?>| \, Y \, Z \, | \, X \, \verb|_.|
\, \verb|<P?>| \, Y)
\end{eqnarray*}
without any denomenators being less than 105, so no backoffs are required.

A potential disadvantage of direct estimation of $P'_{?}$ is that,
without the strict requirement that $P'_{?} = P'_{+} + P'_{-}$, the
conditional probability estimates may not be normalized when count
cutoffs force us to use backoff estimates. Under some
circumstances, our estimates may even imply that $P'_{+} > P'_{?}$. For
example, consider the abstract example $X \, \verb|_.| \, \verb|<P?>|
\, Y$ with these N-gram counts:
\begin{center}
  \begin{tabular}{|l|r|r|}
    \hline
    N-gram pattern & \verb|<P>| count & \verb|<NO-P>| count \\
    \hline
    $X \, \verb|_.| \, \verb|<P?>|$        & 900    & 100 \\
    $X \, \verb|_.| \, \verb|<P?>| \, Y$   & 0      & 0 \\
    $\verb|_.| \, \verb|<P?>|$             & $10^9$ & $10^9$ \\
    $\verb|_.| \, \verb|<P?>| \, Y$        & 900    & 100 \\
    \hline
  \end{tabular}
\end{center}
For any count cutoff $0 < CC \le 1000$, we estimate
\[p(X \, \verb|_.| \, \verb|<P>| \, Y \, | \, X \, \verb|_.| \,
\verb|<P?>| \, Y) = \frac{P'_{+}}{P'_{?}}\]
where
\begin{eqnarray*}
P'_{+} & = & p(X \, \verb|_.| \, \verb|<P>| \, | \, X \, \verb|_.| \,
\verb|<P?>|) \times \\
& & p(X \, \verb|_.| \, \verb|<P>| \, Y \, | \, X \, \verb|_.| \,
\verb|<P>|) \\
\\
P'_{?} & = & p(X \, \verb|_.| \, \verb|<P?>| \, Y \, | \, X \, \verb|_.| \,
\verb|<P?>|)
\end{eqnarray*}
We estimate $P'_{+}$ by
\begin{eqnarray*}
p(X \, \verb|_.| \, \verb|<P>| \, | \, X \, \verb|_.| \,
\verb|<P?>|) & = & C(X \, \verb|_.| \, \verb|<P>|)/(C(X \, \verb|_.|
\, \verb|<P>|)+C(X \, \verb|_.| \, \verb|<NO-P>|)) \\
& = & 900/(900+100) \\
\\
p(X \, \verb|_.| \, \verb|<P>| \, Y \, | \, X \, \verb|_.| \,
\verb|<P>|) & = & C(\verb|_.| \, \verb|<P>| \, Y)/C(\verb|_.| \,
\verb|<P>|) \\
& = & 900/10^9
\end{eqnarray*}
\[P'_{+} = [900/(900+100)] \times [900/10^9] = 810/10^9 \]
and we estimate $P'_{?}$ by
\begin{eqnarray*}
P'_{?} & = & p(X \, \verb|_.| \, \verb|<P?>| \, Y \, | \, X \, \verb|_.| \,
\verb|<P?>|) \\
& = & C(\verb|_.| \, \verb|<P?>| \, Y)/C(\verb|_.| \, \verb|<P?>|) \\
& = & (C(\verb|_.| \, \verb|<P>| \, Y)+C(\verb|_.| \, \verb|<NO-P>| \,
Y))/(C(\verb|_.| \, \verb|<P>|)+C(\verb|_.| \, \verb|<NO-P>|)) \\
& = & (900 + 100)/(10^9 + 10^9) \\
& = & 500/10^9
\end{eqnarray*}
Thus due to the approximations introduced by backing off, the estimate
for
\[p(X \, \verb|_.| \, \verb|<P>| \, Y \, | \, X \, \verb|_.| \,
\verb|<P?>| \, Y) = \frac{P'_{+}}{P'_{?}} = \frac{810/10^9}{500/10^9} = 1.62\]
We will see later, however, that the advantage with this approach of
being able to use longer N-grams in our estimates seems empirically to
outweigh the disadvantage of our conditional probability estimates not
being normalized.

\subsubsection{Averaging left-to-right and right-to-left probability
  estimates}

With both our initial approach and the refinements we have considered
so far, we have decomposed conditional probability estimates for
multi-token sequences into products of single token conditional
probability estimates proceeding left-to-right, conditioning on left
contexts. However, it is equally possible to use products of single
token conditional probability estimates proceeding right-to-left,
conditioning on right contexts.

We can compute right-to-left estimates either with just our initial
approach or with any combination of other refinements. Here we present
right-to-left estimation for our approach combining additional count
cutoff constraints with direct estimation of $P'_{?}$. Adaptation to
other estimation variants is straightforward.

\[p(t_{i-j} \ldots t_{i-1} \, \verb|<P>| \, t_{i+1} \ldots t_{i+k} \,
| \, t_{i-j} \ldots t_{i-1} \, \verb|<P?>| \, t_{i+1} \ldots t_{i+k}) =
\frac{P'_{+}}{P'_{?}}\]
where $j$ is the maximum value such that $j \le 9$ and 
\begin{eqnarray*}
C(t_{i-j} \ldots t_{i-1} \, \verb|<P>|)+C(t_{i-j} \ldots t_{i-1} \,
\verb|<NO-P>|)&\ge&CC \\
C(t_{(i-j)+1} \ldots t_{i-1} \, \verb|<P>|)&\ge&CC
\end{eqnarray*}
and $k$ is the maximum value such that $k \le 9$ and
\[C(\verb|<P>| \, t_{i+1} \ldots t_{i+k})+C(\verb|<NO-P>| \,
t_{i+1} \ldots t_{i+k}) \ge CC\]
Then
\begin{eqnarray*}
P'_{+} & = & p(\verb|<P>| \, t_{i+1} \ldots t_{i+k} \, | \,
\verb|<P?>| \, t_{i+1} \ldots t_{i+k}) \times \\
      &   & \prod_{n=1}^{j} p(t_{i-n} \ldots t_{i-1} \, \verb|<P>|
\, t_{i+1} \ldots t_{i+k} \, | \, t_{(i-n)+1} \ldots t_{i-1} \, \verb|<P>|
\, t_{i+1} \ldots t_{i+k}) \\
\\
P'_{?} & = & \prod_{n=1}^{j} p(t_{i-n} \ldots t_{i-1} \, \verb|<P?>|
\, t_{i+1} \ldots t_{i+k} \, | \, t_{(i-n)+1} \ldots t_{i-1} \, \verb|<P?>| \,
t_{i+1} \ldots t_{i+k})
\end{eqnarray*}
where
\[p(\verb|<P>| \, t_{i+1} \ldots t_{i+k} \, | \, \verb|<P?>| \,
  t_{i+1} \ldots t_{i+k}) = \frac{C_{+}}{C_{+}+C_{-}}\]
\begin{eqnarray*}
C_{+} & = & C(\verb|<P>| \, t_{i+1} \ldots t_{i+k}) \\
C_{-} & = & C(\verb|<NO-P>| \, t_{i+1} \ldots t_{i+k})
\end{eqnarray*}
and for all remaining conditional probabilities 
\[p(t_{i-n} \ldots t_{i+k} \, | \, t_{(i-n)+1} \ldots t_{i+k)}) =
\frac{C(t_{i-n} \ldots t_{(i-n)+m})}{C(t_{(i-n)+1} \ldots t_{(i-n)+m})}\]
where, if $t_i = \verb|<P?>|$,
\begin{eqnarray*}
C(t_{i-n} \ldots t_{i-1} \, \verb|<P?>| \, t_{i+1} \ldots t_{(i-n)+m}) & = &
C(t_{i-n} \ldots t_{i-1} \, \verb|<P>| \, t_{i+1} \ldots t_{(i-n)+m}) + \\
& & C(t_{i-n} \ldots t_{i-1} \, \verb|<NO-P>| \, t_{i+1} \ldots
t_{(i-n)+m}) \\
\\
C(t_{(i-n)+1} \ldots t_{i-1} \, \verb|<P?>| \, t_{i+1} \ldots t_{(i-n)+m}) & = &
C(t_{(i-n)+1} \ldots t_{i-1} \, \verb|<P>| \, t_{i+1} \ldots t_{(i-n)+m}) + \\
& & C(t_{(i-n)+1} \ldots t_{i-1} \, \verb|<NO-P>| \, t_{i+1} \ldots
t_{(i-n)+m})
\end{eqnarray*}
and $m$ is set to the maximum value such that $m \le 9$ and
\begin{eqnarray*}
C(t_{i-n} \ldots t_{i-1} \, \verb|<P>| \, t_{i+1} \ldots t_{(i-n)+m}) +
C(t_{i-n} \ldots t_{i-1} \, \verb|<NO-P>| \, t_{i+1} \ldots t_{(i-n)+m})&\ge&CC \\
C(t_{(i-n)+1} \ldots t_{i-1} \, \verb|<P>| \, t_{i+1} \ldots t_{(i-n)+m})&\ge&CC
\end{eqnarray*}

In cases where count cutoffs do not force backoffs, the right-to-left
and left-to-right decompositions of our conditional probability
estimates for multi-token sequences always produce the same result,
because without backoffs, both decompositions simplify to
\[p(t_{i-j} \ldots t_{i-1} \, \verb|<P>| \, t_{i+1} \ldots t_{i+k} \,
| \, t_{i-j} \ldots t_{i-1} \, \verb|<P?>| \, t_{i+1} \ldots t_{i+k}) =
\frac{C_{+}}{C_{+}+C_{-}}\]
where
\begin{eqnarray*}
C_{+} & = & C(t_{i-j} \ldots t_{i-1} \, \verb|<P>| \, t_{i+1} \ldots
t_{i+k}) \\
C_{-} & = & C(t_{i-j} \ldots t_{i-1} \, \verb|<NO-P>| \, t_{i+1}
\ldots t_{i+k})
\end{eqnarray*}
This simplification holds whether we use direct estimation of $P'_{?}$
or apply $P'_{?} = P'_{+} + P'_{-}$ and estimate $P'_{+}$ and $P'_{-}$
separately.

For more complicated reasons that we omit here, with the count cutoffs
of our original approach, right-to-left and left-to-right estimation
also always lead to the same result, even when backoffs occur. With the
additional count cutoffs presented in Sections~5.4.2.1 and 5.4.2.2,
however, backoffs can sometimes cause right-to-left and left-to-right
estimation to give different results. We can illustrate this using the
abstract example $\verb|_.| \, \verb|<P?>| \, Y \, Z$ with the
following N-gram counts:
\begin{center}
  \begin{tabular}{|l|r|r|}
    \hline
    N-gram pattern & \verb|<P>| count & \verb|<NO-P>| count \\
    \hline
    $\verb|_.| \, \verb|<P?>|$             & $10^9$ & $10^9$ \\
    $\verb|_.| \, \verb|<P?>| \, Y$        & 96    & 96 \\
    $\verb|_.| \, \verb|<P?>| \, Y \, Z$   & 4    & 96 \\
    $\verb|<P?>|$                          & $1.1 \times 10^9$ & $1.1 \times 10^9$ \\
    $\verb|<P?>| \, Y$                     & 96    & 96 \\
    $\verb|<P?>| \, Y \, Z$                & 4    & 96 \\
    \hline
  \end{tabular}
\end{center}

With left-to-right estimation and our additional count cutoff
constraints with $CC = 98$, we have to approximate $p(\verb|_.| \,
\verb|<P>| \, Y \, Z \, | \, \verb|_.| \, \verb|<P?>| \, Y \, Z)$ by
$p(\verb|_.| \, \verb|<P>| \, Y \, | \, \verb|_.| \, \verb|<P?>| \,
Y)$, because we fail to satisfy one of the count cutoff constraints
needed to set $k = 2$, namely $C(\verb|<P>| \, Y) \ge 98$  (since
$C(\verb|<P>| \, Y) = 96$). Hence,

\[p(\verb|_.| \, \verb|<P>| \, Y \, | \, \verb|_.| \,
\verb|<P?>| \, Y) = \frac{P'_{+}}{P'_{?}}\]
where
\begin{eqnarray*}
P'_{+} & = & p(\verb|_.| \, \verb|<P>| \, | \, \verb|_.| \,
\verb|<P?>|) \times \\
& & p(\verb|_.| \, \verb|<P>| \, Y \, | \, \verb|_.| \,
\verb|<P>|) \\
\\
P'_{?} & = & p(\verb|_.| \, \verb|<P?>| \, Y \, | \verb|_.| \,
\verb|<P?>|)
\end{eqnarray*}

We then estimate $P'_{+}$ by
\begin{eqnarray*}
p(\verb|_.| \, \verb|<P>| \, | \, \verb|_.| \, \verb|<P?>|) & = &
C(\verb|_.| \, \verb|<P>|)/(C(\verb|_.| \, \verb|<P>|)+C(\verb|_.| \, \verb|<NO-P>|)) \\
& = & 10^9/(10^9+10^9) \\
\\
p(\verb|_.| \, \verb|<P>| \, Y \, | \, \verb|_.| \,
\verb|<P>|) & = & C(\verb|_.| \, \verb|<P>| \, Y)/C(\verb|_.| \,
\verb|<P>|) \\
& = & 96/10^9
\end{eqnarray*}
\[P'_{+} = [10^9/(10^9+10^9)] \times [96/10^9] = 48/10^9 \]
and using direct estimation of $P'_{?}$,
\begin{eqnarray*}
P'_{?} & = & p(\verb|_.| \, \verb|<P?>| \, Y \, | \, \verb|_.| \, \verb|<P?>|) \\
& = & C(\verb|_.| \, \verb|<P?>| \, Y)/C(\verb|_.| \, \verb|<P?>|) \\
& = & (C(\verb|_.| \, \verb|<P>| \, Y)+C(\verb|_.| \, \verb|<NO-P>| \,
Y))/(C(\verb|_.| \, \verb|<P>|)+C(\verb|_.| \, \verb|<NO-P>|)) \\
& = & (96 + 96)/(10^9 + 10^9) \\
& = & 96/10^9
\end{eqnarray*}
Thus
\[p(X \, \verb|_.| \, \verb|<P>| \, Y \, | \, X \, \verb|_.| \,
\verb|<P?>| \, Y) = \frac{P'_{+}}{P'_{?}} = \frac{48/10^9}{96/10^9} = 0.5\]

On the other hand, if we use right-to-left estimation, even with $CC =
98$ and the additional count cutoff constraints, we can estimate
$p(\verb|_.| \, \verb|<P>| \, Y \, Z \, | \, \verb|_.| \, \verb|<P?>|
\, Y \, Z)$ without approximating it by $p(\verb|_.| \, \verb|<P>| \,
Y \, | \, \verb|_.| \, \verb|<P?>| \, Y)$, because different count
cutoff constraints apply right-to-left than apply left-to-right. Hence
we estimate
\[p(\verb|_.| \, \verb|<P>| \, Y \, Z \, | \, \verb|_.| \,
\verb|<P?>| \, Y \, Z) = \frac{P'_{+}}{P'_{?}}\]
where
\begin{eqnarray*}
P'_{+} & = & p(\verb|<P>| \, Y \, Z \, | \, \verb|<P?>| \, Y \, Z) \times \\
& & p(\verb|_.| \, \verb|<P>| \, Y \, Z \, | \, \verb|<P>| \, Y \, Z) \\ 
\\
P'_{?} & = & p(\verb|_.| \, \verb|<P?>| \, Y \, Z \, | \, \verb|<P?>| \, Y \, Z) \\ 
\end{eqnarray*}

We then estimate $P'_{+}$ by
\begin{eqnarray*}
p(\verb|<P>| \, Y \, Z \, | \, \verb|<P?>| \, Y \, Z)  & = &
C(\verb|<P>| \, Y \, Z)/(C(\verb|<P>| \, Y \, Z) + C(\verb|<NO-P>| \,
Y \, Z)) \\
& = & 4/(4+96) \\
\\
p(\verb|_.| \, \verb|<P>| \, Y \, Z \, | \, \verb|<P>| \, Y \, Z) & = & 
C(\verb|_.| \, \verb|<P>|)/C(\verb|<P>|) \\
& = & 10^9/(1.1 \times 10^9)
\end{eqnarray*}
\[P'_{+} = [4/(4+96)] \times [10^9/(1.1 \times 10^9)] = 4/110\]
and we estimate $P'_{?}$ by
\begin{eqnarray*}
P'_{?} & = & p(\verb|_.| \, \verb|<P?>| \, Y \, Z \, | \, \verb|<P?>| \, Y \, Z) \\ 
& = & C(\verb|_.| \, \verb|<P?>|)/C(\verb|<P?>|) \\
& = & (C(\verb|_.| \, \verb|<P>|)+C(\verb|_.| \, \verb|<NO-P>|))/(C(\verb|<P>|)+C(\verb|<NO-P>|)) \\
& = & [10^9+10^9]/[(1.1 \times 10^9)+(1.1 \times 10^9)] \\
& = & 1/1.1
\end{eqnarray*}
Thus
\[p(X \, \verb|_.| \, \verb|<P>| \, Y \, | \, X \, \verb|_.| \,
\verb|<P?>| \, Y) = \frac{P'_{+}}{P'_{?}} = \frac{4/110}{1/1.1} = 0.04\]

In this example, backoff due to count cutoffs occurs in both
left-to-right and right-to-left estimation, but in different places,
leading to different results: a paragraph break probability estimate
of 0.5 for left-to-right estimation, and 0.04 for right-to-left
estimation. We have constructed this example in such a way that the
right-to-left estimate may appear more reasonable than the
left-to-right estimate, but we could have just as easily constructed
an example where the opposite is the case. We have therefore chosen to
experiment with using the average of left-to-right and right-to-left
estimates, in combination with the other variations on our initial
approach that we consider, hoping to reduce any bias that might result
from using estimation based on only one direction or the other.



\section{Choosing among options for the form of the paragraph break
  probability model}

We initially presented our general approach to predicting sentence
breaks based on paragraph break probability in terms of the formula
$p(\langle {\rm P} \rangle|t) > \beta$ where $p(\langle {\rm P}
\rangle|t)$ represents the probability of a paragraph break at a
certain position given the surrounding text context $t$, and $\beta$
is an empirically optimized threshold above which we predict a
sentence break. We can restate this in more detail using the
notation developed in Section~5.4 as
\[p(t_{i-j} \ldots t_{i-1} \, \verb|<P>| \, t_{i+1} \ldots t_{i+k} \,
| \, t_{i-j} \ldots t_{i-1} \, \verb|<P?>| \, t_{i+1} \ldots t_{i+k})
> \beta\]
Turning this approach into a sentence break predictor requires
answering several questions:
\begin{itemize}
\item
  What combination of refinements from Section~5.4.2, if any, should
  we add to our initial method presented in Section~5.4.1?
\item
  What value of the count cutoff parameter $CC$ should we use?
\item
  What value of the paragraph break probability threshold $\beta$
  should we use?
\end{itemize}

We answer these questions by searching the space of possible answers
to find the minimum number of sentence break prediction errors on a
tuning/development data set annotated with sentence breaks. For this
purpose we use sections 00--02 and 07--21\footnote{Recall that we do
not use sections 03--06 because they overlap with our in-domain test
set.} of the Penn Treebank WSJ corpus.\footnote{We will later use this
data set as a training set for directly supervised sentence break
predictors (retaining sections 22--24 for hyperparameter tuning and
development testing of those predictors), but since we have no need
for data annotated with sentence breaks to train the main parameters
of the models we are currently considering, we can use this data set
for model structure selection and hyperparameter tuning here.}  For
every combination of refinements, we carry out a grid search over
combinations of possible values of $CC$ and $\beta$ to find the lowest
possible number of sentence break errors on the tuning/development
data set. The results are shown in the table below:
\begin{center}
  \begin{tabular}{|l|c|c|c|c|c|c|c|c|}
    \hline
    Additional count cutoff constraints &  & X &  & X &  & X &  & X  \\
    \hline
    Direct estimation of $P'_{?}$  &  &  & X & X &  &  & X & X  \\
    \hline
    Averaging L-to-R and R-to-L estimation  &  &  &  &  & X & X & X & X  \\
    \hline
    Tuning/development set minimum errors  & 337 & 333 & 330 & 318 & 337 & 321 & 330 & 315  \\
    \hline
  \end{tabular}
\end{center}

This table shows the minimum number of sentence break prediction
errors on the tuning/development set (which includes 35,432 actual
sentence breaks) for each possible combination of refinements of our
initial approach, using a separate grid search for each variant over
combinations of values of the count cutoff $CC$ and probability
threshold $\beta$ to find the lowest possible number of errors. The
grid search explores values of $CC$ from 100 to 250, incrementing by
1, and values of $\beta$ from 0.010 to 0.040, incrementing by
0.001. These ranges have been chosen after preliminary experiments
that indicate the optimal values of $CC$ and $\beta$ lie in this
region for all variants of of our model.

As the table shows, both the additional count cutoff constraints and
direct estimation of $P'_{?}$ improve on the 337 errors of our
original approach, and are even better when combined.  Averaging
left-to-right and right-to-left estimates gives an additional benefit
whenever additional count cutoff constraints are used, so that the
best overall approach, yielding 315 errors, uses all three
refinements.  This is the method that we use in all subsequent
experiments.

\section{Tie-breaking in selection of hyperparameter values}

Selection of values for the hyperparameters $CC$ and $\beta$ might
seem to be a simple matter of selecting the values that led to the
lowest number of errors in the experiments described in Section~5.5,
but it is actually more complicated than that, because of the problem
of ties. The following table shows the six combinations of values of $CC$
and $\beta$ that produced the minimum of 315 errors on our
tuning/development set when used with our best paragraph break
probability model structure, including all three refinements of our
initial approach:
\begin{center}
  \begin{tabular}{|c|c|}
    \hline
    $CC$ & $\beta$ \\
    \hline
    178 & 0.032 \\
    199 & 0.025 \\
    222 & 0.025 \\
    223 & 0.025 \\
    225 & 0.025 \\
    226 & 0.025 \\
    \hline
  \end{tabular}
\end{center}

To use a paragraph break probability model as a sentence break
predictor, we need to choose specific values for $CC$ and $\beta$. To
do this, from the six combinations that produce the lowest number of
errors on the tuning/development set, we try to choose one that seems
to be in the center of a region of other good combinations, in the
hope that it will generalize better to unseen data.  We do this by
examining a set of rectangular regions in our search grid that have
our candidate points as their centers, computing the mean number of
errors over all the points in each rectangle:
\[\frac{\sum_{i=-n}^n \sum_{j=-m}^m E(CC + i, \beta + 0.001 j)}{(2n+1)
  \times (2m+1)}\]

In this formula, $CC$ is the candidate count cutoff, $\beta$ is the
candidate probability threshold, $E(x,y)$ is the number of errors on
the tuning/development set for a count cutoff of $x$ and a probability
threshold of $y$, $n$ is the number of steps in the search grid away
from $CC$ that we include in the rectangle, and $m$ is the number of
steps in the search grid away from $\beta$ that we include in the
rectangle. Recalling that the step size in the grid is 1 for count
cutoffs and 0.001 for probability thresholds, we can see that this
gives us the mean number of errors on the tuning/development set over
a $2n+1$ by $2m+1$ sized rectangle of search grid points centered on
the point $(CC,\beta)$.

We compute the mean number of errors for all search grid rectangles
centered on our candidate points with $n$ ranging from 0 to 25 and
$m$ ranging from 0 to 10. We then sort all the rectangles by the
mean number of errors, and find the lowest score having rectangles
for that score centered on only one of our candidate points. That
candidate point gives us our selected values of $CC$ and $\beta$.
In our experiments, the lowest five scores and list of associated
rectangles are shown in the following table:
\begin{center}
  \begin{tabular}{|l|c|c|l|}
    \hline
    Mean Error & $n$ & $m$ & $(CC,\beta)$ Combinations \\
    \hline
    315.000000000000 & 0 & 0 & (178,0.032) (199,0.025) (222,0.025) \\
    & & & (223,0.025) (225,0.025) (226,0.025) \\
    \hline
    315.333333333333 & 1 & 0 & (223,0.025) (224,0.025) (225,0.025) \\
    & & & (226,0.025) \\
    \hline
    315.428571428571 & 3 & 0 & (225,0.025) \\
    \hline
    315.666666666667 & 4 & 0 & (225,0.025) (226,0.025) \\
    \hline
    315.909090909091 & 5 & 0 & (226,0.025) \\
    \hline
  \end{tabular}
\end{center}

As we can see, the two lowest mean error counts 315.000000000000 and
315.333333333333 correspond to rectangles centered on multiple points,
but the third lowest mean error count 315.428571428571 corresponds to
only one point, $CC=225$ and $\beta=0.025$, so those are our selected
values for our hyperparameters. Examining the entire search grid
suggests that the accuracy of our sentence predictor is not terribly
sensitive the exact values of the hyperparameters, but we need some
rational method to choose particular values, and this one seems
reasonable.

Regarding the particular values for $CC$ and $\beta$ selected, the
value of the probability threshold $\beta$ seems surprisinly low. We
have no good explanation for this value, given the high proportion of
sentence breaks that are paragraph breaks in our training data. Given
the low value of $\beta$, however, the high value of the count cutoff
$CC$ is not surprising. If a paragraph break probability of, say, 0.05
indicates that there is very likely a sentence break at a given
position, then we need $CC$ to be high enough to distinguish 0.05 from
0.0. A small count cutoff would not let us do that. If $CC=10$, there
would be probabilities estimated based on N-gram patterns with only 10
training examples, in which case the smallest difference in paragraph
break probability we could distinguish would be 0.1. So, we need much
larger counts to distinguish small differences in paragraph break
probability.

\section{Evaluation on in-domain test data}

We now evaluate sentence break prediction from estimated paragraph
break probabilities on our in-domain test data. We use the version of
our paragraph break probability model that incorporates all three of
our refinements to our original approach: additional count cutoff
constraints, direct estimation of $P'_{?}$, and averaging
left-to-right and right-to-left probability estimates. We apply this
model structure with the selected hyperparameter values $CC=225$ and
$\beta=0.025$.

For in-domain test data, we use the Satz WSJ corpus described in
Chapter~3, which was extracted from the ACL/DCI corpus (ACL/DCI, 1993)
and sentence-break-annotated by Palmer and Hearst (1997). We actually
test on three different versions of this corpus, to be able to compare
our results to previous work, because the version of the corpus used
by Gillick (2009) and, we believe, by Kiss and Strunk (2006) differs
significantly from the original version of Palmer and Hearst. The
three versions of the corpus that we test on are:
\begin{itemize}
\item
  Satz-1, corresponding to Palmer and Hearst's version of the corpus.
\item
  Satz-2, corresponding to Gillick's version of the corpus.
\item
  Satz-3, reconciling the differences between Satz-1 and Satz-2.
\end{itemize}
The differences between these versions of the corpus are described in
detail in Appendix~A.3.

Because Palmer and Hearst test on sentence breaks signaled by periods,
question marks, and exclamation points, while Kiss and Strunk, and
Gillick, test only on sentence breaks signaled by periods, we report
two results on each version of the corpus for our sentence break
predictor, one result for periods alone, and one for periods, question
marks, and exclamation points combined. The results of testing our
paragraph-break-probability-based (PBP-based) sentence break predictor
on in-domain data, compared to previous work, are presented in the
table below:
\begin{center}
  \begin{tabular}{||l|r|r|r|r|r|r||}
    \hline \hline
    Version of corpus & \multicolumn{2}{c|}{Satz-1} & \multicolumn{2}{c|}{Satz-2} &
    \multicolumn{2}{c||}{Satz-3} \\
    \hline
    Sentence breaks considered & \multicolumn{1}{c|}{.} & \multicolumn{1}{c|}{. ? !} & \multicolumn{1}{c|}{.} & \multicolumn{1}{c|}{. ? !}  & \multicolumn{1}{c|}{.} & \multicolumn{1}{c||}{. ? !} \\
    \hline \hline
    Reference sentence breaks & 20068 & 20262 & 20072 & 20302 & 20070
    & 20278 \\
    \hline \hline
    Errors for our PBP-based predictor & 170 & 172 & 159 & 178 & 156 & 158 \\
    \hline
    Errors for Palmer \& Hearst's method & & 409 & & & & \\
    \hline
    Errors for Kiss \& Strunk's method & & & 445 & & & \\
    \hline
    Errors for Gillick's SVM classifier & & & 67 & & & \\
    \hline \hline
    Error rate for our PBP-based predictor & 0.85\% & 0.85\% & 0.79\% & 0.88\% & 0.78\% &
    0.78\% \\
    \hline
    Error rate for Palmer \& Hearst's method & & 2.02\% & & & & \\
    \hline
    Error rate for Kiss \& Strunk's method  & & & 2.22\% & & & \\
    \hline
    Error rate for Gillick's SVM classifier  & & & 0.33\% & & & \\
    \hline \hline
  \end{tabular}
\end{center}

The top part of the table gives results in terms of the absolute
number of errors on the test set, and the bottom part of the table
translates these into error rates. Recall that we measure error rates
in terms of errors per reference sentence break, rather than errors
per sentence break candidate, since there is no general agreement on
what constitutes a sentence break candidate. For this reason, the
error rates we give for previous work are higher than reported in the
original papers.

The lowest error rates for our paragraph-break-probability-based
sentence break predictor are on the Satz-3 corpus, which in our view
is a better version of the Satz corpus for our purposes than the other
two, correcting a number of outright errors in Palmer and Hearst's
original version, Satz-1, and changing annotations reflecting
differences in the interpretation of the task in Gillick's version,
Satz-2. Our sentence break predictor produces less than half as many
errors as Palmer and Hearst's or Kiss and Strunk's methods when
measured on the versions of the Satz corpus they used, but more than
twice as many errors as Gillick's SVM. At under 1\% error, our
sentence break predictor seems quite good in an absolute sense, but it
is still not as good as Gillick's directly supervised SVM classifier
when tested on data from the same source it was trained on.

\chapter{Incorporating paragraph break probability signals in directly
  supervised sentence break predictors}

In the previous chapter, we showed that a paragraph break probability
model trained on a large corpus of naturally occurring English news
wire text from a variety of sources, with no additional annotation,
and with only two hyperparameters tuned on a much smaller WSJ corpus
annotated with sentence breaks, could be used as an effective sentence
break predictor on other WSJ text. In comparison to previous work,
however, we found that it is not as accurate as the results reported
for Gillick's (2009) directly supervised linear SVM binary classifier,
both trained and tested on WSJ text.

In light of those results, in this chapter we investigate whether we
can improve on the accuracy of Gillick's SVM on in-domain data by
adding a feature carrying a paragraph break probability signal to a
directly supervised model similar to Gillick's. We begin by training
and testing a linear SVM classifier similar to Gillick's, to serve as a
baseline. We then explore a number of different ways of adding a
paragraph break probability feature to our SVM baseline classifier,
selecting the one that performs the best on a tuning/development data
set. Finally, we evaluate the selected classifier on our Satz-3 in-domain
test data.

\section{Baseline SVM classifier}

It would perhaps be ideal for us to use Gillick's (2009) linear SVM as
our baseline directly supervised classifier, but Gillick does not give
enough information in his paper for us to replicate it. He describes
the eight indicator feature templates he uses well enough, but they
are defined with respect to a tokenization that is not
well-specified. Regarding this, Gillick says:
\begin{quote}
  While there is not room to catalog our tokenizer rules, we note that
  both untokenized text and mismatched train-test tokenization can
  increase the error rate by a factor 2.
\end{quote}

Rather than trying to reverse engineer Gillick's tokenization and
feature set, we decided to define our own feature set over the
tokenization that we were already using for our paragraph break
probability model. In fact, we use basically the same N-gram templates
to define indicator features in our directly supervised linear SVM
classifier that we use to extract N-gram counts for our paragraph
break probability model. The particular collections of N-grams we use
in the two sentence break predictors are not identical, however, since
the N-grams we use in the paragraph break probability model are those
occuring in the LDC Gigaword Fifth Edition corpus sufficiently often
to satisfy our count cutoff, while the N-grams we use in our SVM are
those that occur in our Penn Treebank training data.

\subsection{Features}

We use as indicator features all tokenized N-grams containing a single
break candidate position token as defined in Section 5.2, from length
1 to 10, occuring in sections 00--02 and 07--21\footnote{Once again,
we do not use sections 03--06 because they overlap with our in-domain
test set.} of the Penn Treebank WSJ corpus, which we use as our SVM
training data, normalized and tokenized as we described in
Sections~5.1 and 5.2. No count cutoff is applied to the selection of
features. In each feature, the break candidate token is represented by
\verb|<S?>|, interpreted as a sentence break candidate. No paragraph
break information is represented. The only length-1 N-gram is the
break candidate token \verb|<S?>| itself, which acts as a bias feature
since it occurs in the set of features for every example. The
collection of features for a particular example is labeled as either
\verb|<S>| for a sentence break, or \verb|<NO-S>|, for a
non-sentence-break.

To illustrate, for the string \verb|U.S. Army|, we would extract the
following features:
\begin{quote}
  \verb|U <SH>A _._ S <SH>A _. <S?> <SH>Aaaa Army| \\
  \verb|<SH>A _._ S <SH>A _. <S?> <SH>Aaaa Army| \\
  \verb|_._ S <SH>A _. <S?> <SH>Aaaa Army| \\
  \verb|S <SH>A _. <S?> <SH>Aaaa Army| \\
  \verb|<SH>A _. <S?> <SH>Aaaa Army| \\
  \verb|_. <S?> <SH>Aaaa Army| \\
  \verb|<S?> <SH>Aaaa Army| \\
  \verb|U <SH>A _._ S <SH>A _. <S?> <SH>Aaaa| \\
  \verb|<SH>A _._ S <SH>A _. <S?> <SH>Aaaa| \\
  \verb|_._ S <SH>A _. <S?> <SH>Aaaa| \\
  \verb|S <SH>A _. <S?> <SH>Aaaa| \\
  \verb|<SH>A _. <S?> <SH>Aaaa| \\
  \verb|_. <S?> <SH>Aaaa| \\
  \verb|<S?> <SH>Aaaa| \\
  \verb|U <SH>A _._ S <SH>A _. <S?>| \\
  \verb|<SH>A _._ S <SH>A _. <S?>| \\
  \verb|_._ S <SH>A _. <S?>| \\
  \verb|S <SH>A _. <S?>| \\
  \verb|<SH>A _. <S?>| \\
  \verb|_. <S?>| \\
  \verb|<S?>|
\end{quote}
In most situations, we would extract even more features by
incorporating more context preceding \verb|U.S.| and following
\verb|Army|, up to N-grams of length 10 either begining or ending with
the break candidate token \verb|<S?>|. Assuming there is no sentence
break between \verb|U.S.| and \verb|Army|, the set of features for
this example would be labeled \verb|<NO-S>|. Note that, as we
explained in Section~5.2, the ``shape'' tokens (prefaced by
\verb|<SH>|) follow their corresponding base tokens when they precede
the sentence break candidate, but precede their corresponding base
tokens when they follow the sentence break candidate.

\subsection{Training}

Our SVM classifier is trained by optimizing the SVM multiclass\footnote{Since
  we have only two classes, we could have used the somewhat simpler
  binary hinge loss objective, but we already had code for the
  multi-class objective, which gives equivalent results in the
  two-class case, so we use that.} (Crammer and Singer, 2001) hinge
loss objective by stochastic subgradient descent, as described by
Zhang (2004), using early stopping and a small constant learning rate.
Our hinge loss training objective does not include any numerical
regularization penalty (e.g., $L_1$ or $L_2$). The combination of
early stopping and a small learning rate provides a kind of
regularization, however, because it limits the magnitude of the
feature weights that can be learned for rare features.

Training begins with all feature weights set to 0, and for each
training example we update the feature weights according to stochastic
subgradient descent for hinge loss. We run multiple training epochs
over sections 00--02 and 07--21 of the Penn Treebank WSJ corpus, which
contain 45,441 examples, according to the definition of sentence break
candidate from Section~5.2.  For each epoch, we randomize the order of
the examples.  To implement early stopping, after each training epoch
we evaluate the resulting model on our tuning/development data set,
consisting of sections 22--24 of the Penn Treebank WSJ corpus, which
contain 6,728 examples. We keep track of the model that performs best
on this data set and which epoch it came from. When we have performed
ten epochs of training beyond the best scoring epoch observed so far,
training stops, or we stop after 200 epochs, whichever happens
first. We then output the model from the epoch that scored the best on
the tuning/development data set.

To select the best model, we use both the number of errors and the
total hinge loss on the tuning/development set. The number of errors
is our primary metric, but this results in many ties, because the
minimum number of errors on the tuning/development set is generally
quite low. So, from among the models with the lowest number of errors,
we select the one with the lowest total hinge loss. If there is still
a tie according to the hinge-loss criterion, we select the model from
the earliest epoch in the set of best models.

The other hyperparameter we have to choose a value for is the learning
rate. To find the best learning rate we perform a systematic search
over nonpositive integer powers of two, from $2^0$ to $2^{-10}$. For
each learning rate, we run our training procedure until our stopping
criterion is met, noting the number of tuning/development set errors
and total hinge loss for the best training epoch. The learning rate
that leads to the lowest number of errors (or lowest total hinge loss,
among learning rates with the same lowest number of errors) becomes
our selected learning rate.

For our baseline SVM sentence break predictor, we found that 28 epochs
of training with a learning rate of $2^{-5}$ produced the best results
on the tuning/development data set: 21 errors and a total hinge loss
of 79.35, over 6,728 sentence break candidate examples.

\subsection{Evaluation on in-domain test data}

We evaluate the baseline SVM sentence break predictor selected by the
training procedure described above on the three versions of our
in-domain test data set. The table below shows the results compared to
our paragraph-break-probability-based sentence break predictor and to
Gillick's SVM sentence break predictor:
\begin{center}
  \begin{tabular}{||l|r|r|r|r|r|r||}
    \hline \hline
    Version of corpus & \multicolumn{2}{c|}{Satz-1} & \multicolumn{2}{c|}{Satz-2} &
    \multicolumn{2}{c||}{Satz-3} \\
    \hline
    Sentence breaks considered & \multicolumn{1}{c|}{.} & \multicolumn{1}{c|}{. ? !} & \multicolumn{1}{c|}{.} & \multicolumn{1}{c|}{. ? !}  & \multicolumn{1}{c|}{.} & \multicolumn{1}{c||}{. ? !} \\
    \hline \hline
    Reference sentence breaks & 20068 & 20262 & 20072 & 20302 & 20070
    & 20278 \\
    \hline \hline
    Errors for PBP-based predictor & 170 & 172 & 159 & 178 & 156 & 158 \\
    \hline
    Errors for baseline SVM classifier & 77 & 79 & 64 & 83 & 63 & 65 \\
    \hline
    Errors for Gillick's SVM classifier & & & 67 & & & \\
    \hline \hline
    Error rate for PBP-based predictor & 0.85\% & 0.85\% & 0.79\% & 0.88\% & 0.78\% &
    0.78\% \\
    \hline
    Error rate for baseline SVM classifier & 0.38\% & 0.39\% & 0.32\% & 0.41\% & 0.31\% & 0.32\% \\
    \hline
    Error rate for Gillick's SVM classifier & & & 0.33\% & & & \\
    \hline \hline
  \end{tabular}
\end{center}

As the table shows, our baseline SVM sentence break predictor
considerably outperforms our paragraph-break-probability-based
sentence break predictor on all three versions of our in-domain test
corpus, both for periods only and for periods, question marks, and
exclamation points. It also performs quite comparably to Gillick's SVM
sentence break predictor under comparable test conditions. In fact, it
performs slightly better than Gillick's predictor, although the
difference is not statistically significant.

\section{Paragraph-break-probability-augmented SVM classifiers}

To get the most out of adding a paragraph-break-probability-based
feature to a linear SVM otherwise consisting of a weighted sum of
indicator feature values, we need to find a good way to
represent paragraph break probability information as a feature
value. On theoretical grounds, one attractive approach is to first
estimate the probability of a sentence break given only the paragraph
break probability estimate, and then to transform this sentence break
probabilty estimate into a ``log odds'' representation to serve as the
value of the added feature.

The optimal threshold on paragraph break probability estimates to use
for predicting sentence breaks turned out to be only 0.025, which
suggests that a paragraph break probability $p_p = 0.025$ implies a
sentence break probability $p_s \approx 0.5$.  Given this point of
correspondence, plus the fact our estimates of paragraph break
probability range from 0 all the way to 1.0 --- and even beyond, due
to the nonnormalized character of our most accurate paragraph break
probability model, which uses direct estimation of $P'_?$ (See
Section~5.4.2.2) --- it seems that the relationship between estimated
paragraph break probability and sentence break probability is likely
to be very nonlinear. We therefore develop a way of empirically
estimating sentence break probabilities from paragraph break
probability estimates, in the hope that a feature value based on a
sentence break probability estimate might work better in an augmented
SVM than one based directly on estimated paragraph break probability.

Once we have a way to map an estimated paragraph break probabiliity
into an estimated sentence break probability, we then need to decide
how to express the sentence break probability as a feature value. For
a number of reasons, we like the log-odds representation defined as
\[LO_s = log(\frac{p_s}{1-p_s})\]
where $p_s$ is the estimated probability of a sentence break, given
the estimated paragraph break probability.

To see why we like this representation, recall that a linear SVM for a
binary classification task, such as sentence break prediction, is
simply a sum of weighted feature values, $w_1 f_1 + \dots + w_n f_n$,
that predicts a positive outcome (a sentence break) if the weighted
sum is greater than 0, and a negative outcome (no sentence break) if
the weighted sum is less than 0.\footnote{An arbitrary choice of what
  to predict is made if the weighted sum is exactly 0.} If the
estimated sentence break probability given the estimated paragraph
break probability for an example is 0.5, this carries no information
about whether the example is a sentence break or not. In this case
$LO_s = 0$, and the prediction will depend only on the other
features. On the other hand, as $p_s$ approaches 1, $LO_s$ approaches
$+\infty$, and as $p_s$ approaches 0, $LO_s$ approaches
$-\infty$. Thus, if the estimate of sentence break probability given
paragraph break probability is certain enough, its log odds
representation can outvote all the other features in the model.

Finally, since $log(p/(1-p)) = log(p) - log(1-p)$, the magnitude of
the feature value does not depend on which possibilty is treated as
the positive outcome. If we reverse the choice, the magnitude of the
value stays the same; only the sign changes. This means that in
training the model (at least in training by stochastic subgradient
descent), the choice of which is the positive case makes no
difference. Reversing the choice simply results a learned feature
weight of the same magnitude but opposite sign, with the predictions
for all examples staying the same.

In the remainder of this section, we first describe how we empirically
map paragraph break probabilities to sentence break log odds. We then
compare, on our in-domain tuning/development data set, sentence break
log odds to several other transforms of both sentence break
probability and paragraph break probability as the value of a feature
added to our baseline linear SVM classifier. Finally, we evaluate a
linear SVM classifier agumented with the best performing feature
derived from paragraph break probability on our in-domain test data.

\subsection{Mapping paragraph break probabilities to sentence break
  log odds}

To estimate the mapping from paragraph break probabilities to sentence
break log odds, we take advantage of the fact that we have 45,441
examples of labeled sentence break candidates in the annotated Penn
Treebank WSJ text that we use as training data for our SVM
classifiers, but which play only a minor role in estimating our
paragraph break probability model, being used only to optimize the two
hyperparameters $CC$ and $\beta$. We begin by computing the estimated
paragraph break probability for all the labeled examples in this data
set. Then we can estimate the sentence break log odds for any
paragraph break probability from a set of nearest neighbors in
paragraph break probability among the labeled examples, according to
the frequencies of sentence breaks and non-sentence-breaks in the
nearest neighbor set.

Above we defined log odds in terms of probabilities, but if the
probabilities are estimated in terms of sentence break and
non-sentence-break frequencies, we can directly estimate log odds from
frequency counts without explicitly computing probabilities. If
$p_s = c_{+}/(c_{+} + c_{-})$, where $c_{+}$ is the number of examples
in the nearest neighbor set labeled as sentence breaks, and $c_{-}$ is
the number of examples in the nearest neighbor set labeled as
non-sentence-breaks, then we can derive $LO_s = log(c_{+}/c_{-})$ as
follows:
\begin{eqnarray*}
  LO_s & = & log \left(\frac{p_s}{1-p_s} \right) \\
       & = & log \left(\frac{c_{+}/(c_{+}+c_{-})}{1-(c_{+}/(c_{+}+c_{-}))} \right) \\
       & = & log \left(\frac{c_{+}/(c_{+}+c_{-})}{((c_{+}+c_{-})/(c_{+}+c_{-}))-(c_{+}/(c_{+}+c_{-}))} \right) \\
       & = & log \left(\frac{c_{+}/(c_{+}+c_{-})}{(c_{+}+c_{-}-c_{+})/(c_{+} + c_{-})} \right) \\
       & = & log \left(\frac{c_{+}/(c_{+}+c_{-})}{c_{-}/(c_{+} + c_{-})} \right) \\
       & = & log(c_{+}/c_{-})
\end{eqnarray*}

The simplest way to define a relevant set of nearest neighbors would
be to take the classic $k$-nearest-neighbors approach. That is, for an
unlabeled example with an estimated paragraph break probability $p_p$,
we could simply take the $k$ labeled examples, for some fixed $k$,
whose estimated paragraph break probability is closest to $p_p$, and
use the frequency of sentence breaks $c_+$ and non-sentence-breaks
$c_-$ in this set to estimate $LO_s$ for the new example. A problem
with this simple approach is that sentence break log odds is not
defined if either $c_+$ or $c_-$ is 0, which might happen for some
examples if we always limit the size of the nearest neighbor set to a
fixed $k$.

Our solution to this problem is to define the nearest neighbor set of
labeled examples so that it must include both positive and negative
cases. Specifically, instead of simply selecting the $k$ labeled
examples closest in paragraph break probability to $p_p$, we find the
smallest distance $d$, such that there are at least $k$ sentence
breaks ($c_+ \ge k$) and $k$ non-sentence-breaks ($c_- \ge k$) among
the labeled examples within distance $d$ of $p_p$, and use all the
labeled examples within that distance $d$ of $p_p$ to estimate $LO_s$.

For example, suppose that we let $k = 10$, and we want to estimate the
log sentence break odds $LO_s$ corresponding to a paragraph break
probability $p_p = 0.05$. Further suppose that $0.002$ is the smallest
$d$ such that $c_+ \ge 10$ and $c_- \ge 10$ for the set of labeled
examples whose estimated paragraph break probability is less than or
equal to $p_p + d$ ($0.52$) and greater than or equal to $p_p - d$
($0.048$). Finally, suppose within this range there are $30$ labeled
labeled sentence breaks and $10$ labeled non-sentence-breaks.  In that
case, we estimate $LO_s = log(30/10)$

Note that due to ties, there might be more than $k$ examples of both
cases in the nearest neighbor set. Suppose there are again $30$
labeled sentence breaks with an estimated paragraph break probability
less than or equal to $p_p + d = 0.052$ and greater than or equal to
$p_p - d = 0.048$, but $9$ labeled non-sentence-breaks with an
estimated paragraph break probability less than or equal to $p_p + d =
0.052$ and strictly greater than $p_p - d = 0.048$, and $3$ labeled
non-sentence-breaks whose estimated paragraph break probability is
exactly $0.048$. In that case we would estimate $LO_s = log(30/12)$.

This approach guarantees that both $c_+$ and $c_-$ will always be
greater than 0, so that there will be no problem using the log-odds
representation. However, it is a biased estimator of the log sentence
break odds, because we select the nearest neighbor set to be just
large enough to include $k$ examples with the minority case
label. There may be larger nearest neighbor sets that also contain $k$
examples with the minority case label, but contain additional examples
with the majority case label. By choosing the smallest nearest
neighbor set that contains at least $k$ examples with the minority
case label, we may underestimate the probability of the majority case
label and overestimate the probability of the minority case label.

We correct for this bias by multiplying the raw log odds estimate by a
factor slightly greater than $1.0$, which effectively increases the
probability of of the majority case and decreases the probability of
the minority case, while doing nothing if there are equal numbers of
of each case. Thus, instead of the raw log odds estimate we get from
our biased estimator for $LO_s$, we use an adjusted estimate $LO'_s =
\gamma LO_s$ with an empirically optimized factor $\gamma$.

This approach uses two hyperparameters $k$ and $\gamma$, which we
optimize on our annotated Penn Treebank training set. We perform a
grid search over integer values of $k$ from $5$ through $100$, and over
values of $\gamma$ from $1.00$ through $2.00$ in increments of $0.01$. For
each combination of values of $k$ and $\gamma$, and each labeled
example in the annotated training set, we compute $LO'_s = \gamma
log(c_{+}/c_{-})$, where $c_{+}$ is the count of sentence break
examples and $c_{-}$ is the count of non-sentence-break examples in
the nearest neighbor set defined by $k$.

In the grid search for optimizing $k$ and $\gamma$, the nearest
neighbor sets are constrained as described above so that at least $k$
members of each set have each label, but we modify the procedure by
``leaving one out''. That is, we omit each example for which we
estimate $LO'_s$ from its own nearest neighbor set. If we included
each example in its own nearest neighbor set, we would end up choosing
1 as the optimal value of $k$, because each example by itself would
always be the best predictor of its own annotated label.

The objective function we optimize for in the grid search is the
estimated log probability of the entire annotated training set. For
each combination of $k$ and $\gamma$, we compute an estimated log
probability for each labeled example from the value of $LO'_s$ for the
example.  If the annotated label is ``sentence break'', the log
probability value is
\[log \left(\frac{exp(LO'_s)}{1+exp(LO'_s)} \right)\]
and if the annotated label is ``no sentence break'', the log
probability value is
\[log \left(1-\frac{exp(LO'_s)}{1+exp(LO'_s)} \right)\]

For each combination of $k$ and $\gamma$, we sum the log probablities
over the training set. The combination of $k$ and $\gamma$ that yields
the highest log probability sum over the entire training set
determines our selected values of these hyperparameters. Optimizing
these hyperparameters in this way on our Penn Treebank annotated
training set produces values of $k = 16$ and $\gamma = 1.41$.

\subsection{Comparison of representations of paragraph break probability information in
  SVMs for sentence break prediction}

Although we hoped to obtain the best results in predicting sentence
breaks by mapping paragraph break probability estimates to sentence
break log odds estimates as described above, we decided to compare
many different ways of encoding paragraph break probability
information as a feature value in a linear SVM:
\begin{itemize}
\item
  Sentence break log odds $ = LO'_s$
\item
  Sentence break log probability $ = log(exp(LO'_s)/(1+exp(LO'_s)))$
\item
  Non-sentence-break log probability $ = log(1-(exp(LO'_s)/(1+exp(LO'_s))))$
\item
  Sentence break probability $ = exp(LO'_s)/(1+exp(LO'_s))$
\item
  Non-sentence-break probability $ = 1-(exp(LO'_s)/(1+exp(LO'_s)))$
\item
  Paragraph break log odds $ = log(p_p/(1-p_p))$
\item
  Paragraph break log probability $ = log(p_p)$
\item
  Non-paragraph-break log probability $ = log(1-p_p)$
\item
  Paragraph break probability $ = p_p$
\item
  Non-paragraph-break probability $ = 1-p_p$
\end{itemize}

The first five ways of encoding paragraph break probability
information are derived by the formulas shown, from the adjusted
sentence break log odds estimate $LO'_s$ computed as described in
Section~6.2.1. The remaining five are derived by the formulas shown,
from the paragraph break probability estimate $p_p$ computed as
described in Chapter~5. For the values based directly on $p_p$ without
mapping to $LO'_s$, we make one small change to the estimation of
$p_p$, by capping its values from above and below, so that we can
compute all the logarithms shown. Any value of $p_p > 0.999999$ is
replaced by $0.999999$, any any value of $p_p < 0.000001$ is replaced
by $0.000001$.

We use each of the ten formulas shown above to provide the value of a
single feature, combined with the indicator features described in
Section~6.1.1, in ten different linear SVMs. These SVMs are trained
using the same procedure we use for the baseline SVM, as described in
Section~6.1.2; again using sections 00--02 and 07--21 of the Penn
Treebank WSJ corpus as the training set, and using sections 22--24 of
the Penn Treebank WSJ corpus as the tuning/development set to choose
the number of training epochs and the learning rate. Note that
sections 00--02 and 07--21 are also used to estimate the mapping from
paragraph break probability estimates to sentence break log odds for
the five SVMs that use features based on $LO'_s$. Accordingly, we use
``leaving one out'' as described in Section~6.2.1 to compute $LO'_s$
values for the examples in sections 00--02 and 07--21. For the
tuning/development set and all other test sets, however, we compute
$LO'_s$ without using ``leaving one out'', since the examples in those
data sets are independent of the examples used to estimate the mapping from
paragraph break probability estimates to sentence break log odds.

The following table compares our basline SVM to SVMs incorporating
each our ten ways of representing paragraph break probability
information as a feature added to the baseline SVM, in terms of errors
on our tuning/development data set:
\begin{center}
  \begin{tabular}{|l|c|r|}
    \hline
    Method & Errors & p-value \\
    \hline
    Baseline & 21 & 0.0523 \\
    Sentence break log odds & 14 & 0.6171 \\
    Sentence break log probability & 14 & 0.4227 \\
    Non-sentence-break log probability & 12 &  \\
    Sentence break probability & 16 & 0.2888 \\
    Non-sentence-break probability & 15 & 0.2482 \\
    Paragraph break log odds & 14 & 0.7137 \\
    Paragraph break log probability & 13 & 1.0000 \\
    Non-paragraph-break log probability & 24 & 0.0190 \\
    Paragraph break probability & 18 & 0.0412 \\
    Non-paragraph-break probability & 17 & 0.1824 \\
    \hline
  \end{tabular}
\end{center}
The table shows that almost any method of representing paragraph break
probability information seems to improve on our baseline SVM (only the
non-paragraph-break log probability feature fails to do so), but only
by very small numbers of errors on this test set. Our favorite
representation on theoretical grounds, sentence break log odds,
performs fairly well, but it is slightly out-performed by
non-sentence-break log probability and paragraph break log
probability.

The differences in numbers of errors on this test set are so small
that we decided we should test to see if any of the differences are
statistically significant. We applied the two-tailed version of
McNemar's (1947) test to compare the method with the smallest number
of errors (the non-sentence-break log probability feature) to all the
others. The resulting p-values are presented in the last column of the
table.\footnote{A McNemar test calculator available at
  https://www.graphpad.com/quickcalcs/McNemar1.cfm was used to compute
  these p-values.} Considered individually, only two of the
differences are significant at the 0.05 level. Moreover, if we apply a
method, such as the Benjamini and Hochberg (1995) procedure, that
takes into account that we picked those two out of a set of ten
comparisons, none of the differences would be significant.

Note that lack of statistical significance does not necessarily mean
that there are no real differences in error rate among these
classifiers; it may only mean that we need a larger test set to
demonstrate the differences with confidence. In any case, we will
proceed to use the model design with the lowest number of errors in
this experiment --- the model incorporating the non-sentence-break log
probability feature --- in the remainder of our experiments.
  
\subsection{Evaluation on in-domain test data}

We now present the results of an evaluation on our Satz-3 in-domain
test set of the augmented SVM that adds a feature whose value is an
estimate of non-sentence-break log probability to the N-gram indicator
features of our baseline SVM.  We will refer to this as an NSBLP
SVM. We evaluate on all sentence breaks signaled by periods, question
marks, and exclamation points in the Satz-3 corpus.

One difference between this evaluation and the results presented above
for our Penn Treebank WSJ tuning/development data set is that here we
approximate the mapping from paragraph break probabilities to
non-sentence-break log probabilities using a precomputed table that
gives values for all paragraph break probabilities truncated to five
decimal points. We use this approximation because finding the nearest
neighbor set for a particular paragraph break probability value is
relatively time consuming. While we have not concerned ourselves
particularly about the speed of our sentence break predictors, we
wanted to be sure that our approach to mapping from paragraph break
probabilities to sentence break probability statistics would not be a
barrier to adopting our methods in practice. We observed no
difference in predictions from using the precomputed mapping table,
and using the mapping table enabled us to do all the mappings for our
largest test sets in less than 1 second. From this point going
forward, all the evaluations we present of sentence break predictors
requiring this kind of mapping make use of a precomputed mapping
table.

The results of the evaluation are shown in the table below, along with
the results previously presented in Section~6.1.3 for the sentence
break predictor based only on our paragraph break probability estimate
and for the baseline SVM:
\begin{center}
  \begin{tabular}{||l|r||}
    \hline \hline
    Reference sentence breaks & 20278 \\
    \hline \hline
    Errors for PBP-based predictor & 158 \\
    \hline
    Errors for baseline SVM classifier & 65 \\
    \hline
    Errors for NSBLP SVM classifier & 62  \\
    \hline \hline
    Error rate for PBP-based predictor & 0.78\% \\
    \hline
    Error rate for baseline SVM classifier & 0.32\% \\
    \hline
    Error rate for NSBLP SVM classifier  & 0.31\% \\
    \hline \hline
  \end{tabular}
\end{center}

As the table shows, the NSBLP SVM classifier produces fewer errors
than the baseline SVM, but only very slightly, and both SVM
classifiers produce many fewer errors than the
paragraph-break-probability-only predictor. We computed the p-value
according to McNemar's test of the differences between the NSBLP SVM
and the baseline SVM, and between the NSBLP SVM and the PBP-based
predictor. The p-value for the comparison of the two SVMs was 0.7946,
and hence, not at all significant. The p-value for the comparison of
the NSBLP SVM to the PBP-based predictor was less than 0.0001, and
hence, highly significant.

\chapter{Evaluation on out-of-domain test data}

In Chapters~5 and 6, we compared three different approaches to sentence
break prediction:
\begin{itemize}
\item
  Predicting a sentence break when a paragraph break probability
  estimate exceeds a given threshold
\item
  Predicting a sentence break based on a linear SVM using a set of
  N-gram indicator features
\item
  Predicting a sentence break based on a linear SVM combining a set of
  N-gram indicator features with an additional feature encoding a
  signal derived from a paragraph break probability estimate
\end{itemize}

We found that, when training on sufficient annotated Penn Treebank WSJ
data (approximately 733,000 words containing 45441 sentence break
candidates) and testing on other WSJ data, the straightforward SVM
based on indicator features is much more accurate than the sentence
break predictor based only on paragraph break probability estimates,
and not significantly less accurate than an SVM combining indicator
features with a paragraph-break-probability-based feature. Thus, so
far, we have seen no significant benefit from using a signal based on
paragraph break probability to predict sentence breaks, compared to
the previous state of the art.

All our evaluations so far, however, have been with test data drawn
from exactly the same source as the annotated training data, The Wall
Street Journal. We now turn our attention to evaluating on
out-of-domain data. As mentioned in Chapter~3, for out-of-domain test
data, we use the LDC English Web Treebank corpus (Bies et al., 2012),
consisting of over 250,000 words of weblogs, newsgroups, email,
reviews, and question-answers. Details of the preparation of this data
are given in Appendix~A.4.

The LDC English Web Treebank corpus is ``out of domain'' with respect
to Wall Street Journal data in at least two ways. First, its
subject matter is very different from and more diverse than The
Wall Street Journal. Second, consisting mostly of user-generated
content, it is much less observant of the orthographic conventions of
commerical publishing. Among other nonstandard features, sentences
often do not begin with capital letters, and sometimes there is no
white space preceding the beginning of a sentence.

The following table shows the accuracy on sentence breaks signaled by
periods, question marks, and exclamation points in this out-of-domain
test set, for our three main sentence break predictors:
\begin{center}
  \begin{tabular}{||l|r||}
    \hline \hline
    Reference sentence breaks & 13112 \\
    \hline \hline
    Errors for PBP-based predictor & 590 \\
    \hline
    Errors for baseline SVM classifier & 957 \\
    \hline
    Errors for NSBLP SVM classifier & 655  \\
    \hline \hline
    Error rate for PBP-based predictor & 4.50\% \\
    \hline
    Error rate for baseline SVM classifier & 7.30\% \\
    \hline
    Error rate for NSBLP SVM classifier & 5.00\% \\
    \hline \hline
  \end{tabular}
\end{center}
Two things are immediately apparent in these results. First, all error
rates are much higher than on the in-domain test set, anywhere from
5.8 to 22.8 times higher, depending on the sentence break
predictor. Second, the ordering of the sentence break predictors by
accuracy is completely different on the out-of-domain data.  The
predictor based only on paragraph break probability estimates has gone
from the least accurate on in-domain data to most accurate on
out-of-domain data, with the baseline SVM being the least accurate by
a wide margin. The augmented NSBLP SVM falls in between the other two,
but closer to the predictor based only on paragraph break probability
estimates than to the baseline SVM.  All the differences between these
three predictors on the out-of-domain test set are highly significant
($p < 0.0001$) according to McNemar's test.

\chapter{Using paragraph breaks at prediction time}

Up to this point, we have ignored one obviously useful signal for
predicting sentence breaks: paragraph breaks at prediction time. That
is, we have not taken advantage of the fact that in the texts we are
trying to break into separate sentences, paragraph breaks are usually
also sentence breaks. Our only use of paragraph breaks so far has been
in the {\em training} data for our paragraph break probability model;
we have not made any use of paragraph breaks in {\em test} data.

We have have adopted this restriction to be consistent with the
previously reported work that we have compared our sentence break
predictors to. None of this previous work appears to have made use of
paragraph breaks at prediction time. We know for certain that Gillick
(2009) did not, because we have worked with his test data, and
observed that paragraph breaks are not represented in it. The other
authors whose work we have compared ours to make no mention of using
paragraph breaks as a signal at prediction time.\footnote{Read et
  al.\ (2012), in their survey of work on sentence break prediction,
  carry out some evaluations in which paragraph breaks are taken into
  account at prediction time, but neither their test data nor
  evaluation metrics are directly comparable to ours.}

Sentence break prediction accuracy on our in-domain data is highly
accurate even without taking paragraph breaks into account. On our
out-of-domain data, however, sentence break prediction accuracy is
much lower, and with such data it would be foolish to ignore this very
strong signal. In this chapter we explore how best to represent
paragraph break information in our models for use at prediction time.

\section{What should count as a paragraph break?}

Just as we considered the question ``What should count as a sentence
break?'' in section~4.2, here we ask, ``What should count as a
paragraph break?'' As a practical matter, our answer is, whatever is
annotated as paragraph break in our data. In all of the data sets for
which we have annotated sentence breaks, the ``raw'' version of the
data is annotated with paragraph breaks represented simply by blank
lines; that is, multiple consecutive new line characters.

A canonical paragraph break separates two canonical paragraphs, each
consisting of a sequence of sentences, with each sentence ending in
standard sentence-final punctuation. To see what else might be
annotated as a paragraph break, we examined our Penn Treebank WSJ
annotated training data (PTB WSJ sections 00--02 and 07--21) to find
paragraphs that were not annotated as ending in a sentence
break. There were 193 such paragraphs out of a total of 16,180.
We can divide these 193 paragraphs that are not annotated as ending in
a sentence break into several categories, discussed below.

\subsection{Breaks following a colon introducing other text}

In our PTB WSJ training data, 17 of the paragraphs not annotated as
ending in a sentence break end in a colon (:) introducing other
text. In one case, the colon introduces a direct quotation, and in the
remaining cases, the colon introduces a sequence of related items. In
some of these cases the sequence of related items is an explicit
itemized or enumerated list. In others, the sequence is simply a set
of separate phrases, sentences, or paragraphs, related in a way
specified by the introductory paragraph that ends in a colon.

It is doubtful that all of these 17 introductory-colon cases are
correctly annotated as not being sentence breaks. There are also 69
paragraphs ending in a colon which are annotated as being sentence
breaks, with no consistent differences between those annotated as
sentence breaks and those not so annotated. One might expect that the
cases not annotated as sentence breaks would be those where there is a
syntactic dependency spanning the paragraph break, but if this is the
intent, there are many annotation errors. For instance, in the
following example, there is no sentence break annotated following the
colon that ends an introductory text, but there is no syntactic
connection to the subsequent list items:
\begin{quote}
  \verb|The following U.S. Treasury, corporate and municipal| \\
  \verb|offerings are tentatively scheduled for sale this week,| \\
  \verb|according to Dow Jones Capital Markets Report:| \\
  \\
  \verb|$15.6 billion three-month and six-month bills.| $\|$ \\
  \\
  \verb|$10 billion of two-year notes.| $\|$
\end{quote}

On the other hand, in this example, each list item completes the
partial sentence preceding the colon, but the colon is nevertheless
annotated as marking a sentence break:
\begin{quote}
  \verb|Among other changes, the White House wants to:| $\|$ \\
  \\
  \verb|-- Give the EPA more flexibility to declare a pesticide| \\
  \verb|   an imminent hazard and pull it from the marketplace.| $\|$ \\
  \\
  \verb|-- Speed up the process for removing a pesticide that isn't| \\
  \verb|   an imminent hazard.| $\|$
\end{quote}

In some cases, different occurences of the very same text are
annotated differently. There are eight instances of
\begin{quote}
  \verb|In major market activity:|
\end{quote}
as an isolated paragraph. Four times there is a sentence break
annotated following the colon, and four times there is not.

We have no disagreement with the annotation of a paragraph break
following the colon in any of the examples discussed here. As we have
seen, however, there is little consistency in whether these paragraph
breaks are annotated as sentence breaks.

\subsection{Elements of an itemized list}

There is one case of an itemized list, consisting of six elements, in
which each element is treated as a separate paragraph, but only the
final list element is annotated as ending in a sentence break:
\begin{quote}
  \verb|They transferred some $28 million from the Community| \\
  \verb|Development Block Grant program designated largely for| \\
  \verb|low- and moderate-income projects and funneled it into| \\
  \verb|such items as:| \\
  \\
  \verb|-- $1.2 million for a performing-arts center in Newark,| \\
  \\
  \verb|-- $1.3 million for "job retention" in Hawaiian sugar| \\
  \verb|   mills.| \\
  \\
  \verb|-- $400,000 for a collapsing utility tunnel in Salisbury,| \\
  \\
  \verb|-- $500,000 for "equipment and landscaping to deter crime| \\
  \verb|   and aid police surveillance" at a Michigan park.| \\
  \\
  \verb|-- $450,000 for "integrated urban data based in seven| \\
  \verb|   cities." No other details.| \\
  \\
  \verb|-- $390,000 for a library and recreation center at Mackinac| \\
  \verb|   Island, Mich.| $\|$ \verb|Rep. Traxler recently purchased an| \\
  \verb|   unimproved building lot on the island.| $\|$
\end{quote}

There are many examples of itemized lists in our PTP WSJ training
data. In all cases, each list element is treated as a separate
paragraph, but this is the only itemized list in which some of the
list elements are annotated as not ending in a sentence break. In all
other itemized lists, each element is annotated as ending in a
sentence break, even if the list elements are not full gramatical
sentences, as in the first example in Section~8.1.1.

We might ask whether ``paragraph break'' is the right way to describe
the breaks preceding and following the elements of an itemized list,
but our data is annotated with only one sort of break involving
vertical white space, so that is the only option available. The
question of when to consider there to be sentence breaks between
itemized list elements is perhaps even more difficult. In all other
itemized lists in the training set, all the list elements are
punctuated with periods, even when they are not full sentences, so
perhaps this is why the PTB annotators decided to treat them as
separate sentences. The example here is unique in having some of the
list elements punctuated with commas, which perhaps explains why the
nonfinal list elements were not treated as ending in sentence breaks.

\subsection{Hard line breaks treated as paragraph breaks}

The largest category of ``paragraph breaks'' not annotated as sentence
breaks appear to be 169 instances of hard line breaks that are not
actually paragraph breaks, but have been annotated as such. These fall
into several subclasses:
\begin{itemize}
\item
  57 cases of nonfinal lines of the signature block of a letter to
  the editor, for example:
  \begin{quote}
    \verb|Leighton E. Cluff M.D.| \\
    \\
    \verb|President| \\
    \\
    \verb|Robert Wood Johnson Foundation| \\
    \\
    \verb|Princeton, N.J.| $\|$
  \end{quote}
\item
  57 lines of humorous verse, for example:
  \begin{quote}
    \verb|Boys with tops, and Frisbee tossers,| \\
    \\
    \verb|And P.R. types with bees in their bonnet,| \\
    \\
    \verb|Have a goal in common, all of them try| \\
    \\
    \verb|To put the right spin on it.| $\|$
  \end{quote}
\item
  35 headlines broken into two lines, for example:
  \begin{quote}
    \verb|Beauty Takes Backseat| \\
    \\
    \verb|To Safety on Bridges|
  \end{quote}
\item
  19 lines in corporate earnings reports, for example:
  \begin{quote}
    \verb|Columbia Savings & Loan| \\
    \\
    \verb|(NYSE; Symbol: CSV)| $\|$ \\
    \\
    \verb|Business: Savings and loan| $\|$ \\
    \\
    \verb|Year ended Dec. 31, 1988:| \\
    \\
    \verb|Net income: $65 million; or $1.49 a share| $\|$ \\
    \\
    \verb|Third quarter, Sept. 30, 1989:| \\
    \\
    \verb|Net loss: $11.57 a share vs. net income:| \\
    \\
    \verb|37 cents a share| $\|$ \\
    \\
    \verb|Average daily trading volume:| \\
    \\
    \verb|83,206 shares| $\|$ \\
    \\
    \verb|Common shares outstanding: 19.6 million| $\|$
  \end{quote}
\item
  One example of a fairly mysterious break that may have been a hard
  line break inserted for formatting reasons, but is simply not
  believable as a paragraph break:
  \begin{quote}
    \verb|NOTE:| $\|$ \verb|NAS is National Advanced Systems,| \\
    \\
    \verb|CDC -- Control Data Corp., Bull NH Information|
    \verb|Systems Inc.| $\|$
  \end{quote}
\end{itemize}

In these examples, all of the blank lines indicating a paragraph break
seem spurious, but except for the last, the reasons for a hard line
break seems clear. This leads us to the conclusion that hard line
breaks and paragraph breaks have been conflated in the raw version of
the Penn Treebank WSJ data. It would be much more useful to have hard
line breaks separately annotated.

\subsection{Humorous verse titles}

The last category of text segments annotated as paragraphs not ending
in sentence breaks are two cases of humorous verse titles. One of
these is in fact the title of the piece of humorous verse we quoted
above: 
\begin{quote}
  \verb|With a Twist of the Wrist| \\
  \\
  \verb|Boys with tops, and Frisbee tossers,|
\end{quote}

A blank line following the title of a piece of verse is certainly
appropriate. Classifying it as a paragraph break is somewhat
questionable, but as we observed previously, we have no other category
of vertical break in the annotation scheme used in our data. Even more
questionable, however, is the fact that these two titles are not
annotated with sentence breaks following them. There are a total of 15
pieces of humourous verse in this training set, and the other 13 are
all annotated as ending in sentence breaks. So, it seems that for
consistency, these two should be annotated as sentence breaks as well.

\subsection{Observations}

Of the 193 instances in our PTB WSJ training data of annotated
paragraph breaks not annotated as sentence breaks, 169 seem not to be
paragraph breaks after all, and the remaining 24 fall into categories
for which sentence breaks are very inconsistently annotated.  However,
only 11 instances of annotated paragraph breaks not annotated as
sentence breaks are marked by standard sentence-final
punctuation. Eight of these were among the hard line breaks
misinterpreted as paragraph breaks, and three are in the one sequence
of list items that were not annotated as being separated by sentence
breaks.

Thus, as long as we confine our efforts to predicting sentence breaks
signaled by standard sentence-final punctuation, although our training
data may be slightly noisy, we can probably get help from taking
paragraph breaks into account. However, any efforts at more
comprehensive prediction of sentence breaks, including those not
signaled by standard sentence-final punctuation, would probably need
more careful annotation of paragraph breaks, and sentence breaks at
paragraph breaks, to make paragraph breaks a more reliable signal for
predicting sentence breaks.

\section{Representing paragraph breaks in sentence break prediction
  models}

Despite finding many problems in our PTB WSJ training set in the
annotation of paragraph breaks and sentence breaks at paragraph
breaks, we nevertheless decided to proceed with experiments trying to
use paragraph breaks to help predict sentence breaks, without
attempting to correct either training or test data in this regard,
because relatively few of the problems we found affected paragraphs
ending in standard sentence-final punctuation, which are the only kind
of sentence breaks we try to predict.

We explore three different methods of using paragraph break
information at prediction time in our sentence break predictors:
\begin{enumerate}
\item
  Making no changes to the statistical models used by the predictors,
  but adding the heuristic of predicting a sentence break following
  every paragraph ending in standard sentence-final punctuation.
\item
  Adding to our previous SVM feature sets a simple indicator feature
  with the value 1, if the sentence break candidate is also a paragraph
  break, and 0 otherwise.
\item
  Including a \verb|<P>| token at the position of each paragraph
  break in the tokenization of our training and test data.
\end{enumerate}

Where possible, we applied each of these methods separately to the
three sentence break predictors we evaluated on in-domain test data in
Section~6.2.3, and on out-of-domain test data in Chapter~7. These
three predictors are (a) predicting a sentence break whenever an
estimated paragraph break probablity exceeds a given threshold, (b) a
baseline SVM binary classifer using N-gram-based indicator features,
and (c) an SVM binary classifier augmenting the indicator features of
(b) with a feature whose value represents an estimate of the
non-sentence-break log probability based on our estimate of the
paragraph break probability (NSBLP SVM).

Applying method 1 to any sentence break predictor is
straightforward. Given that we only attempt to predict sentence breaks
following what might be standard sentence-final punctuation, we simply
first check whether we are at the end of a paragraph, and if so,
predict a sentence break; if we are not at the end of a paragraph, we
apply the unmodified predictor.

Since method 2 involves adding a feature to an SVM, it can only be
applied to the two predictors that use SVMs, predictor (b), the
baseline SVM classifier, and predictor (c), the NSBLP SVM
classifier. For these two predictors, the implementation is simple. We
first add the paragraph break feature to the the feature
representation of the training data and retrain the SVM model. When
the retrained model is applied to make a prediction, the paragraph
break feature is also added to the feature representation of each test
example.

Method 3 is a bit more complicated, because it applies differently to
the paragraph break probability model than it does to the SVMs. The
paragraph break probability model can make only limited use of
paragraph break tokens in test data, because it has to treat a
paragraph break token at the point of prediction as hidden. If a
paragraph break token at the point of prediction were visible to the
model, it would simply predict a paragraph break with probability 1.0
at every point where there was a paragraph break token, and with
probability 0.0 everywhere else. This would be useless as a predictor
of sentence breaks.

The paragraph break probability model can, however, make use of
paragraph break tokens near, but not at, the point of prediction,
which may be useful in certain circumstances. For example, Gillick
(2009) reports that 34\% of the errors made by his sentence break
predictor on his version of the Satz test set follow the token
\verb|U.S.|, noting ``Not only does it appear frequently as a sentence
boundary, but even when it does not, the next word is often
capitalized...'' However, if we observe a paragraph boundary
immediately preceding \verb|U.S.|, we can reasonably infer that this
instance of \verb|U.S.| is not followed by a sentence break, because
that would make \verb|U.S.| by itself a whole sentence, which is
highly unlikely. Gillick, however, does not represent paragaph breaks
in his test set, so he cannot use this information. On examination, we
find that there are 39 instances of paragraph-initial \verb|U.S.| in
our PTB WSJ training set and 19 instances in the Satz test set.  None
of these, in either the training set or test set, also end a sentence.

In Section~5.3, we explained how only one instance of the paragraph
break token \verb|<P>|, marking the point of prediction, is kept in
each N-gram we extract from the training data for the paragraph
break probability model. To apply method 3 to the paragraph break
probability model, we keep all instances of \verb|<P>| that occur
within the span of an N-gram, but if one occurs at the point of
prediction, it is kept hidden from the predictor, for the reasons
discussed above. In creating this modified paragraph break probability
model, we retune the $CC$ and $\beta$ hyperparameters, using our PTB
WSJ training set, as before. We find the optimal values to be $CC =
199$ and $\beta = 0.25$, which produced $305$ errors in predicting
sentence breaks on the PTB WSJ training set. This compares to values
of $CC = 225$ and $\beta = 0.25$, producing $315$ training set errors,
in our earlier experiments without the additional paragraph breaks
represented in the extracted N-grams.

In applying method 3 to the SVM classifiers, all instances of
\verb|<P>| are retained in the N-gram features extracted from both
training and test data.  The models are retrained on this modified
feature set.  Since paragraph break tokens at the point of
prediction are visible to the SVMs, the models easily learn that
paragraph breaks (following standard sentence-final punctuation) are
normally sentence breaks.

In applying method 3 to the NSBLP SVM classifier, we use an estimate
of non-sentence-break log probability derived from the paragraph break
probability model that is sensitive to paragraph breaks near, but not
at, the point of prediction. We use this paragraph break probability
model to re-estmate the mapping from paragraph break probability to
sentence break log odds, reoptimizing the hyper parameters $k$ and
$\gamma$ on our PTB WSJ training set, as described in
Section~6.2.1. The values of these parameters optimized for method 3
are $k = 22$ and $\gamma = 1.40$, compared to $k = 16$ and $\gamma =
1.41$ in our earlier experiments. Sentence break log odds is then
re-expressed as non-sentence-break log probability, as described in
Section~6.2.2.

\section{Comparing ways of representing paragraph break information}

The table below shows the results of applying our three methods of
using paragraph break information at prediction time to three
different sentence break predictors, evaluated on our Satz-3 in-domain
test set:

\begin{center}
  \begin{tabular}{||l|l|r|r|r|r|r||}
    \hline \hline
    Sentence  & Paragraph & & \multicolumn{1}{l|}{Para} & \multicolumn{1}{l|}{Para} & \multicolumn{1}{l|}{Total} & \multicolumn{1}{l||}{P-value} \\
    break     & break & \multicolumn{1}{l|}{Total} & \multicolumn{1}{l|}{internal} & \multicolumn{1}{l|}{final} & \multicolumn{1}{l|}{error} & \multicolumn{1}{l||}{compared} \\
    predictor & signal & \multicolumn{1}{l|}{errors} & \multicolumn{1}{l|}{errors} & \multicolumn{1}{l|}{errors} & \multicolumn{1}{l|}{rate} & \multicolumn{1}{l||}{to heuristic} \\
    \hline \hline
    PBP-based & None               & 158    & 133      & 25     & 0.78\% & $<$ 0.0001 \\
    \hline
                    & Heuristic          & 133    & 133      & 0      & 0.66\% &  \\
    \hline
                    & \verb|<P>| tokens  & 160    & 134      & 26     & 0.79\% & $<$ 0.0001 \\
    \hline \hline
    Baseline SVM    & None               & 65     & 48       & 17     & 0.32\% & $<$ 0.0001 \\
    \hline
                    & Heuristic          & 48     & 48       & 0      & 0.24\% &  \\
    \hline
                    & \verb|<P>| feature & 42     & 42       & 0      & 0.21\% & 0.2386 \\
    \hline
                    & \verb|<P>| tokens  & 45     & 45       & 0      & 0.22\% & 0.6464 \\
    \hline \hline
    NSBLP SVM       & None               & 62     & 44       & 18     & 0.31\% & $<$ 0.0001 \\
    \hline
                    & Heuristic          & 44     & 44       & 0      & 0.22\% &  \\
    \hline
                    & \verb|<P>| feature & 43     & 42       & 1      & 0.21\% & 1.0000 \\
    \hline
                    & \verb|<P>| tokens  & 41     & 41       & 0      & 0.20\% & 0.6056 \\
    \hline \hline
  \end{tabular}
\end{center}

This table displays evaluation results for each combination of sentence
break predictor and method of using paragraph break information,
including a baseline of using no paragraph break information.  For
each combination, the table shows the number of sentence break
prediction errors, in total as well as broken down into
paragraph-internal errors and paragraph-final errors. The table also
shows the total error rate in terms of the ratio of errors to true
sentence breaks (signaled by standard sentence-final punctuation),
expressed as a percentage. In the final column, for each sentence
break predictor, the table shows the p-value according to McNemar's
test for the comparison between the heuristic method of using
paragraph break information and the other methods applicable to that
predictor.

For the sentence break predictor based solely on the paragraph break
probability estimate, the paragraph break indicator feature is not
applicable, so there is no result for that method of using paragraph
break information. Also, since paragraph breaks at the point of
prediction must be kept hidden, adding \verb|<P>| tokens makes almost
no difference to prediction accuracy. Using the heuristic method of
predicting a sentence break whenever a paragraph ends in standard
sentence-final punctuation, however, eliminates all paragraph-final
errors for this test set. For this predictor, the p-value for
McNemar's test comparing the heuristic method to either of the others
is less than 0.0001, indicating that the differences are highly
significant.

The results for the baseline SVM and the NSBLP SVM sentence break
predictors are very similar to each other. With either predictor, any
of the three methods of incorporating paragraph break information is
noticably more accurate than ignoring paragraph break information. The
methods using a \verb|<P>| feature or \verb|<P>| tokens produces
slighly fewer total errors than the heuristic method, but the
differences between them and the heuristic method are not
statistically significant according McNemar's test. The difference
between the heuristic method and ignoring paragaph break information,
however, is highly significant.

The next table, below, shows the results of applying our three methods of
using paragraph break information at prediction time to the same three
sentence break predictors, evaluated on our LDC English Web Treebank
out-of-domain test set:

\begin{center}
  \begin{tabular}{||l|l|r|r|r|r|r||}
    \hline \hline
    Sentence  & Paragraph & & \multicolumn{1}{l|}{Para} & \multicolumn{1}{l|}{Para} & \multicolumn{1}{l|}{Total} & \multicolumn{1}{l||}{P-value} \\
    break     & break & \multicolumn{1}{l|}{Total} & \multicolumn{1}{l|}{internal} & \multicolumn{1}{l|}{final} & \multicolumn{1}{l|}{error} & \multicolumn{1}{l||}{compared} \\
    predictor & signal & \multicolumn{1}{l|}{errors} & \multicolumn{1}{l|}{errors} & \multicolumn{1}{l|}{errors} & \multicolumn{1}{l|}{rate} & \multicolumn{1}{l||}{to heuristic} \\
    \hline \hline
    PBP-based & None               & 590    & 470      & 120    & 4.50\% & $<$ 0.0001 \\
    \hline
                    & Heuristic          & 470    & 470      & 0      & 3.58\% &  \\
    \hline
                    & \verb|<P>| tokens  & 591    & 475      & 116    & 4.51\% & $<$ 0.0001 \\
    \hline \hline
    Baseline SVM    & None               & 957    & 607      & 350    & 7.30\% & $<$ 0.0001 \\
    \hline
                    & Heuristic          & 607    & 607      & 0      & 4.63\% &  \\
    \hline
                    & \verb|<P>| feature & 657    & 652      & 5      & 5.01\% & $<$ 0.0001 \\
    \hline
                    & \verb|<P>| tokens  & 655    & 655      & 0      & 5.00\% & $<$ 0.0001 \\
    \hline \hline
    NSBLP SVM       & None               & 655    & 474      & 181    & 5.00\% & $<$ 0.0001 \\
    \hline
                    & Heuristic          & 474    & 474      & 0      & 3.62\% &  \\
    \hline
                    & \verb|<P>| feature & 499    & 485      & 14     & 3.81\% & 0.0012 \\
    \hline
                    & \verb|<P>| tokens  & 504    & 501      & 3      & 3.84\% & 0.0004 \\
    \hline \hline
  \end{tabular}
\end{center}

The layout of this table is identical to the previous one. The only
difference is that it gives results for the out-of-domain test set,
rather than the in-domain-test set. For the sentence break predictor
based solely on the paragraph break probability estimate, the
comparison among the different approaches to using paragraph break
information is qualitatively similar to what we observed with the
in-domain test set, although the error rates are substantially
higher. The method of adding paragraph break tokens makes almost no
difference compared to ignoring paragraph break information, and the
heuristic method is substantially better, with the differences highly
significant according to McNemar's test.

The results for the baseline SVM and NSBLP SVM sentence break
predictors on the out-of-domain test set, however, are qualitatively
very different from the results on the in-domain test set.  For both
SVMs, the heuristic method is better than any of the other approaches
to using paragraph break information by a statistically significant
margin. Combining the results from both test sets, we observe that
heuristic method is by far the best on the out-of-domain test set, and
not significantly worse than any other method on the in-domain test
set. We therefore conclude that according to these experiments, the
heuristic method of predicting a sentence break whenever a paragraph
ends in standard sentence-final punctuation is better overall than any
other approach we have tried.

\section{Comparing sentence break predictors incorporating paragraph break information}

Having concluded that our heuristic method of using paragraph break
information to help predict sentence breaks is the overall best out of
the ones we have tried, we now turn to the question of whether there
is an overall best approach to sentence break prediction incorporating
our heuristic method of using paragraph break information at
prediction time. Recall that we did not find an overall best approach
to sentence break prediction when paragraph break information was not
used at prediction time. Our sentence break predictor based solely on
a paragraph break probability estimate was the worst when tested on
our in-domain test set (Section~6.2.3), but it was the best when
tested on our out-of-domain test set (Chapter~7), by statistically
significant margins in both cases.

The table below extracts information from the two tables in the
previous section to compare our three principal sentence break
predictors incorporating the heuristic method of using paragraph break
information at prediction time, including additional significance
tests comparing the NSBLP SVM to the other two approaches:

\begin{center}
  \begin{tabular}{||l|l|r|r|r|r|r||}
    \hline \hline
    & Sentence  & & \multicolumn{1}{l|}{Para} & \multicolumn{1}{l|}{Para} & \multicolumn{1}{l|}{Total} & \multicolumn{1}{l||}{P-value} \\
    & break & \multicolumn{1}{l|}{Total} & \multicolumn{1}{l|}{internal} & \multicolumn{1}{l|}{final} & \multicolumn{1}{l|}{error} & \multicolumn{1}{l||}{compared to} \\
    Test set & predictor & \multicolumn{1}{l|}{errors} & \multicolumn{1}{l|}{errors} & \multicolumn{1}{l|}{errors} & \multicolumn{1}{l|}{rate} & \multicolumn{1}{l||}{NSBLP SVM} \\
    \hline \hline
    In domain & PBP-based & 133    & 133      & 0      & 0.66\% & $<$ 0.0001 \\
    \hline
    & Baseline SVM    & 48     & 48       & 0      & 0.24\% & 0.6171 \\
    \hline
    & NSBLP SVM       & 44     & 44       & 0      & 0.22\% &  \\
    \hline \hline
    Out of domain & PBP-based & 470    & 470      & 0      & 3.58\% & 0.8201 \\
    \hline
    & Baseline SVM    & 607    & 607      & 0      & 4.63\% &  $<$ 0.0001 \\
    \hline
    & NSBLP SVM       & 474    & 474      & 0      & 3.62\% &  \\
    \hline \hline
  \end{tabular}
\end{center}

On the in-domain test set, we observe that NSBLP SVM is not
significantly worse than the baseline SVM, and very significantly
better than the predictor based solely on a paragraph break
probability estimate. On the out-of-domain test set, we observe that
NSBLP SVM is not significantly worse than the
paragraph-break-probability-based predictor and very significantly
better than the baseline SVM. All p-values are once again computed
according to McNemar's test. We therefore conclude that the best
overall sentence break predictor in our experiments, over both our
in-domain and out-of-domain test sets, is the NSBLP SVM incorporating
our heuristic method of using paragraph break information at
prediction time.

\chapter{Effects of the amount of annotated data}

In Chapters~5 and 6, we saw that our sentence break predictor based
solely on paragraph break probability estimates is quite accurate in
predicting sentence breaks in clean, well-edited text like that in The
Wall Street Journal, but we also saw that a directly supervised
classifier, such as Gillick's (2009) SVM or our baseline SVM, can be
even more accurate when trained and tested on the same data source.

These results, however, were obtained using a fairly large amount of
data annotated with sentence breaks for training and tuning thousands
of parameters in the directly supervised models. Our
paragraph-break-probability-based predictor, on the other hand, uses
only two hyperparameters that need to be optimized on manually
annotated data. This suggests a possible situation in which a
paragraph-break-probability-based predictor might outperform a
directly supervised classifier, even on in-domain data: building a
sentence break predictor given a large amount of unannotated data, but
only a small amount of annotated data. We explore that situation here.

Since there is a relatively large amount of data annotated with
sentence breaks available in English, this situation is more likely to
arise for other, less studied languages. Nevertheless, we will simply
explore here the effects of cutting down on the amount of manually
annotated English data we use. Our results should therefore be
validated for other languages before being relied on.

Our directly supervised SVM classifiers are trained on 35,433
sentences\footnote{These counts include only sentences ending in
standard sentence-final punctuation.} from the PTB WSJ corpus
annotated with sentence breaks, with another 5,333 PTB WSJ annotated
sentences used for development and hyperparameter tuning. Our
paragraph-break-probability-based sentence break predictors use the
35,433 training sentences to tune two hyperparameters, and the 5,333
sentence tuning/development data set is not used. In this chapter, we
explore how well could we do if we had only a much smaller amount of
text manually annotated with sentence breaks, just 1000 sentences.

We use as our manually annotated 1000 sentences the first 1000
sentences of section 22 of the PTB WSJ corpus, which is included in
our tuning/development data set. We use our version of this data with
paragraph breaks also annotated, to build sentence break predictors
incorporating paragraph break information at prediction time, using our
heuristic method. For the paragraph-break-probability-based sentence
break predictor this is straightforward. As in Chapter~5, we use N-gram
counts from the unannotated data of the LDC English Gigaword Fifth
Edition corpus to define our model, but with the count-cutoff $CC$ and
paragraph break probability threshold $\beta$ optimized on just 1000
sentences of annotated data, instead of all 35,433 sentences of our
PTB WSJ training set. With the smaller annotated data set, we obtain
the optimized values $CC = 250$ and $\beta = 0.04$, compared to $CC =
225$ and $\beta = 0.025$ using the larger annotated data set.

For comparison, we also retrain our baseline SVM sentence break
predictor using only 1000 annotated sentences. This presents us with a
dilemma, since in Chapter~6 we used two sets of annotated data to
train our SVM classifiers, the 35,433-sentence PTB WSJ training set to
optimize model parameters by stochastic subgradient descent, and the
independent tuning set of 5,333 PTB WSJ sentences to optimize the
number of training epochs and the learning rate. With only 1000
total annotated sentences, however, holding out any data for
optimizing hyperparameters seems problematic. We therefore evaluate
two different ways of training the baseline SVM.

We first train the baseline SVM using 700 sentences as the primary
training data, with 300 sentences held out to tune the number of
epochs and learning rate. We next tried using all 1000 sentences both
for training and also for tuning the number of epochs and learning
rate, but this proved unworkable because we found that we eventually
reached convergence on the training data (0 errors and 0 hinge loss)
for any learning rate we tried. So instead, we train to convergence on
all 1000 sentences, using as the learning rate the smallest negative
power of 2 that results in convergence in fewer than 200 epochs, which
turns out to be a learning rate of $2^{-8}$.

The following table shows the accuracy on our Satz-3 in-domain test
set of these three sentence break predictors built using only 1000
manually annotated sentences, with sentence break predictors using the
full PTB WSJ training and tuning/development sets included for
comparison:
\begin{center}
  \begin{tabular}{||l|l|r|r|r||}
    \hline \hline
    Sentence  & Training & & & \multicolumn{1}{l||}{P-value}\\
    break     & and tuning & \multicolumn{1}{l|}{Error} & \multicolumn{1}{l|}{Error} & \multicolumn{1}{l||}{compared to} \\
    predictor & sent counts & \multicolumn{1}{l|}{count} & \multicolumn{1}{l|}{rate} & \multicolumn{1}{l||}{PBP-1000} \\
    \hline \hline
    PBP-based & 0/35433    & 133  & 0.66\% & 0.1853 \\
    \hline
                    & 0/1000     & 122  & 0.60\% &  \\
    \hline \hline
    Baseline SVM    & 35433/5333 & 48   & 0.24\% & $<$ 0.0001 \\
    \hline
                    & 700/300    & 187  & 0.92\% & $<$ 0.0001 \\
    \hline
                    & 1000/0     & 166  & 0.82\% & 0.0059 \\
    \hline \hline
  \end{tabular}
\end{center}

In this table, we show numbers of errors and error rates of the
different sentence break predictors, along with p-values, according to
McNemar's test, compared to the paragraph-break-probability-based
predictor with hyperparameters tuned on 1000 annotated sentences.  The
table shows that reducing from 35,433 to 1000 the number of annotated
sentences used to tune the hyperparameters of the
paragraph-break-probability-based predictor actually reduces the
number of errors slightly on our in-domain test set, although not
significantly so. The table also shows that either way of training the
baseline SVM classifier on only 1000 annotated sentences results in
significantly worse accuracy than the
paragraph-break-probability-based predictor with hyperparameters tuned
on 1000 annotated sentences.

The next table shows the same information when testing on our
out-of-domain LDC English Web Treebank out-of-domain test set:
\begin{center}
  \begin{tabular}{||l|l|r|r|r||}
    \hline \hline
    Sentence  & Training & & & \multicolumn{1}{l||}{P-value}\\
    break     & and tuning & \multicolumn{1}{l|}{Error} & \multicolumn{1}{l|}{Error} & \multicolumn{1}{l||}{compared to} \\
    predictor & sent counts & \multicolumn{1}{l|}{count} & \multicolumn{1}{l|}{rate} & \multicolumn{1}{l||}{PBP-1000} \\
    \hline \hline
    PBP-based & 0/35433    & 470  & 3.58\% & 0.0492 \\
    \hline
                    & 0/1000     & 492  & 3.75\% &  \\
    \hline \hline
    Baseline SVM    & 35433/5333 & 607  & 4.63\% & $<$ 0.0001 \\
    \hline
                    & 700/300    & 776  & 5.92\% & $<$ 0.0001 \\
    \hline
                    & 1000/0     & 836  & 6.38\% & $<$ 0.0001 \\
    \hline \hline
  \end{tabular}
\end{center}
This table shows that reducing the amount of tuning data to 1000
annotated sentences increases the number of errors made by the
paragraph-break-probability-based predictor on this data set by 4.7\%,
which is just barely significant at the 0.05 level according to
McNemar's test. The same reduction in annotated data increases the
number of errors made by the baseline SVM by 27.8\% or 37.7\%,
depending on how the 1000 annotated sentences are used.

With a larger amount of annotated data, we saw previously that the
baseline SVM classifier was much more accurate than
paragraph-break-probability-based predictor on in-domain data, while
on out-of-domain data, the paragraph-break-probability-based predictor
was more accurate. While we might find a better way to divide a
limited amount of annotated data between training and tuning, the
results shown here suggest that the paragraph-break-probability-based
predictor will be the better choice even on in-domain data, when only
a small amount of annotated data is available.

\chapter{Sentence breaks without white space}

As mentioned in Chapter~1 and Appendix~A.4, we noticed in
preprocessing the LDC Web Treebank corpus, which we use as our
out-of-domain test data, that there were many cases of sentence breaks
without white space where we would normally expect it, for example:
\begin{quote}
  \verb|Contact the Indian railways.You might end up as a no show|
  \verb|passenger.|
\end{quote}
Despite the lack of white space, there clearly should be a sentence
break following \verb|railways.| In conventionally formatted English
text, with the exception of following some ellipses (\verb|...|),
every sentence break occurs at white space. In the very informal LDC
Web Treebank corpus, however, we counted 74 sentence breaks that were
neither at white space nor following an ellipsis.

Most sentence break predictors, including ours, do not even consider
the possibility of a sentence break except at white space or
following an ellipsis. In this chapter, we experiment with extending
our preferred sentence break predictor so that it can predict sentence
breaks even when there is no white space or ellipsis. Predicting
sentence breaks without white space\footnote{From this point onward,
we will drop qualifications such as ``unless following an ellipsis''
and take it as given that we are not treating that case here. Recall
that, in fact, we deal with sentence breaks following an ellipsis by
ensuring during text normalization (see Section~5.1) that there is
always white space following an ellipsis.} in out-of-domain data,
when there are almost none in our in-domain training data, presents us
with a problem. If we just retrain on clean in-domain data, but allow
for the possibility of sentence breaks without white space, the model
will simply learn that this never happens.

We address this problem by noting that when a sentence break occurs
without a white space, there would be a space if the text were
formatted conventionally. We therefore treat the problem as that of
inserting spaces where they have been omitted, and afterwards we apply
our normal sentence break predictor trained on text largely without
missing spaces.  For the purposes of this study, we use the predicted
spaces only as an aid to predicting sentence breaks. After predicting
sentence breaks in a piece of text, we project the positions of the
sentence breaks back onto the original text, without the inserted
spaces. Therefore, our missing-space predictor is tuned to optimize
sentence break prediction, rather than missing-space prediction
itself.

\section{Missing-space prediction}

The missing-space problem can be approached in a way nearly identical
to the one we have taken in our paragraph-break-probability-based
sentence break predictor. We motivated the latter by assuming that
paragraph breaks are approximately a random sample of sentence
breaks. Another way of describing the same situation is to say that
there should be a paragraph break at every sentence break, but a
random subset of those paragraph breaks are missing. Similarly, we
now say that there should be white space at every sentence break,
but a random subset of those white spaces are missing. This lets us
predict missing spaces following potential sentence-final
punctuation using exactly the same apparatus we have developed for
predicting sentence breaks based on estimated sentence break
probability.

To predict missing spaces, we create an N-gram model for the
probability of white space following possible sentence-final
punctuation, trained on an unannotated corpus, just as we did for the
paragraph break probability model. As before, we use use the LDC
English Gigaword Fifth Edition corpus to train our model. However,
rather than treating the presence or absence of a paragraph break
following possible sentence-final punctuation as a hidden token to be
predicted, now we treat the presence or absence of white space
following possible sentence-final punctuation as a hidden token.

The context we use to predict hidden white space consists of all the
consecutive non-space characters preceding and following the point of
prediction. So, for positive examples (where white space is present),
we use strings of non-space characters ending in possible standard
sentence-final punctuation\footnote{Recall that we define this as a
period, question mark, or exclamation point, possibly followed by one
or more closing quotes and/or closing brackets.}, followed by white
space, followed by another string of non-space characters, with the
point of prediction at the white space.  Negative examples (where
there is no white space) consist of a single string of non-space
characters between white spaces, containing possible standard
sentence-final punctuation followed by additional non-space
characters, with the point of prediction immediately following the
possible standard sentence-final punctuation.  Note that a single
string of non-space characters can yield more than one negative
example, if more than one substring of possible sentence-final
punctuation occurs within the string of non-space characters.

Both positive and negative examples are filtered by a set of regular
expressions designed to identify certain instances of punctuation
marks not followed by white space that should not be considered
candidates for missing spaces, in hyphenated abbreviations, decimal
numbers, strings of multiple initials, inflected forms of
abbreviations, email addresses, and computer filenames, domain names,
URLs, and related expressions. Details of how the examples are
filtered are presented in Appendix~C.

We build the white-space-probability model by first applying to the
examples extracted as described above from the LDC English Gigaword
Fifth Edition corpus the text normalization exactly as described in
Section~5.1, followed by the tokenization described in Section~5.2,
modified so that (1) special tokens are inserted for the presence or
absence of white space following possible sentence-final punctuation,
rather than for the presence or absence of paragraph breaks following
possible sentence-final punctuation, and (2) there are no underscore
characters attached to the characters immediately preceding and
following the point of prediction when there is no white space.

From the normalized and tokenized examples, we then extract N-gram
counts as described in Section~5.3. We collect counts for N-grams up
to a length of 10, and we do not exclude text occurring within
quotations as we did for paragraph breaks. The probability of white
space at a particular point is estimated from the N-gram counts just
as described in Section~5.4, including all the refinements described
in Section~5.4.2 of our basic approach laid out in
Section~5.4.1. Using this model, we predict a missing space whenever
the estimated probability of white space exceeds a given threshold, at
a text position where there is no white space following possible
sentence-final punctuation.

\section{Selection of training and tuning data}

Using the method described above to predict missing spaces requires
assigning values to two hyperparameters: a count cutoff $CC$ that
limits which N-grams are used in the white-space-probability estimate,
and a probability threshold $\beta$ above which we predict that a
space should be inserted.

To optimize these hyperparameters requires annotated tuning data. If
we wanted to optimize the hyperparamters for predicting whether a
space is actually present, we could use the data we already
have. However, what we want to optimize for is reducing sentence break
prediction error by selecting positions to insert a space where there
is no space in the original text, but where our sentence break
predictor would correctly predict a sentence break if a space were
inserted. This requires specially selected and annotated data.

To create the annotated data we need for tuning the hyperparameters,
we set aside roughly 10\% of the LDC English Gigaword Fifth Edition
corpus. This corpus consists of 1010 files, so we set aside every
tenth file to extract tuning data from, while extracting our training
data from the remaining files. From the tuning data, we extract all
the ``negative examples'' consisting of strings of consecutive
non-space characters containing possible standard sentence-final
punctuation not at the end of the string. These are filtered to remove
cases involving embedded punctuation marks that appear to be something
other than sentence-final punctuation, as described in Appendix~C.  If
the text of the Gigaword corpus were perfectly edited, none of these
examples would be cases where spaces should be inserted, but it turns
out that there are enough examples with spaces missing that we
can use this data for tuning our hyperparameters.

This selection procedure yields 24,176 unique examples, of which 6,336
have more than one occurrence in the tuning data.  In optimizing $CC$
and $\beta$ over this set of tuning examples, we choose to count each
unique example only once, even if it has multiple occurrences in the
tuning data. Our reasoning is that weighting examples by number of
occurrences might overfit to the particular sources that our tuning
data is obtained from. For instance, there are 37 occurrences of
\verb|Ky.Busch| in our tuning data, apparently referring to the race
car driver Kyle Busch. That number of references using that particular
abbreviation of his name seems rather idiosyncratic to a particular
source, so we choose not to give it extra weight for having multiple
occurrences.

For all the tuning examples, we insert a space following the possible
standard sentence-final punctuation and apply our sentence break
predictor, to find out whether we would predict a sentence break if a
space were present.\footnote{The sentence break predictor used in this
process was not the one that ultimately turned out to be our best
version, but it was close enough to the best that it did not seem to
be worth the effort to repeat the tuning data selection and annotation
based on the best version.} Of the 24,176 unique examples in our
tuning data, 14,328 are predicted to be sentence breaks with an
inserted space. Only these examples for which our sentence break
predictor predicts a sentence break are used to tune the missing-space
hyperparameters, because unless a sentence break would be predicted,
inserting a space makes no difference to the results of sentence break
prediction.

\section{Annotating tuning data}

We annotate tuning examples with one of the following three labels:
\begin{itemize}
\item
  Likely a sentence break
\item
  Likely not a sentence break
\item
  Don't care
\end{itemize}
We annotate the examples as to whether they appear to be sentence
breaks or not, rather than missing spaces or not, because sentence
break prediction accuracy is what we are trying to optimize. Thus
``likely not a sentence break'' includes both cases where a space
should not be inserted, and cases where a space should be inserted,
but which are not sentence breaks.

``Don't care'' examples are those we choose to exclude from the tuning
data for a variety of reasons, including
\begin{itemize}
\item
  Markup that should have been removed in preprocessing
\item
  Tables of nonwords or other non-sentence-text
\item
  Nontext binary data
\item
  Languages other than English
\item
  Garbled character strings
\item
  Examples that we find too hard to decide
\end{itemize}
More details about how we classify potential tuning examples are
presented in Appendix~D.

Given a set of tuning examples labeled either ``likely a sentence
break'' or ``likely not a sentence break,'' it is straightforward to
perform a search over possible combinations of the count cutoff $CC$
and probability threshold $\beta$ to find the combinations that yield
the greatest number of correct sentence break predictions on this
set. Even though these $CC$ and $\beta$ hyperparameters directly
control only the insertion of spaces, in our tuning set each predicted
space corresponds to a predicted sentence break, since we are
considering only examples that would be predicted to be sentence
breaks if a space were present.

\subsection{Assumptions that justify reduced data annotation}

Labeling all 14,328 possible tuning examples would be a substantial
amount of work, but it turns out that to optimize $CC$ and $\beta$ we
only need to label a small fraction of the data, given some reasonably
plausible assumptions. The main assumption we rely on is that, at text
positions where potential sentence-final punctuation is not followed
by white space, the probability of a sentence break increases
monotonically with the estimated white-space probability.  This main
assumption follows from three subsidiary assumptions about text
positions where potential sentence-final punctuation is not followed
by white space:
\begin{itemize}
\item
  The probability of a sentence break without a missing space is 0.
\item
  The probability of a sentence break and the probability of the
  estimated white space probability having a particular value are
  conditionally independent, given that a space is missing.
\item
  The probability of a missing space increases monotonically
  with the estimated white-space probability.
\end{itemize}
The first subsidiary assumption is simply a fact about standard
English orthography. The second subsidiary assumption means that
knowing whether there actually is a space missing exhausts the
information that the white-space probability estimate can provide in
predicting whether there is a sentence break. The third subsidiary
assumption should be true if missing spaces are essentially random and
our white-space-probability model is reasonably accurate.

The first and second subsidary assumptions justify the following
derivation, in which $S\!B$ represents the event of a sentence break,
$M\!S = 1$ represents the event of a missing space, $M\!S = 0$ represents
the event of no missing space, and $EW\!S\!P = x$ represents the event of
the estimated white-space probability having the value $x$:
\begin{eqnarray*}
  p(S\!B \, | \, EW\!S\!P = x) & = & p(S\!B, \, M\!S = 1 \, | \, EW\!S\!P = x) + \\
  & & p(S\!B, \, M\!S = 0 \, | \, EW\!S\!P = x) \\
  & = & p(S\!B, \, M\!S = 1 \, | \, EW\!S\!P = x) \\
  & = & p(S\!B \, | \, M\!S = 1 , \, EW\!S\!P = x) \times p(M\!S = 1 \, | \, EW\!S\!P = x) \\
  & = & p(S\!B \, | \, M\!S = 1 ) \times p(M\!S = 1 \, | \, EW\!S\!P = x)
\end{eqnarray*}

The first line of the derivation expresses the conditional probability
of a sentence break, given the value of the estimated white-space
probability, as the sum of the conditional probabilities of two
subcases, with and without a missing space. The second line follows
because the joint probability of a sentence break and no missing space
is 0. The third line factors the joint probability of a sentence break
and a missing space, given the value of the estimated white-space
probability, into the product of two conditional probabilities. The
last line simplifies one of the conditional probabilities, due to the
conditional independence of the probability of a sentence break and
the probability of the estimated white-space probability having a
particular value, given that there is a missing space. This leaves us
with the following equation for the probability of a sentence break
given the value of the estimated white-space probability:
\begin{eqnarray*}
p(S\!B \, | \, EW\!S\!P = x) & = & p(S\!B \, | \, M\!S = 1 ) \times p(M\!S = 1 \, | \, EW\!S\!P = x)
\end{eqnarray*}
Since $p(S\!B \, | \, M\!S = 1 )$ does not vary with the estimated
white-space probability, it follows that if the probability of a
missing space increases monotonically with the estimated white-space
probability, then the probability of a sentence break increases
monotonically with the estimated white-space probability.

For a given value of the count cutoff $CC$ (which affects the
estimation of white-space probabilities), this monotonicity property
implies that the optimal threshold $\beta$ will be the value of the
estimated white-space probability that corresponds to the value for
the probability of a sentence break closest to 0.5.\footnote{Closest
to 0.5, rather than equal to 0.5, because we don't assume continuity,
so there may be no white-space probability that corresponds exactly to
a probability of 0.5 for a sentence break.} To see that this is true,
let $T$ be a representative set of examples that would be predicted to
be sentence breaks if a space were inserted, let $a$ be the estimated
white-space probability corresponding to the value for the probability
of a sentence break closest to 0.5, and let $b$ be another estimated
white-space probability sufficiently greater than $a$ that the subset
of $T$ having estimated white-space probabilities between $a$ and
$b$\footnote{Strictly speaking, by ``between $a$ and $b$'' we mean
greater than $a$ and less than or equal to $b$.} is nonempty.

Let $T'$ be the subset of $T$ having estimated white-space
probabilities between $a$ and $b$.  We expect more than half of of the
members of $T'$ to be sentence breaks, because the members of $T'$ all
have estimated white-space probabilities greater than $a$, and $a$
corresponds to the sentence-break probability closest to 0.5.
Therefore, by our monotonicity property, the members of $T'$ all have
a sentence-break probability greater than 0.5. If we used $b$ instead
of $a$ as the value of the threshold $\beta$, all the examples between
$a$ and $b$ would go from being predicted to be sentence breaks to
being predicted not to be sentence breaks, and since we expect more
than half of these to actually be sentence breaks, the expected number
of sentence break prediction errors over $T'$ would increase.  Now,
with either $a$ or $b$ as the value of our threshold $\beta$, all
examples in $T$ with estimated white-space probability greater than
$b$ will be predicted to be sentence breaks, and all examples in $T$
with estimated white-space probability less than or equal to $a$ will
be predicted not to be sentence breaks. So, if changing the value of
$\beta$ from $a$ to $b$ increases the expected number of sentence
break prediction errors over $T'$, then it also increases the expected
number of errors over all of $T$.

Thus, no value of $\beta$ greater than $a$, corresponding to the value
for the probability of a sentence break closest to 0.5, has a greater
expected number of correct sentence break predictions over $T$ as $a$.
A parallel argument shows that no value of $\beta$ less than $a$ has a
greater expected number of correct sentence break predictions over $T$
as $a$. This is true for any representative set $T$ of examples that
would be predicted to be sentence breaks if a space were
inserted. Hence, the value of $\beta$ corresponding to the probability
of a sentence break closest to 0.5 would be an optimal threshold to
use for improving sentence break prediction accuracy by inserting
missing spaces.

\subsection{How to reduce the amount of annotated data needed to
  optimize the value of $\beta$}

How does all of the above help us limit the amount of data we need to
annotate? It tells us that, if we can find a range of values for
$\beta$ reasonably certain to contain the estimated white-space
probability corresponding to the value closest to 0.5 for the
probability of a sentence break, we don't need to consider, or label,
any examples whose estimated white-space probability falls outside of
that range.

As previously noted, if we are comparing two alternatives $a < b$ as
possible values for the threshold $\beta$, any examples with estimated
white-space probabilities less than $a$ or greater than $b$ will
receive the same prediction with either $a$ or $b$ as the value of
$\beta$, and thus don't need to be looked at to decide whether $a$ or
$b$ is a better choice. So, if we are reasonably certain that the
optimal value of $\beta$ lies within a certain range of values,
labeling and optimizing $\beta$ over the set of tuning examples with
estimated white-space probabilities lying within that range should
give the same result as labeling and optimizing $\beta$ over all our
potential tuning examples.

To find a range of estimated white-space probabilities reasonably
certain to contain the estimated white-space probability corresponding
to the sentence break probability closest to 0.5, we proceed as
follows:
\begin{itemize}
  \item
    First we sort all 14,328 possible tuning examples by their
    estimated white-space probability.
  \item
    We separate the sorted list of examples into two sorted sublists,
    a ``high-probability list'' consisting of all the examples with an
    estimated white-space probability greater than 0.5, and a
    ``low-probability list'' consisting of all the examples with an
    estimated white-space probability less than or equal to 0.5.
  \item
    We then label examples selected in order of increasing probability
    from the high-probability list, and in order of decreasing
    probability from the low-probability list, until we have 150
    examples labeled ``likely a sentence break'' and 150 examples
    labeled ``likely not a sentence break''. Examples labeled ``don't
    care'' are disregarded.
  \item
    We strive to keep the set of labeled examples equally balanced by
    choosing an example to be labeled from the high-probability list
    whenever we have fewer examples labeled ``likely a sentence
    break'' than labeled ``likely not a sentence break''; otherwise,
    we choose an example to be labeled from the low-probability list.
  \item
    If we exceed 150 examples with one label before reaching 150
    examples with the other label, we discard an example with the most
    extreme estimated white-space probability from the larger subset,
    so as not to exceed 150 examples with either label.
  \item
    We then select a value for the threshold $\beta$ that empirically
    optimizes the sentence break prediction error rate over this set
    of 300 labeled examples, which we refer to as the ``critical
    examples''.
  \item
    Finally, we verify that the range of estimated white-space
    probabilities spanned by the critical examples is large enough to
    be very likely to contain the optimal value for $\beta$. To do
    this we first partition the 300 critical examples into those with
    estimated white-space probabilities greater than our selected
    value of $\beta$ and those with estimated estimated white-space
    probabilities less than or equal to the selected value of
    $\beta$. We then look at the number of likely sentence breaks in
    each partition, relative to the number of examples in that
    partition, computing the P-value for the observed counts compared
    to a null hypothesis of a 0.5 sentence break probability,
    according to a one-tailed binomial test. If the P-value for both
    partitions is very low, then the probability is also very low that
    the optimal value of $\beta$ lies outside the range of values
    covered by by the 300 critical examples.
\end{itemize}

To understand the reasoning behind the statistical test in the final
step, consider the results we obtained for $CC = 3$. Annotating all
the examples with estimated white-space probabilities from
approximately 0.1107 through 0.8822, we found 300 critical examples
that we could label either ``likely a sentence break'' or ``likely not
a sentence break'', with another 520 examples that we labeled ``don't
care''. For the 300 critical examples, we found the empirically
optimal value for the threshold $\beta$ to be 0.5902, resulting in 103
sentence break prediction errors. There were 47 likely sentence breaks
out of 141 examples with an estimated white-space probability less
than or equal to 0.5902 and 103 likely sentence breaks out of 159
examples with an estimated white-space probability greater than
0.5902.

The P-value, according to a one-tailed binomial test with a null
hypothesis probability of 0.5, for observing 47 sentence breaks
out of the 141 examples with estimated white-space probabilities
between 0.1107 and 0.5902 is approximately 0.00012. That is, if the
probability of a sentence break is 0.5 for an example with an
estimated white-space probability between 0.1107 and 0.5902, the
chances of observing 47 or fewer sentence breaks out of 141
examples is approximately 0.00012. Recalling that the optimal value of
$\beta$ is the estimated white-space probability corresponding to a
sentence break probability closest to 0.5, and that the sentence break
probability increases monontonically with the white-space probability,
for the optimal value of $\beta$ to be less than 0.1107, the
probability of a sentence break would have to be greater than 0.5 for
all estimated white-space probabilities between 0.1107 and 0.5902,
which would make observing 47 or fewer sentence breaks out of
141 examples even less likely than 0.00012.

Similarly, the P-value, according to a one-tailed binomial test with a
null hypothesis probability of 0.5, for observing 103 sentence breaks
out of the 159 examples with estimated white-space probabilities
between 0.5902 and 0.8822 is approximately 0.000047; meaning that, if
the probability of a sentence break is 0.5 for an example with an
estimated white-space probability between 0.5902 and 0.8822, the
chances of observing 103 or more sentence breaks out of 141 examples
is approximately 0.000047.  For the optimal value of $\beta$ to be
greater than 0.8822, the probability of a sentence break would have to
be less than 0.5 for all estimated white-space probabilities between
0.5902 and 0.8822, which would make observing 103 or more sentence
breaks out of 159 examples even less likely than 0.000047. Thus the
observed distribution of likely sentence breaks over these 300 critical
examples would be extremely unlikely if the optimal value of
$\beta$ were outside the range of estimated white-space probabilities
spanned by these examples, which justifies our not labeling more of
the potential tuning examples.

\section{Optimizing $CC$ and $\beta$ simultaneously}

The procedure outlined above for optimizing the value of the
estimtated-white-space-probability threshold $\beta$, while limiting
the amount of data that we need to label, depends on the value of the
count cutoff $CC$, since changing the count cutoff changes the
estimated white-space probabilities.  To optimize both $CC$ and
$\beta$, we repeat this procedure for all values of $CC$ from 2
through 30. For any of these count-cutoff values, the worst
(i.e. highest) P-values we obtain in the final verifiction step are
0.00324 for the low-probability subset of critical examples and
0.00362 for the high-probability subset (for both $CC = $ 23 and $CC =
$ 24), indicating a reasonably high likelihood that the optimal values
of $\beta$ fall within the ranges spanned by the critical examples for
all tested values of $CC$.

While labeling examples, we maintain a cache of labeled examples
across all values of $CC$, so that when a label is needed for an
example for one value of $CC$, we obtain the label from the cache if
that example has already been labeled for another value of $CC$.  The
critical 300 examples for all values of $CC$ from 2 through 30 include
a total of 561 unique examples, with another 1022 examples labeled
``don't care'' being discarded.

To simultaneously optimize $CC$ and $\beta$, we repeat the search for
an optimal value of $\beta$ for each value of $CC$ from 2 through 30,
over the set of 561 examples containing the critical examples for all
these values of $CC$, selecting a combination of values for $CC$ and
$\beta$ that minimizes the number of sentence break prediction errors
over this set.\footnote{We cannot simply compare the numbers of errors
for the optimal values of $\beta$ for each value of $CC$ in the
original searches, because those error counts come from a different
set of 300 critical examples for each value of $CC$.}  For all values
of $CC$, the optimal values of $\beta$ found over these 561 examples
are identical to those found over each of the individual sets of 300
critical examples (providing some additional support for the data
selection procedure), but the sentence-break-prediction error rates
differ, due to the differences in the sets of examples over which the
error rate is measured.

In Section~5.6, when faced with a similar search for optimal values
of $CC$ and $\beta$ for our paragraph-break-probability-based sentence
break predictor, we described a somewhat complicated tie-breaking
procedure to deal with multiple combinations of values of $CC$ and
$\beta$ that produced the same lowest sentence break prediction error
on the tuning set we used for that optimization problem. A much
simpler tie-breaking method suffices for the missing-space predictor,
because the combination of the value $CC =$ 3 for the count cutoff,
with a single range of values\footnote{Note that in selecting such a
threshold, there will always be at least one continuous range of
values that give the same best result, because any threshold value
between two consecutive members of a sorted list of unique scores of
tuning examples will partition the examples into the same sets of
positive and negative predictions.}  0.753623188405797 $< \beta \leq$
0.756827181059786 for the estimated-white-space-probability threshold,
uniquely produces the lowest error count, 161 errors, on these 561
tuning examples.\footnote{If the number of errors seems high, recall
that the way these examples were selected for labeling makes them, as
a group, the hardest examples to classify in the larger tuning set,
with thousands of easier examples not being labeled. (We don't know
the exact number, because we don't know how many of the remaining
examples would be discarded as ``don't care'' cases.)}  To pick a
specific value for $\beta$ we take the midpoint of the range of
equally optimal values and truncate it to four significant digits,
giving us a selected value of $\beta =$ 0.5902.

\section{Additional heuristic filtering}

As mentioned in Section~10.1, and explained in more detail in
Appendix~C, in applying our missing-space predictor, we use a set of
regular expressions to identify certain instances of punctuation marks
not followed by white space that should not be considered candidates
for missing spaces.  These regular expressions are heuristically
designed to identify non-sentence-final punctuation marks occurring in
hyphenated abbreviations, decimal numbers, strings of multiple
initials, inflected forms of abbreviations, email addresses,
and computer filenames, domain names, URLs, and related
expressions. These regular expressions were intended to have very high
precision, at some loss of recall, and although we did not perform a
formal evaluation, we scanned a few thousand examples caught by these
filters without noticing any cases where a space should, in fact, be
inserted.

On examination of the tuning set errors made in optimizing the $CC$
and $\beta$ hyperparameters for the missing-space predictor, we
noticed another candidate class for a heuristic regular expression
filter. Specifically, we noticed that, even with the optimal values of
$CC$ and $\beta$, 17 of 98 false positives involved periods surrounded
by upper-case letters and numerals, with no lower-case letters between
the period and the closest other punctuation marks or white space;
i.e., contexts where any letters in the context surrounding a period
were all capitalized. Examples include \verb|(WOR.AX)|,
\verb|DAVE.TV|, \verb|USA.NET|, \verb|LETTER.HTML|, and
\verb|WHITEHOUSE.GOV|. We conjecture that the relative rarity of
all-uppercase text in our training data makes these examples difficult
for our model to predict correctly.

This observation suggests adding an additional regular expression
filter to our set of filters. If the text immediately preceding the
point of prediction matches
\begin{quote}
  \verb#/(\W|^)[0-9A-Z]+\.$/#
\end{quote}
and the text immediately following the point of prediction matches
\begin{quote}
  \verb#/^[0-9A-Z]+(\W|$)/#
\end{quote}
we do not treat the point of prediction as a space insertion
candidate. Unlike our other regular expression filters, this filter
does not have near perfect precision. It causes a small number of
false negatives, but not as many as the number of false positives it
blocks.

In incorporating this additional filter into our predictor, we did not
bother to re-extract N-gram counts using this filter. We did, however,
re-optimze $CC$ and $\beta$ on the same set of 561 labeled examples
previously used, setting the white space probability to 0 for all the
examples that would be blocked by the filter. This resulted in the
same optimal value $CC = $ 3 as previously, but the optimal value of
$\beta$ was reduced from 0.5902 to 0.3903. With the additional filter,
the total number of errors for the optimal values of $CC$ and $\beta$
declined from 161 to 147 on the 561 example tuning set. There were
four false negatives caused directly by the additional heuristic
filter, but the total number of false negatives declined from 63 to
48, as a result of reducing $\beta$ from 0.5902 to 0.3903. The number
of false positives increased by one, from 98 to 99.

\section{Evaluation on in-domain and out-of-domain test data}

We evaluate the effect of incorporating missing-space prediction into
our preferred sentence break predictor on both our in-domain and
out-of-domain test sets. As our baseline, we use the NSBLP SVM
sentence break predictor as evaluated in Section~8.4, using the
heuristic method of incorporating paragraph break information at
prediction time by predicting a sentence break whenever a paragraph
ends in standard sentence-final punctuation.

We incorporate missing-white-space prediction into the normalization
process described in Section~5.1. Wherever we find a text position
that is not considered to be a sentence break candidate only because
it is not at a white space, we apply our missing-space predictor. The
missing-space predictor first applies all of our regular expression
filters, including the one that looks for periods in contexts that
include no lowercase letters. If any of these filters match, no space
is predicted. Otherwise, the N-gram-based predictor is applied with
hyperparameter values $CC = $ 3 and $\beta = $ 0.3903. If a missing
space is predicted, a space is inserted. After normalization
incorporating this missing-space insertion step, tokenization and
subsequent steps proceed as in the baseline NSBLP SVM sentence break
predictor.

The following table compares results for our baseline NSBLP SVM
sentence break predictor and the NSBLP SVM predictor plus
missing-space prediction, on our in-domain Satz-3 test set and
out-of-domain English Web Treebank test set:
\begin{center}
  \begin{tabular}{||l|l|r|r||}
    \hline \hline
    Test set & Sentence break predictor & \multicolumn{1}{l|}{Error
      count} & \multicolumn{1}{l||}{Error rate}  \\
    \hline \hline
    In domain & NSBLP SVM & 44    & 0.22\% \\
    \hline
    & $+$ Missing-space prediction & 44     & 0.22\% \\
    \hline \hline
    Out of domain & NSBLP SVM & 474   & 3.62\% \\
    \hline
    & $+$ Missing-space prediction & 442    & 3.37\% \\
    \hline \hline
  \end{tabular}
\end{center}

As the table shows, missing-space prediction produces no change in the
number of errors on the in-domain Satz-3 test set, which is the best
result possible, because this test set contains no sentence breaks
without white space. A detailed examination of the effects of
missing-space prediction on this test set reveals 5052 instances of
potential sentence-final punctuation, other than ellipses, not
followed by white space. All but 6 of these are ruled out as
missing-space candidates by our heuristic regular expression
filters. Of the remaining 6, 3 are predicted, correctly, to be missing
spaces by our N-gram-based model, but each of these 3 is predicted,
also correctly, not to be a sentence break, leaving the sentence break
predictions on this test set unchanged by missing-space prediction.

On the out-of-domain English Web Treebank test set, predicting missing
spaces reduces the number of errors from 474 to 442, a 6.75\% relative
improvement. This reduction of 32 in the total number of errors comes
from a reduction of 43 in the number of false negatives combined with
an increase of 11 in the number of false positives. This result comes
from identifying 1590 instances of potential sentence-final
punctuation, other than ellipses, not followed by white space, of
which 248 survive our heuristic regular expression filters. Of these
248 missing-space candidates, 81 result in spaces being inserted in
the text, and of these, 54 are predicted to be sentence breaks.

Comparing the sentence break predictions on the out-of-domain test
set, with and without missing-space prediction, the two-tailed p-value
according McNemar's test is less than 0.0001, indicating that the
difference in sentence break prediction accuracy is highly
significant. As we previously noted, there are 74 instances in this
test set of sentence breaks not at a white space and not following an
ellipsis. Of these, 70 follow what we regard as standard
sentence-final punctuation, and hence fall under the cases that we are
trying to predict. Thus, the reduction of 43 errors produced by our
missing-space predictor gives us 45.7\% of the maximum improvement
possible in our test scenario.

\chapter{Analysis of sentence break predictions on test data}

In this chapter, we present the results of analyzing the behavior of
our sentence break predictors on our test data, to understand what
types of sentence breaking examples remain difficult to predict
correctly even with our best sentence break predictor, and also to
understand the relative advantages and disadvantages of different
predictors. The analysis is performed by manual inspection of all the
errors made by a particular sentence break predictor on a particular
test set, or all the differences in predictions between two sentence
break predictors on a particular test set. The errors, or differences,
are classified by observation of features of the examples that might
account for the errors or differences. No tracing of exactly how the
predictors come to make particular predictions has been performed.

In the rest of this chapter, we discuss the most frequently occurring
features of examples that appear to be the cause of prediction
errors, or differences between two predictors. We illustrate these
features with actual examples from our test sets, with the symbol
$\|$ used to mark the position where an error or difference occurs.

\section{Error analysis for our best sentence break predictor on our
  out-of-domain test set}

In this section, we look at the errors made on our out-of-domain LDC
English Web Treebank test set by our best overall sentence break
predictor, the SVM with a paragraph-break-probability signal
represented as the log probability of a non-sentence-break (NSBLP SVM),
plus heuristic prediction of sentence breaks at paragraph breaks, plus
prediction of missing spaces to allow prediction of sentence breaks
where there is no white space in the raw text. This is the predictor
evaluated on the LDC English Web Treebank test set in
Section~10.6. Here we discuss the types of examples that lead to the
largest numbers of errors on this test set. A complete categorization
of the errors made by this predictor on this test set is presented in
Appendix~E.1.

This sentence break predictor makes 442 apparent errors on this test
set. We say ``apparent,'' because on inspection, we found 39 cases
that either seemed to be clear annotation errors, or in which a
fairly arbitrary decision could have gone the other way. For example,
in
\begin{quote}
  \verb|Irony is dead...|$\|$\verb|Long live Irony!|
\end{quote}
the position marked by $\|$ seems to us to be an obvious sentence
break, and in
\begin{quote}
  \verb|PS.| $\|$ \verb|Love the new website!|
\end{quote}
\verb|PS.| could be considered either to be part of, or separate from,
the sentence that follows it.

Of the 39 examples of questionable reference annotations, 37 are not
marked as breaks in the reference annotations, but possibly should
have been, and 2 are marked as breaks, but possibly should not have
been.  Twenty of the examples where we find the reference annotation
questionable turn on whether a sentence break follows an ellipsis: 19
of these examples are not marked as breaks, and one of the examples is
marked as a break. We further discuss the annotation of sentence
breaks following ellipses in Chapter~12.

This leaves 403 errors on the out-of-domain test set for which we do
not question the reference annotation. Of these, by far the largest
category are false negative predictions on examples in which a
sentence break is followed by uncapitalized text, for example:
\begin{quote}
    \verb|colorado beat texas a&m.| $\|$ \verb|i know you remember the bet.|
\end{quote}
There are 144 false negative errors where uncapitalized text following
the sentence break appears to be the major reason for the error, and 7
more false negative errors in which the missing space predictor fails
to insert a space at a sentence break preceding uncapitalized text, so
that the possiblity of a sentence break is not considered. Since
sentence breaks are virtually never followed by uncapitalized text in
the standard written English that comprises our training data, it
should not be surprising that our sentence break predictor often fails
to predict a sentence break in this situation. In the user-generated
text that comprises much of the LDC English Web Treebank corpus,
however, the convention of always capitalizing the beginning of a
sentence is often not followed.

The second largest category of examples resulting in prediction errors
are 41 false positive predictions following ellipses used to
anonymize email addresses occurring the the Usenet newsgroup portion
of the LDC English Web Treebank corpus, as follows:
\begin{quote}
  \verb|If you've successfully used any of these supplement to aid| \\
  \verb|weight loss, please write me at gottlie ...| $\|$ \verb|@yahoo.com| \\
  \verb|and let me know, telling me about your experience.|
\end{quote}
This pattern of an ellipsis, followed by a space, followed by
\verb|@|, followed by non-space characters, occurs in this test set
only because of manipulation to anonymize email addresses. The pattern
occurs nowhere in any of our training data, so it is not surprizing
that our predictor fails every time it presented with it.

The next largest category of examples resulting in prediction errors
are 30 instances of false positive errors involving unusual
sentence-final punctuation. These are examples in which a sentence
ends in normal sentence-final punctuation (period, question mark, or
exclamation point), followed by a space, followed by additional
punctuation characters, and the sentence break follows these
additional punctuation characters. In these examples, our
sentence break predictor places the sentence break after the initial
normal sentence-final punctuation character, which gets counted as a
false positive. In most of these examples, the extra punctuation
characters form an emoticon:
\begin{quote}
  \verb|A little help is always great.| $\|$ \verb|=) Thank you or your feed| \\
  \verb|back.|
\end{quote}
but in a few cases, the extra punctuation is simply a hyphen used as a
dash:
\begin{quote}
    \verb|Nothing compares to a home made product that really stands| \\
    \verb|the test of time.| $\|$ \verb|- The Brick, Ikea, and Leon's have their| \\
    \verb|place.|
\end{quote}
(Note that the $\|$ symbol is not part of the text, but simply marks
where the prediction error occurs.)

Emoticons simply do not occur in our training data, which is all from
formal newspaper/newswire sources. The pattern of a period, followed by
a space, followed by a hyphen, followed by a space, does occur
in the English Gigaword Fifth Edition corpus that we use for training our
paragraph break probability model, but in this corpus it
overwhelmingly occurs with a paragraph break between the period and
the hyphen, which leads us to predict a sentence break.

Another category of examples resulting in 30 prediction errors are
sentence breaks predicted following a period that follows an item
number in an enumerated list:
\begin{quote}
  \verb|3.| $\|$ \verb|In Section 6.1, change 20 Business Days to 20 days.|
\end{quote}
We could have grouped this with the examples of questionable reference
annotations, since the decision not to call this a sentence break is
rather arbitrary,\footnote{Gillick's (2009) version of the Satz test
set does treat such cases as sentence breaks.} but it is not a rare
case, and our Penn Treebank training data consistently treats these as
non-sentence-breaks. We therefore decided to consider predicted
sentence breaks in this situation as definite errors, since we would
have hoped our models would have learned to predict a
non-sentence-break. Our best guess is that the prediction of a
sentence break is strongly influenced by the capitalization of the
following word, and by the fact that, since we ignore line breaking,
our models don't pay attention to the fact that the number occurs at
the beginning of a line. This latter fact is a significant signal that
this is not a sentence break where the preceding sentence happens to end
in a number.

In eight cases, our predictor wrongly inserts a sentence break when
our missing space predictor wrongly inserts a space, as in
\begin{quote}
  \verb|- 47K202!.|$\|$\verb|DOC|
\end{quote}

Most of the remaining errors consist of examples that exhibit common
features that simply make it difficult to predict on the basis of
local information whether there is a sentence break or not, or where the
text is nonstandard in some way. Among the common features that make
sentence break prediction difficult are:
\begin{itemize}
\item
  Abbreviations, initials, and ellipses, preceding the point of
  prediction, since those might occur either at the end or in the
  middle of a sentence.
\item
  A word that is usually capitalized whether it begins a sentence or
  not, following the point of prediction, since such words might occur
  either at the beginning or in the middle of a sentence.
\item
  Following an abbreviation or initials, capitalized words that are
  not normally capitalized, but are capitalized in this case because
  they form a capitalized expression such as a company name.
\item
  Sentences ending in \verb|!| or \verb|?|, for which in our training
  data there are simply many fewer examples than sentences ending in
  periods for our models to learn from.
\item
  Sentence breaks followed by text starting with a numerical
  expression, removing the signal of a capitalized word following a
  sentence break.
\item
  Abbreviations in the middle of a sentence that are too rare for our
  models to learn that they don't necessarily end a sentence.
\item
  Words at the end of a sentence ending in a period that might be
  confused with abbreviations.
\end{itemize}
Some of the ways that text is nonstandard in examples our predictor
makes errors on include:
\begin{itemize}
\item
  Sentence breaks not followed by a space, where our missing space
  predictor fails to insert a space.
\item
  Words usually capitalized only in sentence-initial position that are
  capitalized in the middle of a sentence in a nonstandard way.
\item
  Spurious punctuation characters inserted in the middle of sentences.
\end{itemize}

Finally, in only two cases does our predictor make an error when we
observe nothing that should make prediction difficult:
\begin{quote}
  \verb|Great Service, Thanks Don.| $\|$ \verb|Nice Top Lights.| \\
  \\
  \verb|Hi.| $\|$ \verb|Well, wouldn't you know it.|
\end{quote}
In both these examples, the predictor fails to correctly predict a
sentence break where the first sentence ends in a period following a
word that does not resemble any common abbreviation, and the following
sentence starts with a capitalized word that is not normally
capitalized unless it is the first word of a sentence.

\section{Comparison of a PBP-based predictor to an SVM with a PBP signal
  on our out-of-domain test set}

In this section, we look at the differences in sentence break
predictions made on our out-of-domain LDC English Web Treebank test
set, between our predictor based only on the estimated paragraph break
probability (PBP-based) and the one based on an SVM with a
paragraph-break-probability signal represented as the log probability
of a non-sentence-break (NSBLP SVM). Both these predictors use heuristic
prediction of sentence breaks at paragraph breaks, but neither
predicts missing spaces, so no sentence breaks are predicted where
there is no white space in the raw text. These are the PBP-based and NSBLP SVM
predictors evaluated in Section~8.4.

These two predictors make a similar number of errors on this test set,
470 for the PBP-based predictor and 474 for the NSBLP SVM predictor, but they
differ substantially in which examples they make errors on. There are
a total of 174 prediction differences, suggesting that one predictor
may be significantly better than the other in specific
situations. Here we discuss the types of examples on which the two
predictors most often make different predictions on this test set. A
complete categorization of the differences in predictions made by
these two predictors on this test set is presented in Appendix~E.2.

On the out-of-domain test set, the largest class of examples where
these two predictors make different predictions are examples in which
a sentence break is followed by uncapitalized text. These examples can
be divided into three subcases, depending on the punctuation that ends
the sentence preceding the sentence break in question. The largest
subcase consists of 47 examples where the preceding sentence ends in a
question mark or an exclamation point. For 46 of these, according to
the reference annotations, the PBP-based predictor is correct in predicting
a break, as in:
\begin{quote}
  \verb|these guys were fantastic!| $\|$ \verb|they fixed my garage doors in| \\
  \verb|literally less than an hour.|
\end{quote}
In the one remaining example, the reference annotation has no
break, but a sentence break as predicted by the PBP-based predictor seems
correct:
\begin{quote}
  \verb|Come visit irc server: irc.yankeedot.net and join| \\
  \verb|#audiobooks for sharing, discussion and a great trivia| \\
  \verb|game!| $\|$ \verb|large selection of fiction, science fiction and| \\
  \verb|best sellers.|
\end{quote}

The NSBLP SVM predictor gets none of these examples correct. In the error
analysis reported in Section~11.1, we also observed that, for 10 other
examples, the NSBLP SVM predictor fails to predict a sentence break
following a question mark or exclamation point, even when the sentence
break is followed by a capitalized word. Here we observe that the PBP-based
predictor gets 8 of these 10 examples correct. There is only one
example for which the predictions differ following a question mark or
exclamation point and the NSBLP SVM predictor makes the correct prediction
(no break following \verb|Yahoo!| in a capitalized expression). So it
seems that the NSBLP SVM predictor has a general weakness in predicting
sentence breaks following question marks and exclamation points, which
becomes worse when the sentence break is followed by uncapitalized
text.

In the second subcase of the NSBLP SVM and PBP-based predictors making different
predictions when a sentence break is followed by uncapitalized text,
the preceding sentence ends in a single period in 19 examples, such
as:
\begin{quote}
  \verb|your cat is losing his territory.| $\|$ \verb|territory is very| \\
  \verb|important to cats, and they are very nervous when they go| \\
  \verb|into a new place| 
\end{quote}
In 14 of these examples, the NSBLP SVM predictor correctly predicts a
break, and in 5 examples, the PBP-based predictor correctly predicts a
break.

In the final subcase of the NSBLP SVM and PBP-based predictors making different
predictions when a sentence break is followed by uncapitalized text,
the preceding sentence ends in an ellipsis in 12 examples, such as:
\begin{quote}
  \verb|ive never been to the yorkshire moors or penines at| \\
  \verb|all...|$\|$\verb|i can see the penines in the far distance from my| \\
  \verb|bedroom window.|
\end{quote}
In 11 of these 12 examples, it is the PBP-based predictor that correctly
predicts a sentence break.  It should be noted that in 11 of the cases
in which an ellipsis ends a sentence and is followed by uncapitalized
text, the reference annotations fail to annotate the sentence break,
but in our view all of these are clearly sentence breaks. We
previously noted that the majority of cases in which we questioned the
reference annotation in Section~8.1 also involved sentences ending in
ellipses.

The next largest class of examples on which the NSBLP SVM and PBP-based
predictors make different predictions are 16 examples in which a
sentence break is followed by \verb|Please|. In 15 of these cases, the
NSBLP SVM predictor correctly predicts a sentence break, following
sentences ending in a period. In one case, the PBP-based predictor predicts
a sentence break, following a sentence ending in an exclamation
point. There are a total of 128 examples in our out-of-domain test set
in which a sentence break is followed by \verb|Please|, on which the
NSBLP SVM predictor makes 2 errors, and the PBP-based predcitor makes 16
errors. This gives the PBP-based predictor an error rate of 12.50\% on these
examples, compared to an overall error rate of 3.58\%. Without
detailed tracing, we have no clear explanation of why the PBP-based
predictor does so poorly on these examples.

Other classes of examples with multiple cases of the NSBLP SVM and PBP-based
predictors making different predictions on this test set include:
\begin{itemize}
  \item
  7 instances of PBP-based correct, periods following list item numbers (no
  break)
\item
  5 instances of NSBLP SVM correct, sentence breaks followed by
  parenthesized text (break)
\item
  5 instances of NSBLP SVM correct, initials as part of capitalized
  expressions (no break)
\item
  4 instances of PBP-based correct, sentence breaks followed by text starting with a
  numerical expression (break)
\item
  4 instances of NSBLP SVM correct, nonbreaking ordinary
  ellipses\footnote{Most ellipses are indicated with a series of
  either three or four periods. We refer to these as ordinary
  ellipses. Ellipses indicated either by two periods or five or more
  periods are referred to as unusual ellipses.} (no break)
\item
  4 instances of NSBLP SVM correct, abbreviations followed by numerical
  expressions (no break)
\item
  3 instances of PBP-based correct, abbreviations followed by numerical
  expressions (no break)
\item
  3 instances of NSBLP SVM correct, unusual ellipses (no break)
\item
  3 instances of NSBLP SVM correct, ordinary ellipses followed by capitalized words
  that are usually capitalized (no break)
\item
  3 instances of NSBLP SVM correct, sentence breaks followed by text starting with
  capitalized words that are usually capitalized (break)
\item
  2 instances of PBP-based correct, sentence breaks followed by text with unusual
  starting characters (break)
\item
  2 instances of NSBLP SVM correct, quotation-initial ordinary ellipses (no break)
\item
  2 instances of NSBLP SVM correct, abbreviations followed by capitalized
  words that are usually capitalized (no break)
\item
  2 instances of PBP-based correct, initials followed by capitalized words
  that are usually capitalized (no break)
\item
  2 instances of PBP-based correct, sentence-final quotations marked by
  single quotes (break)
\end{itemize}
Examples occuring in the out-of-domain test set for all these classes,
plus all the remaining examples on which the NSBLP SVM and PBP-based
predictors make different predictions on this test set are presented
in Appendix~E.2.

\section{Comparison of a PBP-based predictor to an SVM with a PBP signal
  on our in-domain test set}

In this section, we look at the differences in predictions on the
in-domain Satz-3 test set, between the same PBP-based and NSBLP SVM sentence
break predictors compared in the previous section on our out-of-domain
test set. On the in-domain test set, these two predictors make
different predictions on 107 examples, with the NSBLP SVM predictor being
correct 98 times and the PBP-based predictor being correct only 9 times, when
the predictions differ.  Here we discuss the types of examples on
which the two predictors most often make different predictions on this
test set. A complete categorization of the differences in predictions
made by these two predictors on this test set is presented in
Appendix~E.3.

On the in-domain test set, the largest class of examples for which
these two predictors make different predictions are 37 examples in
which initials form part of a capitalized expression, where the
difference lies in predicting a sentence break following the initials
and preceding a capitalized word forming part of the expression. In 34
cases the NSBLP SVM predictor correctly predicts no break, and in 3 cases
the PBP-based predictor correctly predicts no break. In most cases, the
expressions in questions are names of people, as in:
\begin{quote}
  \verb|Investor Asher B.| $\|$ \verb|Edelman, who bought 30% of Datapoint| \\
  \verb|Corp.'s stock last week to try to head off a battle for| \\
  \verb|control, purchased most of the shares with funds from| \\
  \verb|another company he controls.|
\end{quote}
Other cases include company names, as in:
\begin{quote}
  \verb|Included in the sale are North Coast Bakery, Delight| \\
  \verb|Products, Kenlake Foods, Pontiac Foods, State Avenue,| \\
  \verb|Tara Foods, K.B.| $\|$ \verb|Specialty Foods and Pace Dairy.|
\end{quote}
Examples like this can be especially difficult when the capitalized
word following the initials is one that us usually not capitalized
unless it begins a sentence (e.g., \verb|Specialty|).

The next largest class of examples on which the two predictors make
different predictions consists of 17 examples in which an abbreviation
is followed by a dash introducing the rest of the sentence, such as:
\begin{quote}
  \verb|He points, as an example, to Quaker Oats Co.| $\|$ \verb|--| \\
  \verb|selling at a cool 21 times 12-month trailing earnings.|
\end{quote}
In all 17 of these examples, the NSBLP SVM predictor correctly predicts
that there is no sentence break.

The next largest class of examples on which the two predictors make
different predictions consists of 16 examples in which an abbreviation
forms part of a capitalized expression, where the difference lies in
predicting a sentence break following the abbreviation and preceding a
capitalized word forming part of the expression. In all 16 of these
examples, the NSBLP SVM predictor correctly predicts that there is no
sentence break. Like the similar class involving initials, the
capitalized expression can either be the name of a person, as in:
\begin{quote}
  \verb|However, Mr.| $\|$ \verb|Gorbachev must ensure that within this| \\
  \verb|"alliance" the business sector remains subordinate to the| \\
  \verb|party.|
\end{quote}
or the name of a company, as in:
\begin{quote}
  \verb|The former reflected a drop in Service Corp.| $\|$ \\
  \verb|International, which said it holds a 9.5% stake in| \\
  \verb|Diversified Energies.|
\end{quote}
In the latter case, the prediction becomes more difficult when the
capitalized word following the initials is one that us usually not
capitalized unless it begins a sentence (e.g., \verb|International|).

Other classes of examples with multiple cases of the NSBLP SVM and PBP-based
predictors making different predictions on this test set include:
\begin{itemize}
\item
  7 instances of NSBLP SVM correct, sentence breaks followed by parenthesized text (break)
\item
  5 instances of NSBLP SVM correct, abbreviations followed by parenthesized text (no break)
\item
  4 instances of NSBLP SVM correct, sentence breaks followed by quoted text (break)
\item
  3 instances of NSBLP SVM correct, sentence breaks with nothing unusual (break)
\item
  2 instances of NSBLP SVM correct, abbreviations followed by numerical
  expressions (no break)
\item
  2 instances of PBP-based correct, sentences ending in initials, followed by
  capitalized words that are not usually capitalized (break)
\item
  2 instances of NSBLP SVM correct, initials followed by capitalized words
  that are usually capitalized (no break)
\item
  2 instances of NSBLP SVM correct, non-sentence-breaks with nothing unusual (no break)
\item
  2 instances of NSBLP SVM correct, initials followed by parenthesized
  text (no break)
\end{itemize}
Examples occuring in the in-domain test set for all these classes,
plus all the remaining examples on which the NSBLP SVM and PBP-based
predictors make different predictions on this test set are presented
in Appendix~E.3.

\chapter{Post-analysis corrections to reference annotations}

In the analysis of the errors made by the NSBLP SVM predictor presented in
Section~11.1, we found 39 questionable reference annotations in our
out-of-domain test set. In the analysis of the prediction differences
between the PBP-based and NSBLP SVM sentence break predictors discussed in
Section~11.2, we found an additional 11 cases of what we believe to be
reference annotation errors that were not observed in Section~11.1,
because they agreed with the NSBLP SVM sentence break predictor. Thus we
found 50 wrong or questionable reference annotations in total, which
is a significant number compared to the 442 errors reported for our
best sentence break predictor.

It would be a simple matter to correct just the reference annotations
we encountered in these analyses, and re-run our evaluations, but this
would bias the results in favor of our sentence break predictors,
since there could be many other annotation errors that we never
noticed because they agree with all the sentence break predictors
whose predictions we analyzed.  Ideally, we would like to check our
entire test sets for annotation errors, but with the in-domain Satz-3
test set consisting of more than 420,000 words, and the out-of-domain
English Web Treebank test set consisting of more than 250,000 words,
we consider that to be too great an undertaking.

However, if there are types of examples that are particularly prone to
annotation errors, and which can be identified by objective criteria,
it may be feasible to improve the overall quality of the test set
annotation without checking every possible sentence break candidate.
As long as the criteria for selecting examples to check do not depend
on the predictions of our sentence break predictors, the results
should not be biased in favor of our sentence break predictors, even
though we do not correct all annotation errors.

Of the 50 total questionable annotations in the out-of-domain test set
that we noted in Chapter~11, 30 occur following ellipses, 15 occur
following single periods, 3 occur following question marks, and 2
occur following exclamation points.  The total counts of these
punctuation types in our two test sets\footnote{Although we did not
notice any questionable annotations in analyzing sentence break
predictions on the in-domain Satz-3 test set, any class of examples
that we decide to check and correct needs to be corrected on both test
sets, so that the post-correction results will be comparable on both
test sets.} are as follows:
\begin{center}
  \begin{tabular}{||l|r|r|r|r||}
    \hline \hline
    Test set & \multicolumn{1}{l|}{Single periods} &
    \multicolumn{1}{l|}{Ellipses} & \multicolumn{1}{l|}{Question
      marks} & \multicolumn{1}{l||}{Exclamation points}  \\
    \hline \hline
    In domain & 31,644 & 92 & 215 & 16 \\
    \hline
    Out of domain & 12,950 & 612 & 1,385 & 1,727 \\
    \hline \hline
  \end{tabular}
\end{center}
The counts of question marks and exclamation points are simple counts
of the total number of those characters in the test sets. The counts
of ellipses and single periods are obtained after text normalization
removes any spaces between otherwise consecutive periods. The count of
single periods is then the count of periods not preceded or followed
by another period, and the count of ellipses is the count of periods
preceded by another period but not followed by another period (which
is equal to the number of maximal strings of two or more consectutive
periods).

As the table shows, there are a great many more single periods in each
test set than ellipses, question marks, and exclamation points
combined. We have failed to find any significant, easily identifiable
subcategories of single-period examples in the 15 questionable cases
that we noticed in analyzing the predictions of our sentence break
predictors, and we consider examining more than 44,000 instances of
single periods in the two test sets combined to be infeasible, given
available resources. The number of question marks and exclamation
points is more manageable, but the number of questionable annotations
we observed involving them is so small that we concluded it not to be
worth the effort to examine all of them. This leaves possible sentence
breaks following ellipses, which were responsible for the majority of
the questionable annotations we observed, and for which there are only
704 examples across both test sets. So we decided to examine, and
possibly correct, the annotation of all instances of ellipses in our
in-domain and out-of-domain test sets.

\section{A new annotation standard for sentence breaks following ellipses}

As noted above, in analyzing sentence break predictions on our LDC
English Web Treebank out-of-domain test set, we observed 30
questionable annotations of possible sentence breaks following
ellipses. This may be somewhat surprising, considering that the
annotators were given an standard for annotating sentence breaks
(Laury et al.\ 2011) that addresses how to annotate ellipses:
\begin{quote}
  Ellipsis is used regularly as final and non-final punctuation in
  English writing. Therefore ellipses trigger sentence splits when
  functioning as a period and does [sic] not otherwise, In cases where
  its usage is ambiguous, it will be treated as medial punctuation.
\end{quote}

These instructions, however, are not very specific, as no further
explanation is provided as to what it means for an ellipsis to
function as a period. Moreover, the only example given seems
confusing:
\begin{quote}
  \verb|Come and say hello...you will be warmly welcomed!|
\end{quote}
This example was given as an instance of an ellipsis not functioning
as period, but the text segments preceding and following the ellipsis
are both complete independent clauses, with no syntactic connection
between them, and thus seem like separate sentences. The only way in
which this ellipsis fails to act like a period is that the word
following the ellipsis is not capitalized. In this corpus, however,
lack of capitalization is not a reliable indicator of a
non-sentence-break, with the original annotation of the corpus
including 1,189 sentences beginning with lowercase letters, out of a
total of 16,624 sentences (7.15\%).

To provide more a more detailed annotation standard for sentence
breaks following ellipses, and to avoid making the assumption that a
sentence cannot begin with uncapitalized text, we propose the
following procedure:
\begin{itemize}
\item
  If an ellipsis ends a paragraph, then insert a sentence break
  following the ellipsis.
\item
  If an ellipsis seems to indicate the truncation of an incomplete
  sentence, followed by a completely different sentence (regardless of
  capitalization), then insert a sentence break following the ellipsis.
\item
  Otherwise, compare: \\
  \\
  (a) Replacing the ellipsis with period, question mark, or exclamation
  point (whichever fits best), using appropriate white space, and
  capitalizing the word that follows if not already capitalized. \\
  \\
  (b) Replacing the ellipsis with a comma, using appropriate
  white space. \\
  \\
  (c) Removing the ellipsis, using appropriate white space. \\
  \\
  If (a) seems more preserving of the author's intent than (b) or (c),
  or equally preserving of the author's intent but more idiomatic than
  (b) or (c), then insert a sentence break following the ellipsis.
  Otherwise, do not insert a sentence break.
\end{itemize}

For both our in-domain and out-of-domain test sets, we have applied
this procedure to all occurrences of ellipses (defined as two or more
consecutive periods, optionally separated by spaces). For the
in-domain Satz-3 test set, this resulted in very little change,
removing one sentence break:
\begin{quote}
  \verb|And then there's . . .| $\|$ \verb|The Abyss."|
\end{quote}
and adding one sentence break:
\begin{quote}
  \verb|Further, "none of the new political tasks is ideological| \\
  \verb|. . .| $\|$
  \verb|few if any could be dealt with as adversarial issues;| \\
  \verb|few, if any could be tackled as traditional political issues."|
\end{quote}

For the out-of-domain English Web Treebank test set, this resulted in
removing 2 sentence breaks:
\begin{quote}
 \verb|My questions are..| $\|$ \verb|1) what thes difference between a modern| \\
 \verb|m16 and Vietnam m16?| \\
 \\
 \verb|So my question is..| $\|$ \verb|Is it normal for dogs to throw up blood| \\
 \verb|while having gastroenteritis?|
\end{quote}  
and adding 105 sentence breaks. Some examples of added sentence breaks include:
\begin{quote}
\verb|I have hundreds of VHS movies lying around...| $\|$ \verb|what| \\
\verb|should I do with them?| \\
\\
\verb|My Hamster escaped....| $\|$ \verb|NEED HELP NOW!?| \\
\\
\verb|Ugh my dad is so stupid...| $\|$ \verb|he just doesn't understand| \\
\verb|anything!| \\
\\
\verb|depends what you want from "B&W"...| $\|$ \verb|many programs will| \\
\verb|use a generic B&W...| \\
\\
\verb|deleting a question where you dont like the answer won't| \\
\verb|get you better answers...| $\|$ \verb|i refer to my previous answer:| \\
\\
\verb|lots of important stories out there today jeff....|$\|$\verb|thank god| \\
\verb|and a few good souls for that conviction.|
\end{quote}
These changes increased the number of sentence-final ellipses in the
out-of-domain test set from 246 to 349, out of the total of 612
ellipses in the test set. Of the 105 added sentence breaks, the text
following the ellipsis was capitalized in only 9 cases.

\section{Evaluation on corrected test sets}

Here we evaluate the PBP-based and NSBLP SVM sentence break predictors
on the in-domain and out-of-domain test sets corrected as described
above. Initially, we compare the PBP-based and NSBLP SVM sentence
break predictors evaluated in Section~8.4, whose differences in
predictions we analyzed in Section~11.2. Recall that both these
predictors use heuristic prediction of sentence breaks at paragraph
breaks, but neither predicts missing spaces, so no sentence breaks are
predicted where there is no white space in the raw text. The results
for these predictors on the original annotations and on the corrected
annotations for both test sets are as follows:
\begin{center}
  \begin{tabular}{||l|l|r|r|r|r||}
    \hline \hline
    & Sentence break & \multicolumn{1}{l|}{Original} & \multicolumn{1}{l|}{Original} &
    \multicolumn{1}{l|}{Corrected} & \multicolumn{1}{l||}{Corrected}  \\
    Test set & predictor & \multicolumn{1}{l|}{error
      count} & \multicolumn{1}{l|}{error rate} &  \multicolumn{1}{l|}{error
      count} & \multicolumn{1}{l||}{error rate} \\
    \hline \hline
    In domain & PBP-based & 133 & 0.66\% & 135 & 0.67\% \\
    \hline
    & NSBLP SVM  & 44 & 0.22\% & 46 & 0.23\% \\
    \hline \hline
    Out of domain & PBP-based & 470 & 3.58\% & 535 & 4.05\% \\
    \hline
    & NSBLP SVM & 474 & 3.62\% & 557 & 4.21\% \\
    \hline \hline
  \end{tabular}
\end{center}
The columns labeled ``Original error count'' and ``Original error
rate'' refer to the results reported in Section~8.4 on the in-domain
Satz-3 test set and out-of-domain English Web Treebank test set,
prepared as described in Appendix~A. The columns ``Corrected error
count'' and ``Corrected error rates'' refer to results on the test
sets changed as described in Section~12.1. Recall that we measure
error rate percentages as errors per 100 sentence breaks, so reported
error rates are affected by changes in the number of annotated
sentence breaks, as well as by changes in the number of predictions
that differ from the annotated sentence breaks.

As the table shows, the corrections to the in-domain Satz-3 test set
made little difference to the results, since we changed the annotation
in only two places. We inserted one sentence break in a position
followed by an uncapitalized word, and removed a sentence break
followed by a capitalized word, which made both examples harder to
predict, and the number of errors went up by two for both sentence
break predictors. The situation is quite different for the corrected
out-of-domain test set. The corrections to this test set added 96
sentence breaks followed by uncapitalized text, which makes the test
set much harder, so the number of errors increased substantially for
both sentence break predictors, and the difference in error counts
between the PBP-based predictor and the NSBLP SVM predictor rose from 4 to 22.

Next, using a version of the PBP-based sentence break predictor that
incorporates missing-space prediction (which we have not tested
before), we compare PBP-based and NSBLP SVM sentence break predictors
incorporating missing-space prediction:

\begin{center}
  \begin{tabular}{||l|l|r|r||}
    \hline \hline
    Test set & Sentence break predictor & \multicolumn{1}{l|}{Error
      count} & \multicolumn{1}{l||}{Error rate}  \\
    \hline \hline
    In domain & PBP-based $+$ missing-space prediction & 135     & 0.67\% \\
    \hline
    & NSBLP SVM $+$ missing-space prediction & 46     & 0.23\% \\
    \hline \hline
    Out of domain & PBP-based $+$ missing-space prediction & 496     & 3.75\% \\
    \hline
    & NSBLP SVM $+$ missing-space prediction & 525     & 3.97\% \\
   \hline \hline
  \end{tabular}
\end{center}
As this table shows, adding missing-space prediction makes no
difference to either sentence break predictor on the in-domain test
set, but improves the accuracy of both sentence break predictors on
the out-of-domain test set. The difference in the number of errors
between the PBP-based and NSBLP SVM sentence break predictors increases to 29,
which has a two-tailed p-value of 0.0395 according to McNemar's test.

The comparison of the PBP-based and NSBLP SVM sentence break predictors on the
corrected out-of-domain test set creates a dilemma for us. Before
making the corrections on this test set, we could unequivocally
recommend the NSBLP SVM approach to sentence break prediction regardless
of whether the text to be sentence-broken is more similar to our
in-domain or out-of-domain test set. On our previous test sets the
NSBLP SVM sentence break predictor was never significantly worse than any
other sentence break predictor we tested. On the corrected test sets,
however, the NSBLP SVM predictor is still significantly better than the
PBP-based predictor on the in-domain test set, but the PBP-based predictor is
significantly better than the NSBLP SVM predictor on the out-of-domain
test set.

A review of the analysis reported in Section~11.2 reveals two main
situations in which the PBP-based predictor seems to out-perform the NSBLP SVM
predictor:
\begin{itemize}
\item
  Sentences ending in ellipses followed by uncapitalized text
\item
  Sentences ending in question marks or exclamation points
\end{itemize}
Below, we break down the errors made by the PBP-based and NSBLP SVM sentence
break predictors incorporating missing-space prediction on the
corrected out-of-domain test set, according to whether the predicted
or actual sentence break is at an ellipsis followed by uncapitalized
text, following a question mark or exclamation point, or something
else:
\begin{center}
  \begin{tabular}{||l|l|r|r|r|r||}
    \hline \hline
    Sentence & & \multicolumn{1}{l|}{Lower case} & \multicolumn{1}{l|}{Question mark} &
    \multicolumn{1}{l|}{} & \multicolumn{1}{l||}{}  \\
    break & & \multicolumn{1}{l|}{following} &
    \multicolumn{1}{l|}{or exclamation} &
    \multicolumn{1}{l|}{Other} & \multicolumn{1}{l||}{All}  \\
    predictor & Error type & \multicolumn{1}{l|}{ellipsis} &
    \multicolumn{1}{l|}{point} &  \multicolumn{1}{l|}{cases} & \multicolumn{1}{l||}{cases} \\
    \hline \hline
    PBP-based & False positives & 58 & 25 & 161 & 244 \\
    \hline
    & False negatives  & 91 & 57 & 104 & 252 \\
    \hline
    & All errors  & 149 & 82 & 265 & 496 \\
    \hline \hline
    NSBLP SVM & False positives & 48 & 21 & 154 & 223 \\
    \hline
    & False negatives & 101 & 124 & 77 & 302 \\
    \hline
    & All errors  & 149 & 145 & 231 & 525 \\
    \hline \hline
  \end{tabular}
\end{center}

As we see in this table, the PBP-based predictor indeed makes 10 fewer
errors on actual sentence breaks between ellipses and uncapitalized
following text, but it makes 10 more errors by falsely predicting
sentence breaks where there are none in similar situations. So the PBP-based
predictor is not better than the NSBLP SVM predictor at deciding whether
there is a sentence break between ellipses and uncapitalized following
text; it is simply more likely to predict a sentence break in that
situation. On the other hand, the PBP-based predictor does seem to be more
accurate than the NSBLP SVM predictor at predicting sentence breaks
following a question mark or exclamation point. The PBP-based predictor
correctly predicts 67 additional sentence breaks in that situation, at
the cost of only 4 additional false sentence break predictions.  In
all other cases, in aggregate, the NSBLP SVM predictor is more accurate
than the PBP-based predictor, making 27 more correct sentence break
predictions and 7 fewer false sentence break predictions than the PBP-based
predictor.

The comparison of the PBP-based and NSBLP SVM predictors in these different
situations suggests a simple way of improving overall sentence break
prediction accuracy: For sentence break candidate positions following
question marks or exclamation points, apply the PBP-based predictor to
decide whether to predict a sentence break; in all other cases, apply
the NSBLP SVM predictor. This would result in a total of 462 errors, for a
3.50\% error rate, on the out-of-domain test set compared to 496
errors for the PBP-based predictor alone and 525 errors for the NSBLP SVM
predictor alone. Neither the PBP-based nor NSBLP SVM predictor makes any errors
following question marks or exclamation points on the in-domain test
set, so this combination of predictors would have the same accuracy as
the NSBLP SVM predictor on that test set.

An alternative would be to apply the simple heuristic of always
predicting a sentence break following question marks or exclamation
points. That would result in fewer errors in this situation on the
out-of-domain test set, 61 total errors (all false positives) compared
to 82 total errors for either the PBP-based predictor or the combination of
predictors. This heuristic would be much worse on the in-domain test
set, however, making 21 false positive errors, compared to no errors
for either the PBP-based or NSBLP SVM predictor, or their combination. A
different heuristic of always predicting a sentence break unless the
question mark or exclamation point is followed by an uncapitalized
word would make only one error on the in-domain test set, but would
make 184 errors on the out-of-domain test set.

It should be noted, of course, that any method of addressing the
weakness of the NSBLP SVM predictor in predicting sentence breaks
following question marks or exclamation points needs to be tested on a
new blind test set, since we observed that there is a weakness to be
addressed only after examining our existing test sets in detail.

\chapter{Conclusions}

We group our concluding thoughts into four broad categories:
achievements, limitations, approaches to improving accuracy, and
reconceiving the problem.

\section{Achievements}

This study has produced a number of important results on English
sentence break prediction. We have shown that a sentence break
predictor based almost solely on estimated paragraph break probability
(Chapter~5) can accurately predict the position of sentence breaks in
text. Our sentence break predictor that uses a paragraph break
probability model trained on four billion words of unannotated text
from the LDC English Gigaword Fifth Edition corpus, with only 1000
sentences from the Penn Treebank corpus annotated with sentence breaks
for tuning hyperparameters, and a simple heuristic to predict
sentence breaks at paragraph breaks, achieves error rates of 0.60\% on
our Satz-3 in-domain test set and 3.75\% on our out-of-domain LDC
English Web Treebank test set (Chapter~9).

While we found that a simple SVM classifier using indicator features
trained on 733,000 words from the Penn Treebank corpus has a much
lower error rate, 0.22\%, than our paragraph-break-probability-based
predictor on our in-domain test set (Chapter~6), this classifier has a
much higher error rate, 4.63\%, on our out-of-domain test set than our
paragraph-break-probability-based predictor (Chapter~7). We looked at
various ways of combining paragraph break probability signals with an
SVM classifier (Chapter~6), and found that the best combination we
tested performs very well on both test conditions, with error rates
of 0.23\% on a corrected version of our in-domain test set and 3.50\%
on a corrected version of our out-of-domain test set
(Chapter~12). This hybrid sentence break predictor also incorporates
prediction of missing white spaces to facilitate prediction of
sentence breaks where there is no white space in the original text
(Chapter 10).

In addition to these results on English sentence break prediction, we
believe that we have made a number of other contributions of more
general interest. For example, we have demonstrated some significant
issues with using either $\rm{F_{1}}$ score or $\rm{F_{\beta}}$ score
as an evaluation metric. First we showed that $\rm{F_{1}}$ score
inherrently penalizes false negative predictions more than false
positive predictions. Then we showed that there is no possible choice
of $\beta$ in computing $\rm{F_{\beta}}$ score that equally weights
false positives and false negatives in all situations. Thus if it is
desired to have an evaluation metric that reflects the total number of
prediction errors, but is comparable across test sets of differing
size, neither $\rm{F_{1}}$ score nor $\rm{F_{\beta}}$ score is
suitable (Chapter~4). This is an observation that is relevant for many
different problems beyond sentence break prediction.

We have also developed some technical solutions to problems that arise
in many applications of data-driven prediction. In our work on using a
paragraph break probability estimate to predict sentence break
probability, we developed a general method of empirically mapping a
probability estimate for one type of event into a probability estimate
for another type of event, using a $k$-nearest-neighbors approach with
variable $k$ (Chapter~6). In our work on predicting missing white
space, we developed a general method for limiting the amount of
annotated data needed to pick an optimal threshold when a continuous
signal is used to make a binary prediction (Chapter~10). While our
approach falls under the much-studied topic of active learning, we are
unaware of any previous methods quite like ours.

\section{Limitations}

The most obvious limitation of our study is that we have not looked at
any languages other than English. While we do not make use of very
detailed facts about English, such as lists of common English
abbreviation, we do make use of certain English orthographic
conventions, including:
\begin{enumerate}
\item
  Sentence breaks normally occur only at white spaces.
\item
  The ends of sentences are normally signaled by certain punctuation
  marks, especially periods (including ellipses), question marks, and
  exclamation points.
\item
  The punctuation mark signaling the end of a sentence may be followed
  by a combination of closing quotation marks and brackets that are
  considered part of the sentence that is ending.
\item
  The first word of a sentence is normally capitalized.
\end{enumerate}
Items 1-3 are hard-coded into our algorithms. Item 4 is not
hard-coded, but the features used by our statistical models are
designed to be able to represent generalizations about capitalization,
should they emerge during training as we expect.

For another language that follows these same orthographic conventions,
our algorithms could be applied directly, simply by retraining on data
from that language. To handle languages that follow similar
orthographic conventions, but use different characters for certain
functions, it would not be difficult to abstract from the particular
characters mentioned in our code and introduce language-specific
definitions for the characters used for those functions. For example,
we could define opening and closing double quotation marks to be
\guillemotleft\hspace{0in} and \guillemotright\hspace{0in} in French
and some other languages, instead of the double quote character that
we reduce all English double quotation marks to.

We believe that we could handle most European languages with this
approach. For languages whose orthography differs more significantly
from English, the applicability of our methods is more
doubtful. Languages that do not place sentence breaks at white spaces,
do not divide text into paragraphs, or do not make a distinction
between upper and lower case characters would require substantial
modifications to our approach.

A second limitation is hinted at in item 2, above. Even in English,
sentence breaks can be signaled by characters other than periods,
question marks, and exclamation points. According to the annotation of
the Penn Treebank, many sentence breaks are signaled by colons or
dashes. In the English Web Treebank, many sentence breaks are signaled
by a hard line break with no punctuation at all. Our study, however,
considers only sentence breaks signaled by periods, question marks,
and exclamation points.  Addressing other sentence breaks would
require consistently annotated training and test data that we
currently lack.

A limitation of our testing is that we have evaluated on only two
independent test sets: the Satz test set and the English Web Treebank
test set. We have referred to the Satz test set as in-domain, because
it is drawn from the same source, The Wall Street Journal, as our Penn
Treebank training data; and we have referred to the English Web
Treebank test set as out-of-domain because it is drawn from very
different sources. The English Web Treebank data differs from the WSJ
data in multiple ways, however, and it would be useful to tease apart
these differences in evaluations. Two major differences are that it
differs from the WSJ data both in topics and in the degree to which it
follows standard orthographic conventions. At a minimum, it would be
interesting to see how well our sentence break predictors perform on
text that differs in topics from the WSJ data, but is less noisy in
terms of following standard orthographic conventions than the English
Web Treebank data.

\section{Approaches to improving accuracy}

We group possible ways for improving sentence break prediction
accuracy into three areas: better data, better models, and addressing
specific types of examples.

\subsection{Better data}

One thing that would improve the accuracy of our sentence break
predictors would be having more, and especially cleaner, training
data. Our approach is based on the assumption that paragraph breaks
are naturally annotated and easy to extract from ordinary text, but we
have found that in the annotation of both the English Gigaword corpus
and the Penn Treebank WSJ corpus, there are many obviously spurious
paragraph breaks, which create false positives in our training data.
Unfortunately, our paragraph break probability model is especially
sensitive to false positives in the training data, because the
threshold on paragraph break probability for predicting a sentence
break is so low, only 0.025. Thus false positives in the training data
can lead to false positive predictions, even when the false positive
training examples are only a small proportion of the total examples
matching a particular text sequence.

For instance, our paragraph-break-probability-based predictor often
falsely predicts a sentence break in the name \verb|J.K. Rowling|
between \verb|J.K.| and \verb|Rowling|. This happens because of
spurious paragraph breaks occuring in the annotation of the New York
Times best-seller lists in the English Gigaword corpus:
\begin{quote}
  \verb|9. HARRY POTTER AND THE SORCERER'S STONE, by J.K.| \\
  \verb|</P>| \\
  \verb|<P>| \\
  \verb|Rowling. (Scholastic, $5.99.) A British boy, neglected| \\
  \verb|</P>| \\
  \verb|<P>| \\
  \verb|by his relatives, finds his fortune attending a school| \\
  \verb|</P>| \\
  \verb|<P>| \\
  \verb|for witchcraft.| \\
  \verb|</P>|
\end{quote}
Repetitions of text like this lead to 344 paragraph breaks splitting
\verb|J.K. Rowling| out of 4578 total examples in the English Gigaword
corpus, for a proportion of 0.075, which is well above our prediction
threshold of 0.025.

Our conjecture about these spurious paragraph breaks is that they
actually represent hard (i.e, forced) line breaks, which are not
distinguished from paragraph breaks in the LDC encoding of either the
English Gigaword corpus or the ``raw'' version of the Penn Treebank
WSJ corpus. The main reasons we believe the spurious paragraph breaks are
mostly hard line breaks are that in the Penn Treebank WSJ text there
are numerous examples of humorous verse in which each line is treated
as a separate paragraph, and also many signature blocks in letters to
the editor in which each line is treated as a separate paragraph.

We have no access to the truly original, raw data for either the
English Gigaword corpus or the Penn Treebank WSJ corpus, so we do not
know whether it would have been possible to distinguish hard line
breaks from paragraph breaks in representing these corpora. If we did
have access to such data however, we would not only be able to avoid
the false positive paragraph breaks resulting from conflating these
two types of breaks, but we might also make use of the hard line
breaks to help predict sentence breaks in the absence of
sentence-final punctuation. Hard line breaks appear to be an important
signal for recognizing sentence breaks without sentence-final
punctuation in the English Web Treebank test set, but since we do not
have that signal in any of our training data, we cannot make use of
it. In the English Web Treebank data it does not appear that hard line
breaks are conflated with paragraph breaks, but in some parts of the
corpus hard line breaks and soft\footnote{By a soft line break, we
mean a line break inserted only to prevent text from exceeding a
desired line length.} line breaks are difficult to distinguish, so
that would remain a problem to be solved.

\subsection{Better models}

Turning from data issues to modeling issues, it is natural to ask
whether we could achieve better accuracy by using more
sophisticated machine learning approaches, particularly deep
learning. We use machine learning in three different aspects of our
most accurate sentence break predictors, and the answer to this
question may be different in different cases.

The simplest model we use is the N-gram-based paragraph break
probability model. We deliberately kept this very simple so that it
would scale easily to large amounts of data. Since no special
annotation of the training data for this model is required, and we
only need to extract N-gram counts from the training data, our
approach should easily scale to very large amounts data indeed. As
pointed out by Banko and Brill (2001), the ability to increase the
amount of training data can turn out to be much more important than
the sophistication of the machine learning approach used.

Furthermore, it is not clear that a better paragraph break probability
model would yield a better sentence break predictor. A better
paragraph break probability model might learn to distinguish paragraph
breaks from other sentence breaks, which would be very bad for our
purposes. A perfect paragraph break probability model, which would
assign a probability of 1.0 to every actual paragraph break and a
probability of 0.0 to every non-paragraph-break, would be useless as a
sentence break predictor.

A place that we might want to actually use more sophisticated models
are the SVMs based on indicator features that we use for sentence
break prediction.  We did not do any significant experimentation to
optimize these models, except for how to represent paragraph break
probability information. In particular, for the indicator features, we
simply used the same types of N-grams used in the paragraph break
probability model. When we discovered that an SVM based only on these
indicator features gave a comparable result on in-domain test data to
Gillick's (2009) state-of-the-art result on the same test set, we did
not look for any ways to improve this model, other than to add a
paragraph break probability signal. It might very well turn out that a
better choice of features, whether discovered by human experimentation
or through deep learning, would give better results.

The final area where we use a machine learning approach that might be
improved upon is the N-gram model for missing space prediction. This
model is structurally the same as the paragraph break probability
model, but we would have less hesitation in trying a more
sophisticated model here, because we think the probability of a white
space is likely to be a more reliable predictor of a missing white
space than probability of a paragraph break is of a sentence break. In
other words, we doubt that there is any information to be learned that
distinguishes what we are training to predict, whether there is a
white space, from what we want to predict, whether there should be a
white space.

However, there seems to be a problem with with our existing missing
space predictor that may not be addressed simply by training a better
model on the same data. The missing space predictor seems to suffer
from false positives on examples that do not resemble the training
data either with or without a space. This is an instance of
training/test mismatch, which is always a problem in machine learning,
but it appears we could improve the predictor if we forced it to make
a negative prediction when the example does not appear to be likely
either with or without a space.

\subsection{Addressing specific types of examples}

The largest number of specific problematic examples that our sentence
break predictors get wrong are those where features that carry the
most information about sentence breaks are missing. In our
out-of-domain test set, the largest single class of this type is
sentences that begin with uncapitalized text. This violates the
othographic conventions of English and basically does not occur in our
annotated training data, so it seems hard to do anything about this.

Most of the remaining examples that are missing strong sentence
breaking signals don't violate orthographic conventions, but are still
hard to predict based on the local information that we use. The
largest group of errors in this category are false positives following
an abbreviation ending in a period and preceding a capitalized word
that is usually capitalized wherever it appears. Since abbreviations
ending in periods may either end a sentence or occur in the middle of
a sentence, and words that are normally capitalized may occur either
begin a sentence or in the middle of a sentence, it seems hard to do
anything about these cases either.

Some other groups of examples resulting in significant numbers of
errors are not intrinsically very difficult, but simply never occur in
our training data, so our predictors do not know what to do with
them. One such group of examples involve ellipses used to anonymize
email addresses, such as \verb|gottlie ... @yahoo.com|. These could be
handled by a simple heuristic of never predicting a sentence break
between an ellipsis and the character \verb|@|, but it is not clear
that this is worth worrying about, since it only occurs because of
artificial manipulation of the original text.

Another type of example that does not seem too difficult, but never
appears in our training data is sentences that end in ordinary
sentence-final punctuation followed by a combination of characters
forming an emoticon, e.g., \verb|^_^| or \verb|>:(|. By convention,
the emoticon attaches to the preceding sentence, with the sentence
break following the emoticon. A short list of commonly used emoticons
might be needed to predict these correctly. In more modern unicode
text, emojis would also need to be addressed.

Finally, there are at least two classes of examples that occur in the
training data and are not really unusual, yet our sentence break
predictors find them difficult. One of these classes is sentences
ending in question marks and exclamation points. Actually it is only
the SVM classifiers that seem to have special difficulty with these; our
paragraph-break-probability-based (PBP-based) predictor handles them much
better. We attribute this to be lack of relevant examples in the
annotated Penn Treebank data used to train the SVMs. There are only
419 examples of sentence break candidates signaled by question marks or
exclamation points in our annotated training set. Furthermore, our
SVMs are particularly bad compared to the PBP-based predictor on sentences
ending in question marks or exclamation points with the following
sentence starting with uncapitalized text. There are no examples where
this occurs in our annotated training set, but 141 examples in the
out-of-domain test set. As we pointed out in Section~12.2, we could
handle this simply by using the PBP-based predictor on sentence break
candidates following question marks or exclamation points, but that
seems clumsy, and it would perhaps be better to address the problem by
training the SVMs on a larger number of more diverse examples.

The second class of examples that occur in the training data and are
not really unusual, yet our sentence break predictors find them
difficult, are periods following list item numbers. By convention,
these are annotated in our data as not being sentence breaks, but our
sentence break predictors do not seem to learn that. It is not clear
why they make so many mistakes on these examples, but we think they
could do better if hard line breaks were annotated in the test and
training data, so that we have a chance to learn that list item
numbers normally occur at the beginning of lines. Another approach
would be to attack the problem at the level of document structure,
trying to explicitly identify numbered lists and extract the text
of each list item separately.

\section{Reconceiving the problem}

Although sentence break prediction is one of the classic
natural-language processing problems, with machine learning approaches
dating back more than 30 years (Riley, 1989) and heuristic approaches
in use even before that, the work we report here has led us to
question whether sentence break prediction as it is usually conceived
is the best way to look at the problem.

We start by asking the question ``What is the purpose of sentence
break prediction?'' In Section~4.1, we partially answered this
question by noting that the primary use of sentence prediction is
``dividing an extended text into smaller segments that can be
independently processed for certain purposes, such as parsing,'' but
this leaves open the question, which smaller units? The naive answer
is something like ``complete grammatical sentences,'' but as we
pointed out in Section~4.2, real text often does not consist solely of
complete grammatical sentences. Consider, for example, this piece of
text from the LDC English Web Treebank corpus: \\
\\
{\footnotesize
\verb|----== Posted via Newsfeed.Com - Unlimited-Uncensored-Secure Usenet News==----| \\
\verb|The #1 Newsgroup Service in the World! >100,000 Newsgroups| \\
\verb|---= 19 East/West-Coast Specialized Servers - Total Privacy via Encryption =---|
}
\\
This text consists of many largely independent pieces, but none of
them are complete grammatical sentences.

If the goal of sentence break prediction is to divide texts into
smaller units that can be independently processed for certain
purposes, then perhaps the target units should be the smallest pieces
of text that can be indpendently processed to the same extent that
complete grammatical sentences can be independently processed. This,
in turn, suggests that the target units should be the smallest pieces
that have no syntactic structures or dependencies crossing their
boundaries. This is, in fact, the most general principal articulated
by the sentence splitting guidelines (Laury et al., 2011) for the
English Web Corpus, but this document also defines several arbitrary
exceptions to this principle, such as that a colon does not signal a
sentence break when it occurs at a position other than the end of a
line, and that there are no sentence breaks between the lines of
postal addresses and other blocks of contact information.

Our next observation, which we already noted in Section~4.3.2, is that
if the goal of sentence break prediction is to divide text into smaller
segments that can be independently processed for certain purposes,
then false positives are likely to be much more damaging than false
negatives. If downstream processing relies on sentence break
prediction to provide pieces that can be independently processed, then
a false positive will by definition yield pieces that cannot be
independently processed, resulting in errors. Consider for example the
pair of sentences
\begin{quote}
  \verb|The boat| $\|$ \verb|ran into the bank.| \\
  \verb|The robber| $\|$ \verb|ran into the bank.|
\end{quote}
where $\|$ marks the position of a false positive sentence break
prediction.

If the downstream task is to translate these sentences into another
language, and the target language has the basic word order
verb-subject-object, then neither sentence can be translated correctly
if the pieces separated by the false positive sentence breaks are
translated separately into contiguous segments. Even if the basic word
order is the same as English, it would be impossible to appropriately
translate the word \verb|bank|, without considering the other pieces,
into a language in which the two senses of \verb|bank| are not
homonyms.

On the other hand, downstream processing of a larger piece that could
have been broken into smaller pieces need not necessarily result in
errors as long as the downstream processes are prepared to accomodate
this possibility. For translation, a short sequence of independent
sentences should not require any special accomodation. Even for
parsing, as long as the parser is not hard-wired to require a single
head or a constituent spanning the entire segment, this should not be
particularly difficult.

These observations suggest the following recasting of the sentence
break prediction problem: First, annotate training, tuning, and test
data so that every position between minimal syntactically independent
segments is marked as a sentence break. Second, optimize a sentence
break predictor to minimize false positives by allowing more false
negatives. This can be accomplished by modifying the loss functions we
use in optimizing our models to give a higher cost to false positives
than to false positives. This loss function weight will then be
another hyperparameter of the sentence break predictor. The loss
function weight could be optimized for a particular downstream task,
or a generally good value might be selected by plotting, for a
tuning set, the false positive count against the false negative count
and picking a point corresponding to ``the knee of the curve'', or
perhaps the beginning of the knee of the curve from the side
corresponding to lower false positives.

Of course, this assumes that it is actually necessary to break longer
texts into shorter segments to achieve high accuracy and efficiency on
tasks of interest. Even in the age of deep learning, we think this is
probably still true. We expect that there are problems that can
currently be addressed with greater accuracy and efficiency using
vector representations of individual sentences than using vector
representations of entire paragrahs or entire documents. If a time
comes when that is no longer the case, however, then the problem of
sentence break prediction may become a historical relic.

\chapter*{Acknowledgements}

\addcontentsline{toc}{chapter}{Acknowledgements}

I wish to thank my Google managers, Fernando Pereira and Daphne Luong,
for giving me the freedom to pursue this research wherever it led,
over a far longer period of time than any of us expected. I also wish
to thank Google colleagues Andy Golding, for in-depth discussions over
several years on the problems of sentence break prediction, and Dan
Gillick for reviewing this much-too-long report for publication.

\appendix

\pagebreak

\addcontentsline{toc}{chapter}{Appendices}

\nopagebreak

\chapter{Data preprocessing details}

\section{English Gigaword Fifth Edition corpus}

For training paragraph break probability estimates, we use the LDC
English Gigaword Fifth Edition corpus (Parker et al., 2011).  Since we
do not require sentence break annotations for this data, our
preprocessing is simpler than for our three other corpora.

The Gigaword corpus consists of printable ASCII and white space, with
SGML markup. We extract all text between begin and end paragraph
tags, \verb|<P>| and \verb|</P>|. Within each paragraph, we replace
SGML tags representing special characters with plain text ASCII
equivalents or stand-ins. Tags for which there is no obvious ASCII
replacement are left unchanged. The complete list of SGML special
character tag replacements is as follows:
\begin{center}
  \begin{tabular}{|l|l|c|}
    \hline
    Special character name & Tag & Replacement \\
    \hline
    Ampersand & \verb|&amp;| & \verb|&| \\
    Less than & \verb|&lt;| & \verb|<| \\
    Greater than & \verb|&gt;| & \verb|>| \\
    En dash & \verb|&#8211;| & \verb|-| \\
    Em dash & \verb|&#8212;| & \verb|--| \\
    Horizontal bar & \verb|&#8213;| & \verb|--| \\
    Left single quotation mark & \verb|&#8216;| & \verb|`| \\
    Right single quotation mark & \verb|&#8217;| & \verb|'| \\
    Left double quotation mark & \verb|&#8220;| & \verb|``| \\
    Right double quotation mark & \verb|&#8221;| & \verb|"| \\
    Horizontal ellipsis & \verb|&#8230;| & \verb|...| \\
    \hline
  \end{tabular}
\end{center}

For the tags containing alphabetic characters, all case variants are
given the replacement indicated in the table. For example,
\verb|&amp;|, \verb|&AMP;|, and \verb|&aMp;| are all replaced
with \verb|&|. We noticed some cases that look like double encoding, as
if the ampersand introducing a tag haa been re-encoded as
\verb|&amp;|, for example, \verb|&amp;gt;|. For that reason, whenever
we replace an ampersand tag with a plain text ampersand, we check
whether that replacement creates a tag that we recognize, and we
replace that as well.

In the sequence of extracted paragraphs, we distinguish pairs of
consecutive paragraphs from pairs of nonconsecutive
paragraphs. Paragraphs pairs are considered nonconsecutive if they
come from different documents, or if some non-paragraph material
occurs between them. In collecting statistics about paragraph breaks,
we do not consider the breaks between nonconsecutive paragraph pairs.

\section{Penn Treebank WSJ corpus}

Our supervised training, tuning, and development test data is taken
from the Penn Treebank WSJ corpus.  We use the untokenized, raw form
of the WSJ corpus included in the Treebank-2 release (Marcus et al.,
1995). This version of the corpus is broken into articles, paragraphs,
and sentences, but the sentence breaks were produced automatically,
and are not accurate enough to be used as a gold standard. Many of the
sentence break errors in the raw form of the corpus are corrected in
the parsed version of the corpus, but the parse trees also incorporate
a tokenization that we do not wish to use. To resolve this issue we
semi-automatically align the raw version of the corpus without
sentence breaks to a tokenized, sentence-broken version of the corpus,
to determine where to place the sentence breaks in the raw version.

Prior to alignment, we modify the raw text files to remove lines of
the form \verb|.START| that appear to indicate the beginning of an
article.  To create the tokenized, sentence-broken version of the Penn
Treebank WSJ corpus we first extract the sequence of terminal tokens
from each parse tree, omitting instances of the empty token
\verb|(-NONE-)|. We then convert certain special tokens back to the
plain text representation most commonly used in the raw text
files. The special tokens and their plain text replacements are as
follows:
\begin{center}
  \begin{tabular}{||c|c||c|c||}
    \hline
    Special token & Replacement & Special token & Replacement \\
    \hline
    \verb|``| & \verb|"| & \verb|''| & \verb|"| \\
    \verb|-LRB-| & \verb|(| & \verb|-RRB-| & \verb|)| \\
    \verb|-LCB-| & \verb|{| & \verb|-RCB-| & \verb|}| \\
    \verb|\*| & \verb|*| & \verb|\*\*| & \verb|**| \\
    \verb|\/| & \verb|/| & & \\
    \hline
  \end{tabular}
\end{center}
Finally, we delete tokens consisting of a single period following a
token that ends in a period, which un-does the Penn Treebank
convention of duplicating periods that end an abbreviation and also
end a sentence.

At this point, the modifed raw text and the modified extracted
tokenized text ought to be identical, except for differing line
breaks, the spaces introduced by tokenization, and the fact that for
some sections of the raw Penn Treebank corpus, only an initial segment is
annotated with parse trees. However, we found this was not completely
true, apparently because in producing parse trees from the raw data,
some noise and errors in the raw data were corrected, and other new
errors were introduced in the parsed data. We therefore use a
semi-automated alignment process in which an automatic aligner
proceeds as far as it can within a section of the Treebank,
requesting assistance when it encounters a place where inserting or
removing spaces and line breaks cannot make the two versions of the
data identical. We then manually examine the text where the alignment
stopped, and edit one or the other version of that section to allow
alignment to proceed past that point. We generally edit whichever
version seems less ``correct''. We repeat this process until every
section of the corpus can be aligned exactly to the point where the
tokenized version of the section runs out of text.

The most common corrections we make to the raw version of the corpus
are to remove extraneous non-ASCII characters that may be remnants of
some original markup in the text, to remove spurious white space
resulting from bad sentence breaking, and to remove whole sentences or
fragments that do not appear in the parsed version of the corpus. The
most common corrections we make to the version of the corpus extracted
from the parse trees involve placement or omission of double
quotes. We correct 21 cases in which opening quotes were moved to the
end of the preceding sentence, 20 cases in which closing quotes were
moved to the beginning of the following sentence, and 99 cases in
which closing quotes were simply dropped from the end of a sentence.

In Appendix~F.1, we provide resources for extracting from the LDC
Treebank-2 release, and sentence-break annotating, the version of the
Penn Treebank WSJ corpus that we use.

\section{Satz WSJ corpus}

For in-domain testing of sentence break prediction, we use the Wall
Street Journal text constituting the Satz WSJ corpus, extracted
from the ACL/DCI corpus (ACL/DCI, 1993) and sentence-break-annotated
by Palmer and Hearst (1997).  We obtained two versions this corpus,
the original untokenized version used by Palmer and Hearst, and a
modified, tokenized version used by Gillick (2009), which we believe
is also the version used by Kiss and Strunk (2006).

We compared these two versions of the corpus and found many
substantive differences, over and above differences in markup and
tokenization.  In light of these differences and a desire both to
report results that would be comparable to previous work and also to
report on what we consider to be the most accurate representation and
annotation of the original text, we decided to report results on three
versions:
\begin{itemize}
\item
  Satz-1, corresponding to Palmer and Hearst's version of the corpus.
\item
  Satz-2, corresponding to Gillick's version of the corpus.
\item
  Satz-3, reconciling the differences between Satz-1 and Satz-2.
\end{itemize}
Satz-1 simply omits markup from Palmer and Hearst's version of the
corpus, other than tags indicating the locations of sentence breaks.

To produce Satz-2, Gillick's version of the corpus is automatically
aligned\footnote{Minimizing an edit distance metric, rather than
  requiring an exact match, since there are differences between the
  versions that we want to preserve.} to a modified version of
Satz-1, to enable removing white spaces from Satz-2 introduced by
tokenization. Besides removing tokenization, Satz-2 also differs from
Gillick's version of the corpus in the exact placement of sentence
break tags. In Gillick's version, sentence break tags are placed
immediately following the period, question mark, or exclamation point
that signals the end of a sentence, even when that punctuation mark is
followed by a closing quotation mark, closing parenthesis, or other
punctuation that logically belongs to the preceding sentence. In
Satz-2, we move the sentence break tags past any such sentence-closing
punctuation.

We find 110 differences between Satz-1 and Satz-2.  Of these, 49 are
differences in whether a sentence break is indicated. We consider 23
of these to be clear errors in the annotation of the original version
of the corpus reflected in Satz-1. The other 26 seem to be due to
differences in the interpretation of the task (see Section~4.1) or
cases where it is simply unclear whether to say there is a sentence
break or not. There are 61 differences in the printable characters
between Satz-1 and Satz-2, including 43 deletions in Satz-2 of
non-word tokens representing en-dashes, em-dashes, and ellipses found
in Satz-1, and two differences in capitalization. The remaining 16
differences are the systematic replacement of exclamation points with
right square brackets (\verb|]|) in Satz-1.\footnote{It is noted in
the README.raw file included in the Treebank-2 release that this was
likely due to a processing error in converting the original form of
the WSJ data from EBCDIC to ASCII, which was evidently corrected in
later LDC releases of data from this source.}

To produce Satz-3, we resolve these differences by
\begin{itemize}
\item
  Replacing right square brackets in Satz-1 with exclamation points.
\item
  Keeping all other printable characters as in Satz-1.
\item
  Correcting the 23 outright sentence break errors in Satz-1.
\item
  Keeping all other sentence breaks as in Satz-1, except for
  removing one sentence break common to both Satz-1 and Satz-2 that
  would result in a ``sentence'' consisting of only punctuation
  characters.
\end{itemize}    

In Appendix~F.2, we provide resources for extracting from the LDC
release of the ACL/DCI corpus, and sentence-break annotating, the
versions of the Satz-3 WSJ corpus that we use.

\section{English Web Treebank corpus}

For out-of-domain testing of sentence break prediction, we use the LDC
English Web Treebank corpus (Bies et al., 2012), consisting of
weblogs, newsgroups, email, reviews, and question-answers. This corpus
is provided in three forms: raw, sentence-broken and tokenized, and
parsed. We again base our test set on the raw form, aligning it
with the sentence-broken tokenized form to determine the position of
sentence breaks in the raw text.

We preprocess the sentence-broken tokenized text by removing line
number tags introduced by the LDC.  We preprocess the raw text by
replacing SGML and XML markup lines with blank lines, and replacing
SGML tags representing special characters with plain text ASCII
equivalents as follows:
\begin{center}
  \begin{tabular}{|l|l|c|}
    \hline
    Special character name & Tag & Replacement \\
    \hline
    Apostrophe & \verb|&amp;apos| & \verb|'| \\
    Ampersand & \verb|&amp;| & \verb|&| \\
    Double quote & \verb|&quot;| & \verb|"| \\
    Less than & \verb|&lt;| & \verb|<| \\
    Greater than & \verb|&gt;| & \verb|>| \\
    \hline
  \end{tabular}
\end{center}

In the sentence-broken tokenized form of the corpus, the LDC also
replaced 8-bit ASCII characters and non-ASCII characters with 7-bit
ASCII substitutes. We decided to do the same to the raw corpus, using
the same set of substitutions as the LDC, but we encountered some
problematical issues.  The LDC documented their character replacements
with a set of Perl commands like \verb|tr/\x{00A3}/L/|. However, a
different set of replacements was given for each of the five genres,
and these were not completely consistent with each other. Moreover, in
some cases, multicharacter replacements were attempted that do not
work as intended, for example, \verb|tr/\x{2026}/.../|. The intent
here is clear, because \verb|\x{2026}| represents the hexadecimal
encoding of the unicode horizontal ellipsis character; but the
\verb|tr| command cannot be used to replace one character with a
sequence of characters, and in this case all characters in the
replacement string after the first are ignored. Hence,
\verb|tr/\x{2026}/.../| is equivalent to \verb|tr/\x{2026}/./|, and
has the effect of replacing the unicode horizontal ellipsis character
with a single period.

We resolve these problems by using commands that can replace a single
character with a multicharacter string in cases where that seems to be
intended. We have also chosen what we regard as the single best
replacement for each of the characters that the LDC gave different
replacements for different genres. The resulting set of character
replacement commands is as follows:

\begin{center}
\begin{tabular}{|l|l|l|}
\hline
\verb|tr/\x{00A0}/ /| & \verb|tr/\x{00A3}/L/| & \verb|tr/\x{00AD}/ /| \\
\verb|tr/\x{00B0}/o/| & \verb|tr/\x{00B3}/3/| & \verb|tr/\x{00B4}/'/| \\
\verb|tr/\x{00B7}/*/| & \verb|tr/\x{00C1}/A/| & \verb|tr/\x{00C3}/A/| \\
\verb|tr/\x{00C7}/C/| & \verb|tr/\x{00CD}/I/| & \verb|tr/\x{00E0}/a/| \\
\verb|tr/\x{00E1}/a/| & \verb|tr/\x{00E3}/a/| & \verb|tr/\x{00E4}/a/| \\
\verb|tr/\x{00E7}/c/| & \verb|tr/\x{00E9}/e/| & \verb|tr/\x{00EA}/e/| \\
\verb|tr/\x{00EF}/i/| & \verb|tr/\x{00F3}/o/| & \verb|tr/\x{00F4}/o/| \\
\verb|tr/\x{00F6}/o/| & \verb|tr/\x{00FC}/u/| & \verb|tr/\x{03A5}/Y/| \\
\verb|tr/\x{2000}/ /| & \verb|tr/\x{2013}/-/| & \verb|tr/\x{2014}/-/| \\
\verb|tr/\x{2014}/-/| & \verb|tr/\x{2018}/'/| & \verb|tr/\x{2019}/'/| \\
\verb|tr/\x{201C}/"/| & \verb|tr/\x{201D}/"/| & \verb|s/\x{2026}/.../g| \\
\verb|s/\x{2665}/<3/g| & & \\
\hline
\end{tabular}
\end{center}

We apply these character replacement commands to all genres of the
raw text version of the corpus. We also identify the places in the
sentence-broken tokenized version of the corpus where these
replacements differ from those that the LDC used, and we make
corresponding changes there. There are seven cases where the LDC
replaced unicode horizontal ellipsis characters with single
periods. When we replaced these single periods with three-period
ellipses, it was clear that the LDC's mistake had led the
annotators to incorrectly label two cases as sentence breaks. We
remove these sentence breaks from our referrence annotation of the
corpus.

In aligning the preprocessed raw text to the tokenized sentence-broken
text, all the raw text aligns except for five blocks of keywords in
the weblogs genre that are not included in the tokenized
sentence-broken version. We eliminate these blocks from our test
version of the corpus.

We also found in aligning the two versions of the corpus that there
are a significant number of sentence breaks at places where there is
no white space in the raw text.\footnote{We had to modify our aligner
  to accommodate this, which brought it to our attention.}  We thought
this was a likely situation for annotation errors to occur, so we
manually checked the post-alignment sentence-broken raw text for
errors in annotating sentence breaks without white space. We found
three cases where the LDC annotators missed clear sentence breaks
without white space, so we add these three sentence breaks to the
reference annotation of our test version of the corpus.

In Appendix~F.3, we provide resources for extracting from the standard
LDC release, and sentence-break annotating, the versions of the
English Web Treebank corpus that we use.

\chapter{Text normalization details}

As outlined in Section~5.1, we normalize both training texts and test
texts with respect to the use of spaces and the representation of
single quotes, double quotes, and ellipses. The particular
normalizations we perform are all motivated by variations in in these
aspects of text that we observe in the LDC English Web Treebank corpus.  Here
we give details of our normalization procedure.

Normalization proceeds one paragraph at a time. Most of
the work of normalization is carried out by Perl regular expression
substitution commands. Some of the substitutions depend on others
having been performed first, so within each paragraph, we perform all
possible instances of each substitution before moving on to the next.

We begin by identifying all the text within one paragaph, and
replacing each line break in the paragraph with a single
space. Although LDC English Web Treebank corpus is intended to be in plain
ASCII with SGML markup, we found some instances of unicode space and
line break characters, which we either replace with ASCII space
characters or remove entirely, using the following sequence of
substitution commands (with each command repeated as often as needed,
before moving on to the next):
\begin{itemize}
\item
  \verb|s/\'\\u2009\\u2009\"/\'\"/| Remove pairs of unicode thin
  spaces between a single quote and a double quote.
\item
  \verb|s/\\u2009\\u2009/ /| Replace other pairs of unicode thin spaces
  with one ASCII space.
\item
  \verb|s/\\u2009//| Remove other unicode thin spaces.
\item
  \verb|s/\\u2002/ /| Replace unicode en-spaces with an ASCII space.
\item
  \verb|s/\\u2028/ /| Replace unicode line breaks with an ASCII space.
\end{itemize}
At this point we break the paragraph into a list of non-space substrings,
and then rejoin into a single string, separating the non-space
substrings by single spaces. This ensures that each white space is
represented by a single ASCII space, with no leading or trailing
spaces. We then move on to the next sequence of substitutions.
\begin{itemize}
\item
  \verb|s/\. \./\.\./| Remove spaces between periods (to normalize
  ellipses).
\item
  \verb#s/ \.([^.\w]*( |$))/\.$1/# Remove spaces preceding a single
  period, when possibly followed by other punctuation, then space or
  end of paragraph.
\item
  \verb#s/\.\.([^\.\s\"\'\)|\}|\]|\,|\;|\:|\?|\!])/\.\. $1/# Insert a
  space after an ellipsis, when not already followed by a space or
  certain punctuation.
\item
  \verb#s/ \.\.(\W*( |$))/\.\.$1/# Remove spaces preceding an ellipsis,
  when possibly followed by punctuation, then space or end of paragraph.
\item
  \verb#s/ \,(\W* |\W*$)/\,$1/# Remove spaces preceding a comma.
\item
  \verb#s/ \;(\W* |\W*$)/\;$1/# Remove spaces preceding a semicolon.
\item
  \verb#s/ \:(\W* |\W*$)/\:$1/# Remove spaces preceding a colon.
\item
  \verb#s/ \?(\W* |\W*$)/\?$1/# Remove spaces preceding a question mark.
\item
  \verb#s/ \!(\W* |\W*$)/\!$1/# Remove spaces preceding an exclamation
  point.
\item
  \verb#s/( \W*|^\W*)(\(|\{|\[) /$1$2/# Remove spaces following an opening
  bracket.
\item
  \verb#s/ (\)|\}|\])(\W* |\W*$)/$1$2/# Remove spaces preceding a closing
  bracket.
\item
  \verb#s/' '''( |$)/'"$1/# Replace sequences consisting of a single
  quote, a space, and three single quotes, when followed by a space or
  end of paragraph, with a single quote followed by a double quote.
\item
  \verb#s/'''( |$)/'"$1/# Replace sequences consisting of three single
  quotes, when followed by space or end of paragraph, with a single
  quote followed by a double quote.
\item
  \verb#s/( |^)'''/$1"'/# Replace sequences consisting of three single
  quotes, when preceded by a space or beginning of paragraph, with a
  double quote followed by a single quote.
\item
  \verb#s/( |^)' '/$1''/# Remove spaces between two single quotes,
  when preceded by a space or beginning of paragraph.
\item
  \verb#s/( |^)` `/$1``/# Remove spaces between two back quotes,
  when preceded by a space or beginning of paragraph.
\item
  \verb#s/( |^)` /$1`/# Remove spaces following a back quote, when preceded
  by a space or beginning of paragraph.
\item
  \verb#s/(\?|\.|\!)``([A-Z])/$1 ``$2/# Insert spaces preceding pairs
  of back quotes, when preceded by sentence-final punctuation and
  followed by a capital letter.
\item
  \verb#s/((\?|\.|\!)\'?)`([A-Z])/$1 `$3/# Insert spaces preceding
  back quotes, when preceded by sentence-final punctuation and possibly
  a single quote, and followed by a capital letter.
\item
  \verb#s/``/"/# Replace pairs of back quotes with one double quote.
\item
  \verb#s/`/'/# Replace other back quotes with a single quote.
\item
  \verb#s/''/"/# Replace pairs of single quotes with one double quote.
\end{itemize}

At this point, we are representing both opening and closing quotes by
ordinary quote characters, either single or double, so to distinguish
opening and closing quotes, we depend on opening quotes being preceded
by a space but not followed by a space, and closing quotes being
followed by a space, but not preceded by a space.  This leaves us two
other cases to try to deal with:
\begin{enumerate}
\item
  Quote characters that are neither preceded nor followed by spaces
\item
  Quote characters that are both preceded and followed by spaces
\end{enumerate}
We deal with these by first converting instances of case 1 to case 2,
by inserting spaces both preceding and following the quote
character. We then deal with instances of case 2 by making assumptions
about how opening and closing quotes are ordered. Spaces are then
removed following quote characters identified as opening quotes, and
preceding quote characters identified as closing quotes.

In handling case 1 by inserting spaces around quote characters that
are neither preceded nor followed by spaces, we limit our attention to
quote characters that, on the basis of immediate context, appear
ambiguous between a quote character that ends a sentence and one that
begins a sentence. To identify these, we examine each string of
non-space characters between two spaces, looking for positions that
might look like the end of a sentence if there were a space there. Our
test for this is that the substring before the possible sentence break
(which we call the prefix) has to match either
\verb#^(.*?\w.*?(!|\?|\.)(\"|\'|\)|\}|\])*(\)|\}|\]))$/# or
\verb#/^(.*?\w.*?(!|\?|\.))$/# and the substring after the possible
sentence break (the suffix) has to match
\verb#/([^!\?\.\,\;\:\/\&\)\}\]]*\w.*)$/#.

If the prefix and suffix match those patterns, and the suffix begins
with one or more single and/or double quotes, we insert spaces around
each of those quotes, unless the string falls under certain
exceptions:
\begin{itemize}
\item
  \verb#$suffix =~ /^'[sSdD][^'\w]*$/# \\
  A single quote appears to be an appostrophe in a possessive noun or
  a past tense verb (e.g., \verb|U.S.A.'s| or \verb|c.c.'d|).
\item
  \verb#$suffix =~ /^([^!\?\.\w]*[^\-\w])?\-($|[^\-])/# \\
  The quote appears as part of a hyphenated word (e.g.,
  \verb|"U-S-A!"-chanting|).
\item
  \verb#(($prefix =~ /^\W*(www|WWW|Www|http|HTTP|Http)/) ||# \\
  \verb# (($prefix =~ /^\W*\w(\w|\/|\-)*\w\.$/) &&# \\
  \verb#  ($suffix =~ /^\w(\w|\/|\-)*\w\.\w(\w|\/|\-)*\w/)) ||# \\
  \verb# (($prefix =~ /^\W*\w(\w|\/|\-)*\w\.(\w(\w|\/|\-)*\w\.)+$/) &&# \\
  \verb#  ($suffix =~ /^\w(\w|\/|\-)*\w/))) &&# \\
  \verb#($suffix =~ /\/|\../)# \\
  The string looks like  a URL or a filename.
\item
  \verb#(($prefix =~ /\w(@|\(a\)|\(at\)|\(AT\))\w/) &&# \\
  \verb# ($suffix =~ /\../)) ||# \\
  \verb#($suffix =~ /\w(@|\(a\)|\(at\)|\(AT\))\w/))# \\
  The string looks like an email address.
\item
  \verb#$suffix =~ /^(\"|\')*(\)|\}|\])/# \\
  The quote or quotes are immediately followed by a closing bracket.
\end{itemize}

Next, for each quote character surounded by space characters, we remove
either the preceding or following space, depending on whether the
quote character appears to open or close quoted text. We start this
processes by looking for some cases that can be disambiguated on the
basis of local context:
\begin{itemize}
\item
  \verb#s/(^'|^") (.*)/$1$2/# Remove spaces following a quote
  character at the beginning of a paragraph.
\item
  \verb#s/(.*) ('$|"$)/$1$2/# Remove spaces preceding a quote
  character at the end of a paragraph.
\item
  \verb#s/ '"([^'"\w]*( |$))/'"$1/# Remove spaces preceding a single
  quote followed by a double quote.
\item
  \verb#s/(( |^)[^'"\w]*)"' /$1"'/# Remove spaces following a double
  quote followed by a single quote. 
\item
  \verb#s/ "((\)|\}|\])[^"\w]*( |$))/"$1/# Remove spaces preceding
  a double quote followed by a closing bracket.
\item
  \verb#s/ '((\)|\}|\])[^'\w]*( |$))/'$1/# Remove spaces preceding
  a single quote followed by a closing bracket. 
\item
  \verb#s/(( |^)[^"\w]*(\(|\{|\{))" /$1"/# Remove spaces following an
  opening bracket followed by a double quote.
\item
  \verb#s/(( |^)[^'\w]*(\(|\{|\{))' /$1'/# Remove spaces following an
  opening bracket followed by a single quote.
\item
  \verb#s/ "((\,|\;|\:|\?|\!|\.)[^"\.\w]*( |$))/"$1/# Remove spaces
  preceding a double quote followed by a comma, semicolon, colon,
  question mark, exclamation point, or single period. 
\item
  \verb#s/ '((\,|\;|\:|\?|\!|\.)[^'\.\w]*( |$))/'$1/# Remove spaces
  preceding a single quote followed by a comma, semicolon, colon,
  question mark, exclamation point, or single period. 
\item
  \verb#s/s\. ' /s\.' /# Remove spaces between \verb|s|. and a
  floating single quote (assumed to be a split plural possessive).
\item
  \verb#s/([^']) 's([^'\w]*( |$))/$1's$2/# Remove spaces preceding
  \verb|'s| optionally followed by non-single-quote punctuation marks
  (assumed to be a split possessive).
\end{itemize}

If there are any remaining quote characters surrounded by spaces, we
classify them as either opening quotes or closing quotes by a
procedure that scans the paragraph from left to right, using
heuristics regarding the relative ordering of opening double quotes,
closing double quotes, opening single quotes, and closing single
quotes. For each quote character surrounded by spaces classified as an
opening quote, we remove the following space, and for each one
classified as a closing quote, we remove the preceding space.

We classify quote characters surrounded by spaces as either opening or
closing quotes, by trying to keep track of whether we are inside a
quoted text string marked by double quotes, inside a quoted text
string marked by single quotes, both, or neither. We assume that a
single-quoted text string may occur inside a double-quoted text
string, but not vice-versa. The latter is not completely true, but
double-quoted text strings inside single-quoted text strings are
exceedingly rare, so we choose to ignore that possibility. Our
heuristics refer to a ``double-quote status'' and a ``single-quote
status'', which can be ``open'' or ``closed'' at a particular position
in the paragraph, indicating whether or not the text position seems to
lie within a double-quoted or single-quoted text string. The heuristic
procedure we use is the following:
\begin{itemize}
\item
  At the beginning of a paragraph, the double-quote status and
  single-quote status are both closed.
\item
  From left to right, for each maximal substring of non-space characters:
  \begin{itemize}
  \item
    If the substring consists of one double quote character:
    \begin{itemize}
    \item
      If the double-quote status is ``closed'':
      \begin{itemize}
      \item
        Classify this double quote character as an opening quote.
      \item
        Set the double-quote status to ``open''.
      \end{itemize}
      Otherwise, if the double-quote status is ``open'':
      \begin{itemize}
      \item
        Classify this double quote character as a closing quote.
      \item
        Set the double-quote status to ``closed''.
      \item
        Set the single-quote status to ``closed''.
      \end{itemize}
    \end{itemize}
  \item
    Otherwise, if the substring consists of one single quote character:
    \begin{itemize}
    \item
      If the single-quote status is ``closed'':
      \begin{itemize}
      \item
        Classify this single quote character as an opening quote.
      \item
        Set the single-quote status to ``open''.
      \end{itemize}
    \item
      Otherwise, if the single-quote status is ``open'':
      \begin{itemize}
      \item
        Classify this single quote character as a closing quote.
      \item
        Set the single-quote status to ``closed''.
      \end{itemize}
    \end{itemize}
  \item
    Otherwise, if the substring contains alphanumeric characters:
    \begin{itemize}
    \item
      If the substring contains a double quote character preceding all
      alphanumeric characters: 
      \begin{itemize}
      \item
        Set the double-quote status to ``open''.
      \item
        If the substring contains a single quote character after the
        double quote character and preceding all alphanumeric
        characters, set the single-quote status to ``open''.
      \item
        Otherwise, set the single-quote status to ``closed''.
      \end{itemize}
    \item
      Otherwise, if the substring contains a single quote character
      preceding all alphanumeric characters, set the single-quote status to ``open''.
    \item
      From left to right, for each double quote character in the
      substring that is both preceded and followed by alphanumeric
      characters:
      \begin{itemize}
      \item
        Set the single-quote status to ``closed''.
      \item
        If the double-quote status is ``open'', set the double-quote
        status to ``closed''.
      \item
        Otherwise, If the double-quote status is ``closed'', set the
        double-quote status to ``open''.
      \end{itemize}
    \item
      If the substring contains a double quote character following all
      alphanumeric characters: 
      \begin{itemize}
      \item
        Set the double-quote status to ``closed''.
      \item
        Set the single-quote status to ``closed''.
      \end{itemize}
    \item
      Otherwise, if the substring contains a single quote character following all
      alphanumeric characters, set the single-quote status to ``closed''.
    \end{itemize}
  \item
    Otherwise, if the substring contains no alphanumeric characters:
    \begin{itemize}
    \item
      If the substring begins with a double quote character, with a
      single quote character somewhere later in the substring:
      \begin{itemize}
      \item
        Set the double-quote status to ``open''.
      \item
        Set the single-quote status to ``open''.
      \end{itemize}
    \item
      Otherwise, if the substring begins with a double quote
      character:
      \begin{itemize}
      \item
        Set the double-quote status to ``open''.
      \item
        Set the single-quote status to ``closed''.
      \end{itemize}
    \item
      Otherwise, if the substring begins with a single quote
      character, set the single-quote status to ``open''.
    \item
      If the substring ends with a double quote character:
      \begin{itemize}
      \item
        Set the double-quote status to ``closed''.
      \item
        Set the single-quote status to ``closed''.
      \end{itemize}
    \item
      Otherwise, if the substring ends with a single quote
      character, set the single-quote status to ``closed''.
    \end{itemize}
  \end{itemize}
\end{itemize}
As previously mentioned, for each quote character surrounded by spaces
that is classified as an opening quote by this procedure, we remove
the following space, and for each one classified as closing quote, we
remove the preceding space.

At this point, we make a final check for pairs of consecutive
single quote characters. We replaced these earlier in the
normalization process with double quotes, but new instances may have
been introduced by the subsequent removal of spaces. We again replace
any pairs of consecutive single quote characters with a double quote
character, by repeated application of the substitution command
\verb#s/''/"/#.

\chapter{Filtering training, tuning, and test data for predicting missing
  spaces}

As mentioned in Section~10.1, possible candidates for inserting spaces
are filtered by a set of regular expression tests designed to identify
certain instances of punctuation marks not followed by white space
that should not be considered candidates for inserting spaces.  These
regular expression tests are intended to filter out hyphenated
abbreviations, decimal numbers, strings of multiple initials,
inflected forms of abbreviations, email addresses, and
computer filenames, domain names, URLs, and related expressions.

All these filters are applied to tuning and test data at inference
time. Most are also used to filter the data used to train our
white-space probability model, but the filters for email addresses and
computer filenames, domain names, URLs, and related expressions are
not applied to training data. The reason for this is that most of the
expressions caught by those two filters are identified by substrings
like \verb|http| or \verb|www| for URLs, and the \verb|@| symbol for
email addresses.  Many URLs and internet domain names, however, do not
contain any of these special strings, so sometimes we have to rely on
our missing-space predictor not to insert spaces in these, and the
examples that do match the filters for email addresses and computer
filenames, domain names, URLs, and related expressions are useful
training data for these harder examples.

Below, we present Boolean combinations of regular expression tests
implementing these filters. In these tests, the variable
\verb|$string| refers to the entire sequence of characters between two
white spaces containing a potential missing-space candidate position,
the variable \verb|$prefix| refers to the substring of \verb|$string|
preceding the candidate position, and the variable \verb|$suffix|
refers to the substring of \verb|$string| following the candidate
position. We also present examples of strings from the LDC Gigaword
corpus identified by these tests, with the symbol $\|$ marking the
candidate positions excluded.

\section{Hyphenated abbreviations}

Hyphenated abbreviations are identified by the regular
expression test \\
\\
\verb#($suffix =~ /^([^!\?\.\w]*[^\-\w])?\-($|[^\-])/)# \\
\\
which matches examples such as
\begin{quote}
  \verb#(37-sq.#$\|$\verb#-mile)# \\
  \verb#Lt.#$\|$\verb#-Colonel# \\
  \verb#Corp.#$\|$\verb#-controlled# \\
  \verb#D.C.#$\|$\verb#-native# \\
  \verb#"Prod-vs.#$\|$\verb#-Prod"# \\
  \verb#U.S.#$\|$\verb#-U.K.#
\end{quote}

\section{Decimal numbers}

Decimal numbers are identified by the regular
expression test \\
\\
\verb#(($prefix =~ /\d\.$/) &&# \\
\verb# ($suffix =~ /^\d/))# \\
\\
which matches examples such as
\begin{quote}
  \verb#379.#$\|$\verb#90# \\
  \verb#6.#$\|$\verb#5-million-US)# \\
  \verb#3,808.#$\|$\verb#50# \\
  \verb#380.#$\|$\verb#90/381.40# \\
  \verb#380.90/381.#$\|$\verb#40# \\
  \verb#$46,714.#$\|$\verb#28#
\end{quote}

\section{Strings of multiple initials}

Strings of multiple initials are identified by the regular
expression test \\
\\
\verb#((($prefix =~ /(^|[\W_\d])([A-Z]\.)+$/) &&# \\
\verb#  (($prefix =~ /(^|[\W_\d])[A-Z]\..*\/([A-Z]\.)+$/) ||# \\
\verb#   ($suffix =~ /^(.*\/)?[A-Z]([^\'\w]|$)/) ||# \\
\verb#   ($suffix =~ /^(.*\/)?[A-Z]'[sSdD]?(\W|_|$)/))) ||# \\
\verb# (($prefix =~ /(^|[\W_\d])([a-zA-Z]\.)+$/) &&# \\
\verb#  (($prefix =~ /(^|[\W_\d])[a-zA-Z]\..*\/([a-zA-Z]\.)+$/) ||# \\
\verb#  ($suffix =~ /^(.*\/)?[a-zA-Z]([^\'\w]|$)/) ||# \\
\verb#  ($suffix =~ /^(.*\/)?[a-zA-Z]'[sd]?(\W|_|$)/))))# \\
\\
which matches examples such as
\begin{quote}
  \verb#a.#$\|$\verb#m.# \\
  \verb#P.#$\|$\verb#M.# \\
  \verb#A.#$\|$\verb#R.K.# \\
  \verb#A.R.#$\|$\verb#K.# \\
  \verb#U.#$\|$\verb#S.-sponsored# \\
  \verb#M.#$\|$\verb#P.'s#
\end{quote}

\section{Inflected forms of abbreviations}

Inflected forms of abbreviations are identified by the regular
expression test \\
\\
\verb#(($suffix =~ /^'[sSdD][^'\w]*$/) ||# \\
\verb# (($suffix =~ /^[sSdD][^'\w]*$/) &&# \\
\verb#  ($prefix =~ /'$/)) ||# \\
\verb# (($suffix =~ /^(s|s')(\W|_)*$/) &&# \\
\verb#  ($prefix =~ /(!|\?|\.)$/)))# \\
\\
which matches examples such as
\begin{quote}
  \verb#Corp.#$\|$\verb#'s# \\
  \verb#Corp.'#$\|$\verb#s# \\
  \verb#M.P.#$\|$\verb#'s# \\
  \verb#M.P.'#$\|$\verb#s# \\
  \verb#Corp.#$\|$\verb#s# \\
  \verb#PhD.#$\|$\verb#s# \\
  \verb#c.c.#$\|$\verb#'d# \\
  \verb#c.c.'#$\|$\verb#d#
\end{quote}
  
\section{Email addresses}

Email addresses are identified by the regular
expression test \\
\\
\verb#(($suffix =~ /\w(@|\(a\)|\(at\)|\(AT\))\w/) ||# \\
\verb# (($prefix =~ /\w(@|\(a\)|\(at\)|\(AT\))\w/) &&# \\
\verb#  (($suffix =~ /\..*\w/) ||# \\
\verb#   ((($suffix !~ /^[A-Z][a-z\.]/) ||# \\
\verb#     ($prefix !~ /[\.\/]([0-9]*\-?[a-z][\w\-]*|[0-9]+)\/?\.$/)) &&# \\
\verb#    ($suffix !~ /^["'\(\{\[]/) &&# \\
\verb#    ($prefix !~ /["'\)\}\]][!\?\.]?$/)))))# \\
\\
which matches examples such as
\begin{quote}
  \verb#graphics(a)afp.#$\|$\verb#com# \\
  \verb#john_paul_ii@vatican.#$\|$\verb#va.# \\
  \verb#Ticket@fifa.#$\|$\verb#de# \\
  \verb#ldn.#$\|$\verb#pas@afp.com# \\
  \verb#ldn.pas@afp.#$\|$\verb#com# \\
  \verb#JohnDoe(at)aol.#$\|$\verb#com#
\end{quote}

\section{Computer filenames, domain names, URLs, and related expressions}

Computer filenames, domain names, URLs, and related expressions are
identified by the regular expression test \\
\\
\verb#((($string =~ /\w[\w\(\)\-~\/]*\.([\w\(\)\-~\/]*# \\
\verb#              \w[\w\(\)\-~\/]*\.)+[\w\(\)\-~\/]*\w/x) ||# \\
\verb#  ($string =~ /\w[\w\(\)\-~\/]*\/([\w\(\)\-~\/]*# \\
\verb#              \w[\w\(\)\-~\/]*\.)+[\w\(\)\-~\/]*\w/x) ||# \\
\verb#  ($prefix =~ /(^|\W|_)(www|WWW|Www|http|HTTP|Http)|\:\/\//)) &&# \\
\verb# (($prefix !~ /(^|[\W_\d])[A-Za-z]\.([A-Za-z]\.)+$/) ||# \\
\verb#  ($prefix =~ /(^|\W|_)(www|WWW|Www|http|HTTP|Http)|\:\/\//) ||# \\
\verb#  ($prefix =~ /\.\w\w+\./)) &&# \\
\verb# ($suffix !~ /^([A-Za-z]\.)+[A-Za-z][\W_]*$/) &&# \\
\verb# ((($suffix =~ /\/|[^\.]\.[^\.][^\.]|[^\.][^\.]\.[^\.]/) &&# \\
\verb#   ($suffix !~ /^["'\(\{\[]/)) ||# \\
\verb#  ($prefix =~ /\?$/) ||# \\
\verb#  ($prefix =~ /(^|\W|_)(www|WWW|Www)\w*(\.|\/)$/) ||# \\
\verb#  ((($suffix !~ /^[A-Z][a-z\.]/) ||# \\
\verb#    ($prefix !~ /(^|[\.\/])([0-9]*\-?[a-z][\w\-]*|# \\
\verb#                  [0-9\-]+)\/?\.$/x)) &&# \\
\verb#   ($suffix !~ /^["'\(\{\[]/) &&# \\
\verb#   ($prefix !~ /["'\)\}\]][!\?\.]?$/))))# \\
\\
which matches examples such as
\begin{quote}
  \verb#http://thomas.#$\|$\verb#loc.gov# \\
  \verb#http://thomas.loc.#$\|$\verb#gov# \\
  \verb#"MY.#$\|$\verb#G.NiE"# \\
  \verb#"MY.G.#$\|$\verb#NiE"# \\
  \verb#www.#$\|$\verb#partenia.fr,# \\
  \verb#www.partenia#$\|$\verb#.fr,# \\
  \verb#soc.#$\|$\verb#culture.singapore.# \\
  \verb#soc.culture.#$\|$\verb#singapore.# \\
  \verb#planet-rugby.#$\|$\verb#co.uk.# \\
  \verb#planet-rugby.co.#$\|$\verb#uk.# \\
  \verb#"LOVE-LETTER-FOR-YOU#$\|$\verb#.txt.vbs"# \\
  \verb#"LOVE-LETTER-FOR-YOU.txt.#$\|$\verb#vbs"# \\
  \verb#W32.#$\|$\verb#Mydoom.O,# \\
  \verb#W32.Mydoom.#$\|$\verb#O,# \\
  \verb#bcp/dr/.#$\|$\verb#hd#
\end{quote}

\chapter{Criteria for annotating tuning data for predicting missing
  spaces}

As discussed in Section~10.3, we annotate the potential tuning
examples for predicting missing spaces with one of the following three
labels:
\begin{itemize}
\item
  Likely a sentence break
\item
  Likely not a sentence break
\item
  Don't care
\end{itemize}

In text that is more complex than a simple sequence of sentences, a
text position between a full sentence ending in standard
sentence-final punctuation and subsequent nonsentence text is labeled
as a sentence break. In particular, a position between a full sentence
and a single letter or number, possibly followed by additional
well-formed text, is labeled as a sentence break, since the single
letter or number might be a footnote marker, given that superscript
positions are not indicated in the data. In a numbered or lettered
list item, the boundary between the period following the number or
letter and the list item is labeled as not a sentence break, to be
consistent with the way that LDC treebanks are sentence broken.

``Don't care'' examples are those we choose to exclude from the tuning
data for a variety of reasons, including
\begin{itemize}
\item
  Markup that should have been removed in preprocessing
\item
  Tables of nonwords or other nonsentence text
\item
  Nontext binary data
\item
  Languages other than English
\item
  Garbled character strings
\item
  Examples that we find too hard to decide
\end{itemize}

In the English Gigaword corpus, we find many periods and exclamation
points as part of idiosyncratic markup, or idiosyncratic markup
following sentence final punctuation with no intervening space. The
LDC seems to have made an attempt to replace the various types of
markup used by the different wire services with uniform SGML markup,
but they missed a lot of cases. We do not consider it to be part of
the task of a sentence break predictor to recognize arbitrary
formatting markup, so we mark all resulting examples as ``don't
care''.

There are several types of ``garbled'' text that we chose to exclude
from the data we use to tune the missing space predictor. The most
extreme kind of garbled text is apparently random character sequences
caused by data transmission noise in receiving text from the wire
services. Other types of text we consider garbled include text with
significant substrings missing or text where spaces between words are
missing, resulting in bad tokenization, and text where question marks
are inserted or substituted for non-ASCII characters. Errors that
appear to be human-produced typos in otherwise fluent text are not
considered garbled.

Other rare cases that we treat as ``don't care'' examples include
testing text, e.g.,
\begin{quote}
  \verb|this is a test.this is a test.|
\end{quote}
and the position between a dateline and a dash preceding a sentence.

Below we present some examples from the English Gigaword corpus that
we label ``don't care,'' with the symbol $\|$ marking one of the
candidate positions excluded:

\begin{quote}
  \verb#Robert Dederick, an economiE!#$\|$\verb#Q!9IQ!I9QIUMQ said# \\
  \verb#Christmas is running on par with the rest# \\
  \\
  \verb#!arrow!!off!#$\|$\verb#Kings president Tim Leiweke has already# \\
  \verb#earned his six-figure salary.# \\
  \\
  \verb#(italics)Announcing a sham "private placement" of#
  \verb#stock.#$\|$\verb#(end italics)# \\
  \\
  \verb#Haddon, who ran former!#$\|$\verb#U*S. Sen. Gary Hart's# \\
  \verb#1988 presidential campaign, represented John Ramsey# \\
  \\
  \verb#she'd also like to address such issues as drug# \\
  \verb#abuse.#$\|$\verb#ions of dwarfs.# \\
  \\
  \verb#a 16-nation climate gathering starting Thursday in#
  \verb#Washington, hosted byU.#$\|$\verb#S. President George W. Bush.# \\
  \\
  \verb#can be produced in myriad fabrics for formal or# \\
  \verb#in[\F?$\|$1IW;+kw B "U;L[XU;Q Vv.#$\|$\verb#Zv7ed it because he# \\
  \verb#dislikes multiple-cushion couches.# \\
  \\
  \verb#World Showcase features its holiday traditions through# \\
  \verb#Dec.?#$\|$\verb#30.# \\
  \\
  \verb#16.0-0 Ra7 17.Qh5 Raf7 18.#$\|$\verb#Rad1 Ne7 19.Nxe7+ Qxe7# \\
  \\
  \verb#GREENBELT, Md.#$\|$\verb#--The alleged ringleader of a street# \\
  \verb#racing club# \\
  \\
  \verb#her daughter spoke Vietnamese the girl would say, "Con# \\
  \verb#mu?#$\|$\verb#i n?y n? c q·a. N? c??n con!"# \\
  \\
  \verb#Closer to home, you?#$\|$\verb#ll find quilts made by the Art# \\
  \verb#Quilt Association# \\
\end{quote}

\chapter{Categorization of prediction errors and prediction differences, with examples}

We have analyzed the results of sentence break prediction on our test
data, to understand what types of sentence breaking examples remain
difficult to predict correctly even with our best sentence break
predictor, and also to understand the relative advantages and
disadvantages of different predictors. The analysis was performed by
manual inspection of all the errors made by a sentence break
predictor, or all the differences in predictions between two sentence
break predictors, on a particular test set. The major effects observed
in this analysis are discussed in depth in Chapter~11.

The analysis consists of noting features of each error example or
difference example that might have contributed to the difficulty of
predicting that example correctly. These features are based on
superficial observation of the examples, informed by an understanding
of how the predictors work and how they are trained, but no tracing of
exactly how the predictors come to make particular predictions has
been performed.

We group the examples into categories according to what combination of
features they exhibit. In analyzing errors, we distinguish categories
of examples exhibiting the same features according to whether the
error is a false positive or a false negative. In comparing
differences between predictors, we distinguish categories according to
which predictor made the correct prediction, and whether the correct
prediction was a break or no break. We also create separate categories
when we believe the reference annotation may be in error.  We count
the number of examples falling into each category, and present one or
more examples for each category, with more examples presented for the
most frequent categories. In the presented examples, no attempt has
been made to preserve the original line breaking, except in a few
cases for clarity, as indicated.\footnote{In any case, line breaking
within a paragraph is not taken into account by any of our sentence
break predictors.}

\section{Categories of prediction errors on our out-of-domain test set for our best sentence break predictor}

In this section, we categorize the errors made on our out-of-domain
LDC English Web Treebank test set by our best overall sentence break
predictor, the SVM with a paragraph-break-probability signal
represented as the log probability of a non-sentence-break (NSBLP
SVM), plus heuristic prediction of sentence breaks at paragraph
breaks, plus prediction of missing spaces to allow prediction of
sentence breaks where there is no white space in the raw text. This is
the predictor evaluated on the LDC English Web Treebank test set in
Section~10.6.



\begin{itemize}
\item
  144 instances of sentence breaks followed by uncapitalized text (false
  negative)
  \begin{quote}
    \verb|looking for a surprise spot to take my bf.| $\|$ \verb|a bar| \\
    \verb|would be nice but also something extremely unique.| \\ 
    \\
    \verb|Should I just settle and book on line?| $\|$ \verb|or is it| \\
    \verb|better if I wait till I get there?| \\
    \\
    \verb|will i have to pay customs in NZ.| $\|$ \verb|its a gift from| \\
    \verb|my brother.| \\
    \\
    \verb|Anyone have any suggestion?| $\|$ \verb|thank you very much.| \\
    \\
    \verb|colorado beat texas a&m.| $\|$ \verb|i know you remember the| \\
    \verb|bet.| \\
    \\
    \verb|Let me tell you how this works and most important,| \\
    \verb|why it works..........| $\|$ \verb|also make sure you print this| \\
    \verb|out NOW, so you can get the information off of it,| \\
    \verb|as you will need it.| \\
    \\
    \verb|He didn't take a dislike to the kids for "no" reason!| \\
    $\|$ \verb|cats react to the treatment they receive, they are| \\
    \verb|not toys.|
\end{quote}
\item
  41 instances of ordinary\footnote{Most ellipses are indicated with a
  series of either three or four periods. We refer to these as
  ordinary ellipses. Ellipses indicated either by two periods or five
  or more periods are referred to as unusual ellipses.} ellipses used
  to anonymize email addresses (false positive)
  \begin{quote}
    \verb|If you've successfully used any of these supplement to| \\
    \verb|aid weight loss, please write me at gottlie ...| \\
    $\|$ \verb|@yahoo.com and let me know, telling me about your| \\
    \verb|experience.| \\
    \\
    \verb|Email: "Dharmadeva"<dharmad ...| $\|$ \verb|@gmail.com>| \\
    \\
    \verb|Email: BBC Breaking News Alert <dailyem ...| \\
    $\|$ \verb|@ebs.bbc.co.uk>|
  \end{quote}
\item
  37 instances of possible missing sentence breaks in reference
  annotations (true positive)
  \begin{quote}
    \verb|She puts mayonnaise on aspirin.| $\|$ \verb|(<- clearly the| \\
    \verb|winner) Her cereal bowl came with a lifeguard.| \\
    \\
    \verb|sorry but i can't help my stupidity =( EDIT: I love how| \\
    \verb|she says it's not callum but doesn't deny it's me....|$\|$\verb|i| \\
    \verb|think we all know who she's talking about...| \\
    \\
    \verb|Irony is dead...|$\|$\verb|Long live Irony!| \\
    \\    
    \verb|I caught the third day of Corey's Jeopardy run last night| \\
    \verb|here at the office (yes, I had to come back after the| \\
    \verb|play...|$\|$\verb|I love my job!).| \\
    \\
    \verb|N.O.?| $\|$ \verb|Atlanta?| \\
    \\    
    \verb|Then it's real easy to call Ben Taub to find out a room| \\
    \verb|number.?| $\|$ \verb|The address is below.?| \\
    \\
    \verb|PS.| $\|$ \verb|Love the new website!|
  \end{quote}
\item
  30 instances of unusual sentence-final punctuation (false positive)
  \begin{quote}
    \verb|A little help is always great.| $\|$ \verb|=) Thank you or your| \\
    \verb|feed back.| \\
    \\
    \verb|PORTILLO'S OR WHITE CASTLE!| $\|$ \verb|:D Help findin a restaurant| \\
    \verb|for anniversary in SF?| \\
    \\
    \verb|Patience is the key.| $\|$ \verb|=) he might never be a cuddly,| \\
    \verb|loving, tame bird, but you can still have a friendly bond.| \\
    \\
    \verb|Ignore the answer above me, btw!| $\|$ \verb|^_^ What are good| \\
    \verb|B&W software's Photography?| \\
    \\
    \verb|Nothing compares to a home made product that really stands| \\
    \verb|the test of time.| $\|$ \verb|- The Brick, Ikea, and Leon's have| \\
    \verb|their place.|
  \end{quote}
\item
  30 instances of periods following list item numbers (false positive)
  \begin{quote}
    \verb|1.| $\|$ \verb|I found the title to the 4 wheeler and I was going to| \\
    \verb|change it over to you name.| \\
    \\
    \verb|2.| $\|$ \verb|I'm getting ready to divy up the Dow stock.| \\
    \\
    \verb|3.| $\|$ \verb|In Section 6.1, change 20 Business Days to 20 days.|
  \end{quote}
\item
  19 instances of abbreviations followed by capitalized words that are
  usually capitalized (false positive)
  \begin{quote}
    \verb|Attn.| $\|$ \verb|GCP_London: There's a new EOL Counterparty listed| \\
    \verb|in the UK.| \\
    \\
    \verb|The management from Julie and Janice to the work staff,| \\
    \verb|esp.| $\|$ \verb|Edwin, are just wonderful.| \\
    \\
    \verb|Regards, Vangie McGilloway Constellation Power Source,| \\
    \verb|Inc.| $\|$ \verb|("CPS")|
  \end{quote}
\item
  10 instances of sentences ending in \verb|!| or \verb|?| (false negative)
  \begin{quote}
    \verb|I don't know what type of bird it is, probably a dove?| \\
     $\|$ \verb|What should I feed it?| \\
    \\
    \verb|Hooray, hoorah!| $\|$ \verb|Needs to go to $41 and then I will be| \\
    \verb|happy.| \\
    \\
    \verb|Has the disaster affected the way other countries view| \\
    \verb|the US?| $\|$ \verb|Send us your comments using the form.|
  \end{quote}
\item
  8 instances of sentence breaks followed by text starting with a numerical expression
  (false negative)
  \begin{quote}
    \verb|I have 4 Golden Wonder Killifish.| $\|$ \verb|1 of them is really| \\
    \verb|colorful and is about 6cm long.|
  \end{quote}
\item
  8 instances of spurious inserted spaces (false positive)
  \begin{quote}
    \verb|- 47K202!.|$\|$\verb|DOC|
  \end{quote}
\item
  7 instances of sentence-final missing white spaces not recognized,
  with the break followed by uncapitalized text (false negative)
  \begin{quote}
    \verb|some members of the traveler community bare knuckle| \\
    \verb|box.|$\|$\verb|other than that i don't know.|
  \end{quote}
\item
  7 instances of ordinary ellipses followed by capitalized words that
  are not usually capitalized (false positive)
  \begin{quote}
    \verb|And It's Not Hard To Do....| $\|$ \verb|IF You Know HOW.|
  \end{quote}
\item
  6 instances of ordinary ellipses followed by capitalized words that
  are usually capitalized (false positive)
  \begin{quote}
    \verb|The key is...|$\|$\verb|I have a few problems:|
  \end{quote}
\item
  6 instances of possibly spurious question marks (false negative)
  \begin{quote}
    \verb|KEEP UP THE GOOD WORK.| $\|$ \verb|?|
  \end{quote}
\item
  5 instances of sentences ending in unusual ellipses (false negative)
  \begin{quote}
    \verb|XBOX Joys are soo comfyy.....|$\|$\verb|Performance - You get the| \\
    \verb|same in all, just get a good TV.|
  \end{quote}
\item
  5 instances of a sentence-internal \verb|etc.| (false positive)
  \begin{quote}
    \verb|We envision a format that would provide a brief summary| \\
    \verb|and analysis of the order, filing, etc.| $\|$ \verb|with a link| \\
    \verb|to the source document as well for those who would like| \\
    \verb|more detailed information.|
  \end{quote}
\item
  5 instances of abbreviations followed by numerical expressions (false
  positive)
  \begin{quote}
    \verb|Essie and Leon have proposed xferring Co.| $\|$ \verb|1691 to| \\
    \verb|your world (see below).|
  \end{quote}
\item
  5 instances of possibly spurious sentence-internal periods (false
  positive)
  \begin{quote}
    \verb|The sauce was dry and the enchiladas did not taste| \\
    \verb|good.|$\|$\verb|at all.|
  \end{quote}
\item
  5 instances of unusual abbreviations (false positive)
  \begin{quote}
    \verb|Either way rec.| $\|$ \verb|and dom.| $\|$ \verb|white genes dilute the black| \\
    \verb|gene.|
  \end{quote}
\item
  4 instances of unusual sentence-internal punctuation (false
  positive)
  \begin{quote}
    \verb|VERYYYY!!!!| $\|$ \verb|VERYYY!!| $\|$ \verb|Good auto repair men.|
  \end{quote}
\item
  4 instances of sentences ending in initials, followed by capitalized
  words that are not usually capitalized (false negative)
  \begin{quote}
    \verb|Attached for your further handling is the draft form| \\
    \verb|Deemed ISDA between ENA and Ispat Mexicana S.A. de C.V.| \\
     $\|$ \verb|Please note that the economic provisions have been| \\
    \verb|omitted.|
  \end{quote}
\item
  4 instances of missing spaces not recognized (false negative)
  \begin{quote}
    \verb|Don't give these guys a penny.|$\|$\verb|----cgy|
  \end{quote}
\item
  4 instances of abbreviations as part of capitalized expressions (false
  positive)
  \begin{quote}
    \verb|1579 - EBS Network Co.| $\|$ \verb|Division of 17H.|
  \end{quote}
\item
  4 instances of initials as part of capitalized expressions (false
  positive)
  \begin{quote}
    \verb|Bush also nominated A.| $\|$ \verb|Noel Anketell Kramer for a 15-year| \\
    \verb|term as associate judge of the District of Columbia Court| \\
    \verb|of Appeals, replacing John Montague Steadman.|
  \end{quote}
\item
  3 instances of possibly spurious question marks (false positive)
  \begin{quote}
    \verb|We would like to meet between November 14 and?|$\|$\verb|December| \\
    \verb|1?since construction of our services is already underway.?|
  \end{quote}
\item
  3 instances of paragraph-initial ordinary ellipses (false positive)
  \begin{quote}
    \verb|...|$\|$\verb|Now comes the fun part....|
  \end{quote}
\item
  3 instances of initials followed by capitalized words that are not usually
  capitalized (false positive)
  \begin{quote}
    \verb|i.e.| $\|$ \verb|Are we covered if we proceed a word with "Enron"?|
  \end{quote}
\item
  2 instances of abbreviations followed by capitalized words that are not usually
  capitalized (false positive)
  \begin{quote}
    \verb|This insurance co.| $\|$ \verb|Is a joke !!!|
  \end{quote}
\item
  2 instances of sentence-final quotations marked by single quotes
  (false negative)
  \begin{quote}
    \verb|It's a Western scene by a guy named W.H.S. Koerner called|
    \verb|'A Charge to Keep.'| $\|$ \verb|It's on loan, by the way, from a guy|
    \verb|named Joe O'Neill in Midland, Texas.|
  \end{quote}
\item
  2 instances of unusual ellipses (false positive)
  \begin{quote}
    \verb|When I was Younger In my early 20s My farrier was A hottie| \\
    \verb|named Joby..|$\|$\verb|( for those who don't Know I AM A GIRL read| \\
    \verb|my Bio on my Page) dumb as a post about everything But| \\
    \verb|horses..|
  \end{quote}
\item
  2 instances of sentence breaks with nothing unusual (false negative)
  \begin{quote}
    \verb|Great Service, Thanks Don.| $\|$ \verb|Nice Top Lights.|
  \end{quote}
\item
  2 instances of sentence breaks followed by text with unusual starting
  characters (false negative)
  \begin{quote}
    \verb|The #1 Newsgroup Service in the World!| $\|$ \verb|-----== Over| \\
    \verb|100,000 Newsgroups - 19 Different Servers! =-----|
  \end{quote}
\item
  2 instances of initials followed by capitalized words that are
  usually capitalized (false positive)
  \begin{quote}
    \verb|Attached is a new link for employees unable to attend the| \\
    \verb|all-employee meeting today at 10 a.m.| $\|$ \verb|(CDT) at the| \\
    \verb|Hyatt Regency Houston, Imperial Ballroom.|
  \end{quote}
\item
  2 instances of sentences ending in unusual ellipses, followed by
  capitalized words that are usually capitalized (false negative)
  \begin{quote}
    \verb|Heres my reason.....| $\|$ \verb|I did the same thing that you did| \\
    \verb|for christmas.|
  \end{quote}
\item
  2 instances of unusual ellipses followed by capitalized words that
  are not usually capitalized (false positive)
  \begin{quote}
    \verb|Whatever you do ...............|$\|$\verb|DO NOT try to come here| \\
    \verb|with out legal permission.|
  \end{quote}
\item
  2 instances of possible incorrect sentence breaks in reference
  annotations (true negative)
  \begin{quote}
    \verb|Your mom just called and said your dad's surgery will be| \\
    \verb|at 2:30 p.m.| $\|$ \verb|Tuesday, March 20th.|
  \end{quote}
\item
  1 instance of unusual sentence-final punctuation (false negative)
  \begin{quote}
    \verb|Do You Yahoo!?| $\|$ \verb|Send instant messages & get email alerts| \\
    \verb|with Yahoo! Messenger.|
  \end{quote}
\item
  1 instance of a sentence ending in an initial, followed by a
  capitalized word that is usually capitalized (false negative)
  \begin{quote}
    \verb|Ma said she had a great time in Houston, She was happy to| \\
    \verb|see yourself, as well as C & D.| $\|$ \verb|I think she likes her| \\
    \verb|special room in your house.|
  \end{quote}
\item
  1 instance of a quotation-initial ordinary ellipsis (false positive)
  \begin{quote}
    \verb|The change is mainly on the Settlement Date text, which| \\
    \verb|instead of "...|$\|$\verb|Two business days after the date of the| \\
    \verb|Transaction..." shall read as follow:|
  \end{quote}
\item
  1 instance of an ordinary ellipsis followed by a numerical
  expression (false positive)
  \begin{quote}
    \verb|SMS Me...| $\|$ \verb|09819602175|
  \end{quote}
\item
  1 instance of a quotation-final exclamation point followed by a
  capitalized word that is usually capitalized (false positive)
  \begin{quote}
    \verb|"Wonderful!"| $\|$ \verb|Winston beams.|
  \end{quote}
\item
  1 instance of a sentence break followed by text starting with a capitalized
  word that is usually capitalized (false negative)
  \begin{quote}
    \verb|Yes.| $\|$ \verb|I don't think it matters|
  \end{quote}
\item
  1 instance of a sentence ending in an ordinary ellipsis, followed by
  a capitalized word that is usually capitalized (false negative)
  \begin{quote}
    \verb|some of the tests and procedures that are ran can be| \\
    \verb|costly (just like they would be for any other medical| \\
    \verb|tests elsewhere if you do not have insurance)...|$\|$\verb|I have| \\
    \verb|seen several of the providers from the office and have not| \\
    \verb|once been shown anything but care and consideration.| \\
  \end{quote}
\item
  1 instance of a sentence ending in an ordinary ellipsis, followed by
  a numerical expression (false negative)
  \begin{quote}
    \verb|Tmobile want to Send of my phone and i didn't want to go| \\
    \verb|thru that...| $\|$ \verb|3 Days For get that...|
  \end{quote}
\item
  1 instance of a sentence break followed by text starting with \verb|Please| (false
  negative)
  \begin{quote}
    \verb|I have a question about McDonald's Monopoly!| $\|$ \verb|Please| \\
    \verb|HELP!!!!!!!!!!!!!!!!!!!!!|
  \end{quote}
\item
  1 instance of unusual quotation-final-but-not-sentence-final
  punctuation (false positive)
  \begin{quote}
    \verb|He's pretty much an "I love American Food, good drinks on| \\
    \verb|occasion, laid back."| $\|$ \verb|kind of guy. ;)|
  \end{quote}
\item
  1 instance of an ordinary ellipsis followed by a sentence-internal quotation
  (false positive)
  \begin{quote}
    \verb|I also never have to wait long for a yearly inspection| \\
    \verb|sticker...and never get the usual excuses other shops| \\
    \verb|always gave me...|$\|$\verb|"the inspection guy isn't here| \\
    \verb|today"....for example.|
  \end{quote}
\item
  1 instance of \verb|Ok.| possibly mistaken for an abbreviation of
  \verb|Oklahoma| (false negative)
  \begin{quote}
    \verb|Ok.| $\|$ \verb|Contact Cindy Stark to make an appointment.|
  \end{quote}
\item
  1 instance of a possible confusion of \verb|No.| as a negative
  statement with \verb|No.| as an abbreviation of \verb|Number| (false
  negative)
  \begin{quote}
    \verb|No.| $\|$ \verb|BTA only exists in the mind of SCAMMERS.|
  \end{quote}
\item
  1 instance of a sentence-final single capital letter possibly
  mistaken for an initial (false negative)
  \begin{quote}
    \verb|Every time I go, Kevin, the manager, will always remember| \\
    \verb|my family and I.| $\|$ \verb|Overall, it is very family oriented, and| \\
    \verb|I recommend it to everyone!!!|
  \end{quote}
\item
  1 instance of a sentence ending in \verb|Yahoo!| (false negative)
  \begin{quote}
    \verb|Yahoo!| $\|$ \verb|The word was invented by Jonathan Swift and used| \\
    \verb|in his book 'Gulliver's Travels'.|
  \end{quote}
\item
  1 instance of an unusual ellipsis followed by a capitalized word
  that is usually capitalized (false positive)
  \begin{quote}
    \verb|The Atmosphere is the best..| $\|$ \verb|Italian music, candles,| \\
    \verb|helpful and friendly staff... And the food is beautiful| \\
    \verb|too !|
  \end{quote}
\item
  1 instance of a sentence break followed by text starting with a
  lower-case-lettered list item (false negative)
  \begin{quote}
    \verb|(a) Do you have a particular sensitivity about the| \\
    \verb|reference to bad faith proceedings made by United India?| \\
    $\|$ \verb|(b) Is it right that Enron has 3 sets of bad faith| \\
    \verb|proceedings in Texas?|
  \end{quote}
\end{itemize}

\section{Categories of prediction differences on our out-of-domain test
  set comparing a PBP-based predictor and an SVM with a PBP signal}



In this section, we categorize the differences in sentence break
predictions made on our out-of-domain LDC English Web Treebank test
set, between our predictor based only on the estimated paragraph break
probability (PBP-based) and the predictor based on an SVM with a
paragraph-break-probability signal represented as the log probability
of a non-sentence-break (NSBLP SVM). Both these predictors use heuristic
prediction of sentence breaks at paragraph breaks, but neither
predicts missing spaces, so no sentence breaks are predicted where
there is no white space in the raw text. These are the PBP-based and NSBLP SVM
predictors evaluated in Section~8.4.

\begin{itemize}
\item
  51 instances of PBP-based correct, sentence breaks followed by uncapitalized text
  (break)
  \begin{quote}
    \verb|Should i bring her?| $\|$ \verb|if so how quick will she adapt to a| \\
    \verb|litter box inside?| $\|$ \verb|thanks| \\
    \\
    \verb|Anyone have any suggestion?| $\|$ \verb|thank you very much.| \\
    \\   
    \verb|Call a vet would be a good idea with a sick dog VERY| \\
    \verb|CONCERNED!!!| $\|$ \verb|plz bring your dog to the vet ASAP!!!| \\
    \\
    \verb|these guys were fantastic!| $\|$ \verb|they fixed my garage doors in| \\
    \verb|literally less than an hour.| \\
    \\
    \verb|green curry and red curry is awesome!| $\|$ \verb|remember to ask| \\
    \verb|for extra vege| \\
    \\
    \verb|This office is awesome!| $\|$ \verb|everyone here is super friendly| \\
    \verb|and efficient!| $\|$ \verb|its great to know you can get great| \\
    \verb|service, great product, and for the best price all in one!|
  \end{quote}
\item
  15 instances of NSBLP SVM correct, sentence breaks followed by text starting with
  \verb|Please| (break)
  \begin{quote}
    \verb|One thought is to have just the agenda as a couple topics| \\
    \verb|and then lead in with the first two questions as I've| \\
    \verb|listed.| $\|$ \verb|Please provide me your thoughts asap.| \\
    \\
    \verb|The cost is $567.77 per chair plus tax.| $\|$ \verb|Please| \\
    \verb|approve the purchase of two.| \\
    \\
    \verb|Please send the draft to Troy Black and he will forward| \\
    \verb|to the CP.| $\|$ \verb|Please deliver a copy to me as well.| \\
    \\
    \verb|There's a new EOL Counterparty listed in the UK.| \\
     $\|$ \verb|Please respond.| \\
    \\
    \verb|My family and our colleagues will be forever grateful.| \\
    $\|$ \verb|Please reply in strict confidence to the contact numbers| \\
    \verb|below.|
  \end{quote}
\item
  14 instances of NSBLP SVM correct, sentence breaks followed by uncapitalized text
  (break)
  \begin{quote}
    \verb|answered all my questions, and called me back when I| \\
    \verb|needed something.| $\|$ \verb|highly recommended!| \\
    \\
    \verb|the only down fall of this pho house is the difficulty| \\
    \verb|in finding parking in chinatown.| $\|$ \verb|remember to bring| \\
    \verb|cash since they don't take debit or credit.| \\
    \\
    \verb|your cat is losing his territory.| $\|$ \verb|territory is very| \\
    \verb|important to cats, and they are very nervous when they go| \\
    \verb|into a new place| 
  \end{quote}
\item
  7 instances of PBP-based correct, periods following list item numbers (no
  break)
  \begin{quote}
    \verb|1.| $\|$ \verb|I tested it out by making a figure boat And after 4| \\
    \verb|days when it dried it was very fragile!| \\
    \\
    \verb|1.| $\|$ \verb|I found the title to the 4 wheeler and I was going to| \\
    \verb|change it over to you name.| \\
    \\
    \verb|2.| $\|$ \verb|I'm getting ready to divy up the Dow stock.| \\
    \\
    \verb|2.| $\|$ \verb|I did not include a "Setoff" provision in this draft,| \\
    \verb|mainly because the vast majority of the time, we will not,| \\
    \verb|nor will an affiliate, have another agreement in place| \\
    \verb|with these customers.|
  \end{quote}
\item
  7 instances of PBP-based correct, sentences ending in \verb|!| or
  \verb|?|, followed by capitalized text (break)
  \begin{quote}
    \verb|Hooray, hoorah!| $\|$ \verb|Needs to go to $41 and then I will be| \\
    \verb|happy.| \\
    \\    
    \verb|Can you accept that the road we are travelling points| \\
    \verb|toward a grim and painful future?| $\|$ \verb|Do you have the| \\
    \verb|heart to face monumental failures while bravely struggling| \\
    \verb|beyond where we are now?| \\
    \\
    \verb|Has the disaster affected the way other countries view the| \\
    \verb|US?| $\|$ \verb|Send us your comments using the form.| \\
    \\
    \verb|Why?| $\|$ \verb|Because they cut me good deals if I paid in cash.|
  \end{quote}
\item
  7 instances of NSBLP SVM marked correct, but PBP-based actually correct,
  sentences ending in unusual ellipses, with the break followed by
  uncapitalized text (break)
  \begin{quote}
    \verb|its for a craft project.....|$\|$\verb|i want to use it the way u| \\
    \verb|would use icing....| \\
    \\
    \verb|hi everyone....just hav my hands on my new OLYMPUS X940| \\
    \verb|digital camera..|$\|$\verb|wel,i always wanted 2 hav one by sony..| \\
    \\
    \verb|These guys really know their stuff ..| $\|$ \verb|they have| \\
    \verb|almost anything you could want in terms of spy and| \\
    \verb|surviellance equipment.| \\
  \end{quote}
\item
  5 instances of NSBLP SVM correct, sentence breaks followed by
  parenthesized text (break)
  \begin{quote}
    \verb|Wiki Media Foundation, the group behind the Wikipedia| \\
    \verb|online encyclopedia project, said Friday that search| \\
    \verb|giant Google has volunteered to host some of its content| \\
    \verb|on company servers.| $\|$ \verb|(ZD Net)| \\
    \\
    \verb|Former Russian bioweaponeer Ken Alibek has said that a| \\
    \verb|key Russian scientist assisted Iraq and that Russia had| \\
    \verb|the Ames strain.| $\|$ \verb|(His conclusion may have been based on| \\
    \verb|the fake mobile biolab plans foisted upon the US by the| \\
    \verb|Chalabi associate "Curveball", which Alibek divined to be| \\
    \verb|identical to Russian mobile lab design).| \\
    \\
    \verb|They actively excluded State Department Iraq hands like| \\
    \verb|Tom Warrick.| $\|$ \verb|(Only recently have a few experienced State| \\
    \verb|Department Arabists been allowed in to try to begin| \\
    \verb|mopping up the mess.)|
  \end{quote}
\item
  5 instances of NSBLP SVM correct, initials as part of capitalized
  expressions (no break)
  \begin{quote}
    \verb|It's a Western scene by a guy named W.H.S.| $\|$ \verb|Koerner| \\
    \verb|called 'A Charge to Keep.'| \\
    \\
    \verb|W.| $\|$ \verb|Don Germany, Jr| \\
    \\
    \verb|M.D.| $\|$ \verb|Nalapat, an expert on jihad, is professor of| \\
    \verb|geopolitics at the Manipal Academy of Higher Education,| \\
    \verb|India.|
  \end{quote}
\item
  4 instances of PBP-based correct, sentence breaks followed by text starting with a
  numerical expression (break)
  \begin{quote}
    \verb|If you own a Retail Store or are a Professional Vendor| \\
    \verb|who exhibits at Sport, Hunting, or Craft Shows and are| \\
    \verb|interested in selling our products,please give us a call!| \\
    $\|$ \verb|732-657-3416|\\
    \\
    \verb|The #1 Newsgroup Service in the World!| $\|$ \verb|>100,000| \\
    \verb|Newsgroups|
  \end{quote}
\item
  4 instances of NSBLP SVM correct, nonbreaking ordinary ellipses (no break)
  \begin{quote}
    \verb|We can not wait to go back to Santa Fe and to this great| \\
    \verb|B&B...|$\|$\verb|especially my 4 year old, who made friends with Ms.| \\
    \verb|Sue and all the ladies, and has talked about them since| \\
    \verb|we left!| \\
    \\
    \verb|Not only was it a good cut but my wife and friends comment| \\
    \verb|on my hair every time I leave...|$\|$\verb|saying it's the best look| \\
    \verb|I've ever had.|
  \end{quote}
\item
  4 instances of NSBLP SVM correct, abbreviations followed by numerical
  expressions (no break)
  \begin{quote}
    \verb|If you have any doubts, refer to Title 18 Sec.| $\|$ \verb|1302 &| \\
    \verb|1341 of the Postal Lottery laws.| \\
    \\
    \verb|WHERE: The Front Porch 217 Gray St.| $\|$ \verb|(713) 571-9571|
  \end{quote}
\item
  3 instances of PBP-based correct, abbreviations followed by numerical
  expressions (no break)
  \begin{quote}
    \verb|Leonardo Pacheco ext.| $\|$ \verb|39938|
  \end{quote}
\item
  3 instances of NSBLP SVM correct, unusual ellipses (no break)
  \begin{quote}
    \verb|If you have gotten through ordering, dealing with the rude| \\
    \verb|staff and if you followed the dumb rules, you are finally| \\
    \verb|presented with what you came for..|$\|$\verb|some tacos that are| \\
    \verb| "ok," but definitely not worth putting up with all the| \\
    \verb|hassle.|
  \end{quote}
\item
  3 instances of NSBLP SVM correct, ordinary ellipses followed by capitalized words
  that are usually capitalized (no break)
  \begin{quote}
    \verb|I have a Kodak Camera (10.2 Megapixels)...| $\|$ \verb|Kodak AF| \\
    \verb|5x OPTICAL LENS...|
  \end{quote}
\item
  3 instances of NSBLP SVM marked correct, but PBP-based actually correct,
  sentences ending in ordinary ellipses, with the break followed by
  uncapitalized text (break)
  \begin{quote}
    \verb|ive never been to the yorkshire moors or penines at| \\
    \verb|all...|$\|$\verb|i can see the penines in the far distance from my| \\
    \verb|bedroom window.|
  \end{quote}
\item
  3 instances of NSBLP SVM correct, sentence breaks followed by text starting with
  capitalized words that are usually capitalized (break)
  \begin{quote}
    \verb|DO NOT EVER GO HERE.| $\|$ \verb|I prefer Advanced auto parts over| \\
    \verb|this crappy place with the meanest people.|
  \end{quote}
\item
  2 instances of PBP-based correct, sentence breaks followed by text with unusual
  starting characters (break)
  \begin{quote}
    \verb|Neither was this day less fortunate to| \\
    \verb|his father Philip; for on the same day he| \\
    \verb|took Potidea; >> - JOHN AUBREY, F.R.S.| $\|$ \\
    \verb|--------------------------------------|
  \end{quote}
\item
  2 instances of NSBLP SVM correct, quotation-initial ordinary ellipses (no break)
  \begin{quote}
    \verb|Check these out: "...|$\|$\verb|there is no companion quite so| \\
    \verb|devoted, so communicative, so loving and so mesmerizing| \\
    \verb|as a rat."|
  \end{quote}
\item
  2 instances of NSBLP SVM correct, abbreviations followed by capitalized
  words that are usually capitalized (no break)
  \begin{quote}
    \verb|Sat.| $\|$ \verb|Sept. 23 - Drive to St. Tropez|
  \end{quote}
\item
  2 instances of PBP-based correct, initials followed by capitalized words
  that are usually capitalized (no break)
  \begin{quote}
    \verb|Attached is a new link for employees unable to attend the| \\
    \verb|all-employee meeting today at 10 a.m.| $\|$ \verb|(CDT) at the| \\
    \verb|Hyatt Regency Houston, Imperial Ballroom.|
  \end{quote}
\item
  2 instances of PBP-based correct, sentence-final quotations marked by
  single quotes (break)
  \begin{quote}
    \verb|It's a Western scene by a guy named W.H.S. Koerner called|
    \verb|'A Charge to Keep.'| $\|$ \verb|It's on loan, by the way, from a guy|
    \verb|named Joe O'Neill in Midland, Texas.|
  \end{quote}
\item
  1 instance of NSBLP SVM correct, an unusual abbreviation (no break)
  \begin{quote}
    \verb|John Suttle has OK'd Anthony's latest request w/r.t.| $\|$ \\
    \verb|the CSA.|
  \end{quote}
\item
  1 instance of NSBLP SVM correct, a sentence break followed by text with unusual
  starting characters (break)
  \begin{quote}
    \verb|Ok, I know that Mcdonald's has Monopoly going on right| \\
    \verb|now.| $\|$ \verb|& I wanna know what food has the game pieces.|
  \end{quote}
\item
  1 instance of NSBLP SVM correct, a sentence break followed by text starting with a
  numerical expression (break)
  \begin{quote}
    \verb|Any feedback from Rick Buy? Please, let me know.| $\|$ \\
    \verb|42299| \\
    \verb|Vince|\footnote{Line breaks as in original.}
  \end{quote}
\item
  1 instance of PBP-based correct, a sentence break with nothing unusual (break)
  \begin{quote}
    \verb|Hi.| $\|$ \verb|Well, wouldn't you know it.|
  \end{quote}
\item
  1 instance of NSBLP SVM correct, a sentence break followed by text starting with \verb|PLEASE|
  (break)
  \begin{quote}
    \verb|Keep a copy of these steps for yourself and whenever you| \\
    \verb|need money, you can use it again, and again.| $\|$ \verb|PLEASE| \\
    \verb|REMEMBER that this program remains successful because of| \\
    \verb|the honesty and integrity of the participants and by their| \\
    \verb|carefully adhering to directions.|
  \end{quote}
\item
  1 instance of NSBLP SVM correct, an abbreviation as part of a capitalized
  expression (no break)
  \begin{quote}
    \verb|I suggest that you compare Elena and Sarah's duties/ level| \\
    \verb|to other Sr.| $\|$ \verb|Spec. in your group such as Kenneth Parkhill| \\
    \verb|and Sevil to determine if they are equivalent, or if their| \\
    \verb|scope of responsibilities and experience is not as broad.|
  \end{quote}
\item
  1 instance of NSBLP SVM marked correct, but PBP-based actually correct, a
  sentence break followed by uncapitalized text (break)
  \begin{quote}
    \verb|Come visit irc server: irc.yankeedot.net and join| \\
    \verb|#audiobooks for sharing, discussion and a great trivia| \\
    \verb|game!| $\|$ \verb|large selection of fiction, science fiction and| \\
    \verb|best sellers.|
  \end{quote}
\item
  1 instance of NSBLP SVM correct, initials followed by a capitalized word
  that is usually capitalized (no break)
  \begin{quote}
    \verb|Kindly confirm your availability to attend an Analyst and| \\
    \verb|Associate Recruiting Meeting scheduled for Thursday, July| \\
    \verb|26th at 2:00 p.m. (4:00 p.m.| $\|$ \verb|Toronto time).|
  \end{quote}
\item
  1 instance of NSBLP SVM correct, unusual sentence-final punctuation
  (break)
  \begin{quote}
    \verb|Because he liked making statues of David!| $\|$ \verb|:D| \\
    \verb|Obviously, he should have been arrested and jailed|
  \end{quote}
\item
  1 instance of PBP-based correct, a sentence-final single capital letter
  possibly mistaken for an initial (break)
  \begin{quote}
    \verb|Every time I go, Kevin, the manager, will always remember| \\
    \verb|my family and I.| $\|$ \verb|Overall, it is very family oriented, and| \\
    \verb|I recommend it to everyone!!!|
  \end{quote}
\item
  1 instance of PBP-based correct, an abbreviation followed by a capitalized
  word that is usually capitalized (no break)
  \begin{quote}
    \verb|Tues.| $\|$ \verb|Oct. 3 - Fly London to Houston|
  \end{quote}
\item
  1 instance of NSBLP SVM correct, an ordinary ellipsis followed by a numerical
  expression (no break)
  \begin{quote}
    \verb|Out of the 650k est stolen works of art by the| \\
    \verb|Nazis...|$\|$\verb|70k still remain missing..and there are| \\
    \verb|thousands in musems that haven't been returned to| \\
    \verb|their right full owners and heirs..|
  \end{quote}
\item
  1 instance of NSBLP SVM correct, \verb|Yahoo!| as part of a capitalized
  expression (no break)
  \begin{quote}
    \verb|Send instant messages & get email alerts with Yahoo!| $\|$ \\
    \verb|Messenger.|
  \end{quote}
\item
  1 instance of PBP-based marked correct, but NSBLP SVM actually correct,
  initials followed by a capitalized word that is usually capitalized (no break)
  \begin{quote}
    \verb|Your mom just called and said your dad's surgery will be| \\
    \verb|at 2:30 p.m.| $\|$ \verb|Tuesday, March 20th.|
  \end{quote}
\item
  1 instance of NSBLP SVM correct, an abbreviation as part of a signature block
  (no break)
  \begin{quote}
    \verb|Kriste K. Sullivan| \\
    \verb|Enron Corp.| $\|$ \verb| -  Legal|\footnote{Line breaks as in original.}
  \end{quote}
\item
  1 instance of NSBLP SVM correct, a period following a list item number (no
  break)
  \begin{quote}
    \verb|3.| $\|$ \verb|Former U.S. President George Bush|
  \end{quote}
\item
  1 instance of PBP-based correct, a sentence ending in \verb|!|, with the text
  following the break starting with \verb|Please| (break)
  \begin{quote}
    \verb|I have a question about McDonald's Monopoly!| $\|$ \verb|Please| \\
    \verb|HELP!!!!!!!!!!!!!!!!!!!!!|
  \end{quote}
\item
  1 instance of PBP-based correct, initials followed by a capitalized word that is
  not usually capitalized (no break)
  \begin{quote}
    \verb|1542 LEONARDO, Giovanni (a.k.a.| $\|$ \verb|Il Puttino/the Boy)| \\
    \verb|born in Calaria.|
  \end{quote}
\item
  1 instance of NSBLP SVM correct, a sentence break with nothing unusual (break)
  \begin{quote}
    \verb|If you have any questions or require any assistance at| \\
    \verb|all, please contact Hope Duncan at 202.739.0134 or Jeff| \\
    \verb|Mangold at 703.729.2710.| $\|$ \verb|We are looking forward to seeing| \\
    \verb|you on September 12.|
  \end{quote}
\item
  1 instance of NSBLP SVM correct, an ordinary ellipsis followed by bracketed text (no
  break)
  \begin{quote}
    \verb|Now, FLDS leader Warren Jeffs has been added to the FBI's| \\
    \verb|list of "Ten Most Wanted Fugitives," a move that caps law| \\
    \verb|enforcement's dramatic change of approach toward the| \\
    \verb|polygamous group in recentyears...|$\|$\verb|[because of] the impact| \\
    \verb|that the group's practices, lawenforcement officials say,| \\
    \verb|are having on the most vulnerable within the sect,| \\
    \verb|particularly children and women.|
  \end{quote}
\item
  1 instance of PBP-based marked correct, but NSBLP SVM actually correct, a
  sentence ending in an ordinary ellipsis, with the break followed by
  uncapitalized text (break)
  \begin{quote}
    \verb|Bullet-proof vests, automatic weapons, helicopters, tanks,| \\
    \verb|robots ...| $\|$ \verb|the testosterone is oozing through the streets,| \\
    \verb|more prisons, longer sentences, tighten the belt, spartan| \\
    \verb|conditions, task forces, gang units, gun courts.|
  \end{quote}
\item
  1 instance of PBP-based correct, an unusual ellipsis followed by a
  capitalized word that is not usually capitalized (break)
  \begin{quote}
    \verb|XBOX Joys are soo comfyy.....|$\|$\verb|Performance - You get the| \\
    \verb|same in all, just get a good TV.|
  \end{quote}
\item
  1 instance of PBP-based correct, a sentence ending in initials, followed by a
  capitalized word that is not usually capitalized (break)
  \begin{quote}
    \verb|TO 2:00 A.M.| $\|$ \verb|WHERE: MAIN STREET SALOON|
  \end{quote}
\item
  1 instance of PBP-based correct, \verb|Ok.| possibly mistaken for an
  abbreviation of \verb|Oklahoma| (break)
  \begin{quote}
    \verb|Ok.| $\|$ \verb|Contact Cindy Stark to make an appointment.|
  \end{quote}
\item
  1 instance of PBP-based correct, a sentence ending in an unusual ellipsis,
  with the break followed by text that is uncapitalized (break)
  \begin{quote}
    \verb|Don't worry, I'll take care of you!" .......| $\|$ \verb|the rest| \\
    \verb|was history!|
  \end{quote}
\item
  1 instance of NSBLP SVM correct, a sentence ending in an ordinary ellipsis, followed by
  parenthesized text (break)
  \begin{quote}
    \verb|about our lifestyle...| $\|$ \\
    \verb|(See attached file: TEXT.htm)|\footnote{Line breaks as in original.}
  \end{quote}
\end{itemize}

\section{Categories of prediction differences on our in-domain test
  set comparing a PBP-based predictor and an SVM with a PBP signal}



In this section, we categorize the differences in predictions on the
in-domain Satz-3 test set, between the same PBP-based and NSBLP SVM sentence
break predictors compared in the previous section on our out-of-domain
test set.

\begin{itemize}
\item
  34 instances of NSBLP SVM correct, initials as part of capitalized expressions (no break)
  \begin{quote}
    \verb|Salomon Inc.'s Salomon Brothers Inc., in a move affirming| \\
    \verb|its growing emphasis on investment-banking activities,| \\
    \verb|named Denis A.| $\|$ \verb|Bovin as head of the firm's corporate-| \\
    \verb|coverage group based in New York.| \\
    \\
    \verb|Investor Asher B.| $\|$ \verb|Edelman, who bought 30% of Datapoint| \\
    \verb|Corp.'s stock last week to try to head off a battle for| \\
    \verb|control, purchased most of the shares with funds from| \\
    \verb|another company he controls.| \\
    \\
    \verb|Included in the sale are North Coast Bakery, Delight| \\
    \verb|Products, Kenlake Foods, Pontiac Foods, State Avenue,| \\
    \verb|Tara Foods, K.B.| $\|$ \verb|Specialty Foods and Pace Dairy.| \\
    \\
    \verb|In addition, several European companies, led by N.V.| $\|$ \\
    \verb|Philips of the Netherlands, have made such machines.| \\
    \\
    \verb|Mr. Finkelstein flew to San Francisco the day after the| \\
    \verb|earthquake, and found that 10 to 12 of his company's| \\
    \verb|stores had sustained some damage, including the breakage| \\
    \verb|of most windows at the I.| $\|$ \verb|Magnin store on Union Square.| \\
    \\
    \verb|N.J.| $\|$ \verb|Nicholas, president of the new Time Warner Inc.,| \\
    \verb|says the company will continue to do business "on the| \\
    \verb|merits" with Paramount, which is closely linked in a| \\
    \verb|number of business ventures and contractual relationships.|
  \end{quote}
\item
17 instances of NSBLP SVM correct, abbreviations followed by a dash
introducing the rest of the sentence (no break)
  \begin{quote}
     \verb|He points, as an example, to Quaker Oats Co.| $\|$ \verb|--| \\
     \verb|selling at a cool 21 times 12-month trailing earnings.| \\
     \\     
     \verb|Bowing to a mounting public outcry, three more major| \\
     \verb|securities firms -- Bear, Stearns & Co. Inc., Morgan| \\
     \verb|Stanley & Co. and Oppenheimer & Co.| $\|$ \verb|-- announced| \\
     \verb|Friday they would suspend stock-index arbitrage trading| \\
     \verb|for their own accounts.| \\
     \\
     \verb|Wake County, N.C.| $\|$ \verb|-- $72 million of school bonds through| \\
     \verb|a group led by Wachovia Bank & Trust Co.| \\
     \\
     \verb|Federal Home Loan Mortgage Corp.| $\|$ \verb|-- $450 million of| \\
     \verb|Remic mortgage securities being offered in 10 classes by| \\
     \verb|First Boston Corp.| \\
     \\
     \verb|Collateralized Mortgage Securities Corp.| $\|$ \verb|-- $150| \\
     \verb|million of Remic mortgage securities offered in 12 classes| \\
     \verb|by First Boston Corp.|
  \end{quote}
\item
  16 instances of NSBLP SVM correct, abbreviations as part of capitalized expressions (no break)
  \begin{quote}
    \verb|Both companies are allies of Navigation Mixte in its fight| \\
    \verb|against a hostile takeover bid launched last week by Cie.| \\
    $\|$ \verb|Financiere de Paribas at 1,850 French francs ($297) a| \\
    \verb|share.| \\
    \\
    \verb|However, Mr.| $\|$ \verb|Gorbachev must ensure that within this| \\
    \verb|"alliance" the business sector remains subordinate to the| \\
    \verb|party.| \\
    \\
    \verb|The former reflected a drop in Service Corp.| $\|$ \\
    \verb|International, which said it holds a 9.5% stake in| \\
    \verb|Diversified Energies.| \\ 
    \\
    \verb|Mr. Meyers, who currently is chairman of his own firm,| \\
    \verb|J.A.M. Enterprises, recently retired as chairman of the| \\
    \verb|Time Inc.| $\|$ \verb|Magazine Group.| \\
    \\
    \verb|"That's silly," says Fox Inc.| $\|$ \verb|Chairman Barry Diller.|
  \end{quote}
\item
  7 instances of NSBLP SVM correct, sentence breaks followed by parenthesized text (break)
  \begin{quote}
    \verb|Scientists hope that a chemical process similar to early| \\
    \verb|Earth's occurs today on Triton.| $\|$ \verb|(No one expects life| \\
    \verb|will form there, though.| \\
    \\
    \verb|Verne Q. Powell Flutes Inc. is famed for its flutes and| \\
    \verb|piccolos.| $\|$ \verb|(Its idea of a low-end flute sells for $5,700.)| \\
    \\
    \verb|Many residents say agencies supply housekeepers, then call| \\
    \verb|when the family is out with higher offers.| $\|$ \verb|(The going| \\
    \verb|price ranges from about $200 to $500 a week.)|
  \end{quote}
\item
  5 instances of NSBLP SVM correct, abbreviations followed by parenthesized text (no break)
  \begin{quote}
    \verb|It was established in Delaware to pool $75 million raised| \\
    \verb|from U.S.investors for joint ventures with China| \\
    \verb|International Trust & Investment Corp.| $\|$ \verb|(CITIC), the| \\
    \verb|merchantbanking arm of the Chinese government.| \\
    \\
    \verb|FEDERAL HOME LOAN MORTGAGE CORP.| $\|$ \verb|(Freddie Mac):| \\
    \verb|Posted yields on 30-year mortgage commitments for| \\
    \verb|delivery within 30 days.|
  \end{quote}
\item
  4 instances of NSBLP SVM correct, sentence breaks followed by quoted text (break)
  \begin{quote}
    \verb|"On certain occasions a spirit could be earthbound and| \\
    \verb|make itself known," he says.| $\|$ \verb|"It happens."| \\
    \\
    \verb|"Sometimes I felt pity for them when I read those| \\
    \verb|children's letters," she recalls.| $\|$ \verb|"But I knew they| \\
    \verb|were spies who had tried to ruin this country."|
  \end{quote}
\item
  3 instances of NSBLP SVM correct, sentence breaks with nothing unusual (break)
  \begin{quote}
    \verb|Other Western Union securities were also lower.| $\|$ \verb|The| \\
    \verb|company's 7.90% sinking fund debentures were quoted at a| \\
    \verb|bid price of 14 1/4 and an offered price of 30, while the| \\
    \verb|10 3/4% subordinated debentures of 1997 were being bid for| \\
    \verb|at 28 and offered at around 34 3/4.|
  \end{quote}
\item
  3 instances of PBP-based correct, initials as part of capitalized
  expressions (no break)
  \begin{quote}
    \verb|The bulk of the 897.5 million cubic feet of gas a day| \\
    \verb|would go to two U.S. pipelines proposing to ship gas to| \\
    \verb|the U.S.| $\|$ \verb|Northeast.|
  \end{quote}
\item
  2 instances of NSBLP SVM correct, abbreviations followed by numerical
  expressions (no break)
  \begin{quote}
    \verb|Dealers said New West sold $1 billion of Federal Home| \\
    \verb|Loan Mortgage Corp.| $\|$ \verb|9% securities.|
  \end{quote}
\item
  2 instances of PBP-based correct, sentences ending in initials, followed by
  capitalized words that are not usually capitalized (break)
  \begin{quote}
    \verb|It is difficult for consumers to accept that there is| \\
    \verb|such a supply problem in the U.S.| $\|$ \verb|Electricity| \\
    \verb|consumers historically have accepted, as a right, a high| \\
    \verb|level of service from the utility industry.|
  \end{quote}
\item
  2 instances of NSBLP SVM correct, initials followed by capitalized words
  that are usually capitalized (no break)
  \begin{quote}
    \verb|When Europe took over the arena, activity diminished as| \\
    \verb|participants awaited the U.S.| $\|$ \verb|GNP report.|
  \end{quote}
\item
  2 instances of NSBLP SVM correct, non-sentence-breaks with nothing unusual (no break)
  \begin{quote}
    \verb|Also, Needham & Co.| $\|$ \verb|recommended Sun Microsystems and said| \\
    \verb|that it believes "the worst news is out" on the maker of| \\
    \verb|computer workstations and that its stock will rise from| \\
    \verb|its current price level.|
  \end{quote}
\item
  2 instances of NSBLP SVM correct, initials followed by parenthesized
  text (no break)
  \begin{quote}
    \verb|The control of the airline was transferred to a private| \\
    \verb|group made up of Mexican, U.S.| $\|$ \verb|(Chase Manhattan) and| \\
    \verb|British investors who made the winning bid in the| \\
    \verb|auctioning process that opened in July.|
  \end{quote}
\item
  1 instance of NSBLP SVM correct, a sentence break followed by text starting with a
  capitalized word that is usually capitalized (break)
  \begin{quote}
    \verb|Lyondell Petrochemical dropped 1/2 to 22 1/4 on 2.1| \\
    \verb|million shares.| $\|$ \verb|Goldman Sachs lowered its estimate of| \\
    \verb|1989 earnings by 9%, based on "industry factors related| \\
    \verb|to volume and margins," according to the company.|
  \end{quote}
\item
  1 instance of PBP-based correct, a sentence break followed by text starting with a capitalized word
  that is usually capitalized (break)
  \begin{quote}
    \verb|"I'm predicting a move of 200 to 400 points down," he| \\
    \verb|says, "but the word I'm using is 'detour.'| $\|$ \verb|I don't| \\
    \verb|think the bull market is over.|
  \end{quote}
\item
  1 instance of PBP-based correct, an abbreviation followed by a capitalized
  word that is usually capitalized (no break)
  \begin{quote}
    \verb|At 12:01 a.m.| $\|$ \verb|EST today, the federal government's| \\
    \verb|temporary $2.87 trillion debt limit expired.|
  \end{quote}
\item
  1 instance of NSBLP SVM correct, lower cased initials (no break)
  \begin{quote}
    \verb|The last traces of the October 1987 stock market crash| \\
    \verb|were erased at 4 p.m.| $\|$ \verb|yesterday when the Dow Jones| \\
    \verb|Industrial Average closed at a record high of 2734.64.|
  \end{quote}
\item
  1 instance of NSBLP SVM correct, a quotation ending in a question mark,
  followed by a dash introducing the rest of the sentence (no break)
  \begin{quote}
    \verb|The emergence of Russian corporatism had been anticipated| \\
    \verb|in journalist George Urban's introduction to a series of| \\
    \verb|colloquies -- "Can the Soviet System Survive Reform?"| \\
    $\|$ \verb|-- published this spring.|
  \end{quote}
\item
  1 instance of PBP-based correct, a sentence ending in initials, followed by
  a capitalized word that is usually capitalized (break)
  \begin{quote}
     \verb|The company said the possible sale is part of a long-term| \\
     \verb|strategy to shed refining and marketing operations outside| \\
     \verb|the U.S.| $\|$ \verb|Amoco already has shed such operations in Italy,| \\
     \verb|Australia and India.|
  \end{quote}
\item
  1 instance of PBP-based correct, a sentence break followed by text starting with a
  numerical expression (break)
  \begin{quote}
     \verb|Indianapolis Museum of Art: "Latin and Haitian Folk Art"| \\
     \verb| -- An assortment of santos (hand-carved devotional| \\
     \verb|images), retablos (painted tin votive objects) and| \\
     \verb|colorful voodoo banners (handmade of satin, sequins and| \\
     \verb|beads), brought here from Latin America and Haiti by| \\
     \verb|importer Alden Smith.| $\|$ \verb|200 West 38th St., Nov. 1-Dec. 3.|
  \end{quote}
\end{itemize}

\chapter{Resources for data extraction and annotation}

This appendix provides resources for extracting, and sentence break
annotating, the data we use from the Penn Treebank WSJ corpus, the
Satz-3 WSJ corpus, and the English Web Treebank corpus, using data
published by the Linguistic Data Consortium (LDC). For each of these
three data sets, we provide a Perl script for extracting unannotated
raw text from a coresponding LDC data release, plus lists of byte
offsets for the positions of sentence breaks in the text.

Details about the use of each data extraction script and the names and
format of the text output files it produces are described below under
the section for the data set the script pertains to. We have verified
that, at least under Linux, if these scripts are copied from the pdf
version of this document displayed in the Chrome browser, and pasted
into a plain text file, when they are run as described below, they
produce the expected output. The first line of each script expects the
location of the Perl executable to be \verb|/usr/bin/perl|, so these
lines need to be changed to run on systems where Perl is installed in
some other location.

Each list of sentence break byte offsets includes the positions of all
sentence breaks in a particular text output file produced by our data
extraction script, but sentence breaks signaled by a period, question
mark, or exclamation point (the only sentence breaks we train and test
on) are indicated by an asterisk (\verb|*|) following the byte offset.
For sentence breaks located at white space (one or more space or
newline characters), we position the sentence break between the last
printable character before the white space and the first subsequent
space or newline character.

For the Satz-3 WSJ corpus and the English Web Treebank corpus, we
provide two sets of sentence break byte offsets. The lists of sentence
break byte offsets we refer to as V1 include the sentence breaks used
in the experiments and analyses reported in Chapters~5--11 of this
report. The V2 sentence break byte offsets include the sentence breaks
used in the experiments reported in Chapter~12, after applying
proposed changes in how sentence breaks following ellipses are
annotated. The V1 sentence break byte offsets are listed explicitly,
and the V2 sentence break byte offsets are specfied by sets of
deletions and additions to the corresponding V1 lists of sentence
break byte offsets.

The data extraction scripts and sentence break byte offset
specifications are deliberately published here in a nearly unreadably
small font, to minimize the number of pdf document pages that need to
be copied and pasted.

\section{Penn Treebank WSJ corpus}

\subsection{Data extraction script for Penn Treebank WSJ corpus}

The version of the Penn Treebank WSJ corpus that we use can be
extracted from the LDC Treebank-2 release (LDC95T7) using the Perl
script below. This script takes two command line arguments, the first
being a source directory containing the raw version of the text as
distributed by the LDC, and the second being a target directory for
the extracted text. The desired source directory as distributed in the
LDC Treebank-2 release is the \verb|treebank2/raw/wsj|
subdirectory. This directory in turn contains subdirectories named
\verb|00| through \verb|24|, for each of the 25 sections of the PTB
WSJ corpus. The data extraction script tries to extract data from
every file in each section directory, so no other files should be
placed in these directories.

The data extraction script creates one plain ASCII text output file
for each PTB WSJ section, named \verb|ptb-wsj.00.extracted-text|
through \verb|ptb-wsj.24.extracted-text|, in the target directory
named by the second command line argument. The target directory must
exist prior to running the extraction script. Each text output file
placed in this directory consists of plain text paragraphs without
markup, separated by single blank lines. Each paragraph is itself a
single line with no intraparagraph line breaks.\footnote{We remove
intraparagraph line breaks from the text as distributed by the LDC,
because these line breaks represent a heuristic attempt to divide the
text into sentences, rather than line breaks actually occuring in the
original text.}
\\
\begin{spacing}{0.5}
{\tiny
\noindent
\verb|#!/usr/bin/perl|\\
\verb#($input_dir,$output_dir) = @ARGV;#\\
\verb#@sections = ("00","01","02","03","04","05","06","07","08","09","10","11","12","13","14","15","16","17","18","19","20","21",#\\
\verb#             "22","23","24");#\\
\verb#%last_file_in_section = {};#\\
\verb#$last_file_in_section{"08"} = "wsj_0820";#\\
\verb#$last_file_in_section{"24"} = "wsj_2454";#\\
\verb#%section_token_edits = {};#\\
\verb#$section_token_edits{"00"} = [[1159,2,"S.p.A."],[6635,2,"S.p.A.,"],[37742,2]];#\\
\verb#$section_token_edits{"01"} = [[2855,18],[6793,2,"7/16%"],[7287,2,"Rey/Fawcett,"],[14343,1],[22501,21],#\\
\verb#                              [25020,1,"yesterday\x27s","economic"],[27693,39],[32219,1,"Paper"],[32224,1,"7/8,"],#\\
\verb#                              [32226,2,"International"],[39580,5]];#\\
\verb#$section_token_edits{"02"} = [[42,15],[4485,10],[19215,3],[21558,3],[21716,2],[31770,28]];#\\
\verb#$section_token_edits{"03"} = [[14489,6],[25623,2,"Stovall/Twenty-First"],[28046,5],[31503,10924]];#\\
\verb#$section_token_edits{"04"} = [[585,2,"16/32"],[6074,20],[15143,4],[26633,19],[30586,5],[32687,2,"G.m.b.H."]];#\\
\verb#$section_token_edits{"05"} = [[2620,5],[34035,12]];#\\
\verb#$section_token_edits{"06"} = [[1310,15],[2585,26],[3296,29],[4620,10],[5500,10],[5894,29],[21494,2,"S.p.A."],[38781,2],#\\
\verb#                              [39059,5]];#\\
\verb#$section_token_edits{"07"} = [[9047,3],[10939,16],[15634,2],[17788,8],[25357,1,"\x60Who"],[25367,1,"\x60You"],[28273,1,"\x60Big"]];#\\
\verb#$section_token_edits{"08"} = [[8632,13]];#\\
\verb#$section_token_edits{"09"} = [[382,2],[1775,1,"\x60I\x27ve"],[3569,2,"mystery/comedy.\""],[3856,3],[4224,2,"G.m.b.H.,"],[20524,2],#\\
\verb#                              [26939,29],[30404,3],[34530,2,"5/16%"],[38631,10]];#\\
\verb#$section_token_edits{"10"} = [[25968,5],[26852,1,"\x60We"],[33385,1,"TPA","sales"],[33577,27],[38157,2]];#\\
\verb#$section_token_edits{"11"} = [[3191,2],[16523,2],[31567,1,"Tourism"],[36236,1,"\x60controlled"]];#\\
\verb#$section_token_edits{"12"} = [[14078,5],[14791,2,"Bard/EMS"],[19892,2],[20524,2],[21750,3]];#\\
\verb#$section_token_edits{"13"} = [[18567,2,"G.m.b.H."],[33350,2,"S.p.A."],[34079,2,"S.p.A."],[34110,2,"S.p.A."],#\\
\verb#                              [34248,2,"S.p.A.-controlled"],[34253,2,"S.p.A.,"],[40516,4],[41573,4]];#\\
\verb#$section_token_edits{"14"} = [[124,25],[203,4],[6736,10],[12857,37],[16605,13],[24935,15],[28697,1,"\x60shvartzer\x27"],#\\
\verb#                              [29001,1,"\x60breakup\x27"],[42949,3],[45313,2]];#\\
\verb#$section_token_edits{"15"} = [[10199,2],[20066,5],[20147,5],[23558,6],[39146,2],[40656,3]];#\\
\verb#$section_token_edits{"16"} = [[4066,4],[9585,21],[12433,19],[15398,4],[18618,1,"this.\""],[18647,5],[22604,3],[38330,22],#\\
\verb#                              [52596,1,"\x60We"]];#\\
\verb#$section_token_edits{"17"} = [[3963,1,"\x60We"],[10746,6],[11025,2,"U.S."],[13938,17],[14136,19],[14876,22],[22810,2,"H.F."],#\\
\verb#                              [30064,1,"\x60I\x27m"],[35460,1,"\x60pro-choice\x27"]];#\\
\verb#$section_token_edits{"18"} = [[27080,1,"\x60All"],[27519,2],[29986,29],[34828,1,"and","gains"],[35993,29]];#\\
\verb#$section_token_edits{"19"} = [[19811,5],[26590,22],[27980,30],[28033,27],[29160,4],[30693,2,"5/16%"]];#\\
\verb#$section_token_edits{"20"} = [[2613,3],[7708,8],[9421,2],[27440,2,"illegal","exports","lowered"]];#\\
\verb#$section_token_edits{"21"} = [[5524,2],[12380,5],[17381,24],[24086,4],[34906,14104]];#\\
\verb#$section_token_edits{"22"} = [[4218,4],[15480,24],[23534,1,"\x60wow."],[35034,17423]];#\\
\verb#$section_token_edits{"23"} = [[1720,17],[5454,22],[5505,2,"16/64-inch"],[23647,12],[23816,2],[32382,3],[35443,19],[37044,18],#\\
\verb#                              [37821,1,"Fifth","Amendment"]];#\\
\verb#$section_token_edits{"24"} = [[5112,12],[6564,14],[7495,20],[8808,16],[14713,1,"\x60What"]];#\\
\verb#foreach $section (@sections) {#\\
\verb#  open(OUT,">:utf8","$output_dir/ptb-wsj.$section.extracted-text");#\\
\verb#  @section_tokens = ();#\\
\verb#  opendir(DIR,"$input_dir/$section") ||#\\
\verb#    die("\nCannot open directory $input_dir/$section\n\n");#\\
\verb#  @filenames = ();#\\
\verb#  while ($filename = readdir(DIR)) {#\\
\verb#    unless ($filename =~ /^\./) {#\\
\verb#      push(@filenames,$filename);}}#\\
\verb#  foreach $textfile (sort @filenames) {#\\
\verb#    open(IN,"<:utf8","$input_dir/$section/$textfile") ||#\\
\verb#      die("\nCannot open file $input_dir/$section/$textfile $!\n\n");#\\
\verb#    @para_tokens = ();#\\
\verb#    while ($line = <IN>) {#\\
\verb#      chomp($line);#\\
\verb#      if (($line eq "") || ($line eq ".START ")) {#\\
\verb#        if (@para_tokens) {#\\
\verb#          push(@section_tokens,@para_tokens,"");#\\
\verb#          @para_tokens = ();}}#\\
\verb#      else {#\\
\verb#        push (@para_tokens,split(" ",$line));}}#\\
\verb#    if (@para_tokens) {#\\
\verb#      push(@section_tokens,@para_tokens,"");}#\\
\verb#    close(IN);#\\
\verb#    if ($textfile eq $last_file_in_section{$section}) {#\\
\verb#      last;}}#\\
\verb#  @edits = @{$section_token_edits{$section}};#\\
\verb#  @para_tokens = ();#\\
\verb#  $input_pos = 0;#\\
\verb#  while (@edits) {#\\
\verb#    $edit = shift(@edits);#\\
\verb#    ($delete_pos,$delete_length,@insert_tokens) = @$edit;#\\
\verb#    while ($input_pos < $delete_pos) {#\\
\verb#      $next_token = shift(@section_tokens);#\\
\verb#      $input_pos++;#\\
\verb#      if ($next_token ne "") {#\\
\verb#        push(@para_tokens,$next_token);}#\\
\verb#      elsif (@para_tokens) {#\\
\verb#        print OUT "@para_tokens\n\n";#\\
\verb#        @para_tokens = ();}#\\
\verb#      else {#\\
\verb#        die("\nUnexpected lack of paragraph tokens\n\n");}}#\\
\verb#    while ($delete_length) {#\\
\verb#      shift(@section_tokens);#\\
\verb#      $input_pos++;#\\
\verb#      $delete_length--;}#\\
\verb#    if (@insert_tokens) {#\\
\verb#      push(@para_tokens,@insert_tokens);}}#\\
\verb#  while (@section_tokens) {#\\
\verb#    $next_token = shift(@section_tokens);#\\
\verb#    if ($next_token) {#\\
\verb#      push(@para_tokens,$next_token);}#\\
\verb#    else {#\\
\verb#      print OUT "@para_tokens\n\n";#\\
\verb#      @para_tokens = ();}}#\\
\verb#  close(OUT);}#}
\end{spacing}
The version of the Satz WSJ corpus that we call Satz-3 can be
extracted from the LDC ACL/DCI release (LDC93T1) using the Perl script
below. This script takes two command line arguments, the first being a
source directory containing the text as distributed by the LDC, and
the second being a target directory for the extracted text. The
desired source directory as distributed in the LDC ACL/DCI release is
the \verb|LDC93T1/wsj/1989| subdirectory. This directory contains 41
files of text with SGML markup, only a fraction of which is used for
the Satz WSJ corpus. The data extraction script creates one plain
ASCII text output file, named \verb|satz3-extracted-text|, in the
target directory named by the second command line argument. The target
directory must exist prior to running the extraction script. The text
output file placed in this directory consists of plain text
paragraphs, without markup, separated by single blank lines. Each
paragraph is itself a single line with no intraparagraph line
breaks.\footnote{We remove intraparagraph line breaks from the text as
distributed by the LDC, because these line breaks represent a
heuristic attempt to divide the text into sentences, rather than line
breaks actually occuring in the original text.}
\\
\begin{spacing}{0.5}
{\tiny
\noindent
\verb|#!/usr/bin/perl|\\
\verb#($input_dir,$output_dir) = @ARGV;#\\
\verb#@acl_dci_lines = ([1709,1722],[10484,10535],[10550,10567],[10572,10586],[10599,10606],[10611,10624],[10629,10656],#\\
\verb#                  [10660,10835],[10838,10857],[10860,10936],[10965,10971],[10977,11048],[11061,11065],[11068,11074],#\\
\verb#                  [11077,11079],[11082,11084],[11087,11089],[11092,11093],[11096,11099],[11102,11103],[11106,11107],#\\
\verb#                  [11110,11111],[11114,11115],[11129,11170],[11173,11211],[11214,11226],[11229,11363],[11366,11372],#\\
\verb#                  [11375,11382],[11385,11387],[11390,11392],[11395,11397],[11400,11402],[11405,11407],[11410,11413],#\\
\verb#                  [11416,11419],[11422,11424],[11427,11429],[11432,11610],[11613,11663],[11666,12120],[12125,12154],#\\
\verb#                  [12157,12198],[12207,12248],[12252,12549],[12552,12582],[12585,12596],[12599,12674],[12677,12853],#\\
\verb#                  [12856,13269],[13307,13513],[13552,13684],[13695,13823],[13825,13893],[13895,13912],[13914,13918],#\\
\verb#                  [13920,13920],[13924,13925],[13933,14081],[14084,14101],[14104,14108],[14111,14118],[14121,14273],#\\
\verb#                  [14276,14316],[17765,18278],[18303,18628],[18631,18633],[18636,18638],[18641,18643],[18646,18648],#\\
\verb#                  [18651,18653],[18656,18658],[18661,18664],[18667,18669],[18672,18674],[18677,18679],[18682,18684],#\\
\verb#                  [18687,18799],[18802,18804],[18807,18813],[18816,18821],[18824,18825],[18828,18829],[18832,18834],#\\
\verb#                  [18837,18838],[18841,18842],[18845,18846],[18849,18850],[18853,18854],[18857,18859],[18864,18916],#\\
\verb#                  [18919,18927],[18930,18952],[18955,18966],[18969,18984],[18987,19189],[19196,19218],[19231,19243],#\\
\verb#                  [19252,19290],[19293,19323],[19347,19538],[19551,19687],[19690,19705],[19713,19816],[19819,20155],#\\
\verb#                  [20158,20169],[20173,20175],[20178,20182],[20185,20186],[20190,20191],[20194,20196],[20199,20209],#\\
\verb#                  [20211,20214],[20216,20220],[20222,20224],[20226,20232],[20239,20247],[20254,20259],[20264,20399],#\\
\verb#                  [20420,20513],[20560,20659],[20662,20720],[20723,20842],[20852,20966],[20976,21104],[21109,21131],#\\
\verb#                  [21136,21151],[21156,21290],[21306,21506],[21511,21524],[21535,21756],[21759,21786],[21808,22128],#\\
\verb#                  [22131,22161],[22164,22265],[22268,22269],[22285,22291],[22294,22316],[22319,22325],[22328,22334],#\\
\verb#                  [22337,22345],[22348,22465],[22470,22528],[22531,22632],[22646,22939],[22953,23016],[23019,23054],#\\
\verb#                  [23057,23091],[23094,23098],[23101,23311],[23313,23318],[23320,23323],[23325,23335],[23337,23500],#\\
\verb#                  [23503,23592],[23595,23598],[23601,23604],[23607,23610],[23613,23618],[23621,23624],[23627,23629],#\\
\verb#                  [23632,23635],[23638,23640],[23643,23646],[23649,23651],[23654,23657],[23660,23683],[23686,23689],#\\
\verb#                  [23692,23697],[23700,23702],[23705,23707],[23710,23712],[23715,23719],[23722,23723],[23726,23727],#\\
\verb#                  [23730,23731],[23734,23735],[23738,23739],[23742,23743],[23746,23945],[23947,23970],[23973,23982],#\\
\verb#                  [23985,23990],[23995,24042],[24060,24915],[24918,25106],[25109,25119],[25122,25448],[25451,25818],#\\
\verb#                  [25821,25854],[25858,25859],[25871,25952],[25955,25968],[25971,25979],[25984,26251],[26254,26380],#\\
\verb#                  [26383,26482],[26496,26497],[26521,26533],[26536,26543],[26546,26549],[26552,26557],[26560,26561],#\\
\verb#                  [26564,26566],[26569,26571],[26574,26576],[26579,26581],[26584,26587],[26590,26592],[26595,26847],#\\
\verb#                  [26917,27018],[27021,27094],[27098,27530],[27533,28537],[28540,28544],[28547,28549],[28552,28554],#\\
\verb#                  [28557,28558],[28561,28563],[28566,28568],[28571,28575],[28578,28579],[28582,28583],[28586,28587],#\\
\verb#                  [28590,28591],[28594,28597],[28600,28601],[28604,28605],[28608,28714],[28727,28794],[28797,28805],#\\
\verb#                  [28808,28811],[28814,28821],[28824,28827],[28830,28832],[28835,28837],[28840,28843],[28846,28849],#\\
\verb#                  [28852,28854],[28857,28895],[28897,28899],[28901,28910],[28912,28918],[28920,29134],[29137,29747],#\\
\verb#                  [29782,30265],[30268,30309],[30312,30326],[30329,31205],[31208,31210],[31221,31321],[31324,31328],#\\
\verb#                  [31335,31335],[31340,31341],[31345,31345],[31348,31349],[31353,31353],[31357,31375],[31379,31379],#\\
\verb#                  [31394,31461],[31464,31615],[31618,31642],[31645,31654],[31657,31668],[31671,31682],[31685,31941],#\\
\verb#                  [31944,32095],[32117,32131],[32236,32362],[32365,32534],[32544,32665],[32705,32793],[32796,32903],#\\
\verb#                  [32906,33043],[33046,33047],[33050,33051],[33054,33055],[33058,33059],[33062,33063],[33066,33067],#\\
\verb#                  [33070,33071],[33074,33075],[33078,33079],[33082,33084],[33087,33089],[33092,33094],[33097,33099],#\\
\verb#                  [33102,33103],[33106,33107],[33110,33111],[33114,33115],[33118,33119],[33122,33123],[33126,33191],#\\
\verb#                  [33194,33381],[33394,33717],[33719,33723],[33725,33757],[33760,33768],[33771,33774],[33777,33786],#\\
\verb#                  [33789,33792],[33795,33798],[33801,33803],[33806,33809],[33812,33815],[33818,33820],[33823,34063],#\\
\verb#                  [34066,34243],[34252,34703],[34706,34781],[34784,34820],[34823,35251],[35254,35288],[35323,35424],#\\
\verb#                  [35464,35604],[35607,35882],[35885,35885],[35887,35953],[35956,36019],[36023,36023],[36042,36090],#\\
\verb#                  [36888,37037],[37050,37248],[37267,37518],[37537,37939],[37943,38313],[38316,38318],[38321,38323],#\\
\verb#                  [38326,38327],[38330,38332],[38335,38340],[38343,38346],[38349,38351],[38354,38355],[38358,38359],#\\
\verb#                  [38362,38363],[38366,38367],[38370,38371],[38374,38375],[38380,38409],[38412,38414],[38417,38421],#\\
\verb#                  [38424,38427],[38430,38433],[38436,38438],[38441,38443],[38446,38448],[38451,38453],[38456,38458],#\\
\verb#                  [38461,38464],[38467,38469],[38472,38474],[38477,38479],[38482,38841],[38850,38869],[38874,39292],#\\
\verb#                  [39294,39370],[39390,39407],[39412,39447],[39452,39585],[39594,39633],[39638,39741],[39782,40011],#\\
\verb#                  [40014,40029],[40032,40183],[40186,40273],[40276,40442],[40445,40529],[40532,40772],[40792,40842],#\\
\verb#                  [40845,40871],[40928,41008],[41010,41010],[41012,41144],[41167,41237],[41240,41600],[41603,41769],#\\
\verb#                  [41772,41784],[41789,42165],[42168,42735],[42737,42750],[42752,42763],[42765,42767],[42769,42964],#\\
\verb#                  [42967,42980],[42983,42989],[42992,42994],[42997,43009],[43012,43595],[43605,43840],[43853,43965],#\\
\verb#                  [43968,44355],[44358,44359],[44362,44364],[44367,44369],[44372,44374],[44377,44380],[44383,44384],#\\
\verb#                  [44387,44388],[44391,44393],[44396,44397],[44400,44401],[44404,44406],[44409,44411],[44414,44415],#\\
\verb#                  [44418,44420],[44423,44497],[44500,44508],[44511,44513],[44516,44518],[44521,44523],[44526,44528],#\\
\verb#                  [44531,44533],[44536,44538],[44541,44543],[44546,44548],[44551,44553],[44556,44559],[44562,44564],#\\
\verb#                  [44567,45067],[45071,45071],[45081,45217],[45224,45292],[45299,45353],[45355,45462],[45465,45572],#\\
\verb#                  [45623,46256],[46259,46488]);#\\
\verb#@token_edits = (["d",1169,1],["d",35141,1],["d",46106,2],["d",46109,2],["c",54015,1,"U.S.","concerns."],["d",66740,1],#\\
\verb#                ["c",85970,1,"<P>"],["c",88688,1,"Co."],["d",95617,3],["c",102171,1,"record,\""],["d",117519,3],#\\
\verb#                ["c",118894,1,"par).","9.76%,"],["d",204920,1],["d",270610,1],["c",304494,1,"said.","Yamanouchi"],#\\
\verb#                ["d",324147,1],["a",385139,"<P>"]);#\\
\verb#@file_list = ("w9_21.","w9_21.","w9_2.","w9_30.","w9_41.");#\\
\verb#open(OUT,">","$output_dir/satz3-extracted-text");#\\
\verb#$next_edit = shift(@token_edits);#\\
\verb#$next_edit_type = shift(@$next_edit);#\\
\verb#$next_edit_pos = shift(@$next_edit);#\\
\verb#if ($next_edit_type eq "d") {#\\
\verb#  $next_num_delete_tokens = shift(@$next_edit);#\\
\verb#  @insert_tokens = ();}#\\
\verb#elsif ($next_edit_type eq "a") {#\\
\verb#  @insert_tokens = @$next_edit;#\\
\verb#  $next_num_delete_tokens = 0;}#\\
\verb#elsif ($next_edit_type eq "c") {#\\
\verb#  $next_num_delete_tokens = shift(@$next_edit);#\\
\verb#  @insert_tokens = @$next_edit;}#\\
\verb#$line_pair_ref = shift(@acl_dci_lines);#\\
\verb#($start_line,$end_line) = @$line_pair_ref;#\\
\verb#$current_line = 0;#\\
\verb#$prev_line_not_blank = 0;#\\
\verb#$current_token = 0;#\\
\verb#foreach $textfile (@file_list) {#\\
\verb#  open(IN,"<","$input_dir/$textfile") ||#\\
\verb#    die("\nCannot open file $input_dir/$textfile $!\n\n");#\\
\verb#  @para_tokens = ();#\\
\verb#  @edited_tokens = ();#\\
\verb#  $num_delete_tokens = 0;#\\
\verb#  while ($line = <IN>) {#\\
\verb#    chomp($line);#\\
\verb#    if ($line =~ /^<s> (.*) <\/s>$/) {#\\
\verb#      $current_line++;#\\
\verb#      if (($current_line >= $start_line) && ($current_line <= $end_line)) {#\\
\verb#        foreach $token (split(" ",$1)) {#\\
\verb#          $token =~ s/\]/!/g;#\\
\verb#          push (@para_tokens,$token);}#\\
\verb#        if ($current_line == $end_line) {#\\
\verb#          $line_pair_ref = shift(@acl_dci_lines);#\\
\verb#          if ($line_pair_ref) {#\\
\verb#            ($start_line,$end_line) = @$line_pair_ref;}#\\
\verb#          else {#\\
\verb#            last;}}}#\\
\verb#      $prev_line_not_blank = 1;}#\\
\verb#    else {#\\
\verb#      if (@para_tokens) {#\\
\verb#        foreach $token (@para_tokens) {#\\
\verb#          $current_token++;#\\
\verb#          if ($num_delete_tokens) {#\\
\verb#            $num_delete_tokens--;}#\\
\verb#          else {#\\
\verb#            push(@edited_tokens,$token);}#\\
\verb#          if ($current_token == $next_edit_pos) {#\\
\verb#            if ($next_edit_type eq "a") {#\\
\verb#              foreach $insert_token (@insert_tokens) {#\\
\verb#                if ($insert_token eq "<P>") {#\\
\verb#                  print OUT "@edited_tokens\n\n";#\\
\verb#                  @edited_tokens = ();}#\\
\verb#                else {#\\
\verb#                  push(@edited_tokens,$insert_token);}}#\\
\verb#              $next_num_delete_tokens = 0;}#\\
\verb#            elsif ($next_edit_type eq "d") {#\\
\verb#              $num_delete_tokens += $next_num_delete_tokens;}#\\
\verb#            elsif ($next_edit_type eq "c") {#\\
\verb#              foreach $insert_token (@insert_tokens) {#\\
\verb#                if ($insert_token eq "<P>") {#\\
\verb#                  print OUT "@edited_tokens\n\n";#\\
\verb#                  @edited_tokens = ();}#\\
\verb#                else {#\\
\verb#                  push(@edited_tokens,$insert_token);}}#\\
\verb#              $num_delete_tokens += $next_num_delete_tokens;}#\\
\verb#            if (@token_edits) {#\\
\verb#              $next_edit = shift(@token_edits);#\\
\verb#              $next_edit_type = shift(@$next_edit);#\\
\verb#              $next_edit_pos = shift(@$next_edit);#\\
\verb#              if ($next_edit_type eq "d") {#\\
\verb#                $next_num_delete_tokens = shift(@$next_edit);#\\
\verb#                @insert_tokens = ();}#\\
\verb#              elsif ($next_edit_type eq "a") {#\\
\verb#                @insert_tokens = @$next_edit;#\\
\verb#                $next_num_delete_tokens = 0;}#\\
\verb#              elsif ($next_edit_type eq "c") {#\\
\verb#                $next_num_delete_tokens = shift(@$next_edit);#\\
\verb#                @insert_tokens = @$next_edit;}}#\\
\verb#            else {#\\
\verb#              $next_edit_pos = 0;}}}#\\
\verb#        @para_tokens = ();}#\\
\verb#      if ($prev_line_not_blank) {#\\
\verb#        $current_line++;#\\
\verb#        if (($current_line >= $start_line) && ($current_line <= $end_line)) {#\\
\verb#          $current_token++;#\\
\verb#          if ($num_delete_tokens) {#\\
\verb#            $num_delete_tokens--;}#\\
\verb#          else {#\\
\verb#            print OUT "@edited_tokens\n\n";#\\
\verb#            @edited_tokens = ();}#\\
\verb#          if ($current_token == $next_edit_pos) {#\\
\verb#            if ($next_edit_type eq "a") {#\\
\verb#              foreach $insert_token (@insert_tokens) {#\\
\verb#                if ($insert_token eq "<P>") {#\\
\verb#                  print OUT "@edited_tokens\n\n";#\\
\verb#                  @edited_tokens = ();}#\\
\verb#                else {#\\
\verb#                  push(@edited_tokens,$insert_token);}}}#\\
\verb#            elsif ($next_edit_type eq "d") {#\\
\verb#              $num_delete_tokens += $next_num_delete_tokens;}#\\
\verb#            elsif ($next_edit_type eq "c") {#\\
\verb#              foreach $insert_token (@insert_tokens) {#\\
\verb#                if ($insert_token eq "<P>") {#\\
\verb#                  print OUT "@edited_tokens\n\n";#\\
\verb#                  @edited_tokens = ();}#\\
\verb#                else {#\\
\verb#                  push(@edited_tokens,$insert_token);}}#\\
\verb#              $num_delete_tokens += $next_num_delete_tokens;}#\\
\verb#            if (@token_edits) {#\\
\verb#              $next_edit = shift(@token_edits);#\\
\verb#              $next_edit_type = shift(@$next_edit);#\\
\verb#              $next_edit_pos = shift(@$next_edit);#\\
\verb#              if ($next_edit_type eq "d") {#\\
\verb#                $next_num_delete_tokens = shift(@$next_edit);#\\
\verb#                @insert_tokens = ();}#\\
\verb#              elsif ($next_edit_type eq "a") {#\\
\verb#                @insert_tokens = @$next_edit;#\\
\verb#                $next_num_delete_tokens = 0;}#\\
\verb#              elsif ($next_edit_type eq "c") {#\\
\verb#                $next_num_delete_tokens = shift(@$next_edit);#\\
\verb#                @insert_tokens = @$next_edit;}}#\\
\verb#            else {#\\
\verb#              $next_edit_pos = 0;}}#\\
\verb#          if ($current_line == $end_line) {#\\
\verb#            $line_pair_ref = shift(@acl_dci_lines);#\\
\verb#            if ($line_pair_ref) {#\\
\verb#              ($start_line,$end_line) = @$line_pair_ref;}#\\
\verb#            else {#\\
\verb#              last;}}}}#\\
\verb#      $prev_line_not_blank = 0;}}#\\
\verb#  if (@para_tokens) {#\\
\verb#    die("\nNot expecting file $textfile to end in a sentence line\n\n");}#\\
\verb#  close(IN);}#\\
\verb#close(OUT);#}
\end{spacing}
The version of the English Web Treebank corpus that we use can be
extracted from the LDC English Web Treebank release (LDC2012T13) using
the Perl script below. This script takes two command line arguments,
the first being a file specifying source directories containing the
raw version of the text as distributed by the LDC, and the second
being a target directory for the extracted text. The first command
line argument is a file listing multiple directories because the
English Web Treebank contains five genres, and the raw versions of the
text in the various genres are not grouped together in the LDC
distribution. The format of the source directories file is five lines
of the form
\begin{quote}
  $\langle$\verb|genre|$\rangle$ $\langle$\verb|source-directory|$\rangle$
\end{quote}
where $\langle$\verb|genre|$\rangle$ is one of the strings
\verb|answers|, \verb|email|, \verb|newsgroup|, \verb|reviews|, and
\verb|weblog|, and $\langle$\verb|source-directory|$\rangle$ is the
corresponding source directory, which in the LDC distribution are
the following subdirectories:
\begin{quote}
  \verb|eng_web_tbk/data/answers/source/source_original| \\
  \verb|eng_web_tbk/data/email/source/source_original| \\
  \verb|eng_web_tbk/data/newsgroup/source/source_original| \\
  \verb|eng_web_tbk/data/reviews/source/source_original| \\
  \verb|eng_web_tbk/data/weblog/source/source_original|
\end{quote}
The data extraction script tries to extract data from every file in
each genre directory, so no other files should be placed in these
directories.

In the source directories file, full path file names should be used,
e.g,

\noindent
\\
\verb|answers /home/username/eng_web_tbk/data/answers/source/source_original|\\
\\
depending on exactly where in the local file system the data is
stored. Note that the notations \verb|~/| or \verb|~username/| cannot
be used, because the script does not invoke the shell, which is needed
to expand the \verb|~| notation into a full path. The two fields on
each line of the source directories file can be separated by any
combination of spaces and tabs.

The data extraction script creates one plain ASCII text output file
for each genre, named
\begin{quote}
\verb|english-web-treebank.answers.extracted-text| \\
\verb|english-web-treebank.email.extracted-text| \\
\verb|english-web-treebank.newsgroup.extracted-text| \\
\verb|english-web-treebank.reviews.extracted-text| \\
\verb|english-web-treebank.weblog.extracted-text|
\end{quote}
in the target directory named by the second command line argument.
The target directory must exist prior to running the extraction
script. Each text output file placed in this directory consists of
plain text paragraphs without markup, separated by one or more blank
lines. Unlike the PTB WSJ and Satz-3 WSJ corpora, the LDC distribution
of the English Web Treebank preserves the original intraparagraph line
breaks in the source text, so we also preserve intraparagraph line
breaks in these text output files.
\\
\begin{spacing}{0.5}
{\tiny
\noindent
\verb|#!/usr/bin/perl|\\
\verb#($genre_dir_spec,$output_dir) = @ARGV;#\\
\verb#open(DIRS,"<:utf8",$genre_dir_spec) ||#\\
\verb#  die("\nCannot open file $genre_dir_spec\n\n");#\\
\verb#%genre_dir = ();#\\
\verb#while ($line = <DIRS>) {#\\
\verb#  ($genre,$dir) = split(" ",$line);#\\
\verb#  $genre_dir{$genre} = $dir;}#\\
\verb#close(DIRS);#\\
\verb#$weblogs_delete_begin_line = 2852;#\\
\verb#$weblogs_delete_end_line = 3032;#\\
\verb#foreach $genre ("answers","email","newsgroup","reviews","weblog") {#\\
\verb#  if (!exists($genre_dir{$genre})) {#\\
\verb#    die("\nNo directory specified for genre $genre\n\n");}#\\
\verb#  else {#\\
\verb#    $input_dir = $genre_dir{$genre};}#\\
\verb#  open(OUT,">:utf8","$output_dir/english-web-treebank.$genre.extracted-text");#\\
\verb#  $output_line_count = 0;#\\
\verb#  opendir(DIR,$input_dir) ||#\\
\verb#    die("\nCannot open directory $input_dir\n\n");#\\
\verb#  @filenames = ();#\\
\verb#  while ($filename = readdir(DIR)) {#\\
\verb#    unless ($filename =~ /^\./) {#\\
\verb#      push(@filenames,$filename);}}#\\
\verb#  closedir(DIR);#\\
\verb#    if ($genre eq "newsgroup") {#\\
\verb#    @presorted_filenames = sort {lc($a) cmp lc($b)} @filenames;#\\
\verb#    @sorted_filenames = ();#\\
\verb#    for ($i=1; $i<=6; $i++) {#\\
\verb#      push(@sorted_filenames,shift(@presorted_filenames));}#\\
\verb#    @saved_filenames = ();#\\
\verb#    for ($i=1; $i<=16; $i++) {#\\
\verb#      push(@saved_filenames,shift(@presorted_filenames));}#\\
\verb#    for ($i=1; $i<=7; $i++) {#\\
\verb#      push(@sorted_filenames,shift(@presorted_filenames));}#\\
\verb#    push(@sorted_filenames,@saved_filenames,@presorted_filenames);}#\\
\verb#  else {#\\
\verb#    @sorted_filenames = sort {lc($a) cmp lc($b)} @filenames;}#\\
\verb#  foreach $textfile (@sorted_filenames) {#\\
\verb#    open(IN,"<:utf8","$input_dir/$textfile") ||#\\
\verb#      die("\nCannot open file $input_dir/$textfile $!\n\n");#\\
\verb#    while ($line = <IN>) {#\\
\verb#      if ($line =~ /^\s*<.*>\s*$/) {#\\
\verb#        if ($genre ne "email") {#\\
\verb#          $line = "\n";}}#\\
\verb#      else {#\\
\verb#        while ($line =~ /&amp;apos/) {#\\
\verb#          $line =~ s/&amp;apos/\x27/;}#\\
\verb#        while ($line =~ /&amp;/) {#\\
\verb#          $line =~ s/&amp;/&/;}#\\
\verb#        while ($line =~ /&quot;/) {#\\
\verb#          $line =~ s/&quot;/"/;}#\\
\verb#        while ($line =~ /&lt;/) {#\\
\verb#          $line =~ s/&lt;/</;}#\\
\verb#        while ($line =~ /&gt;/) {#\\
\verb#          $line =~ s/&gt;/>/;}#\\
\verb#        $line =~ tr/\x{00A0}/ /;#\\
\verb#        $line =~ tr/\x{00A3}/L/;#\\
\verb#        $line =~ tr/\x{00AD}/ /;#\\
\verb#        $line =~ tr/\x{00B0}/o/;#\\
\verb#        $line =~ tr/\x{00B3}/3/;#\\
\verb#        $line =~ tr/\x{00B4}/\x27/;#\\
\verb#        $line =~ tr/\x{00B7}/*/;#\\
\verb#        $line =~ tr/\x{00C1}/A/;#\\
\verb#        $line =~ tr/\x{00C3}/A/;#\\
\verb#        $line =~ tr/\x{00C7}/C/;#\\
\verb#        $line =~ tr/\x{00CD}/I/;#\\
\verb#        $line =~ tr/\x{00E0}/a/;#\\
\verb#        $line =~ tr/\x{00E1}/a/;#\\
\verb#        $line =~ tr/\x{00E3}/a/;#\\
\verb#        $line =~ tr/\x{00E4}/a/;#\\
\verb#        $line =~ tr/\x{00E7}/c/;#\\
\verb#        $line =~ tr/\x{00E9}/e/;#\\
\verb#        $line =~ tr/\x{00EA}/e/;#\\
\verb#        $line =~ tr/\x{00EF}/i/;#\\
\verb#        $line =~ tr/\x{00F3}/o/;#\\
\verb#        $line =~ tr/\x{00F4}/o/;#\\
\verb#        $line =~ tr/\x{00F6}/o/;#\\
\verb#        $line =~ tr/\x{00FC}/u/;#\\
\verb#        $line =~ tr/\x{03A5}/Y/;#\\
\verb#        $line =~ tr/\x{2000}/ /;#\\
\verb#        $line =~ tr/\x{2013}/-/;#\\
\verb#        $line =~ tr/\x{2014}/-/;#\\
\verb#        $line =~ tr/\x{2014}/-/;#\\
\verb#        $line =~ tr/\x{2018}/\x27/;#\\
\verb#        $line =~ tr/\x{2019}/\x27/;#\\
\verb#        $line =~ tr/\x{201C}/"/;#\\
\verb#        $line =~ tr/\x{201D}/"/;#\\
\verb#        $line =~ s/\x{2026}/.../g;#\\
\verb#        $line =~ s/\x{2665}/<3/g;}#\\
\verb#      $output_line_count++;#\\
\verb#      if (($genre ne "weblog") ||#\\
\verb#          ($output_line_count < $weblogs_delete_begin_line) ||#\\
\verb#          ($output_line_count > $weblogs_delete_end_line)) {#\\
\verb#        print OUT $line;}#\\
\verb#      $prev_line = $line;}#\\
\verb#    if ($prev_line ne "\n") {#\\
\verb#      $output_line_count++;#\\
\verb#      print OUT "\n";}#\\
\verb#    close(IN);}#\\
\verb#  close(OUT);}#}
\end{spacing}
\end{document}